\documentclass{article}

\PassOptionsToPackage{numbers, compress}{natbib}

\usepackage[preprint]{neurips_2021}

\usepackage[utf8]{inputenc} %
\usepackage[T1]{fontenc}    %
\usepackage{hyperref}       %
\usepackage{url}            %
\usepackage{booktabs}       %
\usepackage{amsfonts}       %
\usepackage{nicefrac}       %
\usepackage{microtype}      %
\usepackage{xcolor}         %

\usepackage{enumitem} %
\usepackage{multirow}
\usepackage{makecell}
\usepackage{array} 
\usepackage{graphicx, xcolor, amsmath, amssymb} 
\usepackage{comment} %

\usepackage[utf8]{inputenc}
\usepackage{pgfplots}
\DeclareUnicodeCharacter{2212}{−}
\usepgfplotslibrary{groupplots,dateplot}
\usetikzlibrary{patterns,shapes.arrows}
\pgfplotsset{compat=newest}

\def\gt{GainTuning}
\def\yt{$\mathbf{y}_{\text{test}}$}

\DeclareMathOperator*{\argmin}{arg\,min}
\title{Adaptive Denoising via \gt}

{
\author{Sreyas Mohan$^{1}$, ~Joshua L. Vincent$^{2}$, ~Ramon Manzorro$^{2}$,  ~Peter A. Crozier $^{2}$,\And
Eero P. Simoncelli $^{1, 3, 4}$,~Carlos Fernandez-Granda$^{1, 4}$ 
\\\\~~~~~~~~~$^1$Center For Data Science, NYU, \\
$^2$School for Engineering of Matter, Transport and Energy, ASU\\ 
~~~~~~~~~$^3$Center for Neural Science, NYU and Flatiron Institute, Simons Foundation \\
~~~~~~~~~$^4$Courant Institute of Mathematical Sciences, NYU
}}

\begin{document}

\maketitle

\begin{abstract}
  Deep convolutional neural networks (CNNs) for image denoising are usually trained on large datasets. These models achieve the current state of the art, but they have difficulties generalizing when applied to data that deviate from the training distribution. Recent work has shown that it is possible to train denoisers on a single noisy image. These models adapt to the features of the test image, but their performance is limited by the small amount of information used to train them. Here we propose “\gt”, in which CNN models pre-trained on large datasets are adaptively and selectively adjusted for individual test images. To avoid overfitting, \gt\ optimizes a single multiplicative scaling parameter (the “Gain”) of each channel in the convolutional layers of the CNN. We show that \gt\ improves state-of-the-art CNNs on standard image-denoising benchmarks, boosting their denoising performance on nearly every image in a held-out test set. These adaptive improvements are even more substantial for test images differing systematically from the training data, either in noise level or image type. We illustrate the potential of adaptive denoising in a scientific application, in which a CNN is trained on synthetic data, and tested on real transmission-electron-microscope images. In contrast to the existing methodology, \gt\ is able to faithfully reconstruct the structure of catalytic nanoparticles from these data at extremely low signal-to-noise ratios. 

\end{abstract}

\section{Introduction}

Like many problems in image processing, the recovery of signals from noisy measurements has been revolutionized by the development of convolutional neural networks (CNNs)~\citep{dncnn, tnrd, ffdnet}. These models are typically trained on large databases of images, either in a supervised~\cite{biasfree, dncnn, tnrd, durr, ffdnet} or an unsupervised fashion~\cite{noise2same, noise2self, n2v, blindspotnet}. Once trained, these solutions are  evaluated on noisy test images. This approach achieves state-of-the-art performance when the test images and the training data belong to the same distribution. However, when this is not the case, the performance of these models is often substantially degraded~\cite{lidia, biasfree, durr}. This is an important limitation for many practical applications, in which it is challenging (or even impossible) to gather a training dataset that is comparable in noise and signal content to the images encountered at test time. Overcoming this limitation requires \emph{adaptation} to the test data.

A recent unsupervised method (Self2Self) has shown that CNNs can be trained exclusively on individual test images, producing impressive results~\cite{self2self}. Self2Self adapts to the test image, but its performance is limited by the small amount of information available for learning. As a result it generally underperforms CNN models trained on large databases. 

In this work, we propose \emph{\gt}, a framework to bridge the gap between models trained on large datasets without adaptation, and models trained exclusively on test images. 
In the spirit of two recent methods~\cite{surekoreanarxiv,lidia}, \gt~  adapts pre-trained CNN models to individual test images by minimizing an unsupervised denoising cost function, thus fusing the generic capabilities obtained from the training data with specific refinements matched to the structure of the test data. 
\gt\ does not adapt the full parameter set (filter weights and additive constants) to the test image, but instead optimizes a single multiplicative scaling parameter (the “Gain”) for each channel within each layer of the CNN. The dimensionality of this reduced parameter set is a small fraction ($\approx 0.1\%$) of that of the full parameter set. We demonstrate through extensive examples that this %
prevents overfitting to the test data. %
The \gt\ procedure is general, and can be applied to any CNN denoising model, regardless of the architecture or pre-training process.

\begin{figure}[t]
    \centering
    \begin{tabular}{c@{\hskip 0.01in}c@{\hskip 0.01in}c@{\hskip 0.01in}c@{\hskip 0.01in}c@{\hskip 0.01in}c}
    \scriptsize{(a) Noisy image} &  
    \thead{ \scriptsize{(b) Unsupervised training} \\ \scriptsize{only on (a) \cite{self2self}}} &
    \thead{\scriptsize{(c) Supervised training} \\ \scriptsize{on simulated data~\cite{mohan2020deep}}} &
    \thead{\scriptsize{(d) \gt\ on CNN} \\ \scriptsize{trained on sim. data (c) }} &
    \thead{ \scriptsize{(e) Estimated reference} \\ \scriptsize{image} } \\
    \includegraphics[width=1.05in]{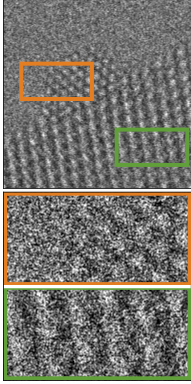}&
    \includegraphics[width=1.05in]{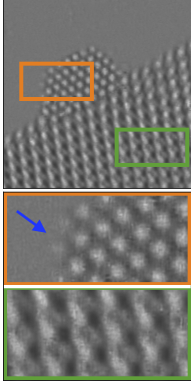}&
    \includegraphics[width=1.05in]{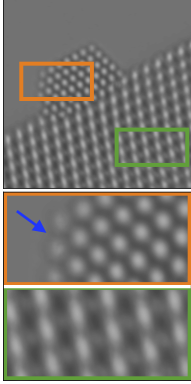}&
    \includegraphics[width=1.05in]{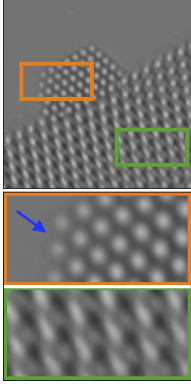}&
    \includegraphics[width=1.05in]{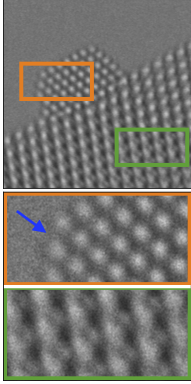}\\
    \end{tabular}
    \vspace{-0.2cm}
    \caption{\textbf{Denoising results for real-world data.} (a) An experimentally-acquired atomic-resolution transmission electron microscope image of a CeO$_2$-supported Pt nanoparticle. The image has a very low signal to noise ratio (PSNR of $\approx 3  dB$). (b) Denoised image obtained using Self2Self~\cite{self2self}, which fails to reconstruct three atoms (blue arrow, second row). (c) Denoised image obtained via a CNN trained on a simulated dataset, where the pattern of the supporting atoms is not recovered faithfully (third row). (d) Denoised image obtained by adapting the CNN in (c) to the noisy test image in (a) using \gt. Both the nanoparticle and the support are recovered without artefacts. (e) Reference image, estimated by averaging $40$ different noisy images of the same nanoparticle. 
    }
    \label{fig:nano}
\end{figure}

\textbf{Our contributions}. GainTuning provides a novel method for adapting CNN denoisers trained on large datasets to a single test image. GainTuning improves state-of-the-art CNNs on standard image-denoising benchmarks, boosting their denoising performance on nearly every image in held-out test sets. Performance improvements are even more substantial when the test images differ systematically from the training data. We showcase this ability through controlled experiments in which we vary the distribution of the noise and image structure of the test data. Finally, we evaluate GainTuning in a real scientific-imaging application where adaptivity is crucial: denoising transmission-electron-microscope data at extremely low signal-to-noise ratios. As shown in Figure~\ref{fig:nano}, both CNNs pre-trained on simulated images and CNNs trained only on the test data produce denoised images with substantial artefacts. In contrast, GainTuning achieves effective denoising, accurately revealing the atomic structure in the real data.

\begin{figure}[t]
    \centering
    \begin{tabular}{c@{\hskip 0.1in}c@{\hskip 0.1in}c}
    \scriptsize{(a)~Pre-training on a database} &  \scriptsize{(b)~Adaptive training on test image} & 
    \scriptsize{(c)~Combined pre-training and test-adaptive training}\\
    \includegraphics[height=1.9in]{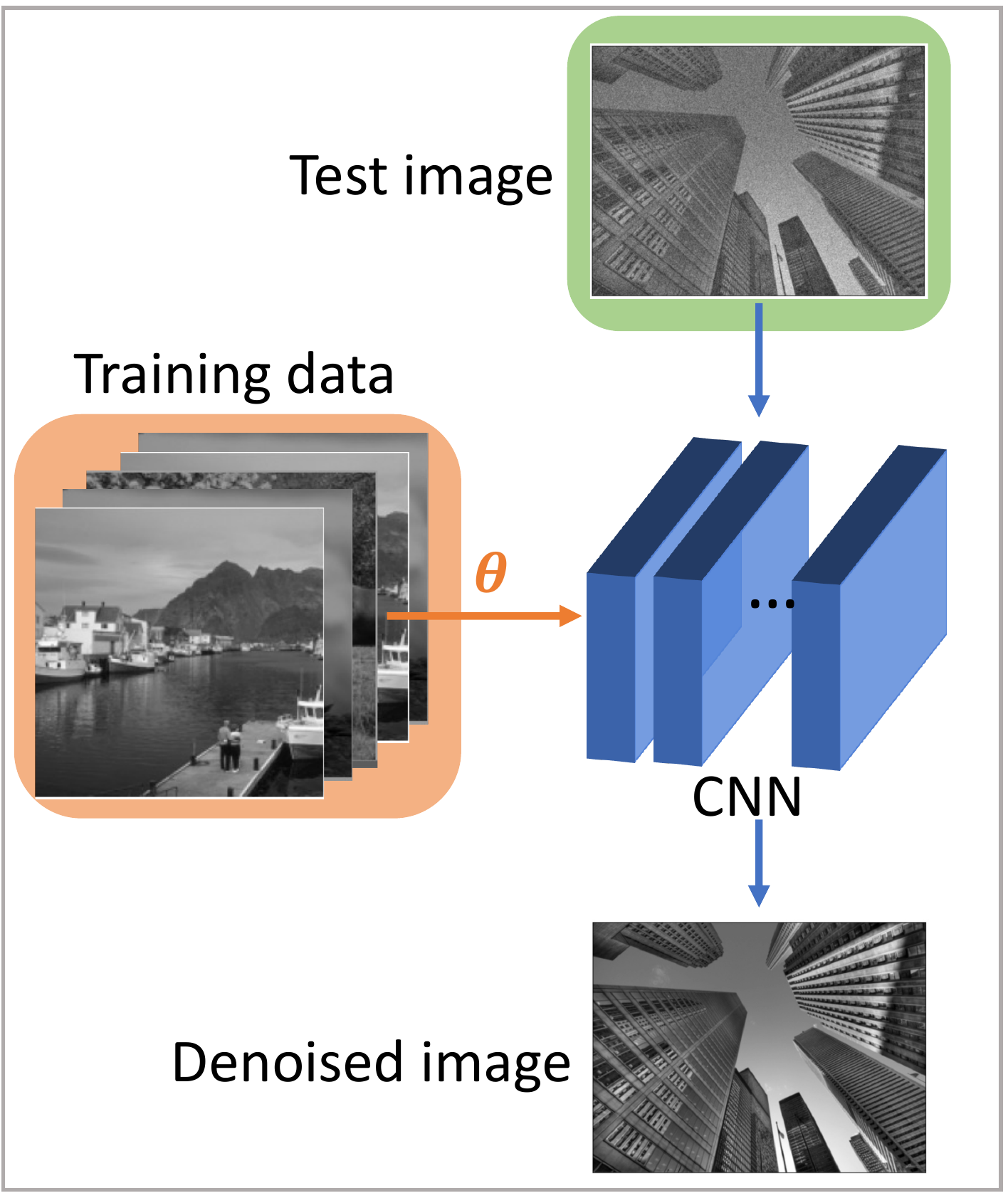}&
    \includegraphics[height=1.9in]{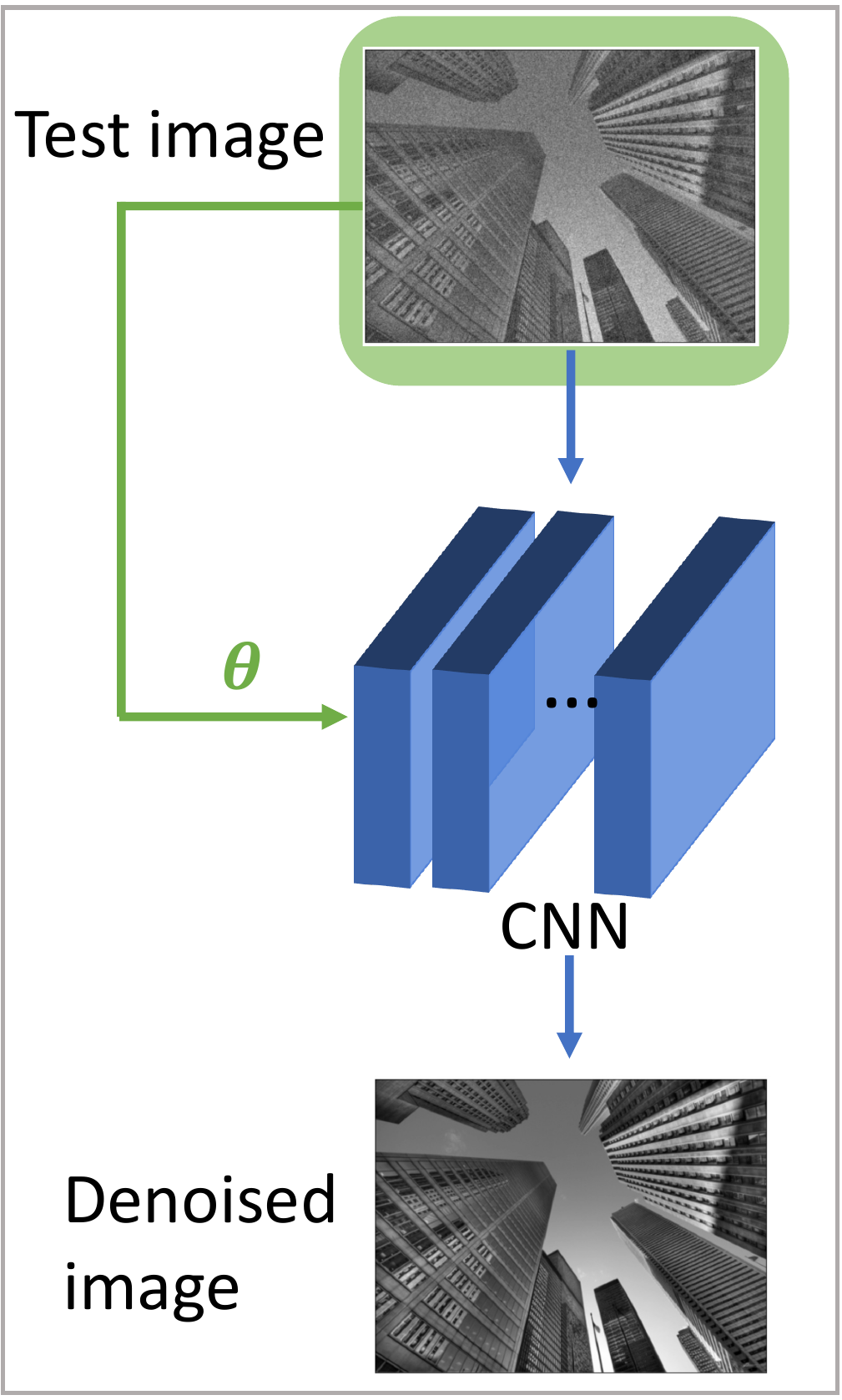}&
    \includegraphics[height=1.9in]{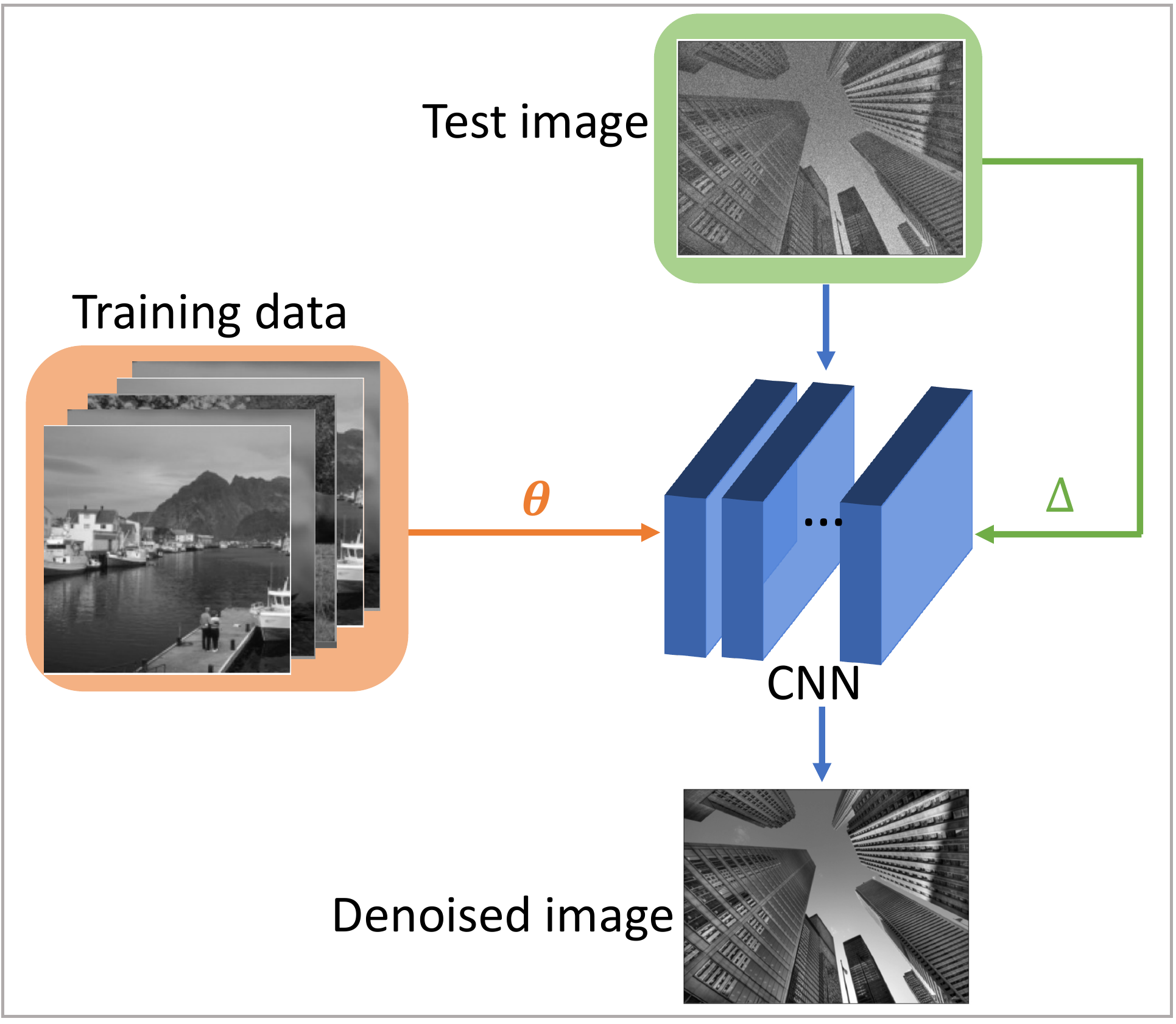}\\
    \end{tabular}
    \caption{\textbf{Proposed denoising paradigm}. (a) Typically, CNNs are trained on a large dataset and evaluated directly on a test image. %
    (b) Recent unsupervised methods perform training on a single test image.
    (c) We propose \gt, a framework which bridges the gap between both of these paradigms: a CNN pre-trained on a large training database is adapted to the test image.}
    \label{fig:paradigms}
\end{figure}

\section{Related Work}
\label{sec:related_work}

\textbf{Denoising via deep learning}. In the last five years, CNN-based methods have clearly outperformed prior state of the art~\cite{Donoho95a,Simoncelli96c,chang2000adaptive,portilla2003image, elad2006image, HelOr2008,bm3d}. Denoising CNNs are typically trained in a supervised fashion, minimizing mean squared error between (MSE) over a large database of example ground-truth images and their noisy counterparts~\cite{dncnn, biasfree, tnrd}. Unsupervised denoising is an alternative approach, which does not rely on ground-truth clean images. There are two main strategies to achieve this: blind-spot methods~\cite{n2v, blindspotnet, noise2self, noise2same} and techniques based on Stein's unbiased risk estimator (SURE)~\cite{Donoho95a,lusier2007SURE,raphan2008optimal,surebaranuik, surekoreanarxiv, surekoeanneurips} (see Section~\ref{sec:unsup_loss} for a more detailed description).

\textbf{Generalization to out-of-distribution images}. In order to adapt CNNs to operate on test data with characteristics differing from the training set, recent publications propose fine-tuning the networks using an additional training dataset that is more aligned with the test data~\cite{lidia,petfinetuning}. This is a form of transfer learning, a popular technique in classification problems~\cite{donahue2014decaf, yosinski2014transferable}. However, it is often challenging to obtain relevant additional training data. %
Here, we show that GainTuning can adapt CNN denoisers using only the test image.

\textbf{Connection to feature normalization in CNNs}. Feature normalization techniques such as batch normalization (BN)~\cite{batchnorm} are a standard component of deep CNNs. BN consists of two stages: (1) centering and normalizing the features corresponding to each channel, %
(2) scaling and shifting the normalized features using two learned parameters per channel (a scaling factor and a shift). The scaling parameter is analogous to the gain parameter introduced in GainTuning. However, in BN this parameter is fixed during test time, whereas \gt\ explicitly adjusts it based on the test data. 

\textbf{Gain normalization}. Motivated by normalization observed in biological sensory neurons~\cite{carandini2012normalization}, adaptive normalization of channel gains has been previously applied in object recognition~\cite{jarrett2009best}, density estimation~\cite{balle2015density}, and compression~\cite{balle2016end}. In contrast to these approaches, which adjust gains based on channel responses, \gt\ adjusts them to optimize an unsupervised cost function.

\textbf{Adapting CNN denoisers to test data}. To the best of our knowledge, adaptation of CNN denoisers to test data was pioneered by~\cite{surekoeanneurips,lidia}. Ref.~\cite{surekoeanneurips} proposes to %
include the noisy test images in the training set. In a recent extension, the authors propose %
fine-tuning all the parameters in a pre-trained CNN on a single test image using the SURE cost function~\cite{surekoreanarxiv}. %
Ref.~\cite{lidia} proposes to %
do the same %
using a novel cost function based on noise resampling (see Section~\ref{sec:unsup_loss} for a detailed description). As shown in Section~\ref{sec:overfitting} fine-tuning all the parameters in a denoiser using only a single test image can lead to overfitting. Ref.~\cite{surekoreanarxiv} avoids this using early stopping, selecting the number of fine-tuning steps beforehand. Ref.~\cite{lidia} uses a specialized architecture with a reduced number of parameters. Here, we propose a  framework for adaptive denoising that generalizes these previous methods. We show that several unsupervised cost functions can be used to perform adaptation without overfitting, as long as we only optimize a subset of the model parameters (the gain of each channel).

\textbf{Adjustment of channel parameters to improve generalization in other tasks}. Adjustment of channel parameters, such as gains and biases, has been shown to improve generalization in multiple machine-learning tasks, such as the vision-language problems~\cite{perez2018film, conditionalbn}, image generation~\cite{chen2018self}, style transfer~\cite{ghiasi2017exploring}, and image restoration~\cite{he2019modulating}. In these methods, the adjustment is carried out while training the model by minimizing a supervised cost function. In image classification, recent studies have proposed performing adaptive normalization~\cite{belikewater, nado2020evaluating, schneider2020improving} and optimization~\cite{tent} of channel parameters during test time, in the same spirit as GainTuning. %

\section{Proposed Methodology: GainTuning}

In this section we describe the \gt~framework. %
Let $f_{\theta}$ be a CNN denoiser parameterized by weight and bias parameters, $\theta$. We assume that we have available a training database and a test image \yt~ that we aim to denoise. First, the networks parameters are optimized on the training database 
\begin{align}
        \theta_{\text{pre-trained}} &= \argmin_{\theta} \sum_{y \in \text{training database}} \mathcal{L}_{\text{pre-training}}(y, f_{\theta}(y)).
    \end{align}
The cost function $\mathcal{L}_{\text{pre-training}}$ used for pre-training can be supervised, if the database contains clean and noisy examples, or unsupervised, if it only contains noisy data.%

A natural approach to adapt the pre-trained CNN to the test data is to finetune all the parameters in the CNN, %
as is done in all prior works on test-time adaptation~\cite{lidia, surekoreanarxiv,petfinetuning}. Unfortunately this can lead to \emph{overfitting} the test data (see Section~\ref{sec:overfitting}). Due to the large number of degrees of freedom, the model is able to minimize the unsupervised cost function without denoising the noisy test data effectively. This can be avoided to some extent by employing CNN architectures with a small number of parameters~\cite{lidia}, or by only optimizing for a short time (``early stopping'')~\cite{surekoreanarxiv}. Unfortunately, using a CNN with reduced parameters can limit  performance (see Section~\ref{sec:experiments}), and it is unclear how to choose a single criterion for early stopping that can operate correctly for each individual test image. Here, we propose a different strategy: tuning a single parameter (the \emph{gain}) in each channel of the CNN. \gt~ can be applied to any pre-trained CNN. %
 
We denote the gain parameter of a channel in a particular layer by $\gamma [\text{layer},\text{channel}]$, and the parameters of that channel by $\theta_{\text{pre-trained}}[\scriptsize{\text{layer}},\scriptsize{\text{channel}}]$ (this is a vector that contains the weight and bias filter parameters). We define the GainTuning parameters as
\begin{align}
    \theta_{\text{GainTuning}}(\gamma)[\scriptsize{\text{layer}},\scriptsize{\text{channel}}] & = \gamma [\scriptsize{\text{layer}},\scriptsize{\text{channel}}] \;  \theta_{\text{pre-trained}}[\scriptsize{\text{layer}},\scriptsize{\text{channel}}] .
    \end{align}
We estimate the gains by minimizing an unsupervised loss that only depends on the noisy image:
\begin{align}
        \hat{\gamma} 
        &= \argmin_{\gamma} \quad  %
        \mathcal{L}_{\text{\gt}}( \mathbf{y}_{\text{test}}, \theta_{\text{GainTuning}}(\gamma))
    \end{align}
The final denoised image is $f_{\theta_{\text{GainTuning}}(\hat{\gamma})}(\mathbf{y}_{\text{test}})$. Section~\ref{sec:unsup_loss} describes several possible choices for the cost function $\mathcal{L}_{\text{\gt}}$. Since we use only one scalar parameter per channel, the adjustment performed by \gt~ is very low-dimensional ($\approx 0.1\%$ of the dimensionality of $\theta$). This makes optimization quite efficient, and prevents overfitting (see Section~\ref{sec:overfitting}).

\section{Cost Functions for \gt\ }
\label{sec:unsup_loss}
A critical element of \gt~ is the use of an unsupervised cost function, which is minimized in order to adapt the pre-trained CNN to the test data. %
Here, we describe three different choices,
each of which are effective when integrated in the \gt~ framework, but which have different benefits and limitations.

\textbf{Blind-spot loss}. This loss measures the ability of the denoiser to reproduce the noisy observation, while excluding the identity solution. To achieve this, the CNN must estimate the $j$th pixel $y_{j}$ of the noisy image $y$ as a function of the other pixels $y_{\{j\}^c}$, but \emph{excluding the pixel itself}. As long as the noise degrades pixels \emph{independently}, this will force the network to learn a nontrivial denoising function that exploits the relationships between pixels that are present in the underlying clean image(s). The resulting loss can be written as
\begin{equation} 
\label{eq:blind-spot}
    \mathcal{L}_{\text{blind-spot}}(\mathbf{y}, \theta) = \mathbb{E}\left[ ( f_\theta( \mathbf{y}_{\{j\}^c} )_j - \mathbf{y}_j )^2 \right].
\end{equation}
Here the expectation is over the data distribution and the selected pixel. We have slightly abused notation to indicate that the network $f_\theta$ does not use $y_{j}$. This ``blind spot'' can be enforced through architecture design~\cite{blindspotnet}, or by masking~\cite{noise2self,n2v} (see also~\cite{self2self} and~\cite{noise2same} for related approaches). The blind-spot loss has a key property that makes it very powerful in practical applications: it makes no assumption about the noise distribution (other than pixel-wise independence). When combined with \gt\ it achieves effective denoising of real electron-microscope data at very low SNRs (see Figure~\ref{fig:nano} and Section~\ref{sec:exp_nano},~\ref{sec:suppl_nano}). %

\textbf{Stein's Unbiased Risk Estimator (SURE)}. Let $\mathbf{x}$ be an $N$-dimensional ground-truth random vector $\mathbf{x}$ and let $\mathbf{y} := \mathbf{x} + \mathbf{n}$ be a corresponding noisy observation, where $\mathbf{n} \sim \mathcal{N}(0, \sigma_n^2\mathbf{I})$. SURE provides an expression for the MSE between $\mathbf{x}$ and a denoised estimate $f_{\theta}(\mathbf{y})$, which \emph{only depends on the noisy observation $\mathbf{y}$}: 
\begin{equation}
\label{eq:sure}
\mathbb{E}\left[\frac{1}{N}\left\|\mathbf{x}-f_{\theta}(\mathbf{y})\right\|^{2}\right]
=\mathbb{E}\left[\frac{1}{N}\left\|\mathbf{y}-f_{\theta}(\mathbf{y})\right\|^{2}-\sigma^{2}+\frac{2 \sigma^{2}}{N} \sum_{k=1}^{N} \frac{\partial (f_{\theta}(\mathbf{y})_k)}{\partial \mathbf{y}_{k}}\right] 
:= \mathcal{L}_{\text{SURE}}(\mathbf{y}, \theta) .
\end{equation}
The last term in Equation~\ref{eq:sure} is the divergence of $f_{\theta}$, which %
can be approximated using Monte Carlo techniques~\cite{sureapprox} (Section~\ref{sec:sureapprox}). The divergence is the sum of the partial derivatives of each denoised pixel with respect to the corresponding input pixel. Intuitively, penalizing it forces the denoiser to not rely as heavily on the $j$th noisy pixel to estimate the $j$th clean pixel. This is similar to the blind-spot strategy, with the added benefit that the $j$th noisy pixel is not ignored completely. 
To further illustrate this connection, consider a linear convolutional denoising function $f_{\theta}(\mathbf{y}) = \mathbf{\theta} \circledast \mathbf{y}$, where the center-indexed parameter vector is $\mathbf{\theta} = [\theta_{-k}, \theta_{-k+1}, \dots, \theta_{0}, \dots, \theta_{k-1}, \theta_{k}]$. The SURE cost function (Equation~\ref{eq:sure}) reduces to
\begin{equation}
\label{eq:linear_sure}
\mathbb{E}_{\mathbf{n}}\left[\frac{1}{N}\left\|\mathbf{y}- \mathbf{\theta} \circledast \mathbf{y}\right\|^{2}\right]-\sigma^{2}+ 2 \sigma^{2} \theta_{0} 
\end{equation}
The SURE loss equals the MSE between the denoised output and the noisy image, with a penalty on the ``self'' pixel. As this penalty is increased, the self pixel will be ignored, so the loss tends towards the blind-spot cost function. When integrated in the \gt~ framework, the SURE loss achieves effective denoising in the case of additive Gaussian noise, outperforming the blind-spot loss. %

\textbf{Noise Resampling}. Ref.~\cite{lidia} introduced a novel procedure for adaptation which we call \emph{noise resampling}.  Given a pre-trained denoiser $f_{\theta}$ and a test image $\mathbf{y}$, first we obtain an initial denoised image by applying $f_{\theta}$ to $\mathbf{y}$, $\mathbf{\hat{x}} := f_{\theta_{\text{pre-trained}}}(\mathbf{y})$.
Then we artificially corrupt $\mathbf{\hat{x}}$ by simulating noise from the same distribution as the data of interest to create synthetic noisy examples. Finally, the denoiser is fine-tuned by minimizing the MSE between $\mathbf{\hat{x}}$ and the synthetic examples. If we assume additive noise, the resulting loss is of the form
\begin{align}
\label{eq:noiseresample}
\mathcal{L}_{\text{noise resampling}}(\mathbf{y}, \theta) &= \mathbb{E}_n\left[ \| (f_\theta(\mathbf{\hat{x}} + \mathbf{n}) - \mathbf{\hat{x}} \|^2 \right].
\end{align}
Noise resampling is reminiscent of Refs.~\cite{noisier2noise,noisyasclean}, which add noise to an already noisy image. When integrated in the \gt~ framework, the noise-resampling loss results in effective denoising in the case of additive Gaussian noise, although it tends to underperform the SURE loss.

\section{Experiments and Results}
\label{sec:experiments}

In order to evaluate the performance of \gt~ we performed three different types of experiment: \textbf{In-distribution} (test examples held out from the training set); \textbf{out-of-distribution noise} (noise level or distribution differs between training and test examples); and \textbf{out-of-distribution signal} (clean images drawn from a different set than the training set). We also apply \gt~ to \textbf{real data} %
from a transmission electron microscope.  

Our experiments make use of four \textbf{datasets}: The BSD400 natural image database~\cite{bsd400} with test sets Set12 and Set68~\cite{dncnn}, the Urban100 images of urban environments~\cite{urban100}, the IUPR dataset of scanned documents~\cite{iupr}, and a set of synthetic piecewise constant  images~\cite{lee2001occlusion}~(see Section~\ref{sec:datasets}). 
We demonstrate the broad applicability of \gt\  by using it in conjunction with multiple \textbf{architectures for image denoising}: DnCNN~\cite{dncnn}, BFCNN~\cite{biasfree}, UNet~\cite{unet} and Blind-spot net~\cite{blindspotnet} (see Section~\ref{sec:architectures}). 
Finally, we compare our results to several \textbf{baselines}: (1) models trained on the training database, (3) CNN models adapted by fine-tuning all parameters (as opposed to just the gains), (3) a model trained only on the test data, (4) LIDIA, a specialized architecture and adaptation strategy proposed in~\cite{lidia}. We provide details on training and optimization in Section~\ref{sec:training}.

\begin{figure}[t]
    \centering
    \begin{tabular}{c@{\hskip 0.02in}c}
    
    \scriptsize{
    \begin{tabular}{ccccccccc}
        \toprule
        
        &
        \multirow{2}{*}{Model } & 
        \multicolumn{1}{c}{\phantom} &
        \multicolumn{3}{c}{Set12}    &
        \multicolumn{3}{c}{BSD68}    \\
        
        \cmidrule(lr){4-6}
        \cmidrule(lr){7-9}

        & & &
        \multicolumn{1}{c}{$\sigma=30$} &
        \multicolumn{1}{c}{$40$} &
        \multicolumn{1}{c}{$50$} &
        \multicolumn{1}{c}{$30$} &
        \multicolumn{1}{c}{$40$} &
        \multicolumn{1}{c}{$50$} \\

        \midrule
        \multirow{4}{*}{\rotatebox[origin=c]{90}{\gt}} & \multirow{2}{*}{DnCNN} 
        & \scriptsize{Pre-trained} & 29.52 &  28.21 &  27.19 &   28.39  & 27.16 & 26.27  \\
         & & \scriptsize{\gt\ } &  \textbf{29.62} &  \textbf{28.30} &  \textbf{27.29} &   \textbf{28.47}  & \textbf{27.23} & \textbf{26.33} \\
        \cmidrule(lr){2-9}
        & \multirow{2}{*}{UNet} 
        & \scriptsize{Pre-trained} & 29.34 &  28.05 &  27.05 &   28.27  & 27.05 & 26.15  \\
        &  & \scriptsize{\gt\ } &  29.46 &  28.15 &  27.13 &   28.34 & 27.12 & 26.22 \\
        \midrule
        \multirow{3}{*}{\rotatebox[origin=c]{90}{Baseline}} & \multirow{2}{*}{LIDIA} 
        & \scriptsize{Pre-trained} & 29.46 &  27.95 &  26.58 &   28.24  & 26.91 & 25.74  \\
        & & \scriptsize{Adapted} &  29.50 &  28.10 &  26.95 &   28.23 & 26.97 & 26.02 \\
        \cmidrule(lr){2-9}
        & Self2Self & & 29.21 & 27.80 & 26.58 & 27.83 & 26.67 & 25.73 \\
        \bottomrule
    \end{tabular}
    }
    
    &

    \begin{tabular}{c}
    \scriptsize{(Top) Set12; (Bottom) BSD68}\\ 
    \includegraphics[height=0.75in]{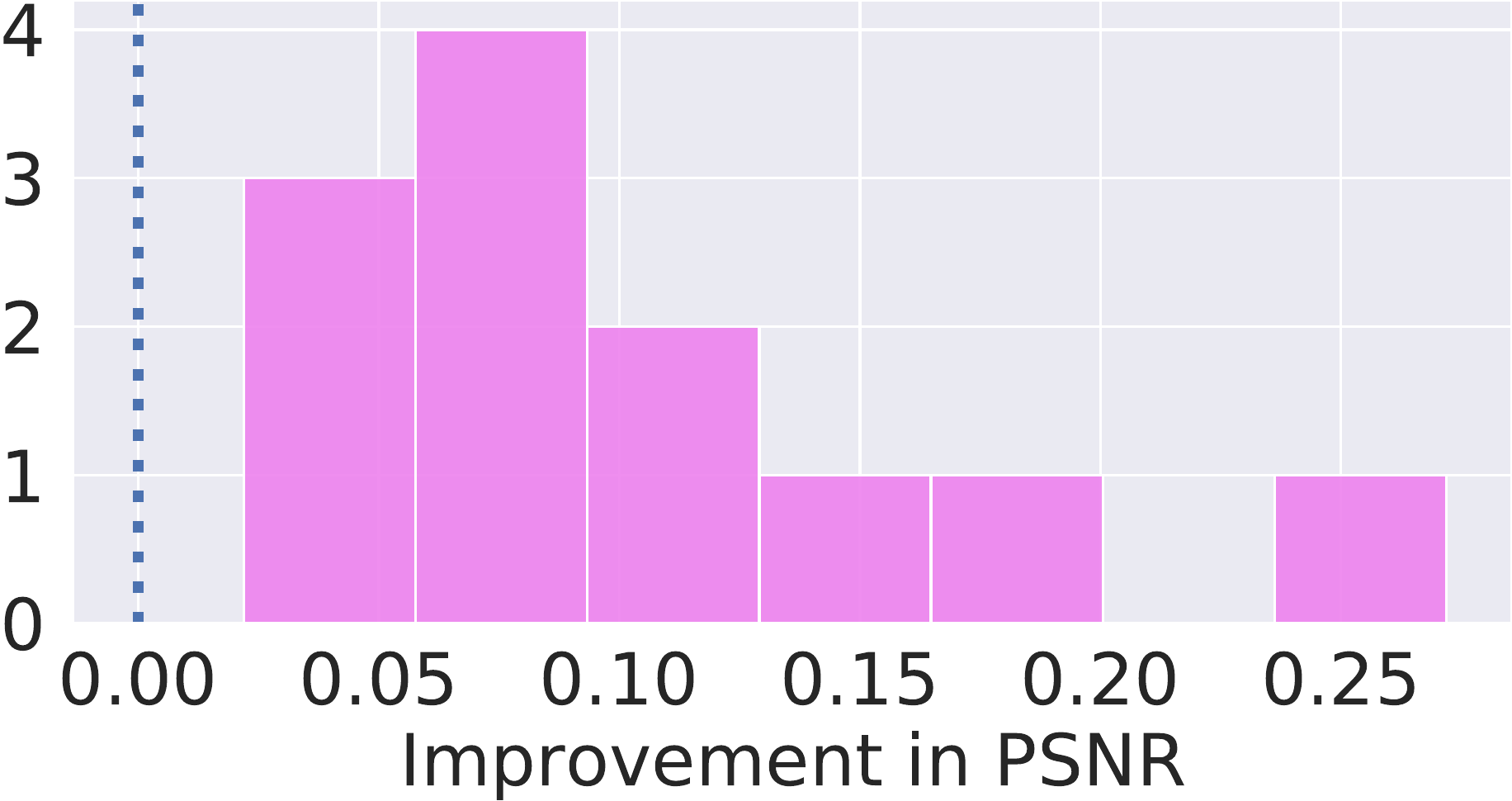} \\
    \includegraphics[height=0.75in]{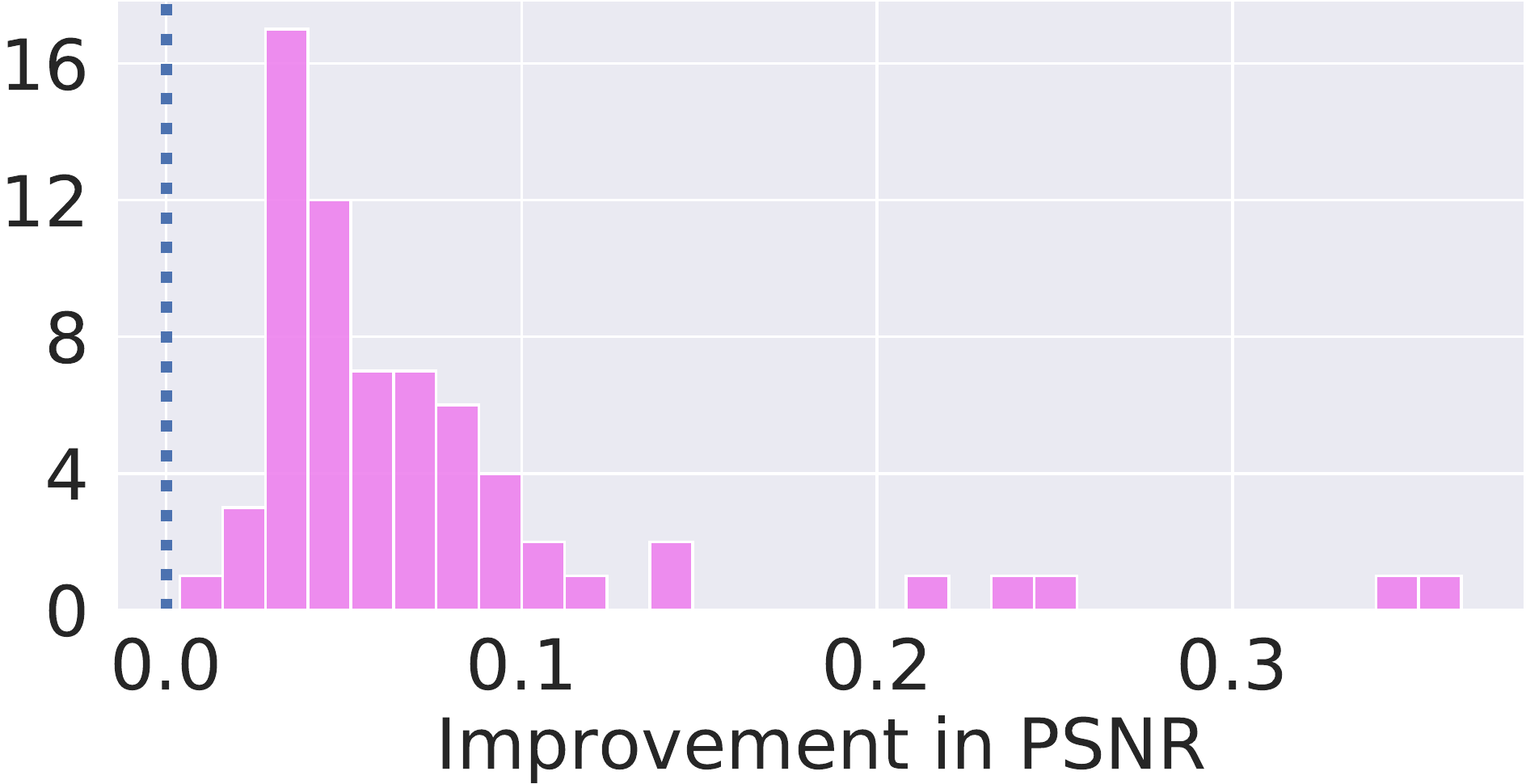}\\
    \end{tabular}
    
    \end{tabular}
    
    \caption{\textbf{\gt\ achieves state-of-the-art performance.} (Left) The average PSNR on two test set of generic natural images improves after \gt\ for different architectures across multiple noise levels. The CNNs are trained on generic natural images (BSD400). (Right) Histograms showing improvement in performance for each image in each of the two test sets at $\sigma=30$.}
    \label{fig:sota}
\end{figure}

\subsection{\gt\ surpasses state-of-the-art performance for in-distribution data}
\label{sec:exp_in_distr}

\textbf{Experimental set-up}. %
We use BSD400, a standard natural-image benchmark, corrupted with Gaussian white noise with standard deviation $\sigma$ sampled uniformly from $[0, 55]$ (relative to pixel intensity range $[0, 255]$). Following~\cite{dncnn}, we evaluate performance on two independent test sets: Set12 and BSD68, corrupted with Gaussian noise with $\sigma \in \{30, 40, 50\}$.

\textbf{Comparison to pre-trained CNNs}. \gt~consistently improves the performance of pre-trained CNN models. Figure~\ref{fig:sota} shows this for two different models, DnCNN~\cite{dncnn} and UNet~\cite{unet} (see also Section~\ref{sec:suppl_in_distr}). The SURE loss outperforms the blind-spot loss, and is slightly better than noise resampling (Table~\ref{tab:lossfun_comparison}). The same holds for other architectures, as reported in Section~\ref{sec:suppl_in_distr}. On average the improvement is modest, but for some images it is quite substantial (up to 0.3 dB in PSNR for $\sigma=30$, see histogram in Figure~\ref{fig:sota}).  

\textbf{Comparison to other baselines}. \gt~outperforms fine-tuning based on optimizing all the parameters for different architectures and loss functions (see Section~\ref{sec:overfitting}). \gt~clearly outperforms a Self2Self model, which is trained exclusively on the test data (Figure~\ref{fig:sota}). It also outperforms the specialized architecture and adaptation process introduced in \cite{lidia}, with a larger gap in performance for higher noise levels.

\subsection{\gt~ generalizes to new noise distributions}
\label{sec:exp_out_noise}
\textbf{Experimental set-up}. The same set-up as Section~\ref{sec:exp_in_distr} is used, except that the test sets are corrupted with Gaussian noise with $\sigma \in \{70, 80\}$ (recall that training occurs with $\sigma \in [0, 55]$). 

\begin{figure}[t]
    \centering
    \begin{tabular}{c@{\hskip 0.01in}c}
    
    \multicolumn{2}{c}{\footnotesize{\textbf{Out-of-distribution test noise}}} \\
    
    \scriptsize{
    \begin{tabular}{cccccc}
        \toprule
        \multirow{2}{*}{Test set} & 
        \multirow{2}{*}{$\sigma$} & 
        \multicolumn{2}{c}{Trained on $\sigma \in [0, 55]$}    &
        \multirow{2}{*}{\thead{\scriptsize{Bias Free} \\ \scriptsize{Model~\cite{biasfree}} } } & 
        \multirow{2}{*}{\thead{\scriptsize{Trained on} \\ \scriptsize{$\sigma \in [0, 100]$} } } \\

        \cmidrule(lr){3-4}
        
        & &
        \multicolumn{1}{c}{Pre-trained} &
        \multicolumn{1}{c}{Gaintuning} &
        & \\

        \midrule
        \multirow{2}{*}{Set12} 
        & $70$ & 22.45 & 25.54 & 25.59  & 25.50  \\
        & $80$ & 18.48 & 24.57 & 24.94 & 24.88 \\
        \midrule
        \multirow{2}{*}{BSD68} 
        & $70$ & 22.15 & 24.89 & 24.87     & 24.88   \\
        & $80$ & 18.72  & 24.14 &  24.38     &  24.36  \\
        \bottomrule
    \end{tabular}
    }&
    \begin{tabular}{c}
    \includegraphics[height=0.9in]{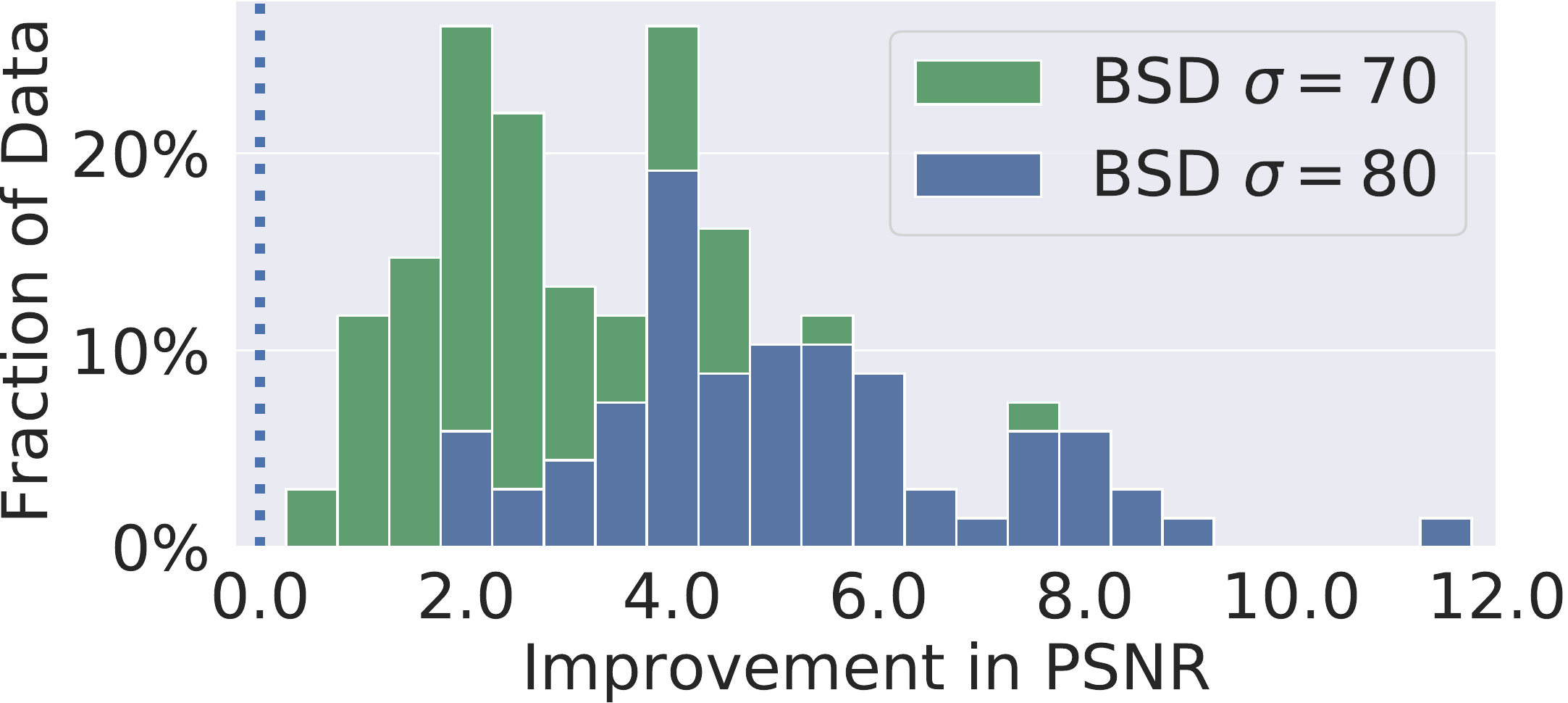}\\
    \end{tabular} \\[1em]
    
    \multicolumn{2}{c}{\footnotesize{\textbf{Out-of-distribution test image}}} \\
   { %
    \renewcommand{\arraystretch}{1.5}
   \scriptsize{
    \begin{tabular}{cccccc}
        \toprule

        & Training data &
        Test data &
        Pre-trained &
        Gaintuning \\
        \midrule

        (a) & Piecewise constant  & Natural images & 27.31 & 28.60 \\
        (b) & Natural images & Urban images & 28.35 & 28.79 \\
        (c) & Natural images &Scanned documents & 30.02 & 30.73 \\
        \bottomrule
    \end{tabular}
    }
    }&
    \begin{tabular}{c}
    \includegraphics[height=0.9in]{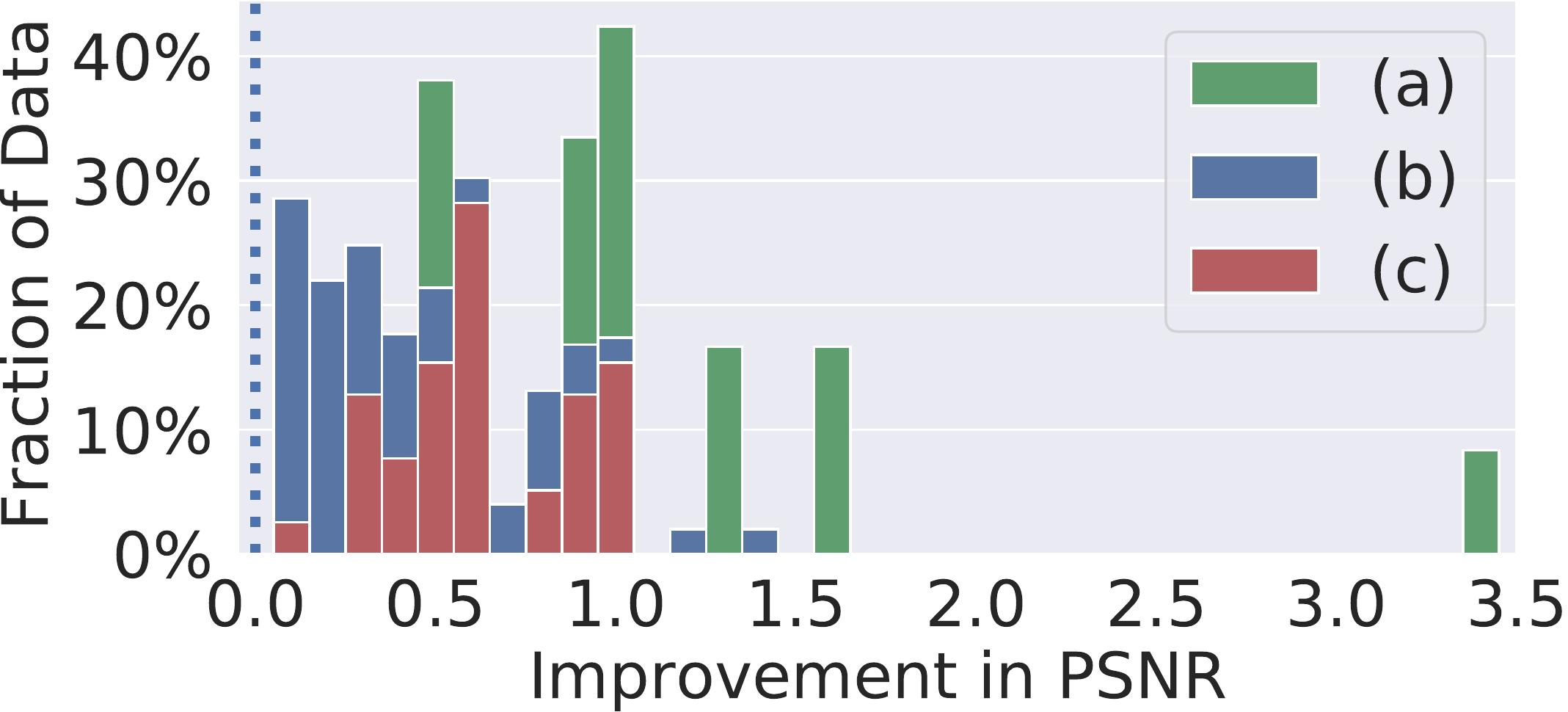}\\
    \end{tabular}
    
    \end{tabular}
    \caption{\textbf{\gt\ generalizes to out-of-distribution data.}  Average performance of a CNN trained to denoise at noise levels $\sigma \in [0, 55]$ improves significantly on test image with noise outside the training range, $\sigma=70, 80$ (top) and on images with different characteristics than training data (bottom) after \gt . Capability of \gt\ to generalize to out-of-distribution noise is comparable to that of Bias-Free CNN~\cite{biasfree}, which is an architecture explicitly designed to generalize to noise levels outside the training range, and to that of a denoiser trained with supervision at all noise levels. (Right) Histogram showing improvement in performance for each image in the test set. The improvement is substantial across most images, reaching nearly 12dB improvement in one example.}
    \label{fig:all_gen}
\end{figure}

\textbf{Comparison to pre-trained CNNs}. Pre-trained CNN denoisers fail to generalize in this setting. \gt~consistently improves their performance (see Figure~\ref{fig:all_gen} and).

The SURE loss again outperforms the blind-spot loss, and is slightly better than noise resampling (see Section~\ref{sec:suppl_out_noise}).
The same holds for other architectures, as reported in Section~\ref{sec:suppl_out_noise}. The improvement in performance for all images is substantial (up to 12 dB in PSNR for $\sigma=80$, see histogram in Figure~\ref{fig:all_gen}).  

\textbf{Comparison to other baselines}. \gt\ achieves comparable performance to a gold-standard CNN trained with supervision at all noise levels~(Figure~\ref{fig:all_gen}). 
\gt\ matches the performance of a bias-free CNN~\cite{biasfree} specifically designed to generalize to out-of-distribution noise~(Figure~\ref{fig:all_gen}).
\gt~outperforms fine-tuning based on optimizing all the parameters for different architectures and loss functions (see Section~\ref{sec:overfitting}). \gt~clearly outperforms a Self2Self model trained exclusively on the test data (Section~\ref{sec:suppl_out_noise}), and the LIDIA adaptation method~\cite{lidia}.

\subsection{\gt\ generalizes to out-of-distribution image content}
\label{sec:exp_out_signal}

\textbf{Experimental set-up}. We evaluate the performance on of \gt\ on test images that have different characteristics from the training images. We perform the following controlled experiments:
\begin{itemize}[leftmargin=*,topsep=0pt]
\item[] \textbf{(a) Simulated piecewise constant images $\rightarrow$ Natural images}. We pre-train CNN denoisers on simulated piecewise constant images. These images consists of constant regions (of different intensities values) with the boundaries having varied shapes such as circle and lines with different orientations (see Section~\ref{sec:datasets} for some examples). Piecewise constant images provide a crude model for natural images~\cite{matheron1975random, pitkow2010exact,lee2001occlusion}. We use \gt\ to adapt a CNN trained on this dataset to generic natural images (Set12). This experiment demonstrates the ability of \gt\ to adapt from a simple simulated dataset to a significantly more complex real dataset.

\item[] \textbf{(b) Generic natural images $\rightarrow$ Images with high self-similarity}. We apply \gt\ to adapt a CNN trained on generic natural images to images in Urban100 dataset. Urban100 consists of images of buildings and other structures typically found in an urban setting, which contain substantially more repeating/periodic structure (see Section~\ref{sec:datasets}) than generic natural images. 

\item[] \textbf{(c) Generic natural images $\rightarrow$ Images of scanned documents}. We apply \gt\ to adapt a CNN trained on generic natural images to images of scanned documents in IUPR dataset (see Section~\ref{sec:datasets}).
\end{itemize}
All CNNs were trained for denoising Gaussian white noise with standard deviation $\sigma \in [0, 55]$ and evaluated at $\sigma=30$.

\textbf{Comparison to pre-trained CNNs}. \gt\ consistently improves the performance of pre-trained CNNs in all the three experiments. Figure~\ref{fig:all_gen} shows this for DnCNN when \gt\ is based on SURE loss. We obtain similar results with other architectures (see Section~\ref{sec:suppl_out_image}). In experiment (a), all test images show substantial improvements, with one image improving as much as 3 dB in PNSR (at $\sigma=30$). We observe similar trends for experiments (b) and (c) as well, with improvements being better on an average for experiment (c). Note that we obtain similar performance increases when both  \emph{image and noise are out-of-distribution} as discussed in Section~\ref{sec:suppl_out_noise_and_image}. 

\textbf{Comparison to other baselines}. Experiment (a): (1)  \gt\ outperforms optimizing all the parameters over different architectures and loss functions for experiment (Section~\ref{sec:overfitting}). (2) Self2Self trained only on test data outperforms \gt in these cases, even though \gt\ improves the performance of pre-trained CNN about 1.3 dB on an average. 
This is because the test images contain content that differs substantially from the training images. 
Self2Self provides the strongest form of adaptation, since it is trained exclusively on the test image, whereas the denoising properties of \gt\ are partially due to the pretraining (see Sections~\ref{sec:limitations},~\ref{sec:suppl_out_image}). (3) We did not evaluate LIDIA~\cite{lidia} for this experiment. Experiments (b) and (c). (1) Training all parameters clearly outperforms \gt\ for case (b), but has similar performance for (c). \gt\ outperforms LIDIA on experiments (b) and (c). Self2Self trained exclusively on test data outperforms \gt (and LIDIA) on (b) and (c) (see Sections~\ref{sec:limitations},~\ref{sec:suppl_out_image}).

\subsection{Application to Electron microscopy}
\label{sec:exp_nano}
\textbf{Scientific motivation.} Transmission electron microscopy (TEM) is a popular imaging technique in materials science~\cite{Smith2015, Crozier-insitu2016}. Recent advancements in TEM enable to image at high frame rates~\cite{FARUQI2018, ercius2020}. These images can for example capture the dynamic, atomic-level rearrangements of catalytic systems~\cite{Sun2020, Guo2020, lawrence_levin_miller_crozier_2020, LEVIN2020, crozier2019dynamics}, which is critical to advance our understanding of functional materials. Acquiring image series at such high temporal resolution produces data severely degraded by shot noise.  Consequently, there is an acute need for denoising in this domain. 

\textbf{The need for adaptive denoising.} Ground-truth images are not available in TEM, because measuring at high SNR is often impossible. Prior works have addressed this by using simulated training data~\cite{mohan2020deep, vincent2021developing}, whereas others have trained CNNs directly on noisy real data~\cite{udvd}.  %

\textbf{Dataset.} We use the training set of $5583$ simulated images and the test set of 40 real TEM images from~\cite{mohan2020deep, vincent2021developing}. The data correspond to a catalytic platinum nanoparticle on a CeO$_2$ support~(Section~\ref{sec:datasets}). %

\textbf{Comparison to pre-trained CNN.} A CNN~\cite{blindspotnet} pre-trained on the simulated data fails to reconstruct the pattern of atoms faithfully (green box in Figure~\ref{fig:nano} (c),~(e)). \gt\ applied to this CNN  using the blind-spot loss correctly recovers this pattern (green box in Figure~\ref{fig:nano} (d),~(e)) reconstructing the small oxygen atoms in the CeO$_2$ support.  \gt\ with noise resampling failed to reproduce the support pattern (probably because it is absent from the initial denoised estimate) (Section~\ref{sec:suppl_nano}). %

\textbf{Comparison to other baselines.} \gt\ clearly outperforms Self2Self, which is trained exclusively on the real data. The denoised image from Self2Self shows missing atoms and substantial artefacts (see Section~\ref{sec:suppl_nano}). We also compare \gt\ dataset to blind-spot methods using the 40 test frames~\cite{blindspotnet,udvd}. \gt\ clearly outperforms these methods (see Section~\ref{sec:suppl_nano}). Finally, \gt~outperforms fine-tuning based on optimizing all the parameters, which overfits heavily (see Section~\ref{sec:overfitting}).

\def\f1ht{0.52in}%
\begin{figure}[t]
    \centering
    \begin{tabular}{c@{\hskip 0.01in}c@{\hskip 0.01in}c@{\hskip 0.01in}c}
    \scriptsize{(a) Noisy image} &  \scriptsize{(b) Trained on piecewise constant } & \scriptsize{(c) After \gt\  on (b)}& \scriptsize{(d) Difference b/w (b) and (c)}\\
    \includegraphics[width=1.33in]{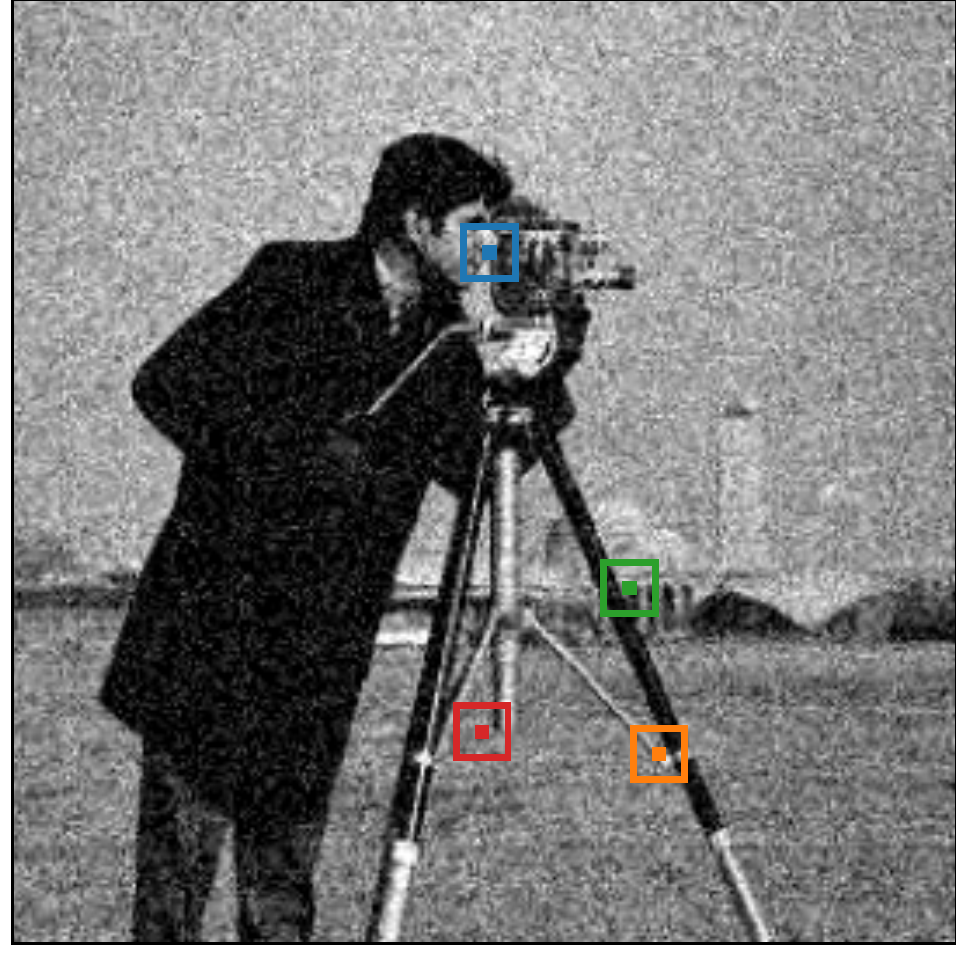}&
    \includegraphics[width=1.33in]{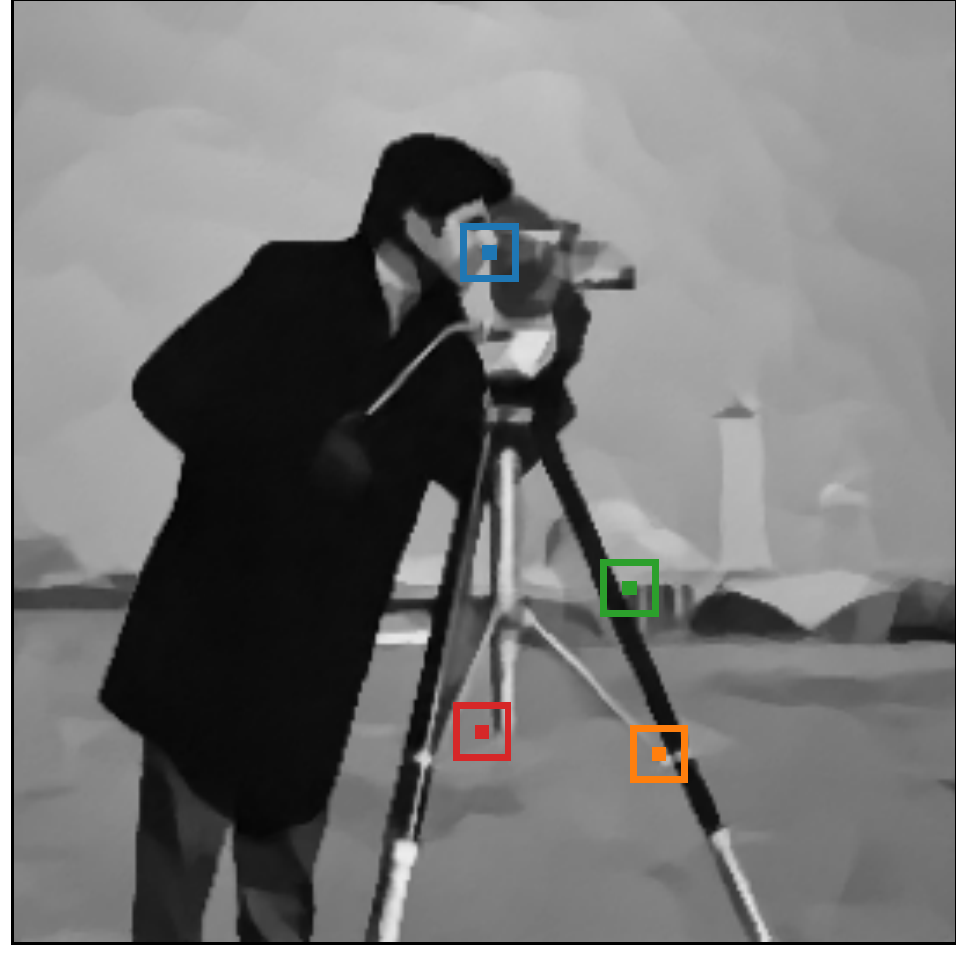}&
    \includegraphics[width=1.33in]{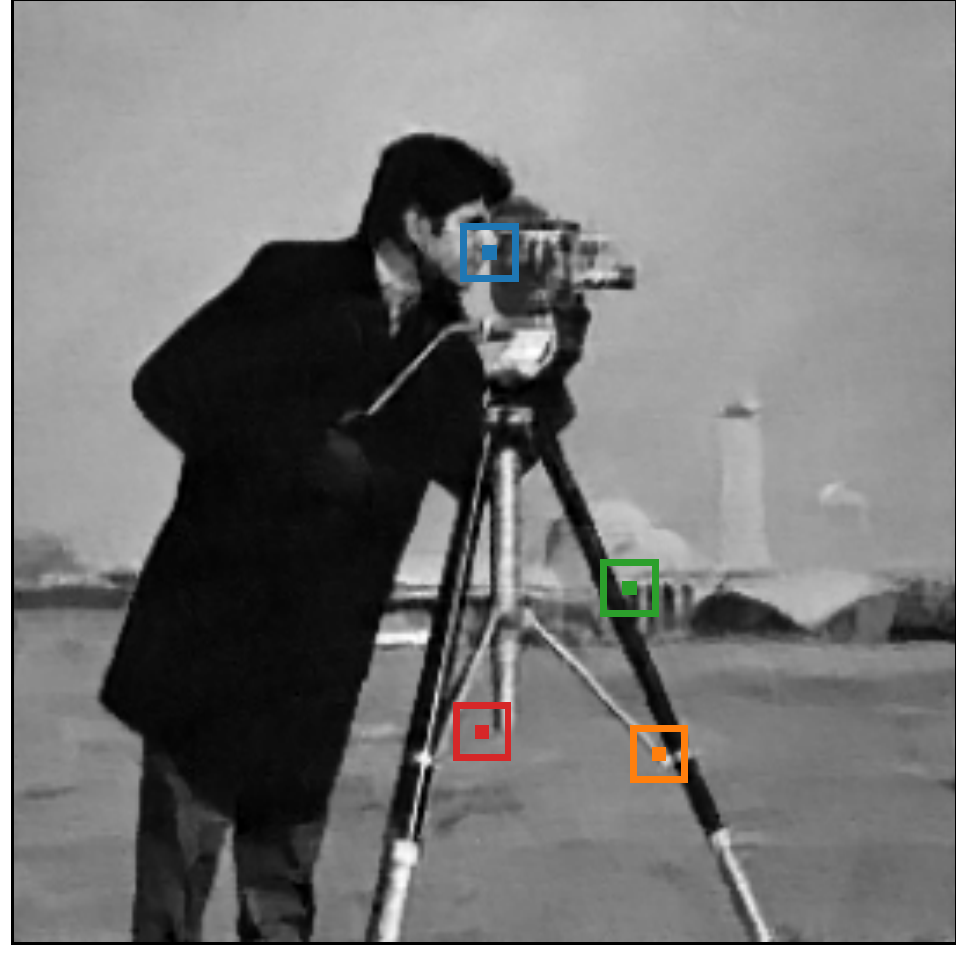}&
    \includegraphics[width=1.33in]{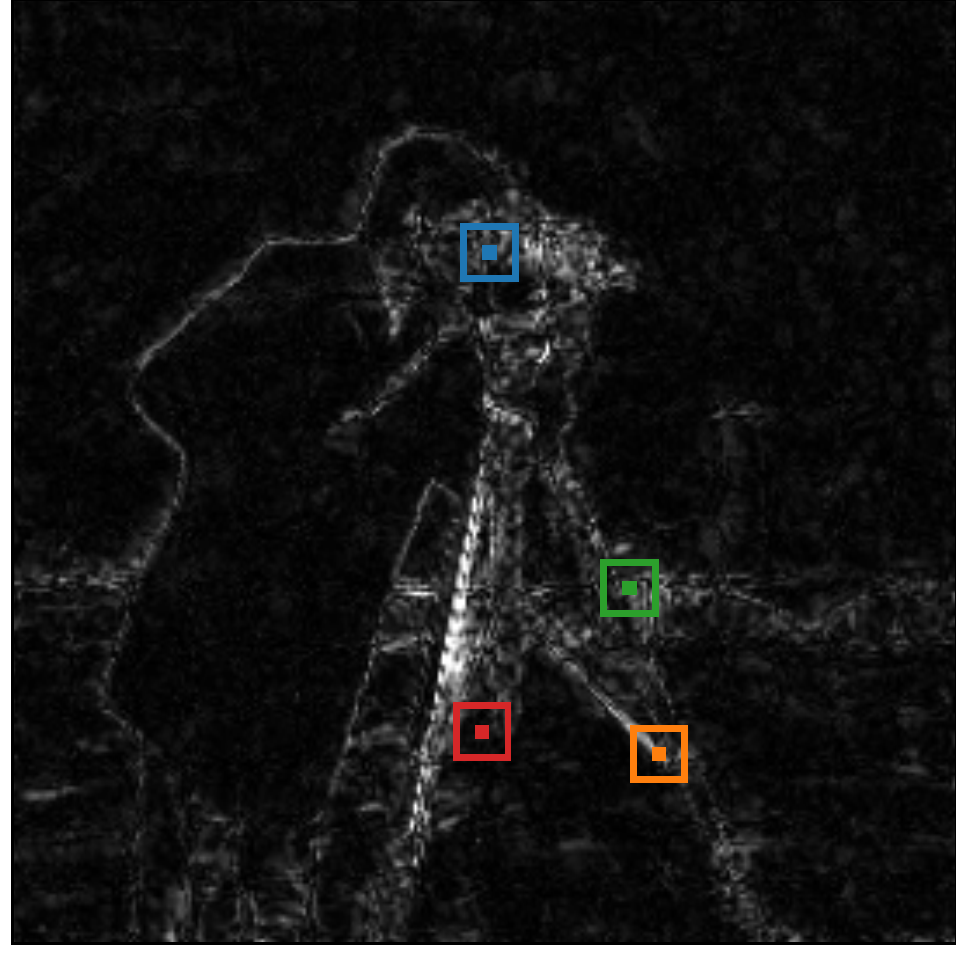}\\
    \end{tabular}

    \centering
    \begin{tabular}{c@{\hskip 0.01in}c@{\hskip 0.01in}c@{\hskip 0.01in}c@{\hskip 0.01in}c@{\hskip 0.10in}c@{\hskip 0.01in}c@{\hskip 0.01in}c@{\hskip 0.01in}c@{\hskip 0.01in}c}
   
    \includegraphics[width=\f1ht]{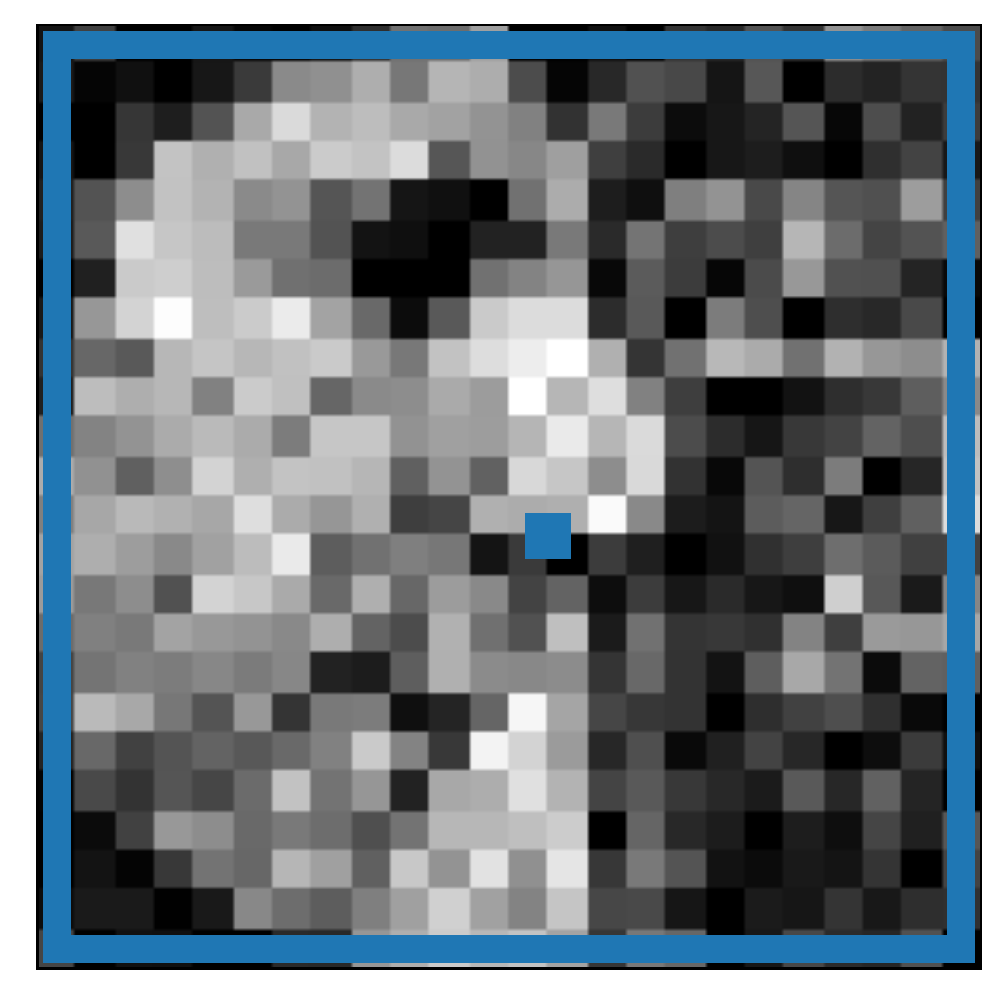}&
    \includegraphics[width=\f1ht]{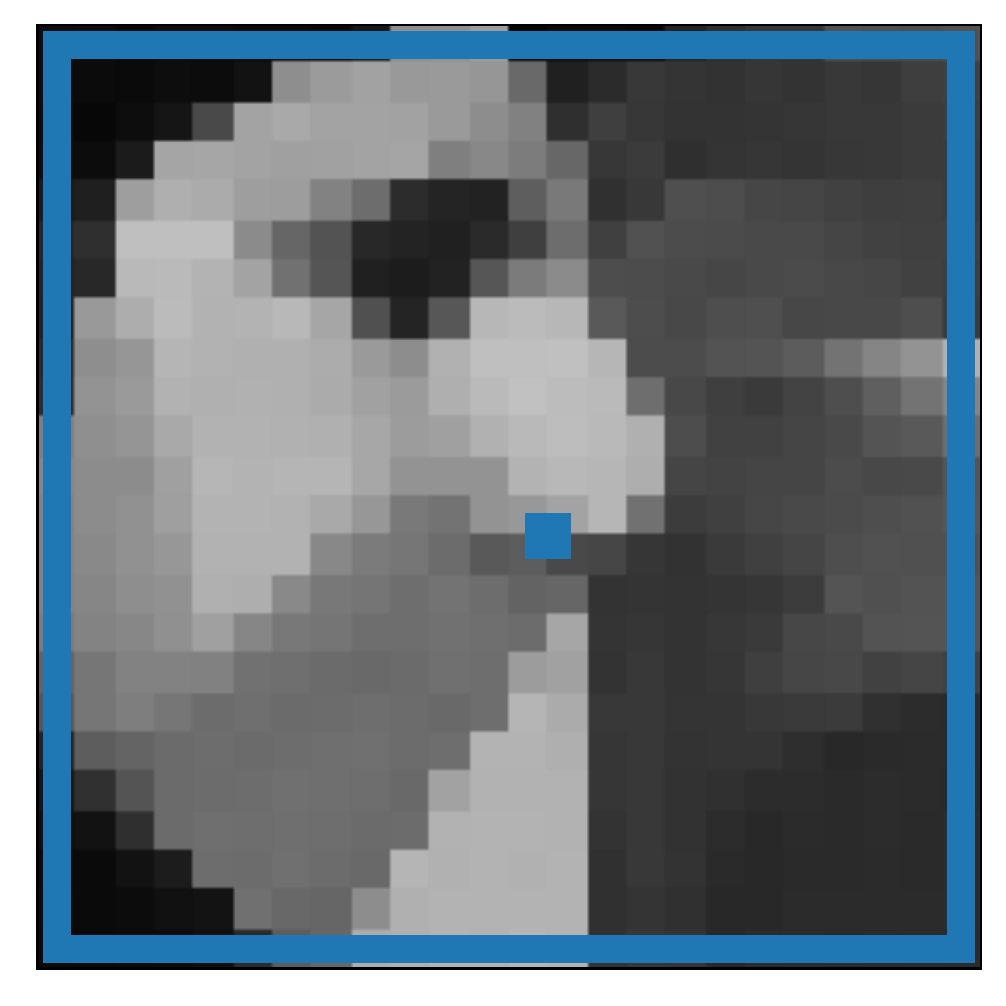}&
    \includegraphics[width=\f1ht]{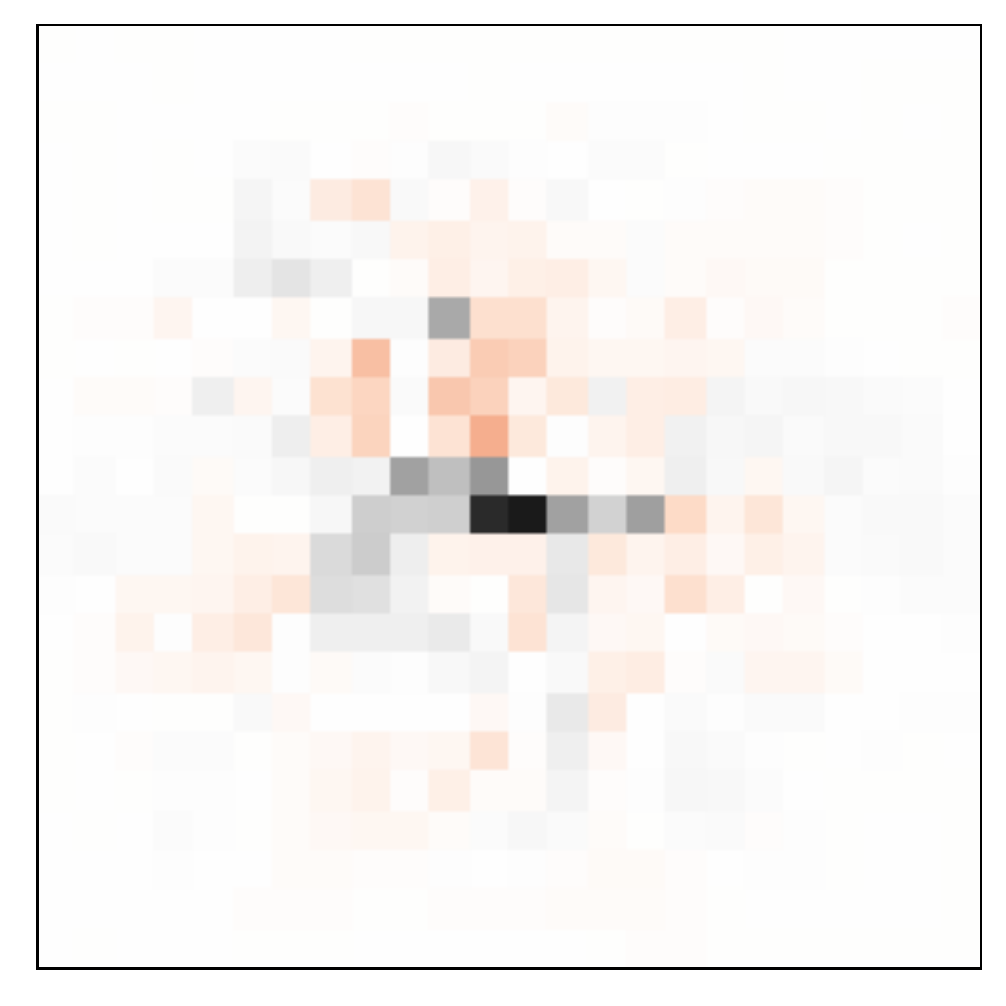}&
    \includegraphics[width=\f1ht]{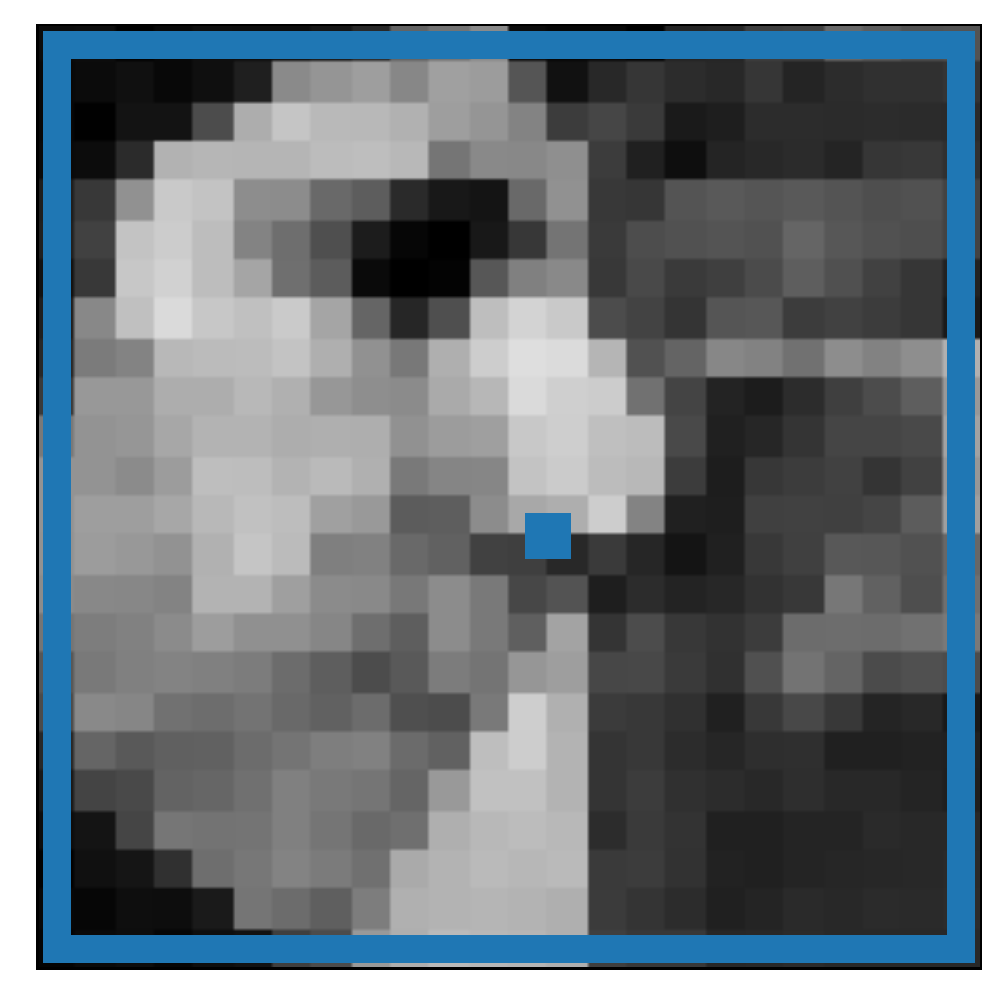}&
    \includegraphics[width=\f1ht]{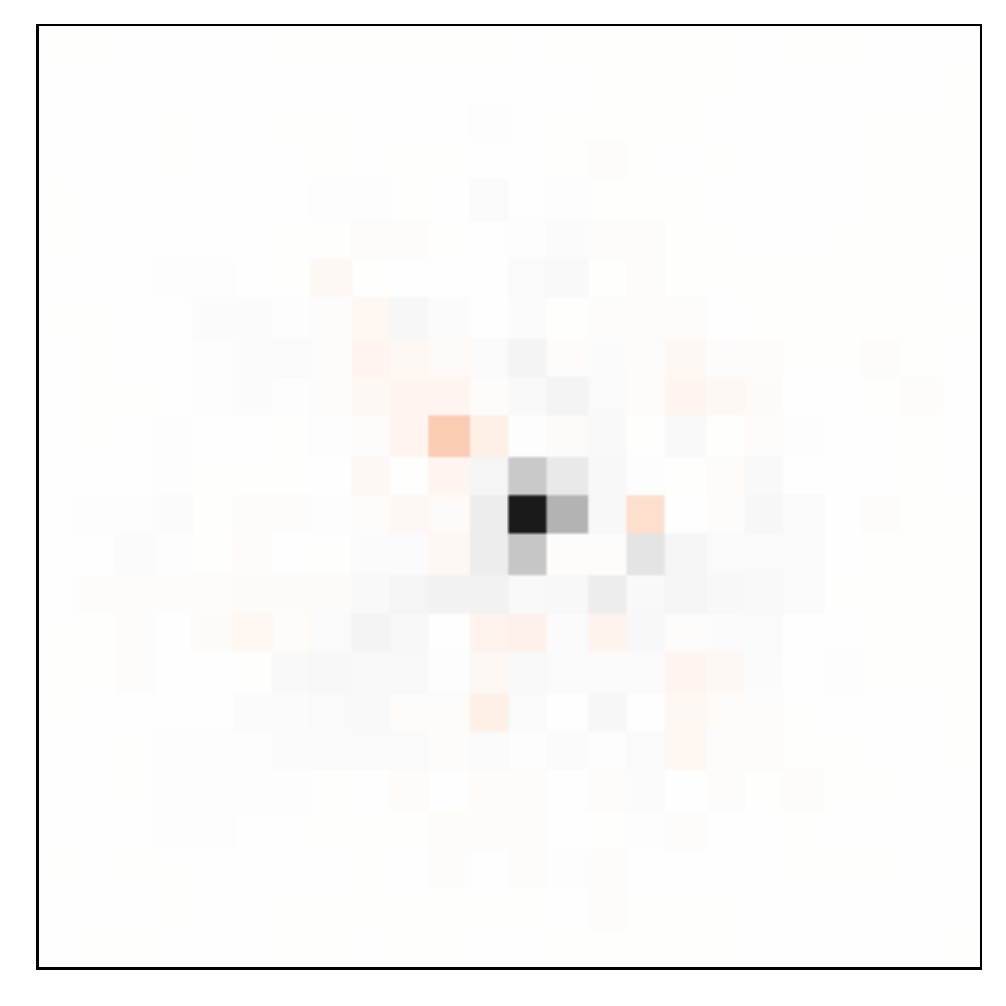}&
    \includegraphics[width=\f1ht]{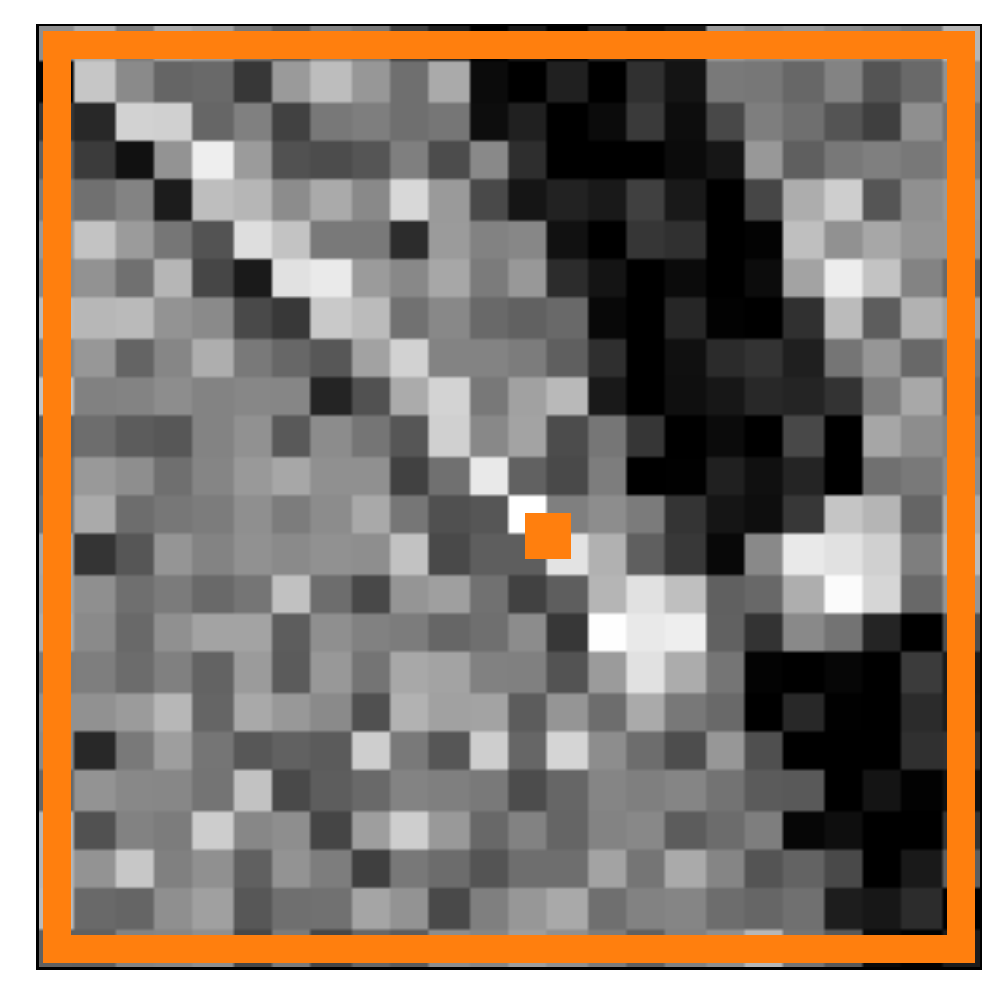}&
    \includegraphics[width=\f1ht]{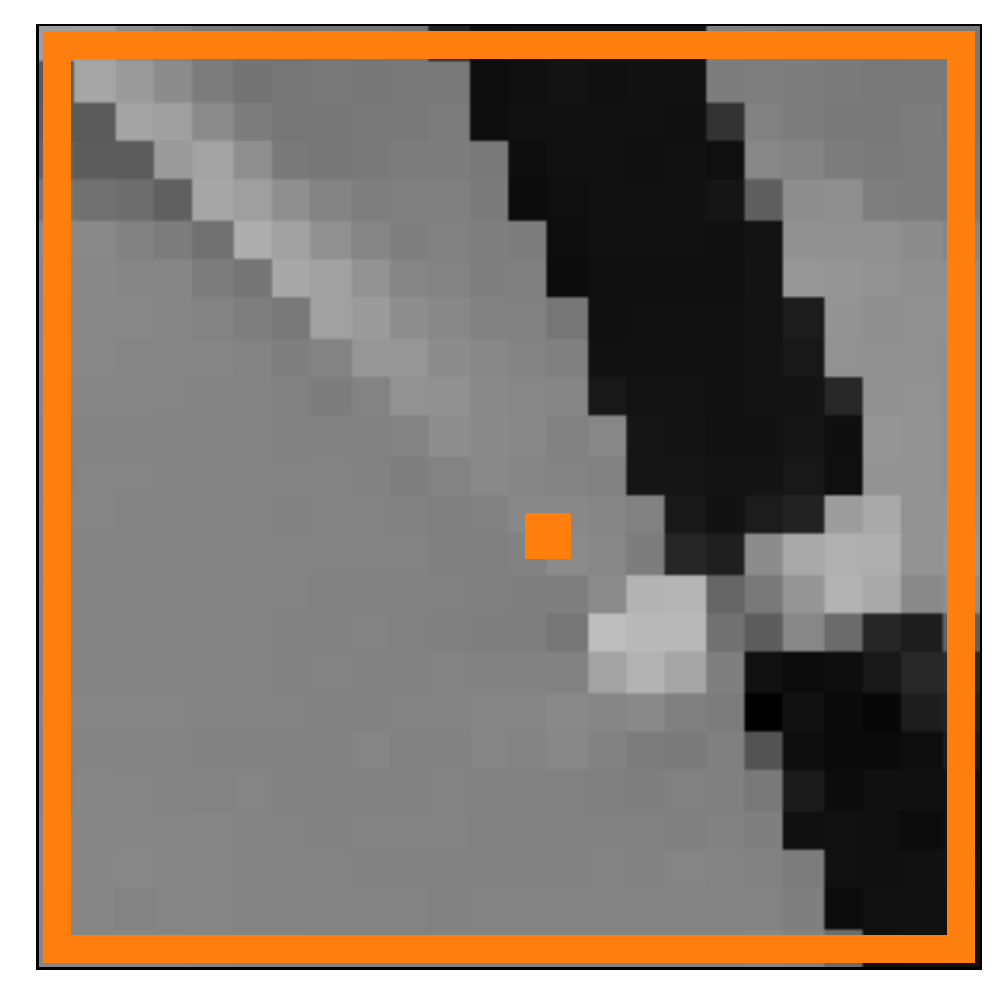}&
    \includegraphics[width=\f1ht]{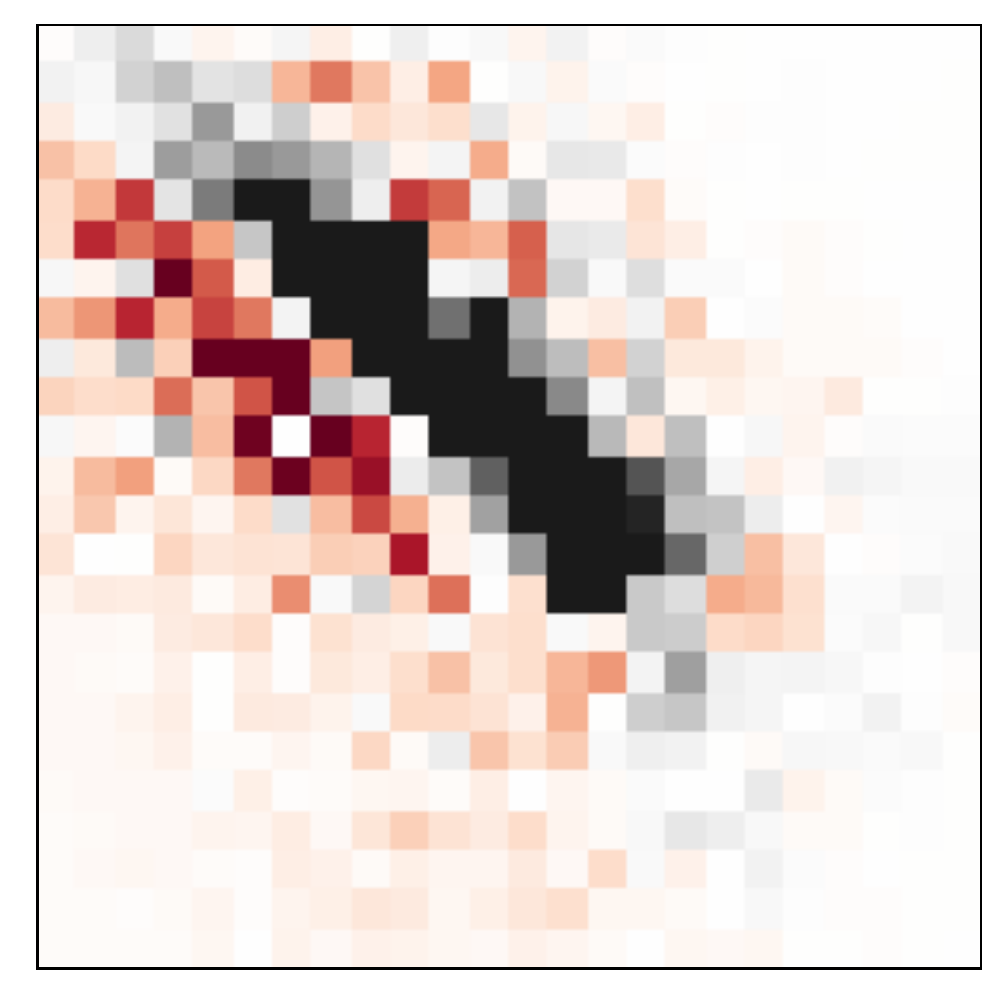}&
    \includegraphics[width=\f1ht]{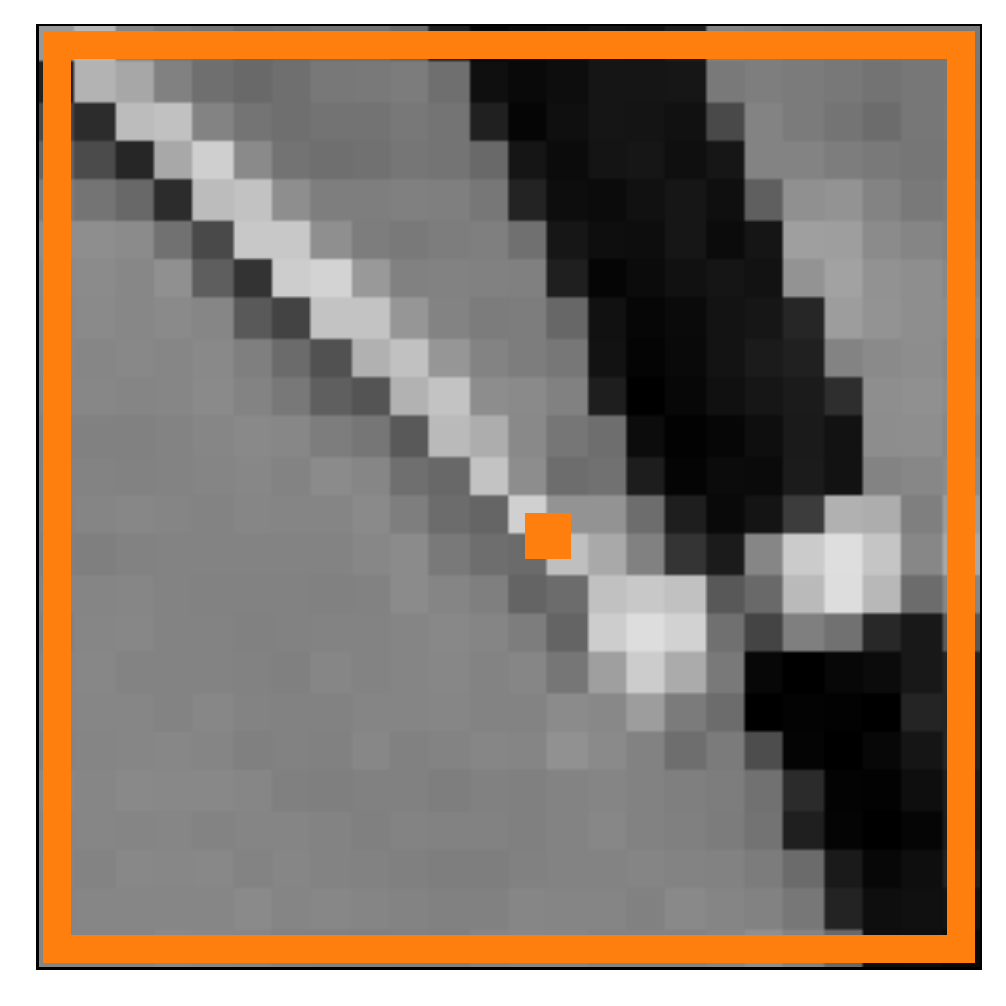}&
    \includegraphics[width=\f1ht]{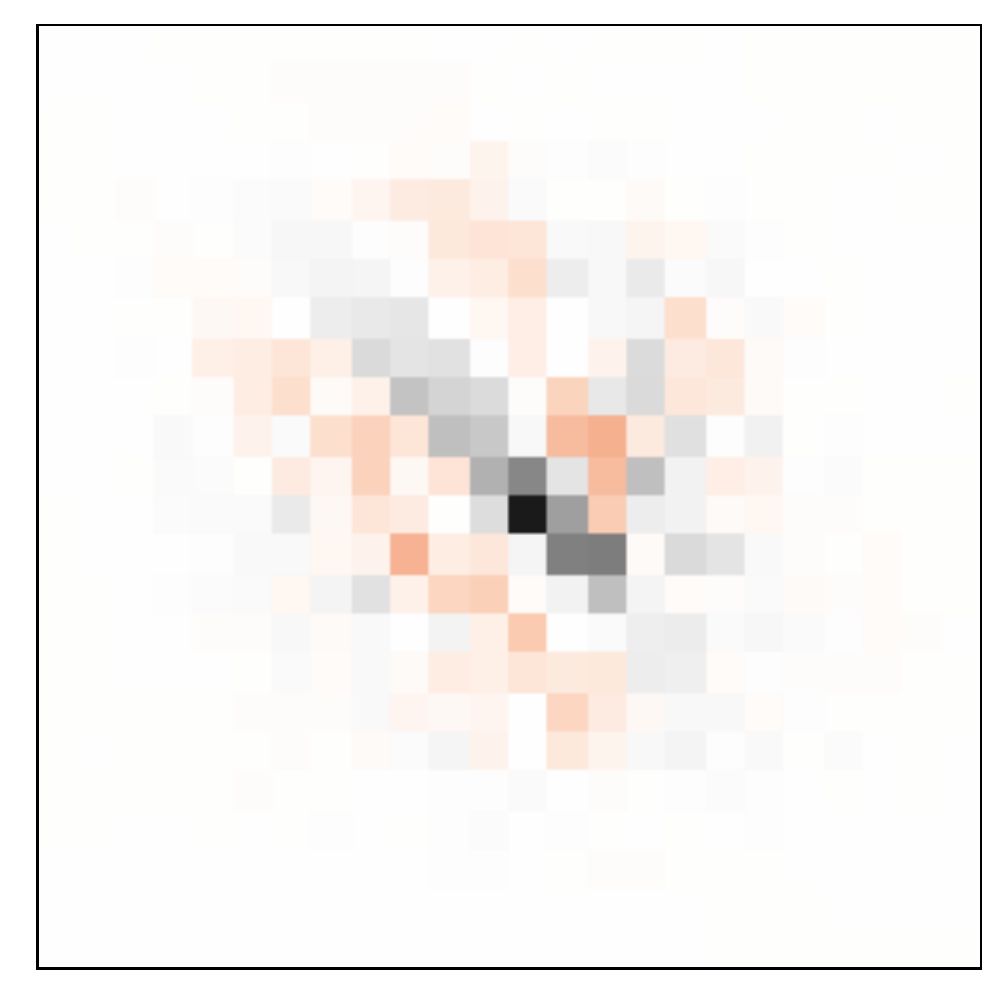}\\
    
    \includegraphics[width=\f1ht]{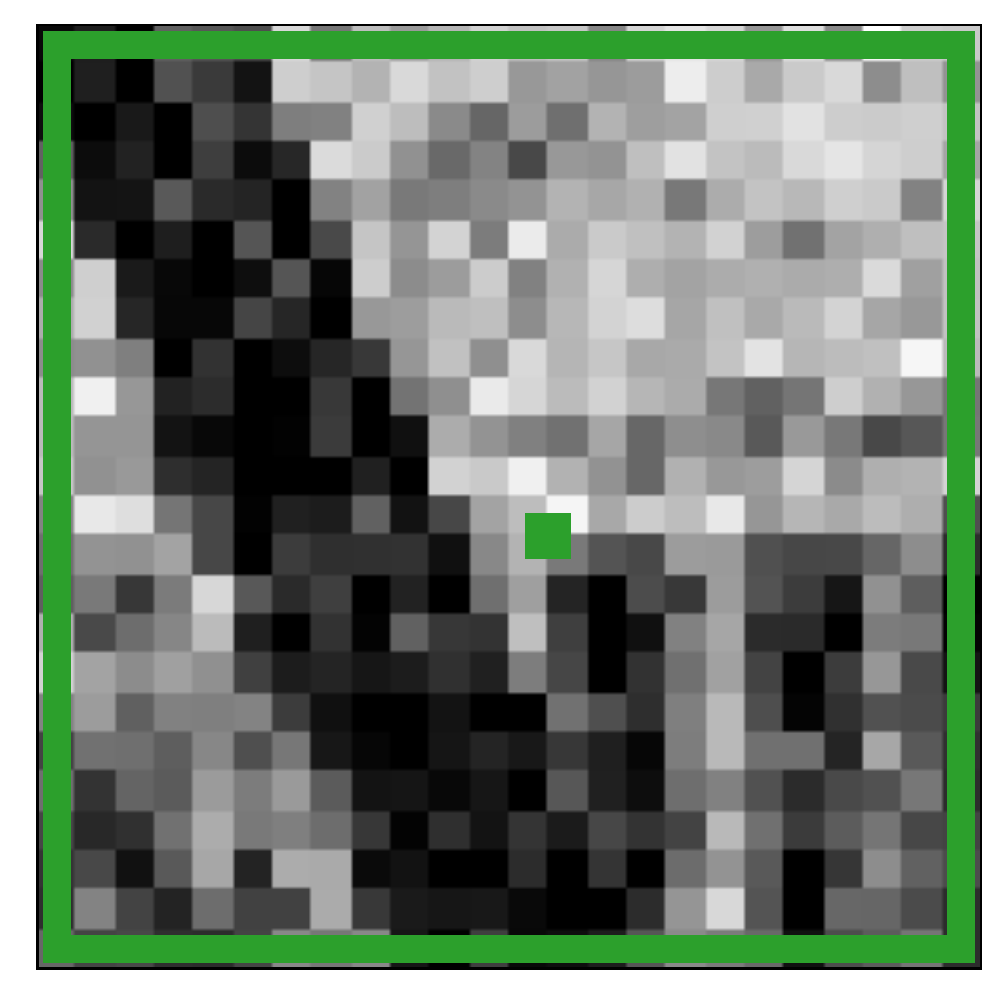}&
    \includegraphics[width=\f1ht]{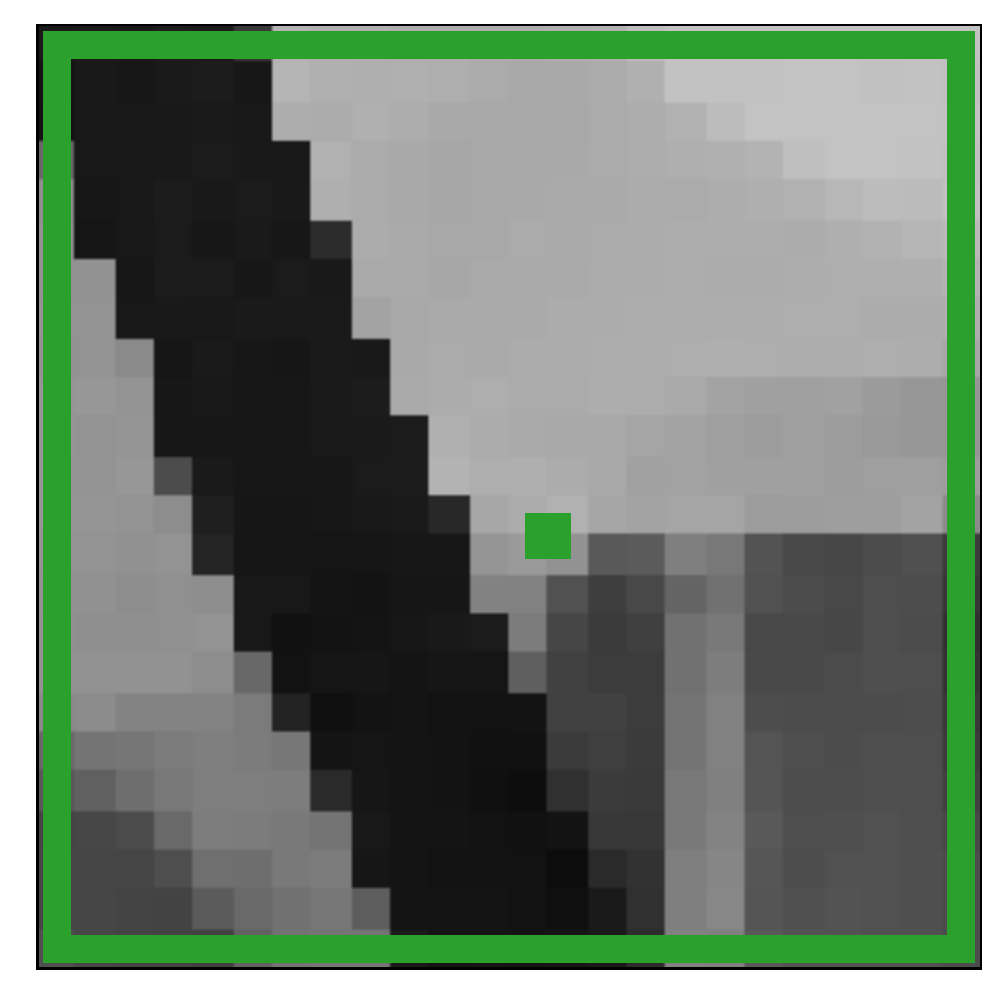}&
    \includegraphics[width=\f1ht]{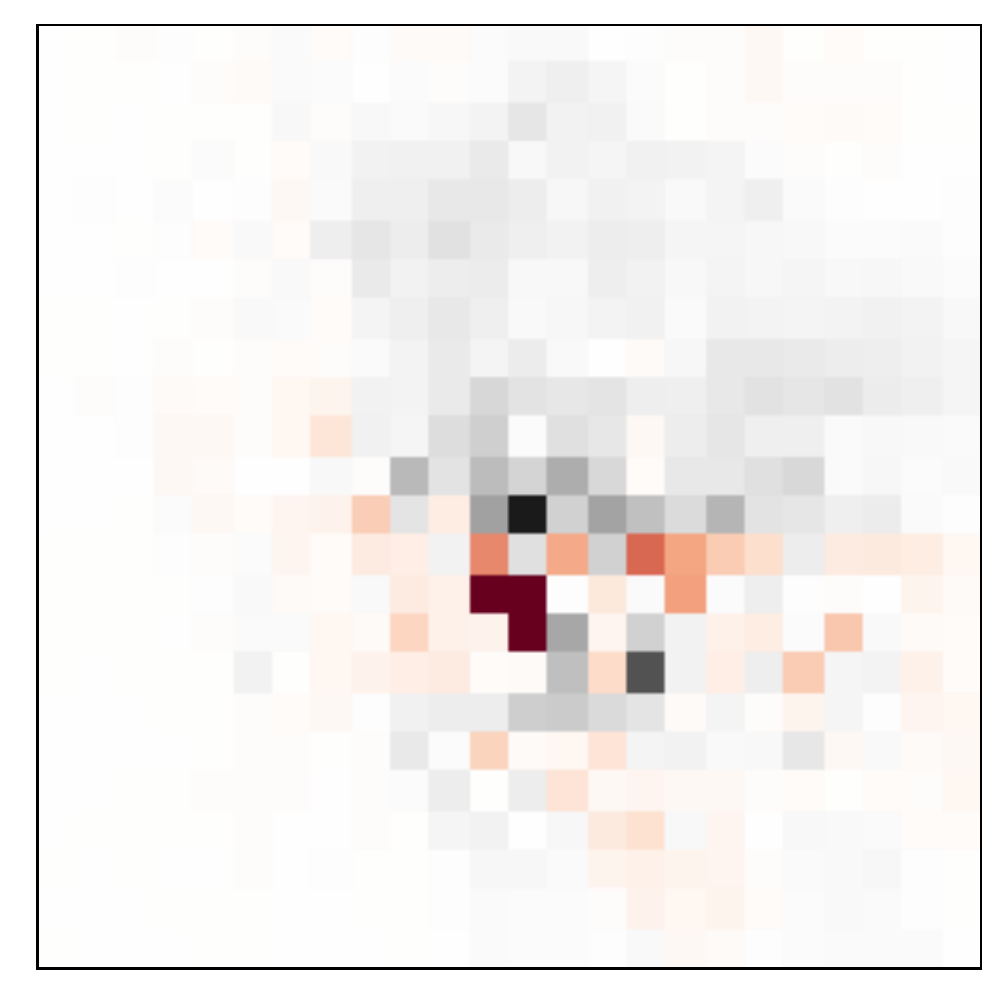}&
    \includegraphics[width=\f1ht]{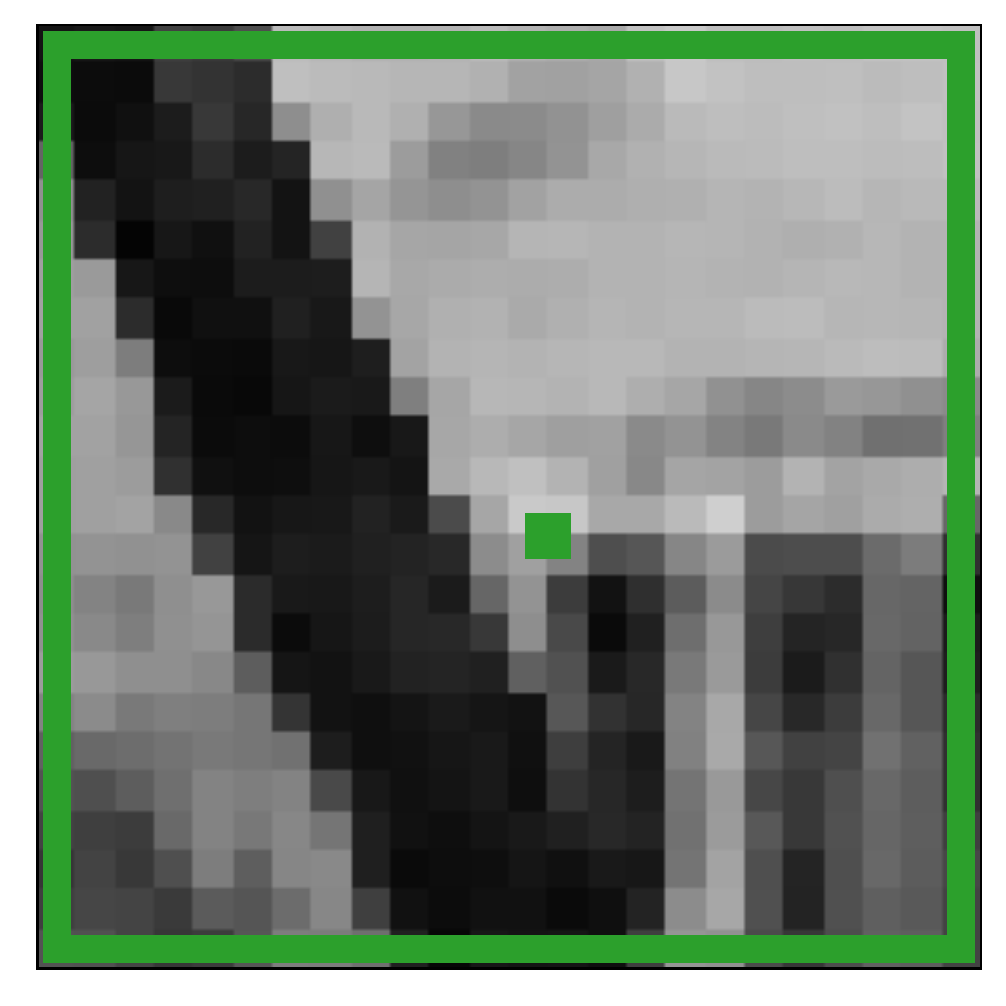}&
    \includegraphics[width=\f1ht]{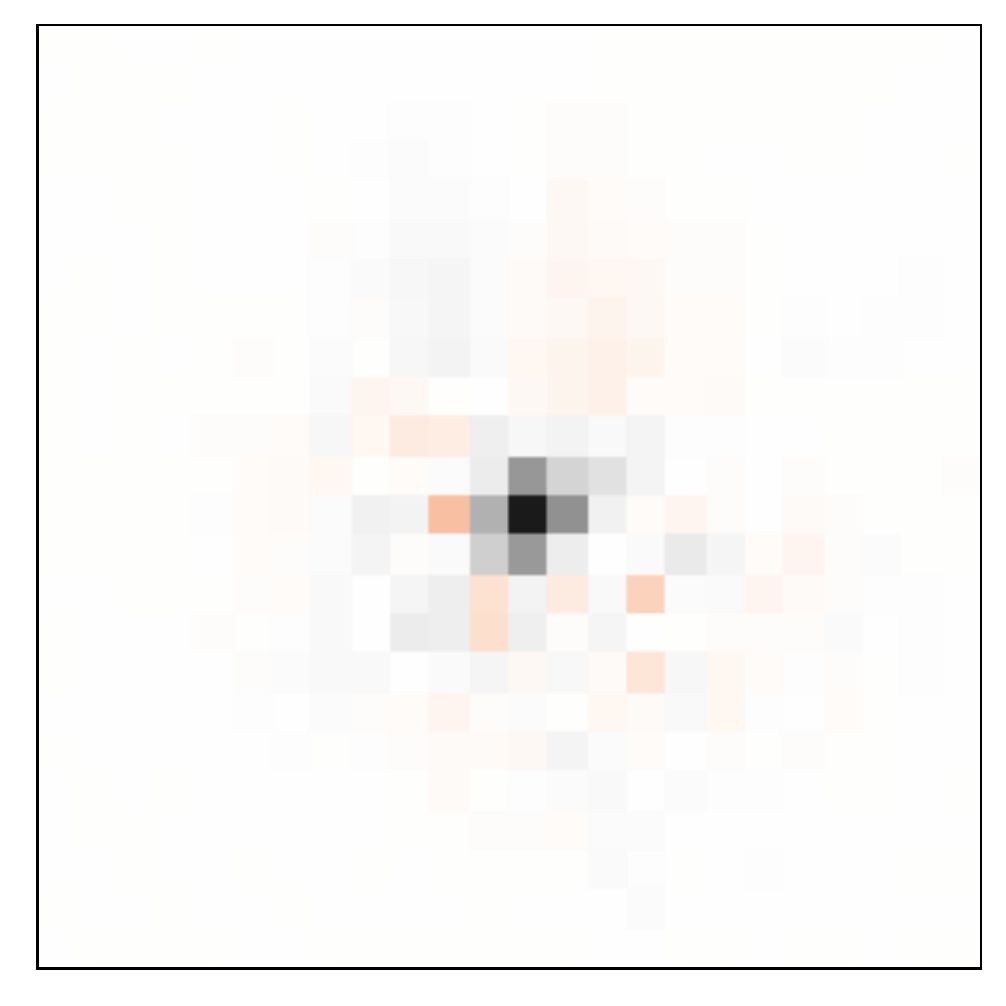}&
    \includegraphics[width=\f1ht]{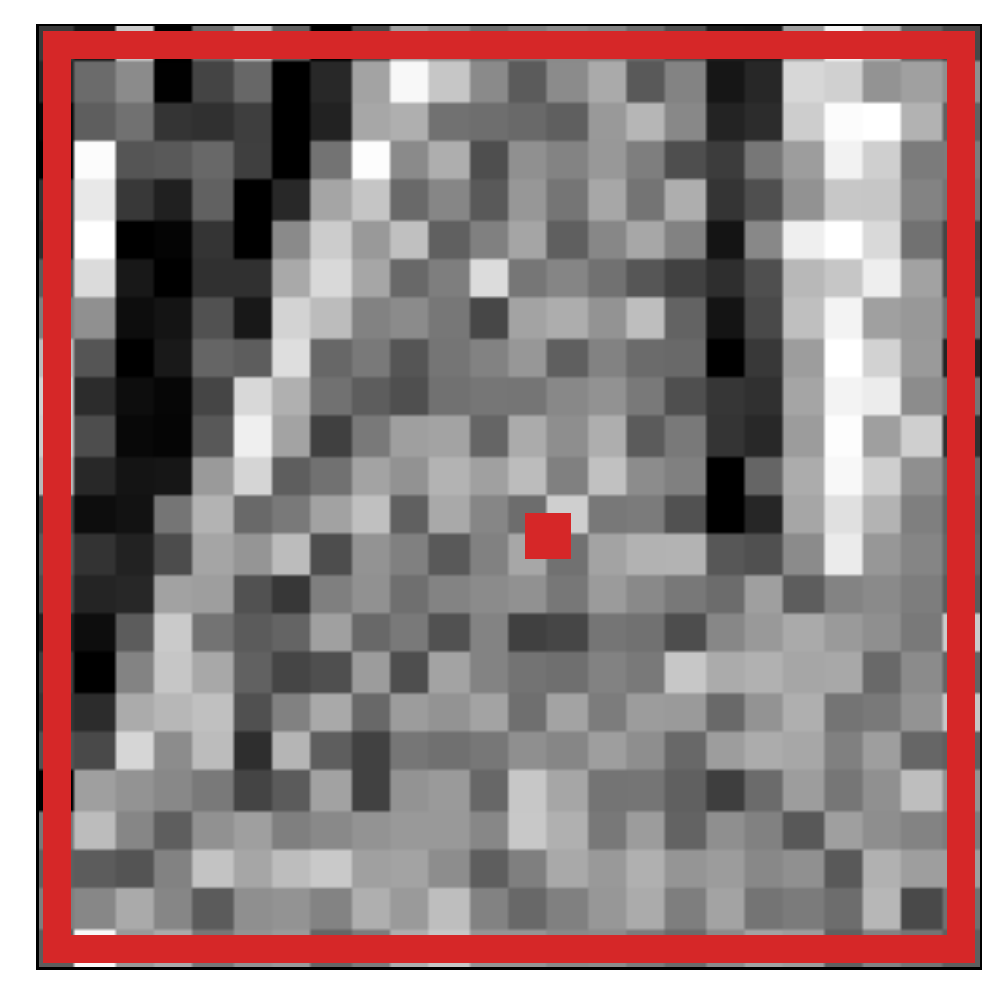}&
    \includegraphics[width=\f1ht]{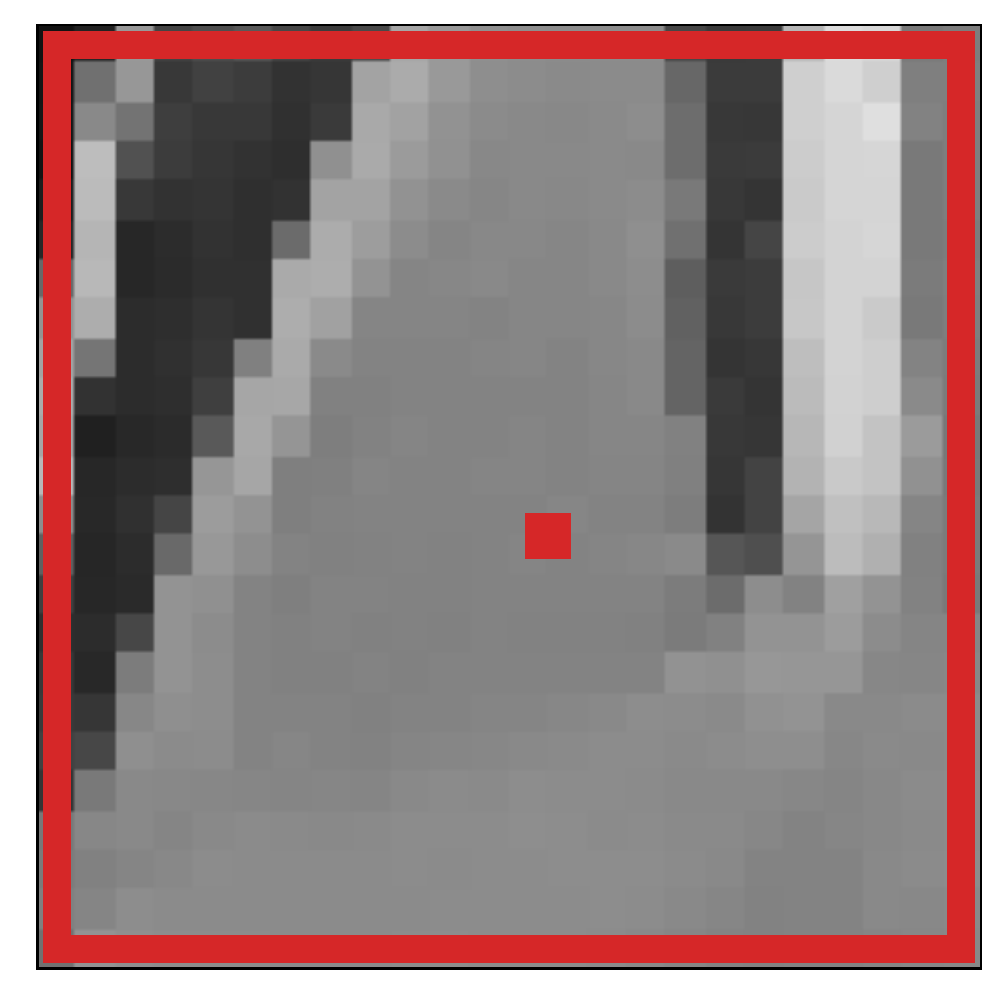}&
    \includegraphics[width=\f1ht]{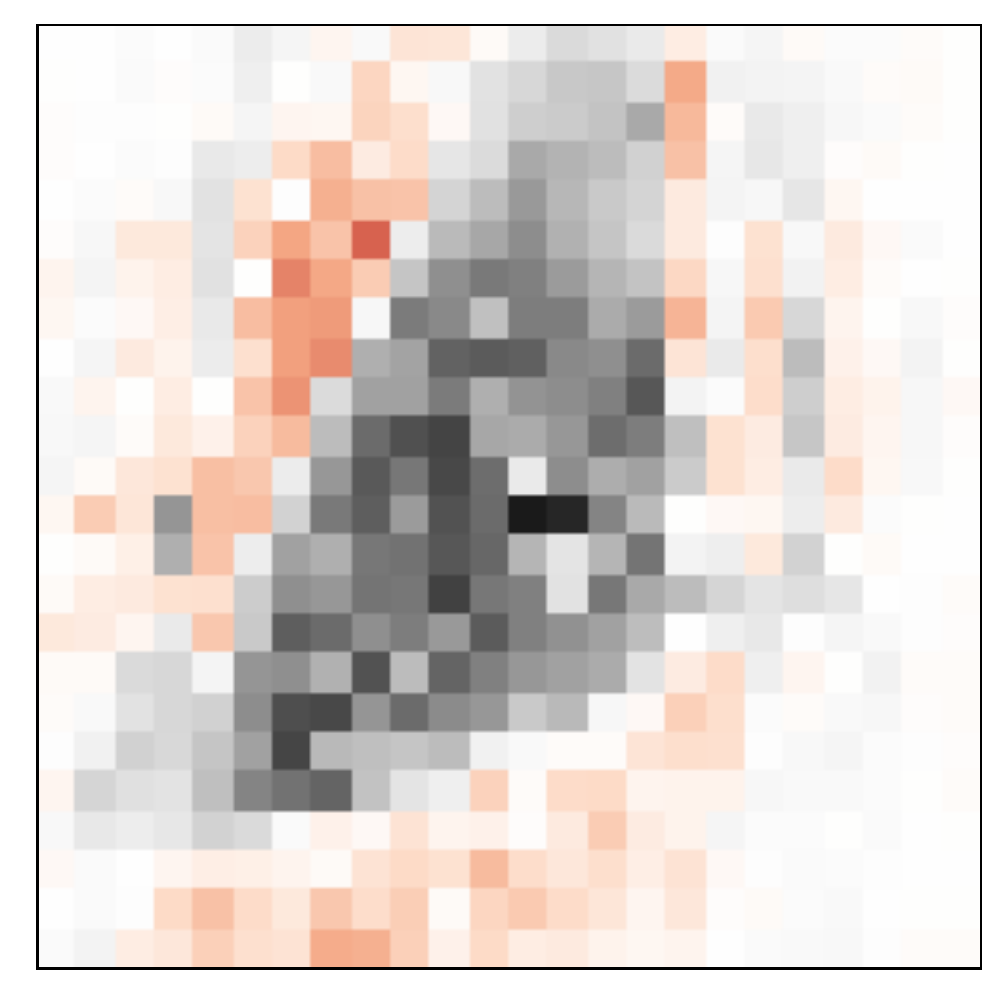}&
    \includegraphics[width=\f1ht]{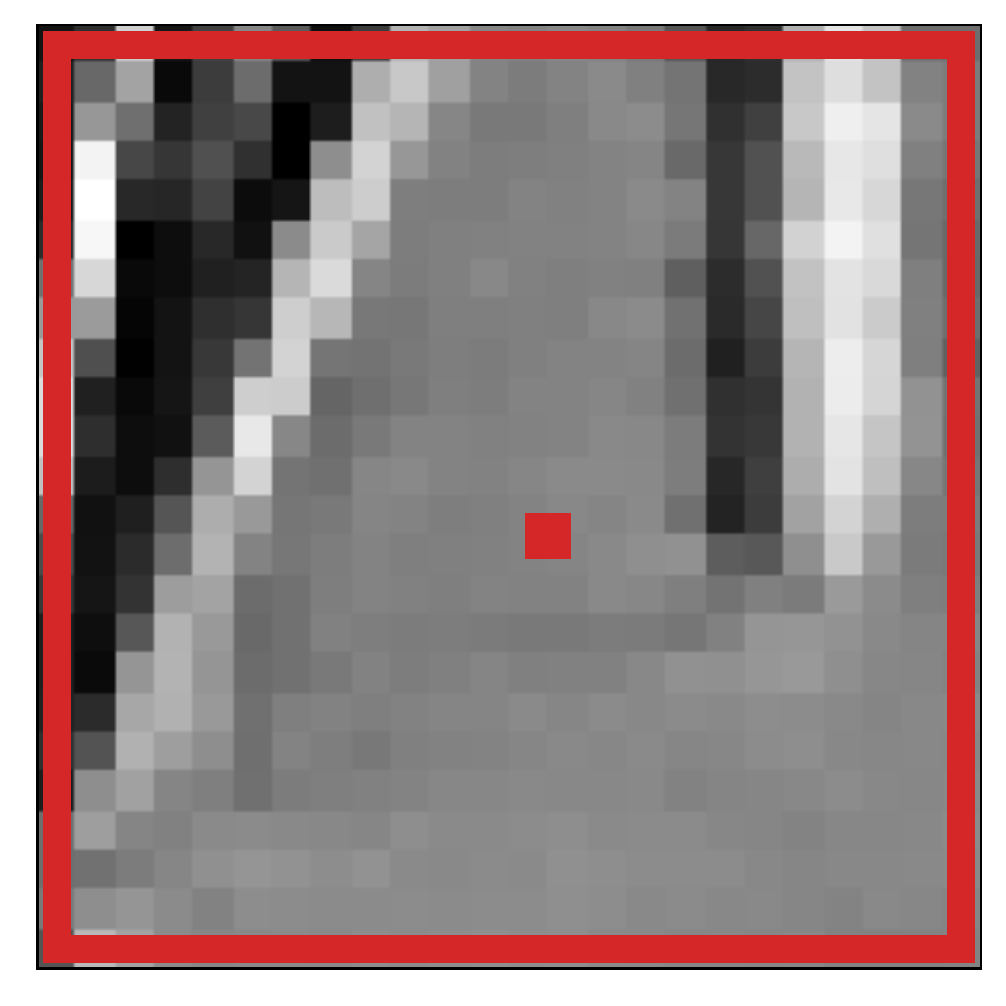}&
    \includegraphics[width=\f1ht]{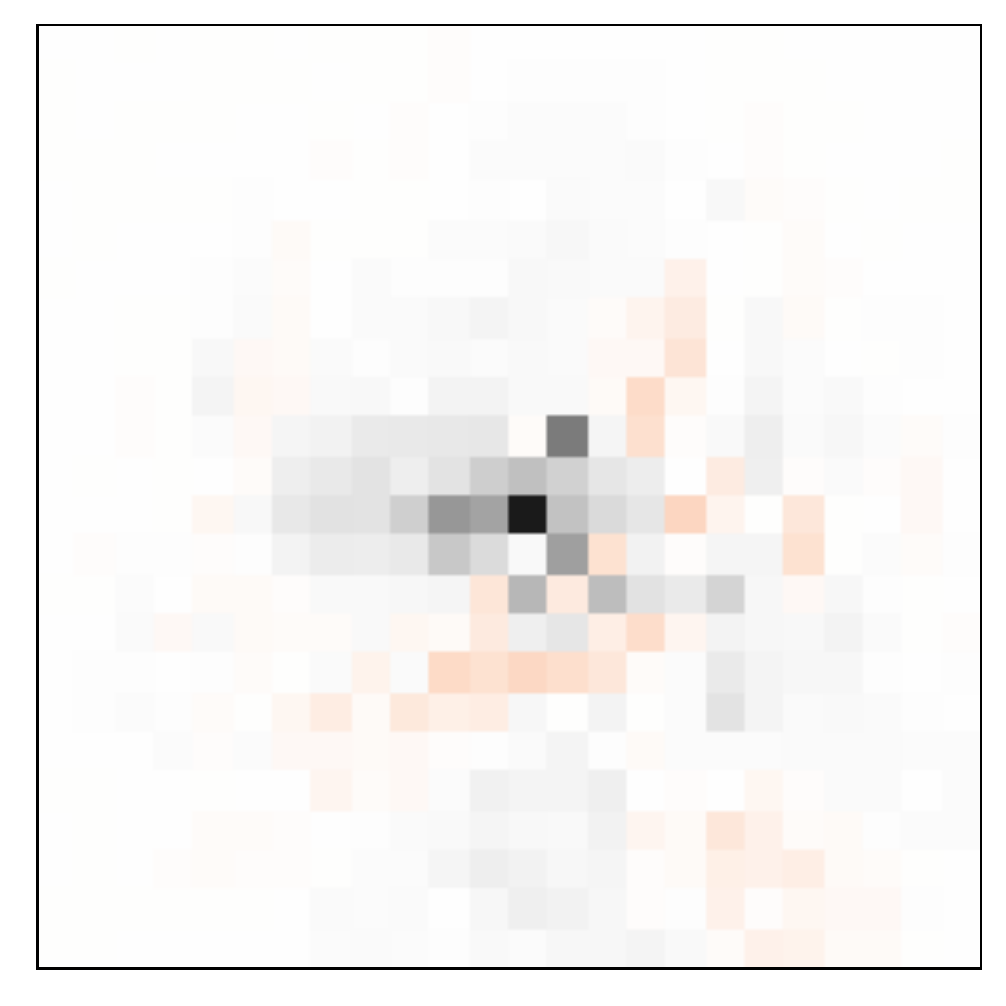}\\

    \thead{\scriptsize{Noisy} } & 
    \thead{\scriptsize{Before GT}}&
    \thead{\scriptsize{Filter bef.}}&
    \thead{\scriptsize{After GT}}&
    \thead{\scriptsize{Filter af.}}&
    \thead{\scriptsize{Noisy} } & 
    \thead{\scriptsize{Before GT}}&
    \thead{\scriptsize{Filter bef}}&
    \thead{\scriptsize{After GT}}&
    \thead{\scriptsize{Filter af.}}\\
    \end{tabular}
    \caption{\textbf{Adaptation to new image content}. (Top) A CNN pre-trained on piecewise constant images applied to a natural test image (a) oversmooths the image and blurs the details (b), but is able to recover more detail after applying \gt\ (c). (Bottom) The CNN estimates a denoised pixel (dot at the center of each image) as a linear combination of the noisy input pixels. The weighting functions (filters) of pre-trained CNN are more dispersed, consistent with the training set. However, after \gt, the weighting functions are more precisely targeted to the local features, resulting in a denoised image (c) with more details.}
    \label{fig:filter}
\end{figure}

\section{Analysis}
\label{sec:analysis}
In this section, we perform a qualitative analysis of the properties of \gt.

\textbf{What kind of images benefit the most from adaptive denoising?} Section~\ref{sec:suppl_analysis_top5} shows the images in the different test datasets for which \gt\ achieves the most and the least improvement in PSNR. For different architecture, the result is quite consistent: \gt\ is more beneficial for images with highly repetitive patterns. This makes intuitive sense; the repetitions effectively provide multiple examples from which to learn these patterns during the unsupervised refinement.

\textbf{Generalization via \gt}. %
Section~\ref{sec:suppl_analysis_gen} shows that \gt\ can achieve generalization to images that are similar to the test image used for adaptation. 

\textbf{How does \gt\ adapt to out-of-distribution noise?} Generalization to out-of-distribution noise provides a unique opportunity to understand how \gt\ modifies the denoising function. Ref.~\cite{biasfree} shows that the first-order Taylor approximation of denoising CNNs trained on multiple noise levels tend to have a negligible constant term, and that the growth of this term is the primary culprit for the failure of these models when tested on new noise levels.  \gt\ reduces the amplitude of this constant term, facilitating generalization (See Section~\ref{sec:suppl_analysis_bias} for more details).

\textbf{How does \gt\ adapt to out-of-distribution images?} %
Figure~\ref{fig:filter} shows the result of applying a
CNN trained on piecewise-constant images to natural images. Due to its learned prior, the CNN averages over large areas, ignoring fine textures. This is apparent in the equivalent linear filters obtained from a local linear approximation of the denoising function~\cite{biasfree}. After \gt\ the model learns to preserve the fine features much better, which is reflected in the equivalent filters (see Section~\ref{sec:suppl_analysis_filter} for more details).

\section{Limitations}
\label{sec:limitations}

As shown in Section~\ref{sec:experiments}, \gt\ improves the state-of-the-art on benchmark datasets, adapts well to out-of-distribution noise and image content, and outperforms all existing methods on an application to real world electron-microscope data. A crucial component in the success of \gt\ is restricting the parameters that are optimized at test time. However, this constraint also limits the potential improvement in performance one can achieve, as seen when fine-tuning for test images from the Urban100 and IUPR datasets, each of which contain many images with highly repetitive structure. 
In these cases, we observe that fine-tuning all parameters, and even training only on the test data using Self2Self often outperforms GainTuning. This raises the question of how to effectively leverage training datasets for such images. %

In addition, when the pre-trained denoiser is highly optimized, and the test image is within distribution, \gt\ can sometimes slightly degrade the performance of the denoiser. This is atypical (3 occurrences in 412 \gt\ experiments using DnCNN and SURE), and the decreases are quite small (maximum PSNR degradation of about 0.02dB, compared to maximum improvement of nearly 12dB).

\section{Conclusions}
\label{sec:conclusion}
We've introduced \gt\, an adaptive denoising methodology for adaptively fine-tuning a pre-trained CNN denoiser on individual test images. The method, which is general enough to be used with any denoising CNN, improves the performance of state-of-the-art CNNs on standard denoising benchmarks, and provides even more substantial improvements when the test data differ systematically from the training data, either in noise level or image type. %
We demonstrate the potential of adaptive denoising in scientific imaging through an application to electron microscopy. Here, \gt\ is able to jointly exploit synthetic data and test-time adaptation to reconstruct meaningful structure (the atomic configuration of a nanoparticle and its support), which cannot be recovered through alternative approaches. %
We hope that these results will motivate further development of techniques for test-time adaptation in denoising and other image-processing tasks. A concrete challenge for future research is how to exploit the unsupervised denoising strategy in Self2Self, which relies heavily on dropout and ensembling, in combination with pre-trained models. Finally, we would like to comment on the potential negative societal outcomes of our work. The training of CNN models on large clusters contributes to carbon emissions, and therefore global warming. In addition, the black-box characteristics of these models may result in unintended biases. We hope that these effects may be offset to some extent by the potential applications of these approaches to tackle challenges such as global warming. In fact, the catalytic system studied in this work is representative of catalysts used in clean energy conversion and environmental remediation~\cite{montini2016fundamentals, yu2012review, nie2015recent}.

\begin{ack}
We gratefully acknowledge financial support from the National Science Foundation (NSF). NSF NRT HDR Award 1922658 partially supported SM. NSF CBET 1604971 supported JLV and PAC, and NSF OAC-1940263 supported RM and PAC. NSF OAC-1940097 supported CFG.  Simons Foundation supported EPS. The authors acknowledge ASU Research Computing and NYU HPC for providing high performance computing resources, and the John M. Cowley Center for High Resolution Electron Microscopy at Arizona State University.
\end{ack}

\small
\bibliographystyle{acm}
\bibliography{bibli.bib}

\appendix

\section{CNN architectures}
\label{sec:architectures}
In this section we describe the denoising architectures used for our computational experiments. 

\subsection{DnCNN and BFCNN}

DnCNN~\cite{dncnn} consists of $20$ convolutional layers, each consisting of $3 \times 3$ filters and $64$ channels, batch normalization~\citep{batchnorm}, and a ReLU nonlinearity. It has a skip connection from the initial layer to the final layer, which has no nonlinear units. We use BFCNN~\cite{biasfree} based on DnCNN architecture, i.e, we remove all sources of additive bias, including the mean parameter of the batch-normalization in every layer (note however that the scaling parameter is preserved).

\subsection{UNet}
\label{sec:unet}
Our UNet model~\citep{unet} has the following layers:
\begin{enumerate}
    \item \emph{conv1} - Takes in input image and maps to $32$ channels with $5 \times 5$ convolutional kernels.
    \item \emph{conv2} - Input: $32$ channels. Output: $32$ channels. $3 \times 3$ convolutional kernels. 
    \item \emph{conv3} -  Input: $32$ channels. Output: $64$ channels. $3 \times 3$ convolutional kernels with stride 2.
    \item \emph{conv4}-  Input: $64$ channels. Output: $64$ channels. $3 \times 3$ convolutional kernels.
    \item \emph{conv5}-  Input: $64$ channels. Output: $64$ channels. $3 \times 3$ convolutional kernels with dilation factor of 2.
    \item \emph{conv6}-  Input: $64$ channels. Output: $64$ channels. $3 \times 3$ convolutional kernels with dilation factor of 4.
    \item \emph{conv7}-  Transpose Convolution layer. Input: $64$ channels. Output: $64$ channels. $4 \times 4$ filters with stride $2$.
    \item \emph{conv8}-  Input: $96$ channels. Output: $64$ channels. $3 \times 3$ convolutional kernels. The input to this layer is the concatenation of the outputs of layer \emph{conv7} and \emph{conv2}.
    \item \emph{conv9}-  Input: $32$ channels. Output: $1$ channels. $5 \times 5$ convolutional kernels.
\end{enumerate}

The structure is the same as in~\cite{durr}. This configuration of UNet assumes even width and height, so we remove one row or column from images in with odd height or width.

\subsection{Blind-spot network}
\begin{table}[]
    \centering
    \begin{tabular}{lcr}
        \toprule
        Name & $N_{out}$ & Function \\
        \midrule
        Input & 1 & \\
        enc\_conv\_0 & 48 & Convolution $3 \times 3$ \\
        enc\_conv\_1 & 48 & Convolution $3 \times 3$ \\
        enc\_conv\_2 & 48 & Convolution $3 \times 3$ \\
        pool\_1 & 48 & MaxPool $2 \times 2$ \\
        enc\_conv\_3 & 48 & Convolution $3 \times 3$ \\
        enc\_conv\_4 & 48 & Convolution $3 \times 3$ \\
        enc\_conv\_5 & 48 & Convolution $3 \times 3$ \\
        pool\_2 & 48 & MaxPool $2 \times 2$ \\
        enc\_conv\_6 & 96 & Convolution $3 \times 3$ \\
        enc\_conv\_7 & 96 & Convolution $3 \times 3$ \\
        enc\_conv\_8 & 48 & Convolution $3 \times 3$ \\
        upsample\_1 & 48 & NearestUpsample $2 \times 2$ \\
        concat\_1 & 96 & Concatenate output of pool\_1 \\
        dec\_conv\_0 & 96 & Convolution $3 \times 3$ \\
        dec\_conv\_1 & 96 & Convolution $3 \times 3$ \\
        dec\_conv\_2 & 96 & Convolution $3 \times 3$ \\
        dec\_conv\_3 & 96 & Convolution $3 \times 3$ \\
        upsample\_2 & 96 & NearestUpsample $2 \times 2$ \\
        concat\_2 & 96+$k_1$ & Concatenate output of Input \\
        dec\_conv\_4 & 96 & Convolution $3 \times 3$ \\
        dec\_conv\_5 & 96 & Convolution $3 \times 3$ \\
        dec\_conv\_6 & 96 & Convolution $3 \times 3$ \\
        dec\_conv\_7 & 1 & Convolution $3 \times 3$ \\ 
        \bottomrule
    \end{tabular}
    \caption{\textbf{Blind-spot network}. The convolution and pooling layers are the blind-spot variants described in Ref.~\cite{blindspotnet}.}
    \label{tab:blind_net_arch}
\end{table}

We use a modified version of the blind-spot network architecture introduced in Ref.~\cite{blindspotnet}. We rotate the input frames by multiples of $90^{\circ}$ and process them through four separate branches (with shared weights) containing asymmetric convolutional filters that are \emph{vertically causal}. The architecture of a branch is described in Table~\ref{tab:blind_net_arch}. Each branch has one input channel and one output channel. Each branch is followed by a de-rotation and the output is passed to a series of three cascaded $1 \times 1$ convolutions and non-linearity for reconstruction with 4 and 96 intermediate output channels, as in \cite{blindspotnet}. The final convolutional layer is linear and has 1 output channel. \\ 

\section{Datasets}
\label{sec:datasets}

\begin{figure}
    \centering
    \begin{tabular}{c@{\hskip 0.1in}c}
    BSD400 &  Set12  \\
    \includegraphics[width=2.6in]{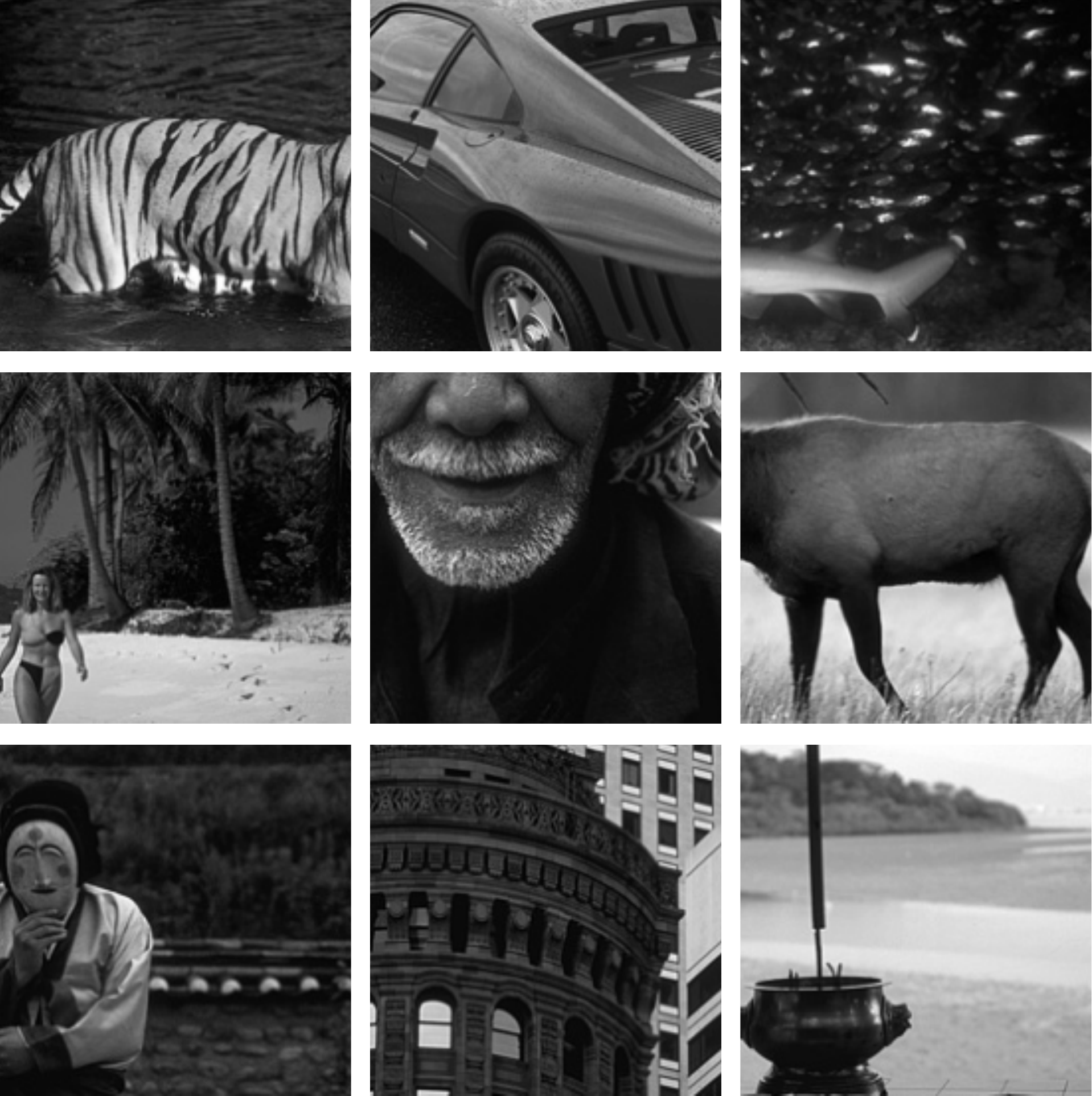}&
    \includegraphics[width=2.6in]{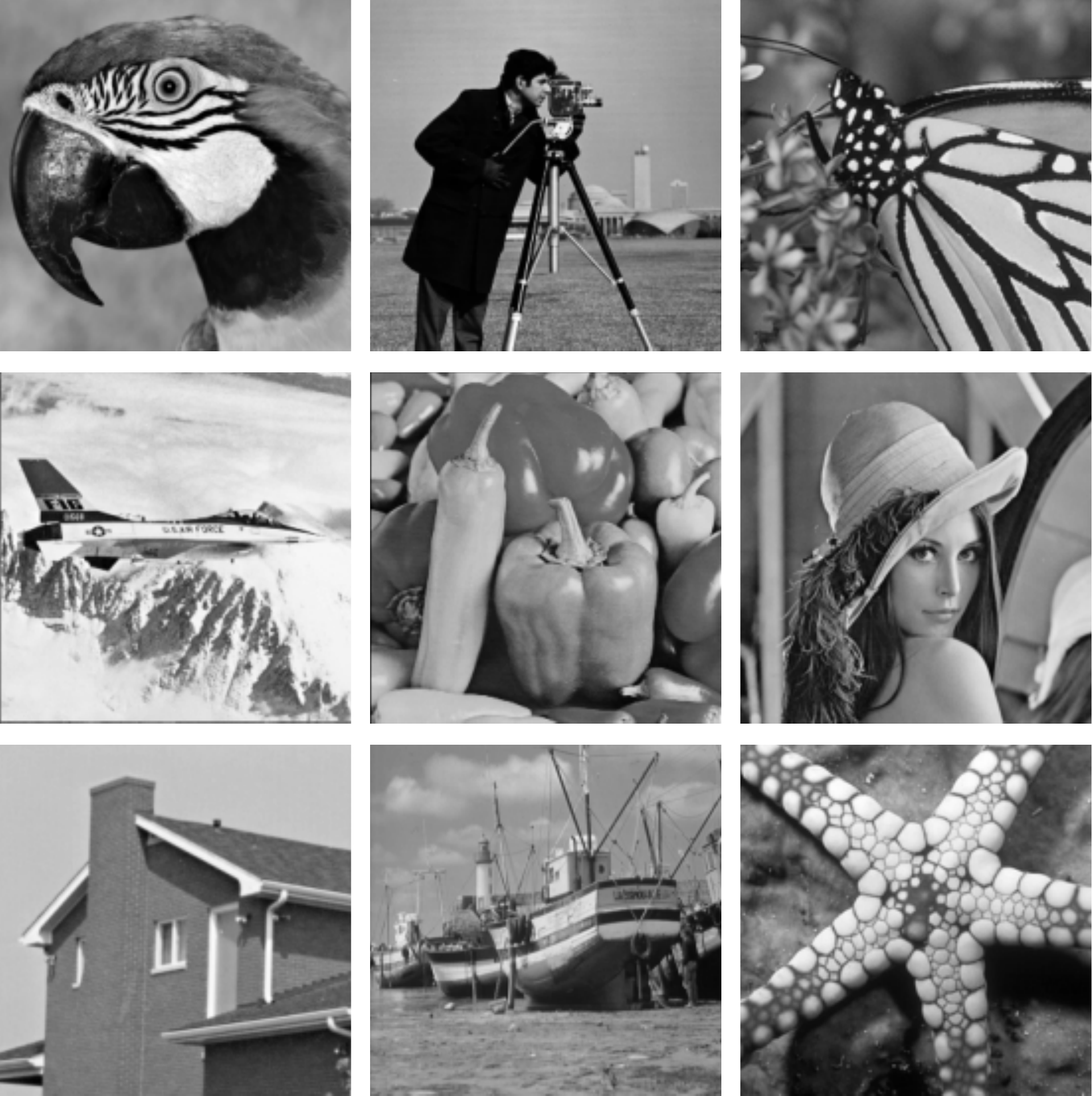}\\
    BSD68 & Urban100 \\
    \includegraphics[width=2.6in]{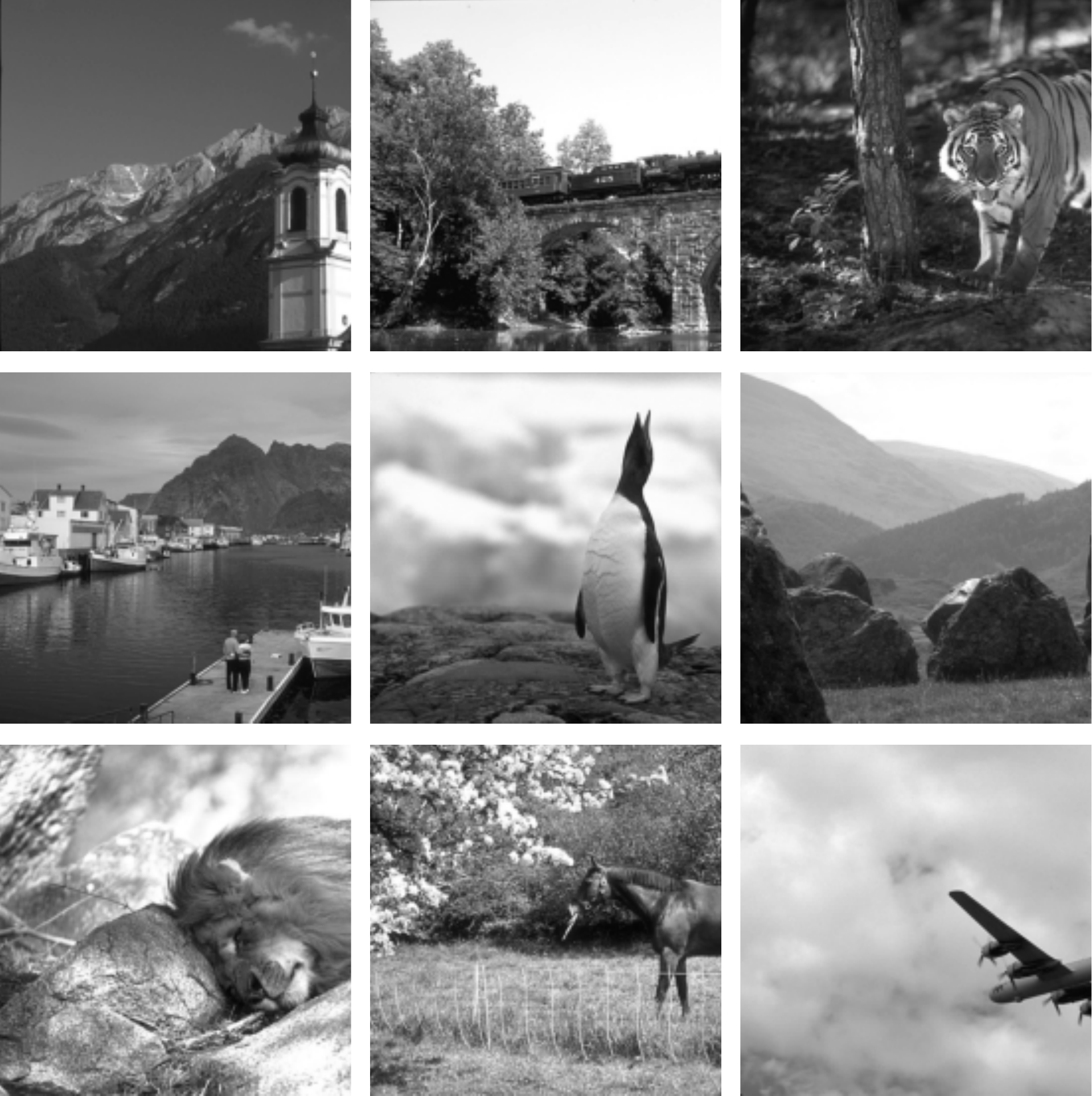}&
    \includegraphics[width=2.6in]{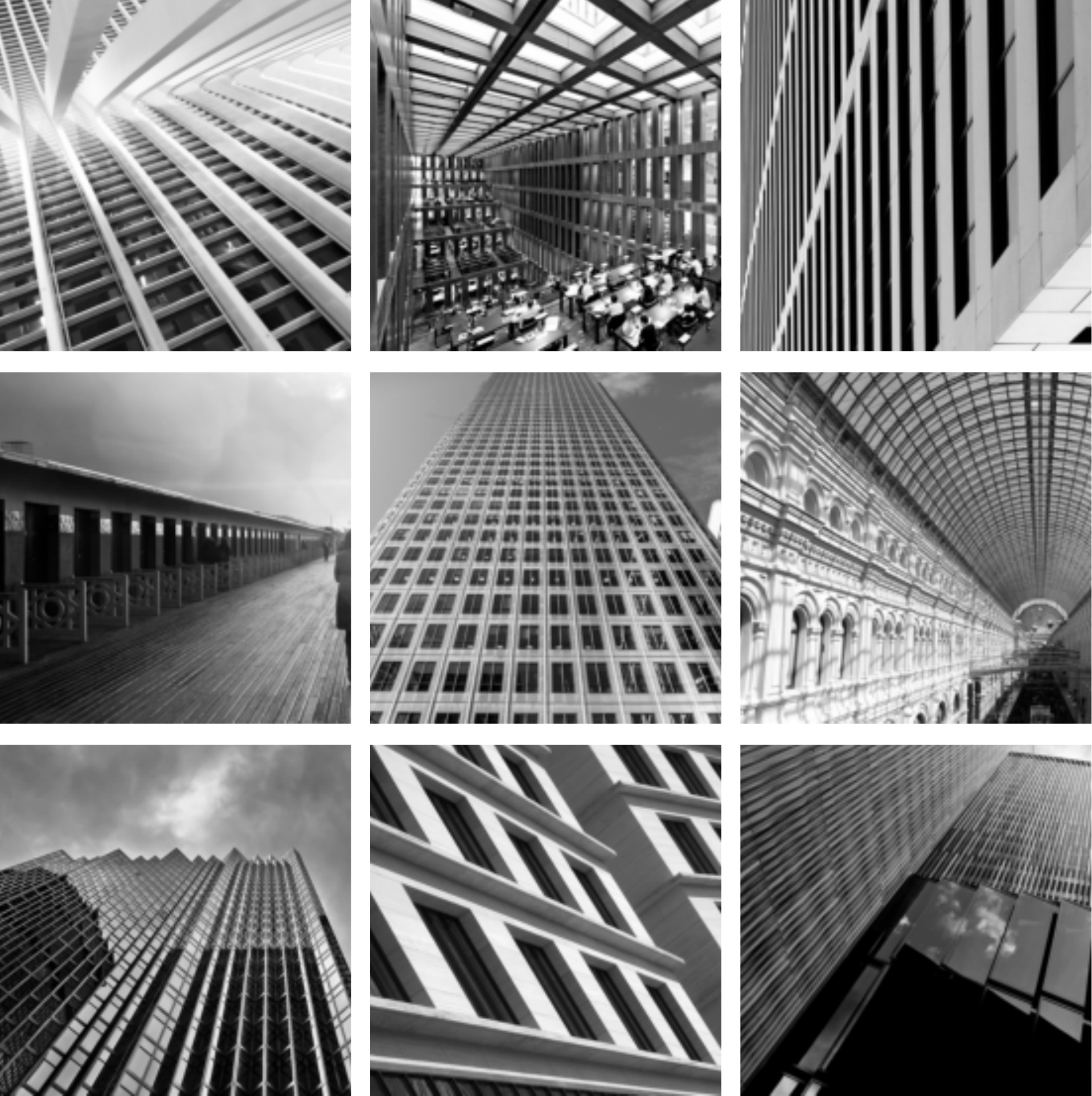}\\
    IUPR & Piecewise constant images\\
    \includegraphics[width=2.6in]{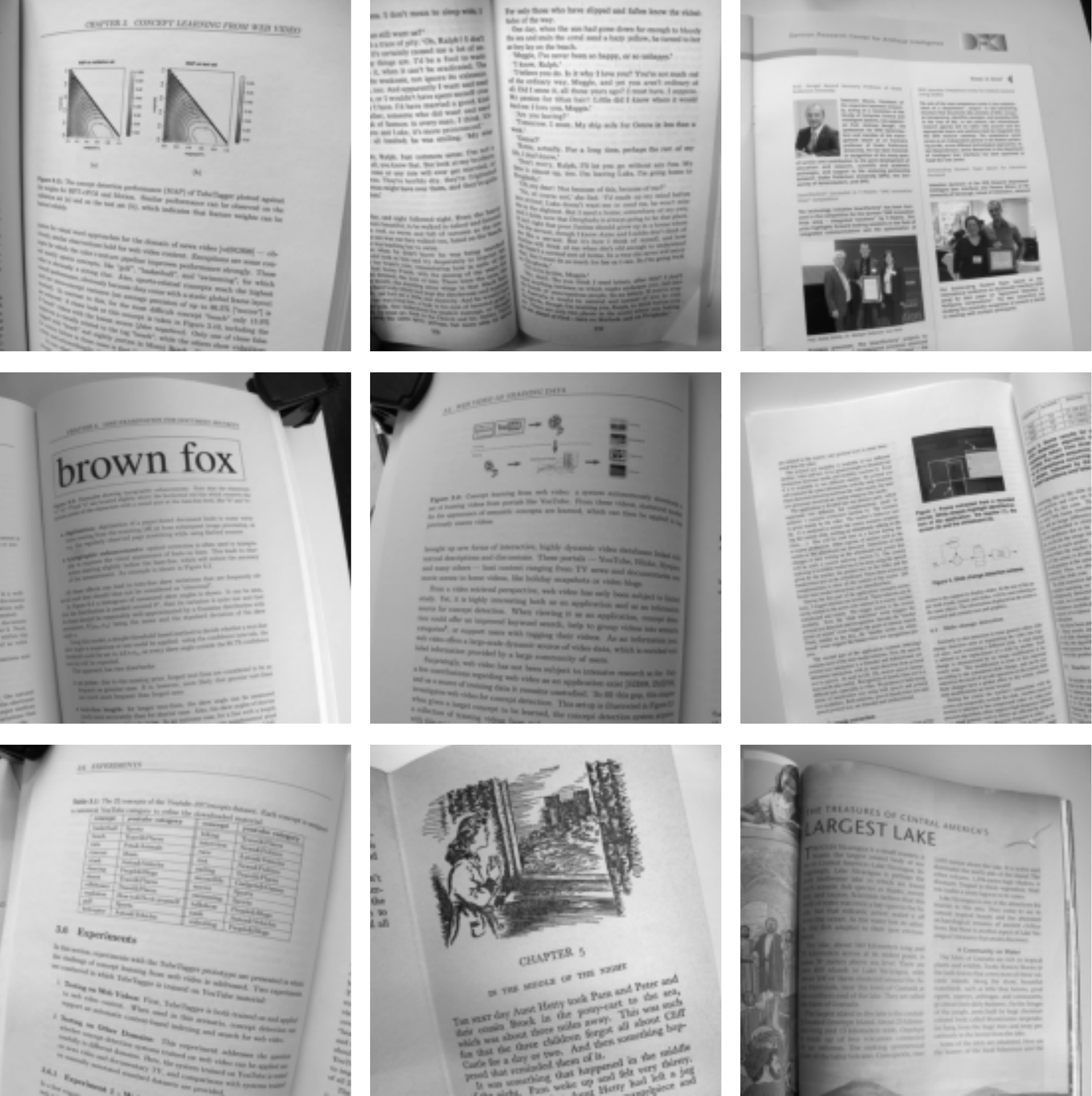}&
    \includegraphics[width=2.6in]{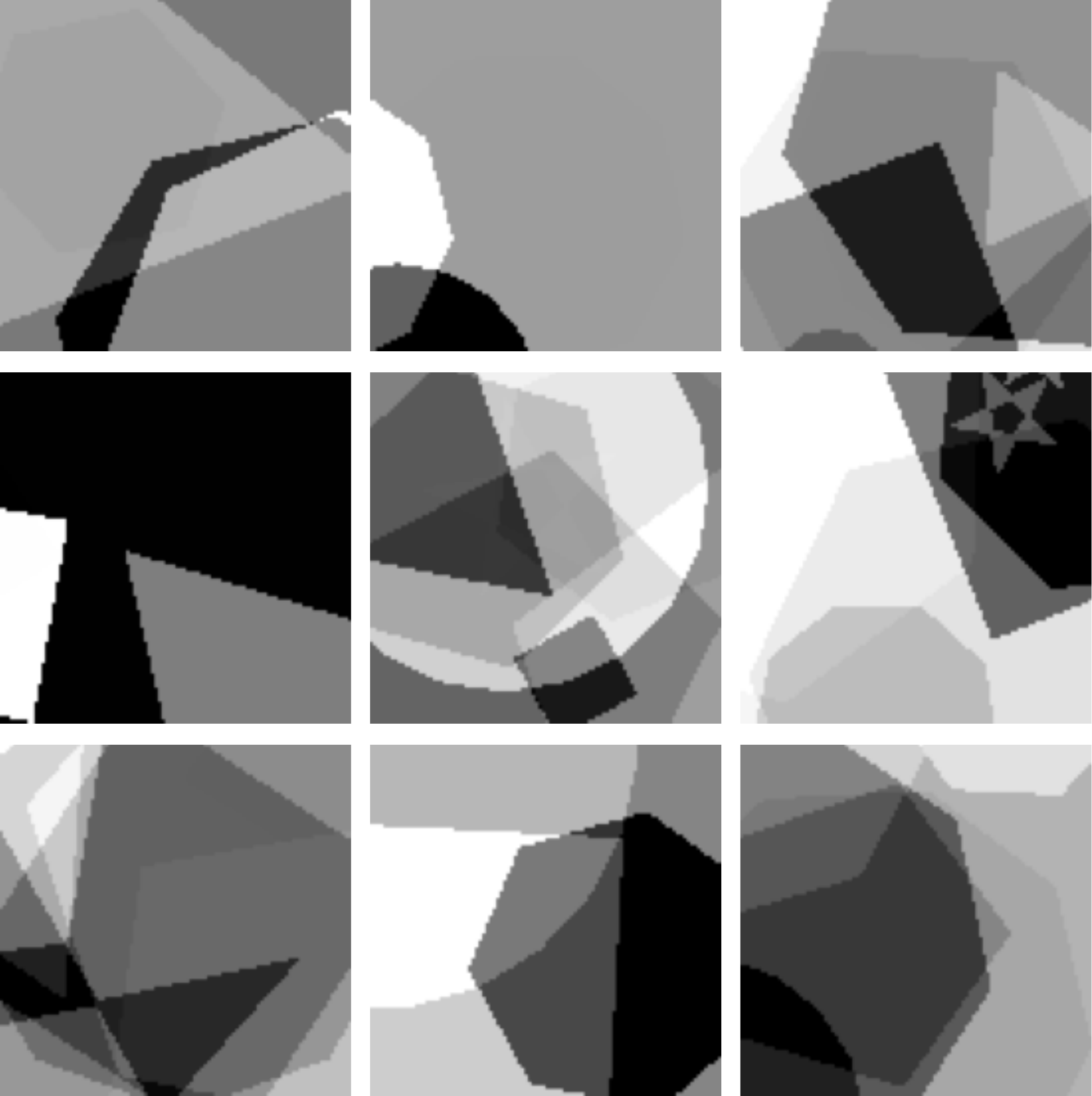}\\
    \end{tabular}
    \caption{\textbf{Datasets}. Nine images chosen at random from each dataset is visualized.}
    \label{fig:datasets}
\end{figure}

We perform controlled experiments on datasets with different signal and noise structure to evaluate the broad applicability of \gt\ (see Figure~\ref{fig:datasets} for a visual summary of datasets). We describe each dataset below:

\textbf{Generic natural images.} We use 400 images from BSD400~\cite{bsd400} dataset for pre-training CNNs. We evaluate two test sets, Set12 and Set68, with 12 and 68 images for testing~\cite{dncnn}. 

\textbf{Images of urban scenes.} We evaluate generalization capabilities of \gt\ using a dataset of images captured in urban settings, Urban100~\cite{urban100}. These images often contain repeating patterns and structures, unlike generic natural images (see Figure~\ref{fig:datasets}). We evaluate \gt\ on the first 50 images from this dataset.

\textbf{Images of scanned documents.} We use images of scanned documents from the IUPR dataset~\cite{iupr}. The original images are very high resolution, and one of our baseline methods, Self2Self~\cite{self2self}, requires long computational time (>24 hours per image) for such high resolution images. Therefore, we resized the images in IUPR dataset by a factor of 6. We used the first 50 images from the dataset for evaluation. 

\textbf{Simulated piecewise constant images.}  We use a dataset of simulated piecewise constant images. These images have constant regions with boundaries consisting of various shapes such as circles and lines with different orientations. The constant region has an intensity value sampled from a uniform distribution between $0$ and $1$ (see Figure~\ref{fig:datasets}). Piecewise constant images provide a crude model for natural images~\cite{matheron1975random, pitkow2010exact,lee2001occlusion}. We use \gt\ to adapt a CNN trained on this dataset to generic natural images (Set12). This experiment demonstrates the ability of \gt\ to adapt from a simple simulated dataset to a significantly more complex real dataset.
 
\textbf{Simulated transmission electron microscopy data}. The TEM image data used in this work correspond to images from a widely utilized catalytic system, which consist of platinum (Pt) nanoparticles supported on a larger cerium (IV) oxide (CeO$_2$) nanoparticle. We use the simulated TEM image dataset introduced in Ref.~\cite{mohan2020deep} for pre-training CNNs.  The simulated dataset contains 1024 x 1024 images, which are binned to match the approximate pixel size of the experimentally acquired real image series (described below). To equate the intensity range of the simulated images with those acquired experimentally, the intensities of the simulated images were scaled by a factor which equalized the vacuum intensity in a single simulation to the average intensity measured over a large area of the vacuum in a single 0.025 second experimental frame (i.e., 0.45 counts per pixel in the vacuum region). Furthermore, during TEM imaging multiple electron-optical and specimen parameters can give rise to complex, non-linear modulations of the image contrast. These parameters include the objective lens defocus, the specimen thickness, the orientation of the specimen, and its crystallographic shape/structure. Various combinations of these parameters may cause the contrast of atomic columns in the image to appear as black, white, or an intermediate mixture of the two. To account for this, the simulated dataset contains various instances of defocus, tilt, thickness, and shape/structure. We refer interested readers to Ref.~\cite{mohan2020deep} for more details.

\textbf{Real transmission electron microscopy data}. The real data consist of a series of images of the Pt/CeO$_2$ catalyst. The images were acquired in a N$_2$ gas atmosphere using an aberration-corrected FEI Titan transmission electron microscope (TEM), operated at 300 kV and coupled with a Gatan K2 IS direct electron detector~\cite{mohan2020deep}. The detector was operated in electron counting mode with a time resolution of 0.025 sec/frame and an incident electron dose rate of 5,000 e$^-$/{\AA}$^2$/s. The electromagnetic lens system of the microscope was tuned to achieve a highly coherent parallel beam configuration with minimal low-order aberrations (e.g., astigmatism, coma), and a third-order spherical aberration coefficient of approximately -13 $\mu$m. We refer interested readers to Ref.~\cite{mohan2020deep} for more details.

\section{Details of pre-training and \gt\ }
\label{sec:training}

In this section, we describe the implementation details of the pre-training process and our proposed \gt\ framework. %
\subsection{Overview}
\label{sec:suppl_pre_gt}
  While performing \gt, we include a scalar multiplier parameter (gain) in every channel of the convolutional layers in the denoising CNN. We do not optimize the gain for the last layer of the network. We describe the general optimization process for \gt\ here, and describe any additional modifications for specific datasets in the respective subsections. 

\textbf{Data}. We perform \gt\ on patches extracted from the noisy image. We extracted $400 \times 400$ patches for the electron microscopy dataset, and $50 \times 50$ patches for all other datasets. We do not perform any data augmentation on the extracted patches. 

\textbf{BatchNorm layers during \gt}. If the denoising CNN contains batch normalization (BN) layers (only DnCNN~\cite{dncnn} and BFCNN~\cite{biasfree} in our experiments), we freeze these parameters during \gt . That is, we do not re-estimate the mean and standard deviation parameter for each layer from the test data. Instead, we re-use the original values estimated from pre-training dataset, and re-purposed the learnable parameter in the batch norm layer as the Gain parameter. 

\textbf{Optimization}. We train with Adam~\cite{adam} optimizer for 100 steps with a starting learning rate of $10^{-4}$ which was reduced to $10^{-5}$ after the $20^{\text{th}}$ step. Here, we define each step as a pass through $5000$ random patches extracted from the test image. When performing experiments to compare optimizing all parameters vs optimizing only gain during the adaptation process, we kept the learning rate constant at $10^{-5}$ for both options, and trained for 1000 steps.  

\subsection{Natural images}
\label{sec:suppl_pre_nat}
\textbf{Pre-training dataset}. Our experiments are carried out on $180 \times 180$ natural images from the Berkeley Segmentation Dataset~\cite{bsd400}.  We use  a training set of $400$ images. The training set is augmented via downsampling, random flips, and random rotations of patches in these images~\citep{dncnn}. We train the CNNs on patches of size $50 \times 50$, which yields a total of 541,600 clean training patches. 

\textbf{Pre-training process}. We train DnCNN, BFCNN and UNet using the Adam Optimizer~\cite{adam} for $100$ epochs with an initial learning rate of $10^{-3}$ and a decay factor of $0.5$ for every 10 epochs after the $50^{th}$, with no early stopping~\cite{biasfree}.

\textbf{\gt}. We follow the same procedure as Section~\ref{sec:suppl_pre_gt}.

\subsection{Piecewise constant images}
\label{sec:suppl_pre_piece}
\textbf{Pre-training dataset}. We generated a synthetic dataset of piecewise constant images with the varied boundary shapes like slanted lines and circles (see Figure~\ref{fig:datasets}). The intensity values of the constant regions were uniformly sampled between $0$ and $1$. The generated patches were of size $50 \times 50$ to mimic the training process for natural images~\cite{dncnn}. 

\textbf{Pre-training}. We train DnCNN, BFCNN and UNet using the Adam Optimizer~\cite{adam} using the same process as in Section~\ref{sec:suppl_pre_nat}. For each epoch, we generated $50,000$ random patches. 

\textbf{\gt\ }. We used Adam~\cite{adam} with a starting learning rate of $10^{-4}$ reduced to $10^{-5}$ after 20 steps for adapting CNNs trained on piecewise constant function to natural images.  We trained for 100 steps in total. 

\subsection{Electron microscope data}
\label{sec:suppl_pre_nano}
\textbf{Pre-training dataset}. Our experiments are carried out on $400 \times 400$ patches extracted from about $5000$ simulated TEM introduced in Ref.~\cite{mohan2020deep}. The training set is augmented via downsampling, random flips, and random rotations of patches in these images~\cite{mohan2020deep, vincent2021developing}. 

\noindent \textbf{Optimization Details:} We trained using Adam \cite{adam} optimizer with a starting learning of $10^{-4}$. The learning rate was decreased by a factor of $2$ at checkpoints $[20, 25, 30]$ during a total training of $40$ epochs~\cite{mohan2020deep}. 

\textbf{GainTuning}. We performed \gt\ using Adam~\cite{adam} optimizer with a constant learning rate of $10^{-5}$ for 100 steps. Each step consisted of 1000 randomly sampled patches of size $400 \times 400$ extracted from the test image.

\subsection{Computational resources used}
The computations were performed on an internal cluster equipped with NVIDIA RTX8000 and NVIDIA V100 GPUs. We used open-source pre-trained networks when available.%

\section{Approximation for SURE}
\label{sec:sureapprox}

Let $\mathbf{x}$ be an $N$-dimensional ground-truth random vector $\mathbf{x}$ and let $\mathbf{y} := \mathbf{x} + \mathbf{n}$ be a corresponding noisy observation, where $\mathbf{n} \sim \mathcal{N}(0, \sigma_n^2\mathbf{I})$. Stein's Unbiased Risk Estimator (SURE) provides an expression for the mean-squared error between $\mathbf{x}$ and the denoised estimate $f_{\theta}(\mathbf{y})$ (where $f_{\theta}$ denotes an arbitrary denoising function), which \emph{only depends on the distribution of noisy observations $\mathbf{y}$}: 
\begin{equation}
\label{eq:sure}
\mathbb{E}_{\mathbf{x,y}}\left[\frac{1}{N}\left\|\mathbf{x}-f_{\theta}(\mathbf{y})\right\|^{2}\right]=
\mathbb{E}_{\mathbf{y}}\left[\frac{1}{N}\left\|\mathbf{y}-f_{\theta}(\mathbf{y})\right\|^{2}-\sigma^{2}+\frac{2 \sigma^{2}}{N} \sum_{k=1}^{N} \frac{\partial (f_{\theta}(\mathbf{y})_k)}{\partial \mathbf{y}_{k}}\right]
\end{equation}
The last term in the equation, called divergence, is costly to compute. Therefore, we use a Monte Carlo approximation of SURE introduced by Ref.~\cite{sureapprox} in our implementation. The approximation is given by:
\begin{equation}
\label{eq:sureapprox}
    \sum_{k=1}^{N} \frac{\partial (f_{\theta}(\mathbf{y})_k)}{\partial \mathbf{y}_{k}} \approx \frac{1}{\epsilon N} \langle \mathbf{\Tilde{n}}, f_{\theta}(\mathbf{y} + \epsilon \mathbf{\Tilde{n}}) - f_{\theta}(\mathbf{y}) \rangle
\end{equation}

where $\langle \textbf{x}, \textbf{y} \rangle$ represents the dot product between  $\textbf{x}$ and $\textbf{y}$, $\Tilde{n}$ represents a sample from $\mathcal{N}(0, 1)$, and $\epsilon$ represents a fixed, small, positive number. We set $\epsilon = \sigma \times 1.4 \times 10^{-4}$ for our computational experiments~\cite{surekoreanarxiv}. Equation~\eqref{eq:sureapprox} has been used in the implementation of several traditional~\cite{sureapprox}, and deep learning based~\cite{surebaranuik, surekoeanneurips, surekoreanarxiv} denoisers.

\section{\gt\ prevents overfitting}
\label{sec:overfitting}

We perform controlled experiments to compare test-time updating of (1) all parameters of a CNN, and (2) only the gain parameters. We briefly describe each experiment and our findings in this section.

\begin{figure}[t]
\def\f1ht{1.05in}%
\centering 
\footnotesize{
\begin{tabular}{ >{\centering\arraybackslash}m{.5in} >{\centering\arraybackslash}m{1.4in} >{\centering\arraybackslash}m{1.4in} >{\centering\arraybackslash}m{1.4in}}
\centering 
    Loss  &  In-distribution  & Out-of-distribution noise  & Out-of-distribution signal \\[0.2cm]
    SURE &
    \includegraphics[height=\f1ht]{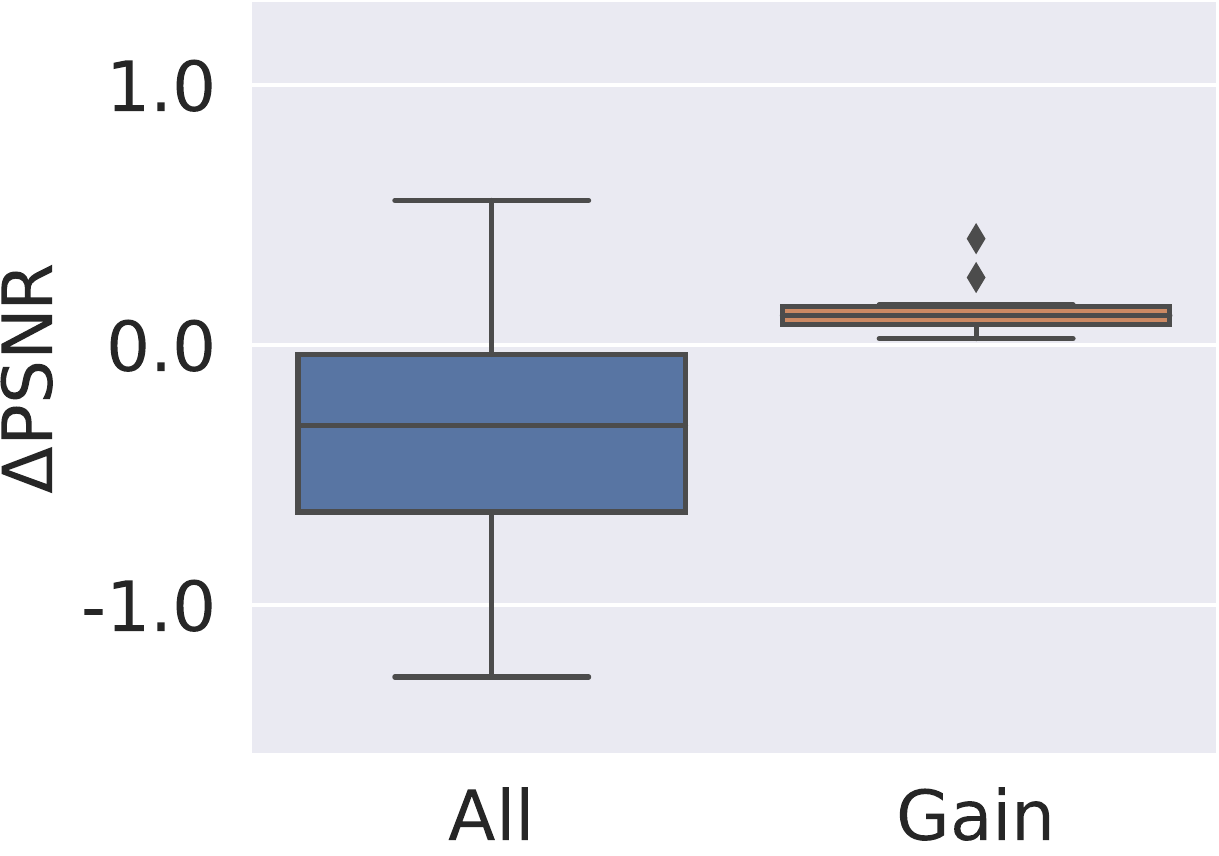}&
    \includegraphics[height=\f1ht]{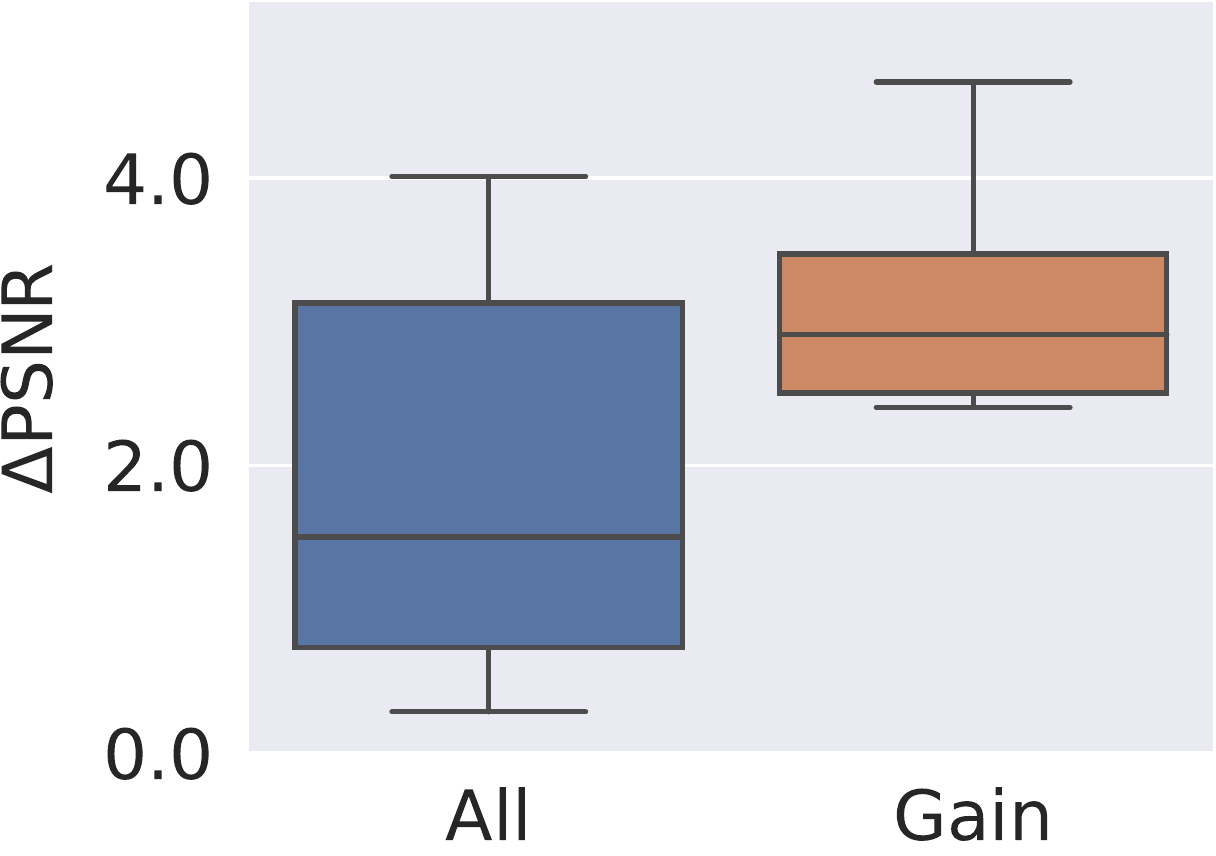}&
    \includegraphics[height=\f1ht]{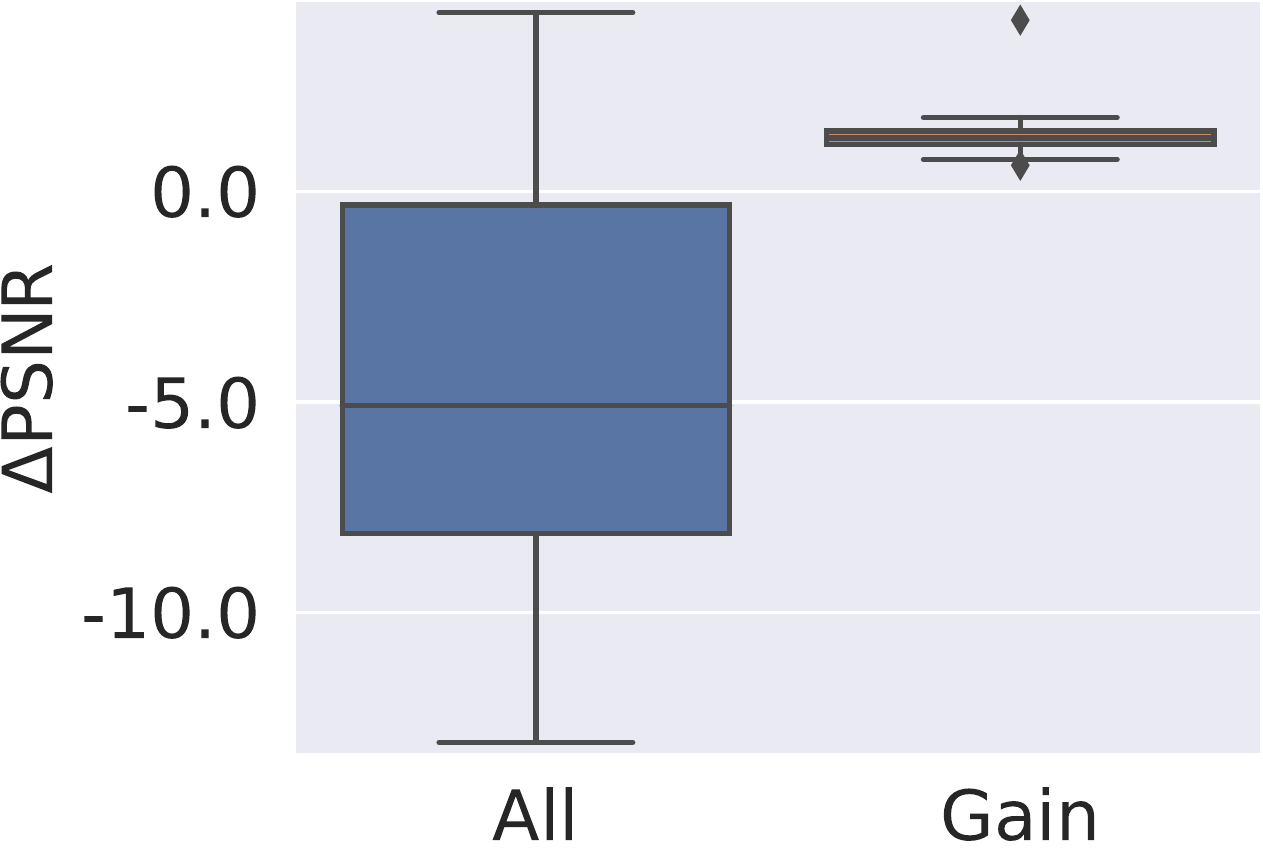}\\
    Noise Resampling &
    \includegraphics[height=\f1ht]{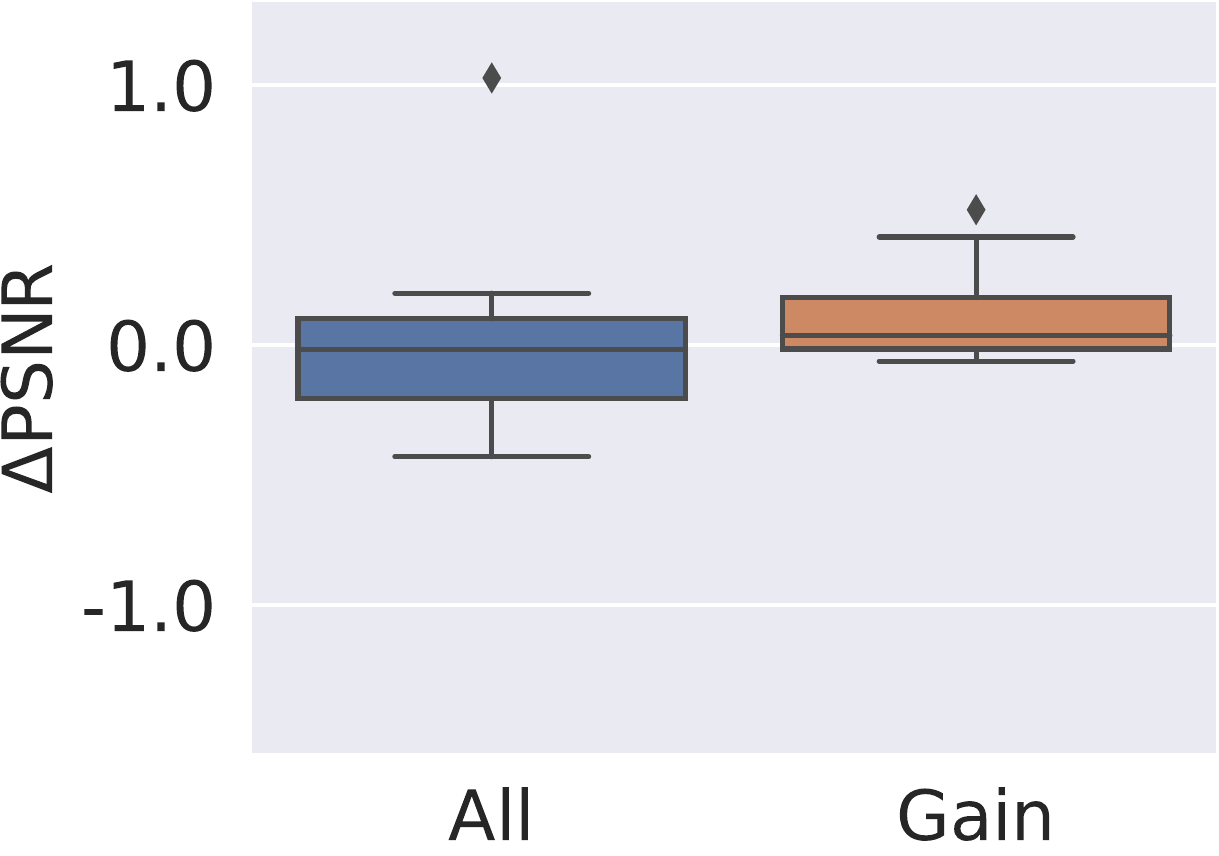}&
    \includegraphics[height=\f1ht]{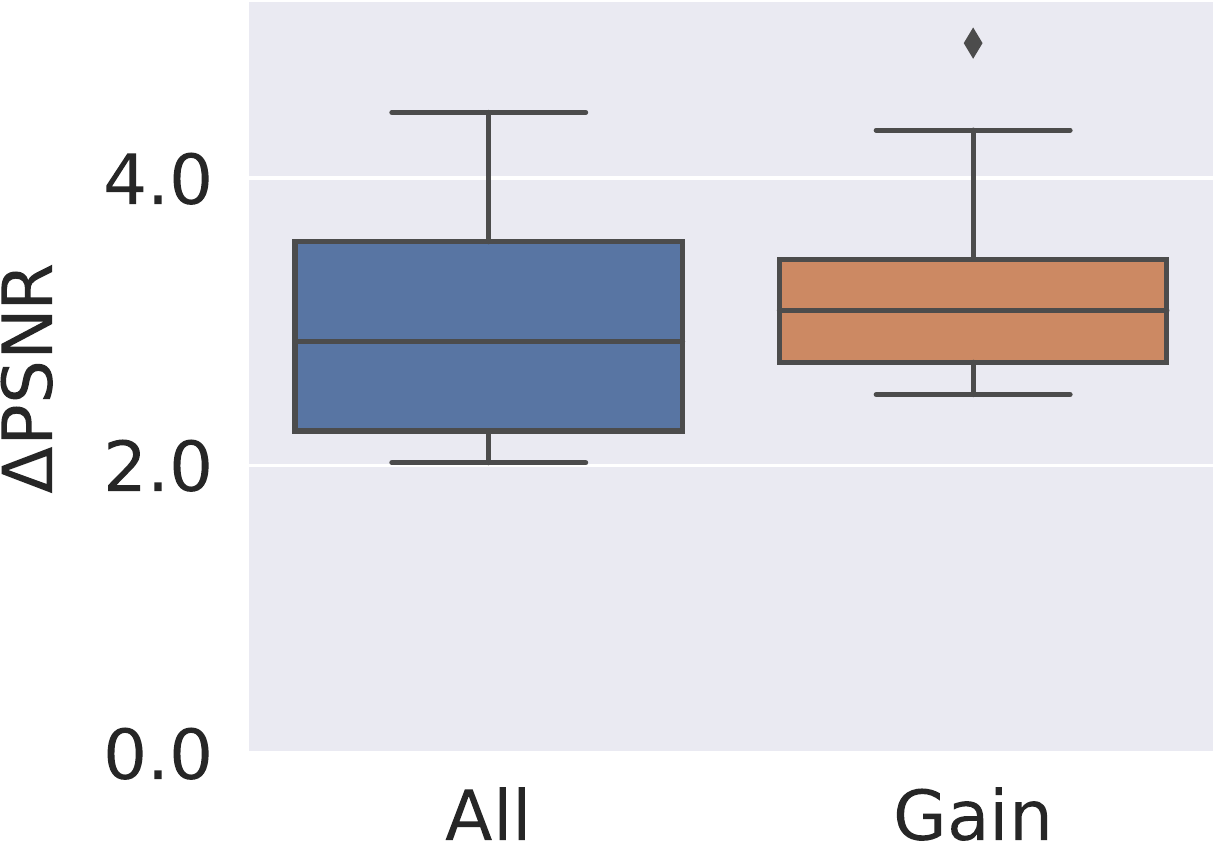}&
    \includegraphics[height=\f1ht]{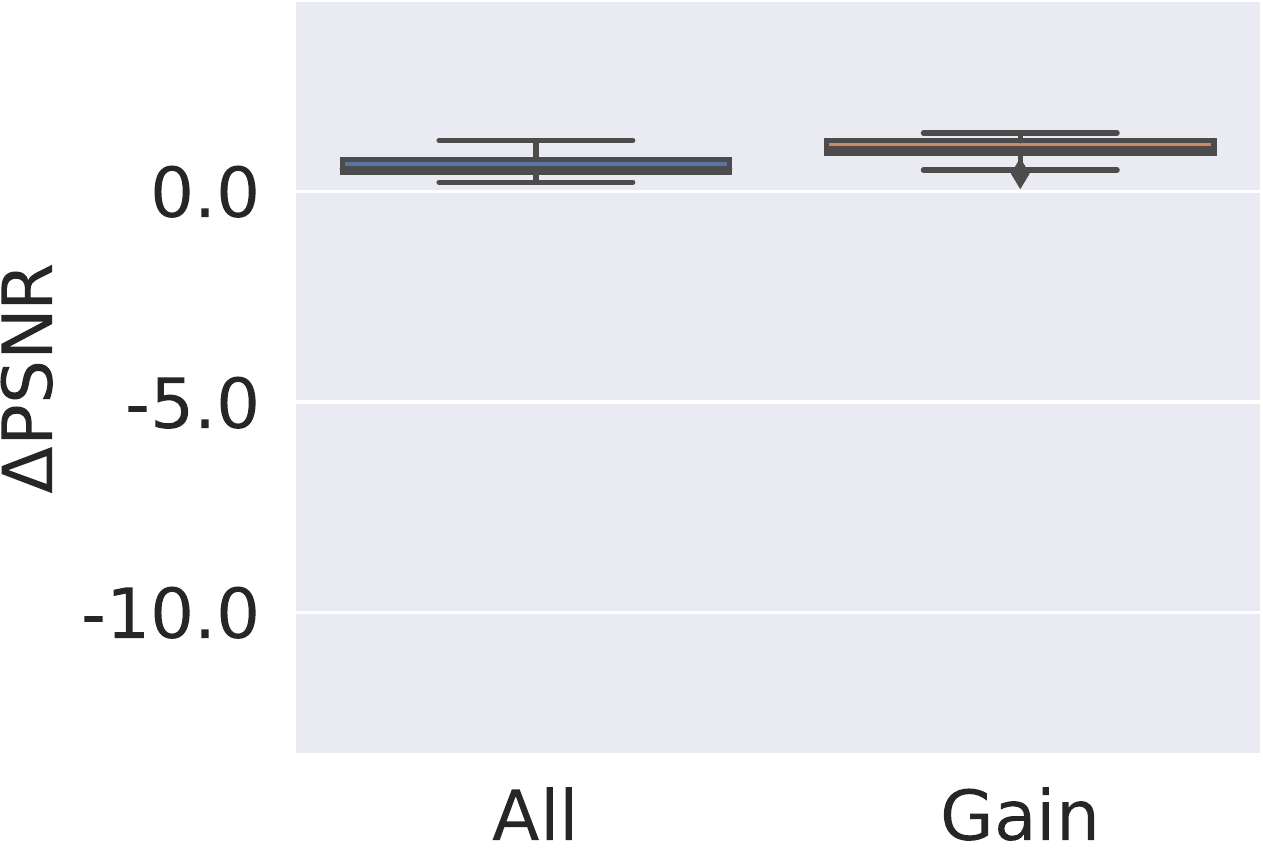}\\
\end{tabular}
}
\caption{\textbf{\gt~prevents overfitting.} Comparison of adaptive training of all network parameters, and \gt\ (training of gains only), using two different unsupervised objectives - SURE (top) and noise resampling (bottom). The distribution of the gain in performance is visualized as a box plot. See Figure~\ref{fig:all_vs_gain_scatter} for a visualization as a scatterplot. For \emph{in-distribution}, we evaluate a CNN pre-trained on natural images corrupted with Gaussian noise of standard deviation $\sigma \in [0, 55]$ on natural images (Set12) at $\sigma=30$. %
For \emph{out-of-distribution noise} we evaluate  natural images (Set12) at $\sigma=70$. For  \emph{out-of-distribution signal} we evaluate a CNN trained on piecewise constant images at $\sigma \in [0, 55]$ on natural images (set12) at $\sigma=30$. Please refer to Section~\ref{sec:suppl_experiments} for details. }
\label{fig:all_vs_gain}
\end{figure}

\begin{figure}
\def\f1ht{1.5in}%
\centering 
\footnotesize{
\begin{tabular}{ >{\centering\arraybackslash}m{.5in} >{\centering\arraybackslash}m{1.4in} >{\centering\arraybackslash}m{1.4in} >{\centering\arraybackslash}m{1.4in}}
\centering 
    Loss  &  In-distribution  & Out-of-distribution noise  & Out-of-distribution signal \\[0.2cm]
    SURE &
    \includegraphics[height=\f1ht]{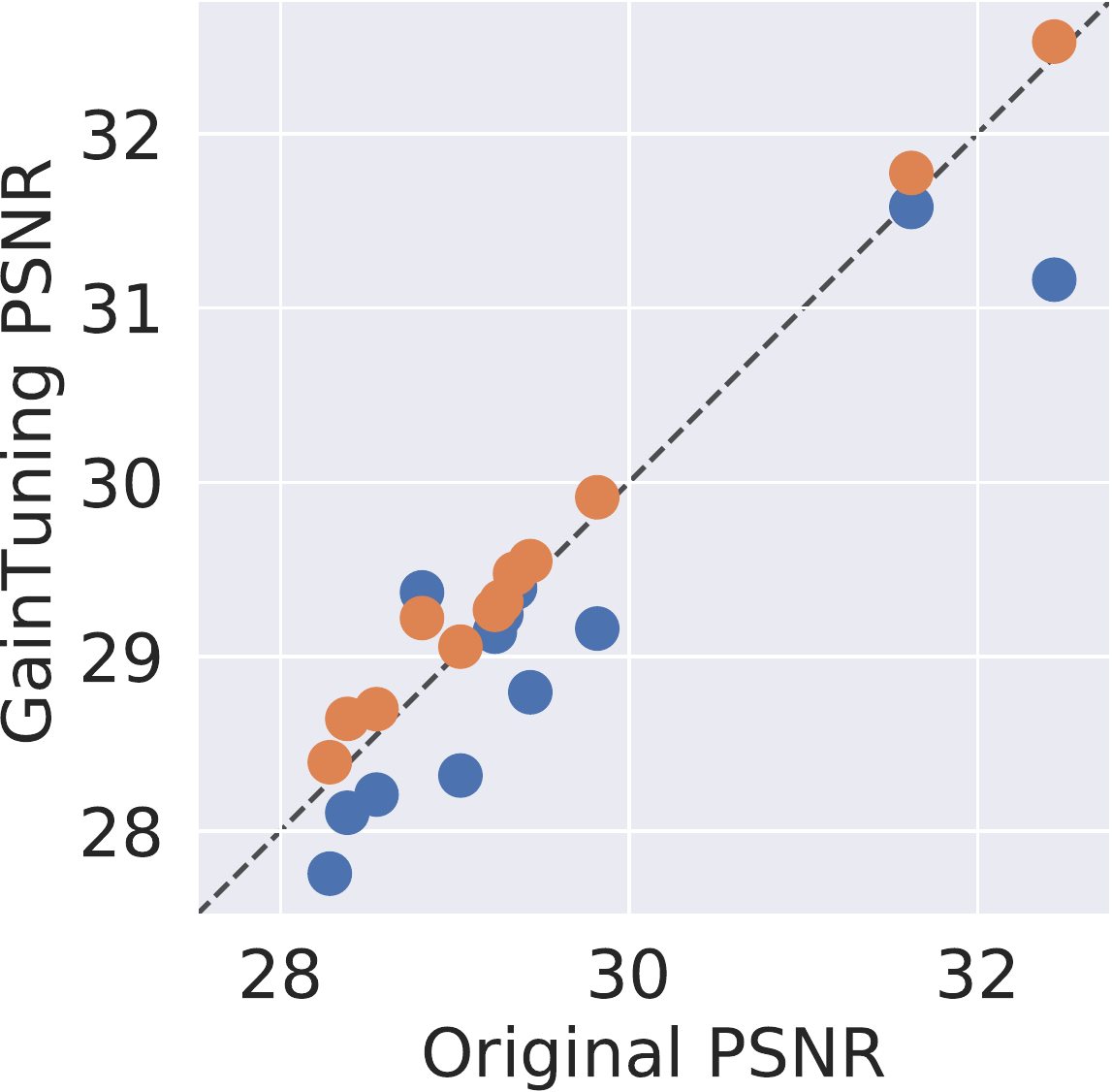}&
    \includegraphics[height=\f1ht]{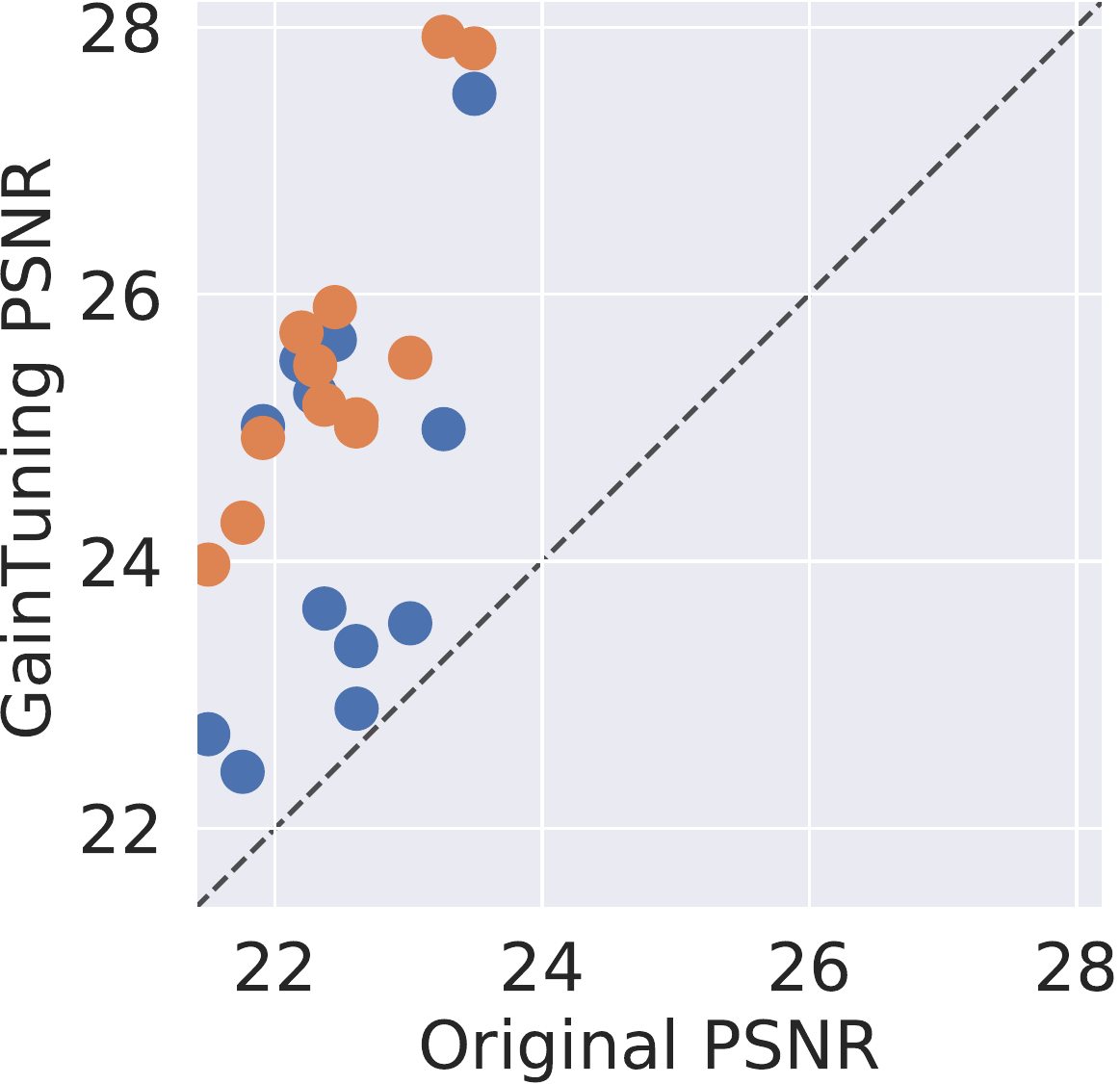}&
    \includegraphics[height=\f1ht]{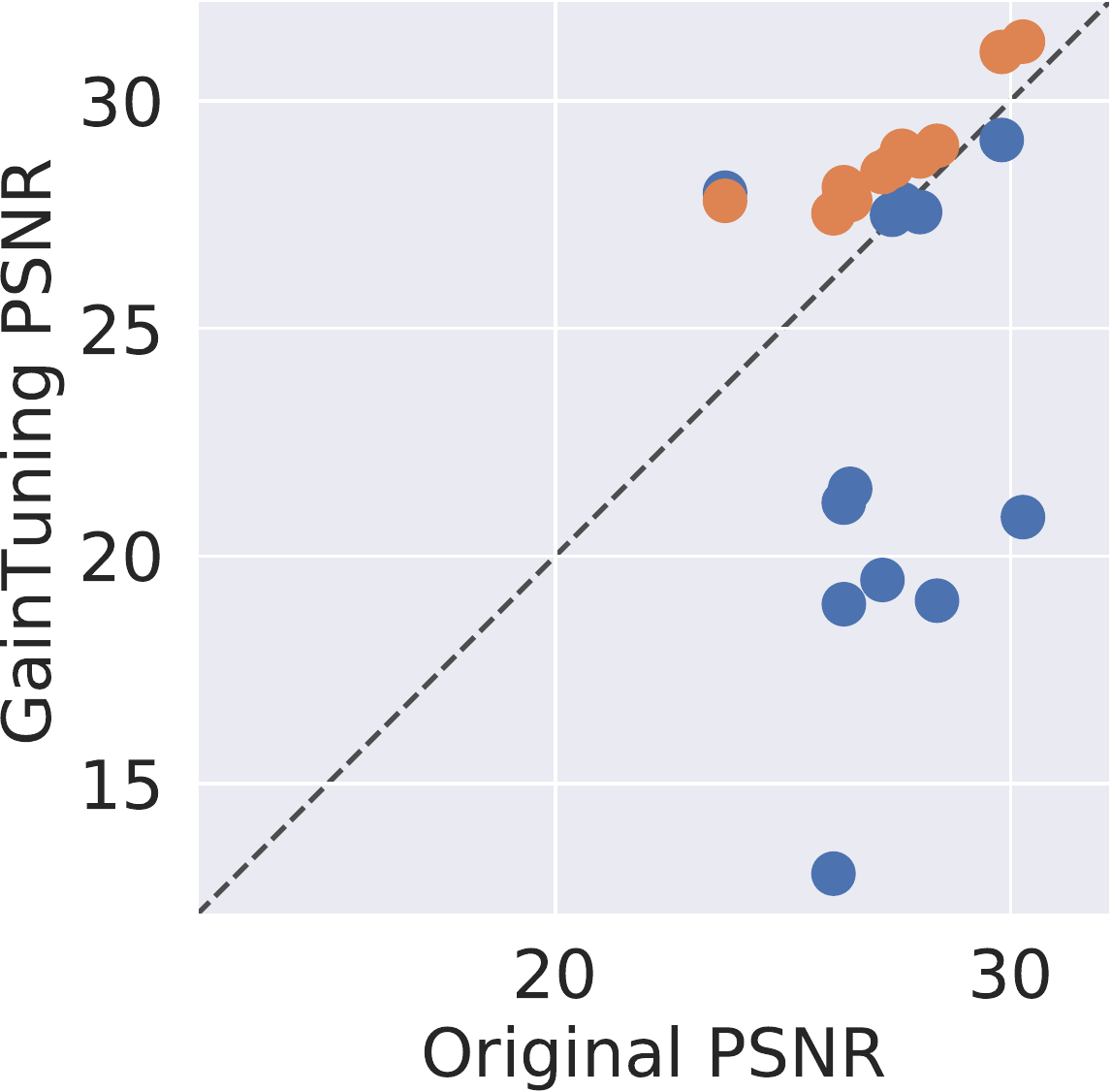}\\
    Noise Resampling &
    \includegraphics[height=\f1ht]{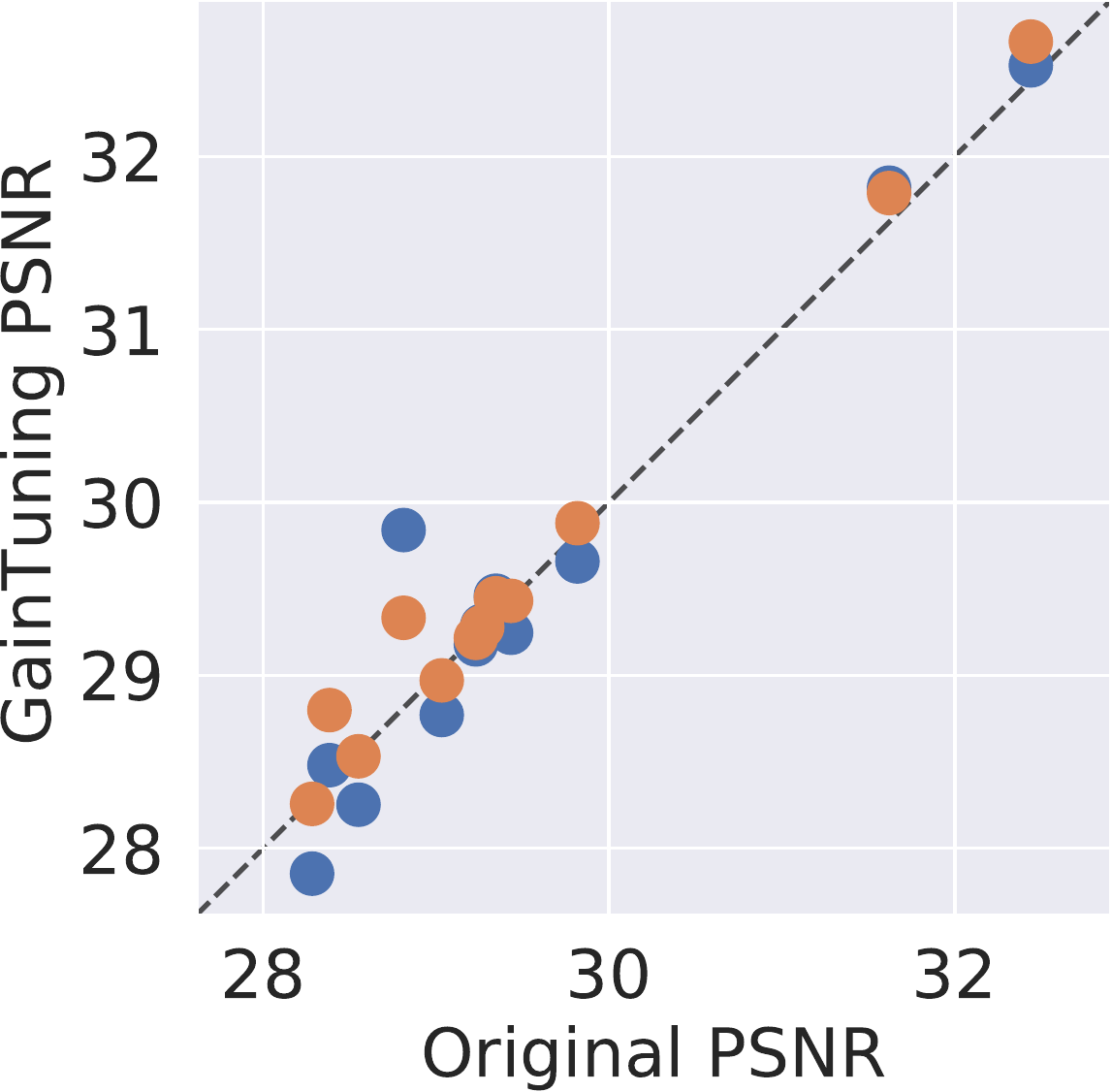}&
    \includegraphics[height=\f1ht]{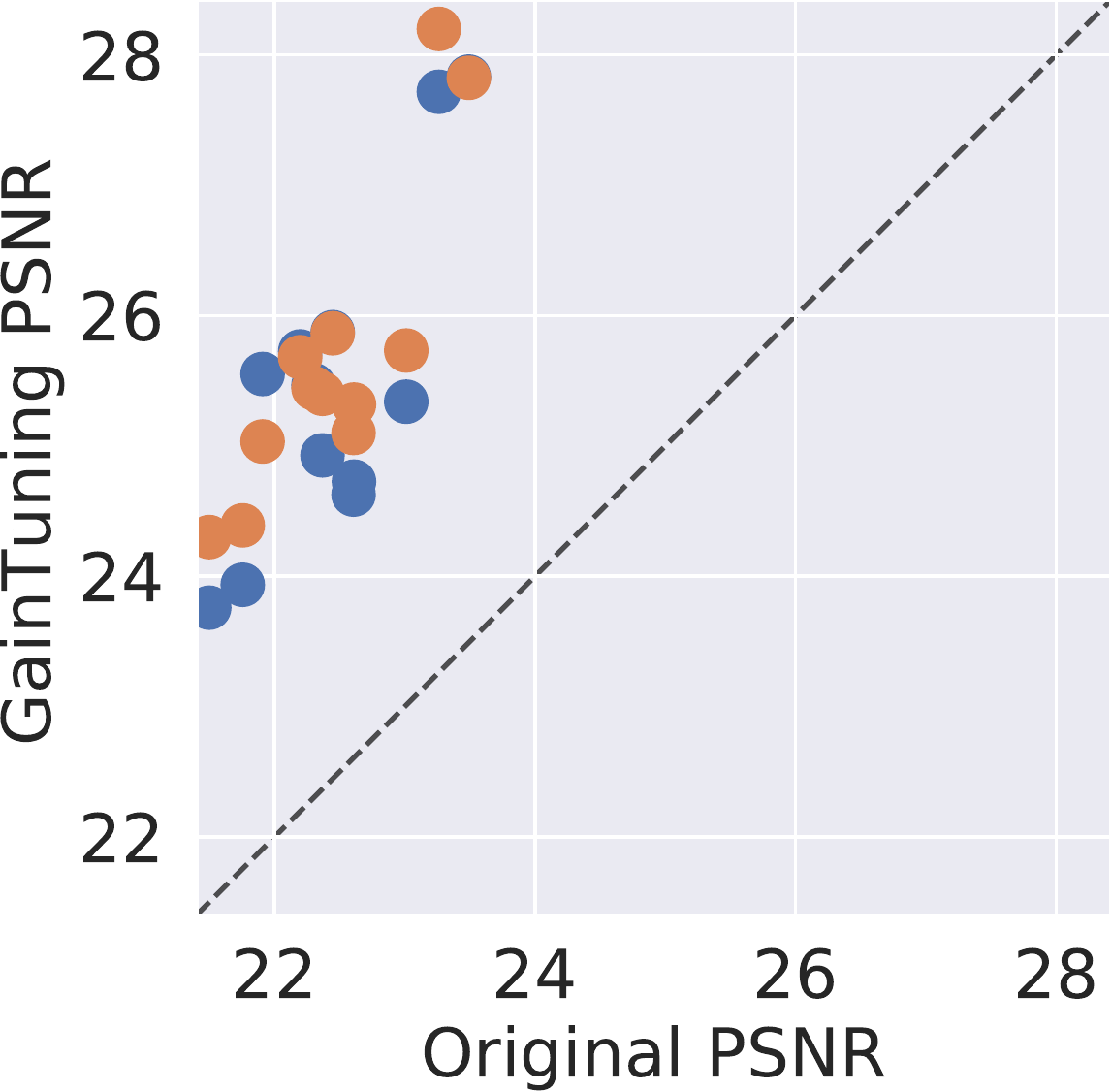}&
    \includegraphics[height=\f1ht]{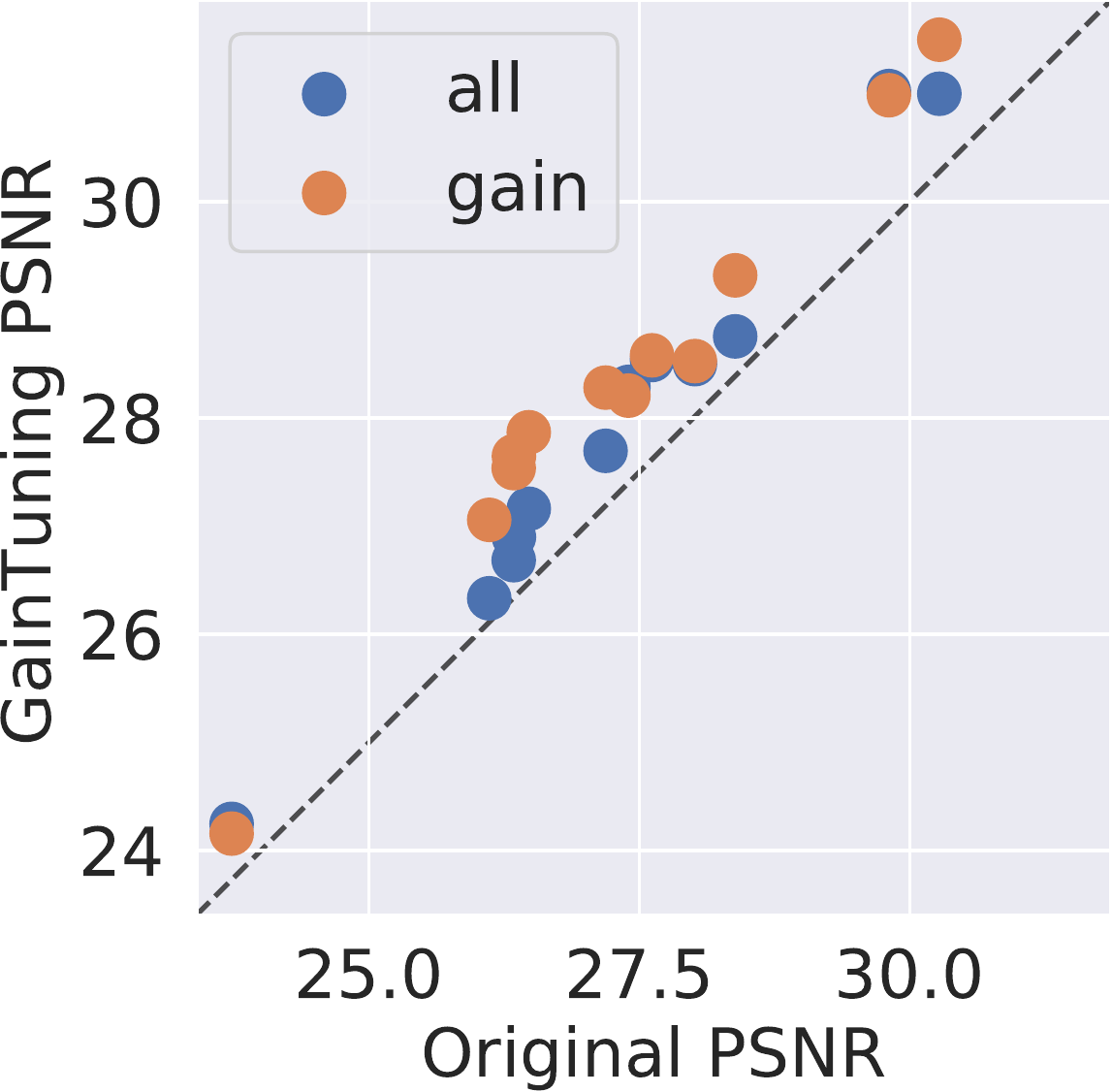}\\
\end{tabular}
}
\caption{\textbf{\gt\ prevents overfitting.} Performance obtained from adaptively training all network parameters (blue points), compared to \gt\ (training of gains only - orange points) using the SURE loss, plotted against performance of the originally trained network. Each data point corresponds to one image in the dataset. The dashed  line represents the identity (i.e., no improvement). Training all parameters (blue points) often leads to degraded performance, but training only the gains (orange points), leads to an improvement. For \emph{in-distribution} test images, we evaluate a CNN pre-trained on natural images corrupted with Gaussian noise of standard deviation $\sigma \in [0, 55]$ on natural images (Set12) at $\sigma=30$. %
For \emph{out-of-distribution noise} we test on natural images (Set12) at $\sigma=70$. 
For  \emph{out-of-distribution signal} we test a CNN trained on piecewise constant images at $\sigma \in [0, 55]$ on natural images (set12) at $\sigma=30$. 
Please refer to Section~\ref{sec:suppl_experiments} for details. }
\label{fig:all_vs_gain_scatter}
\end{figure}

\textbf{Comparison across different cost functions}. We fine-tune (a) all parameters, and (b) only gain parameters of a DnCNN~\cite{dncnn} model when the test image is (1) in-distribution, (2) corrupted with out-of-distribution noise and (c) contains image features which are different from the training set. Fine-tuning only the gain parameters outperforms fine-tuning all parameters in all of these situations for different choices of cost functions (see Figure~\ref{fig:all_vs_gain} for summary statistics as a box plot and Figure~\ref{fig:all_vs_gain_scatter} for improvements on individual data points visualized as a scatter plot)

\textbf{Comparison across different architectures}. We fine-tune (a) all parameters, and (b) only gain parameters of a DnCNN~\cite{dncnn}, BFCNN~\cite{biasfree} and, UNet~\cite{unet} model when the test image is (a) in-distribution, (b) corrupted with out-of-distribution noise and (c) contains image features which are different from the training set. Fine-tuning only the gain parameters often outperforms fine-tuning all parameters in all of these situations for different choices of cost functions (see Figure~\ref{fig:all_vs_gain_suppl}). Figure~\ref{fig:all_vs_gain_suppl} shows results for a CNN trained on generic natural images and tested on images of urban scenes. In this case, training all parameters of the CNN outperforms training only the gains (see Section~\ref{sec:limitations} for a discussion). Interestingly, training gains is comparable to training all parameters when we corrupt the images from urban scenes with a noise level that is also outside the training range (see Figure~\ref{fig:suppl_out_noise_signal}). 

\textbf{\gt\ does not require early stopping}. Optimizing all parameters of a CNN during adaptation often results in overfitting (see Figure~\ref{fig:all_vs_gain_suppl}). In contrast,  optimizing only the gain parameters for longer periods of time results improves performance without overfitting (Figure\ref{fig:training_curve_dncnn}). 

\textbf{Real electron microscopy data}. We fine-tune (a) all parameters, and (b) gain parameters to adapt a CNN to real images of nanoparticle acquired through an electron microscope. The CNN was pre-trained on the simulated data described in Section~\ref{sec:datasets}. Optimizing only the gain parameters outperforms optimizing all parameter and does not require early stopping (Figure~\ref{fig:all_vs_gain_nano_blindspot})

\def\f1ht{1.25in}%
\begin{figure}
\centering 
\footnotesize{
\begin{tabular}{c@{\hskip 0.1in}c@{\hskip 0.1in}c}
      \multicolumn{3}{c}{In distribution. Natural images ($\sigma \in [0, 55]$) $\rightarrow$ Set12 ($\sigma=30$) } \\[0.1 cm]
    \includegraphics[height=\f1ht]{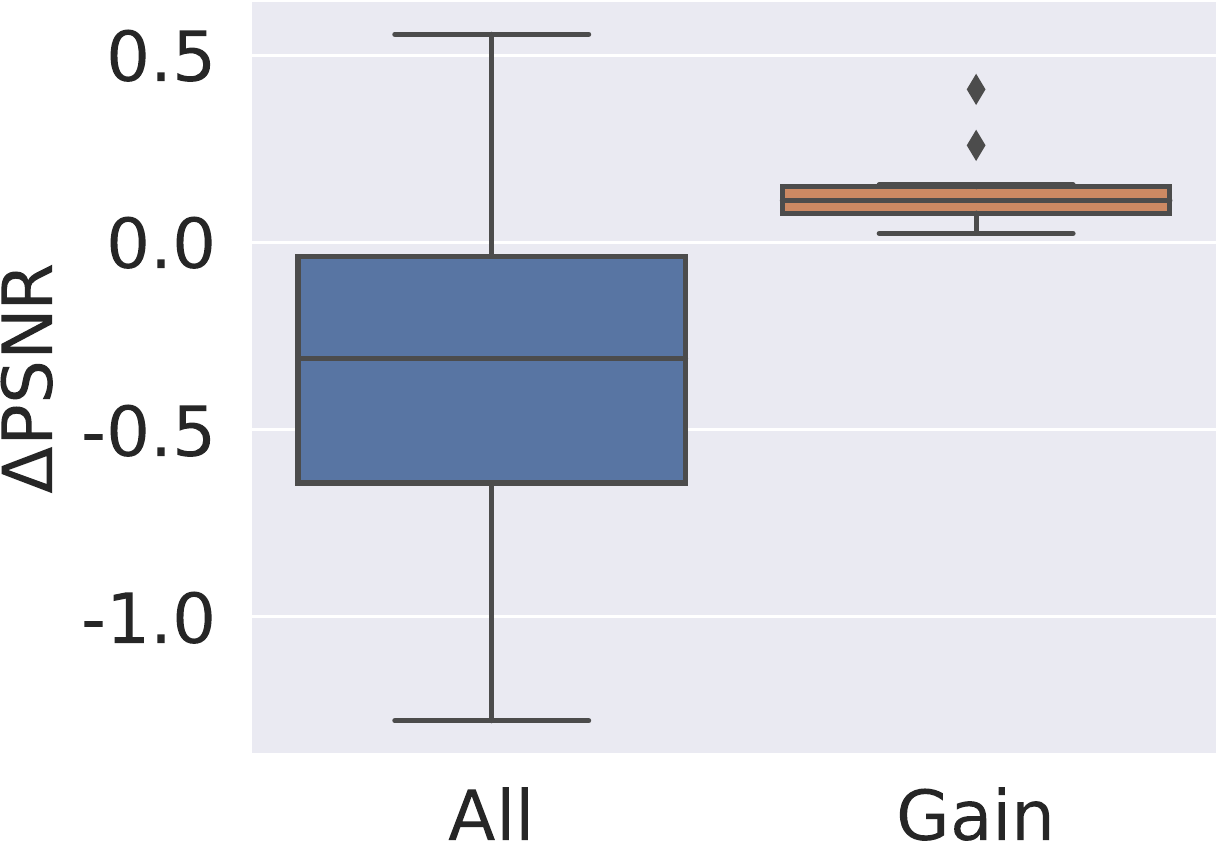}&
    \includegraphics[height=\f1ht]{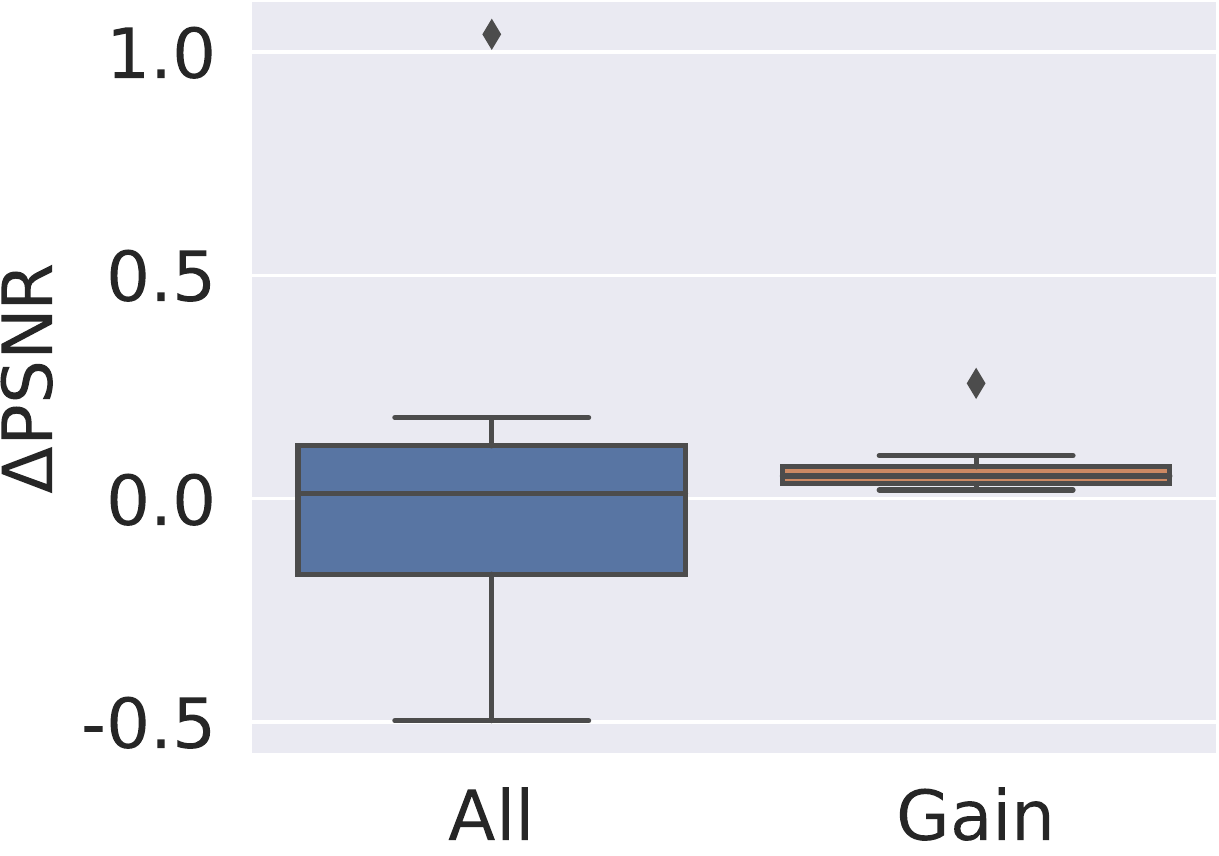}&
    \includegraphics[height=\f1ht]{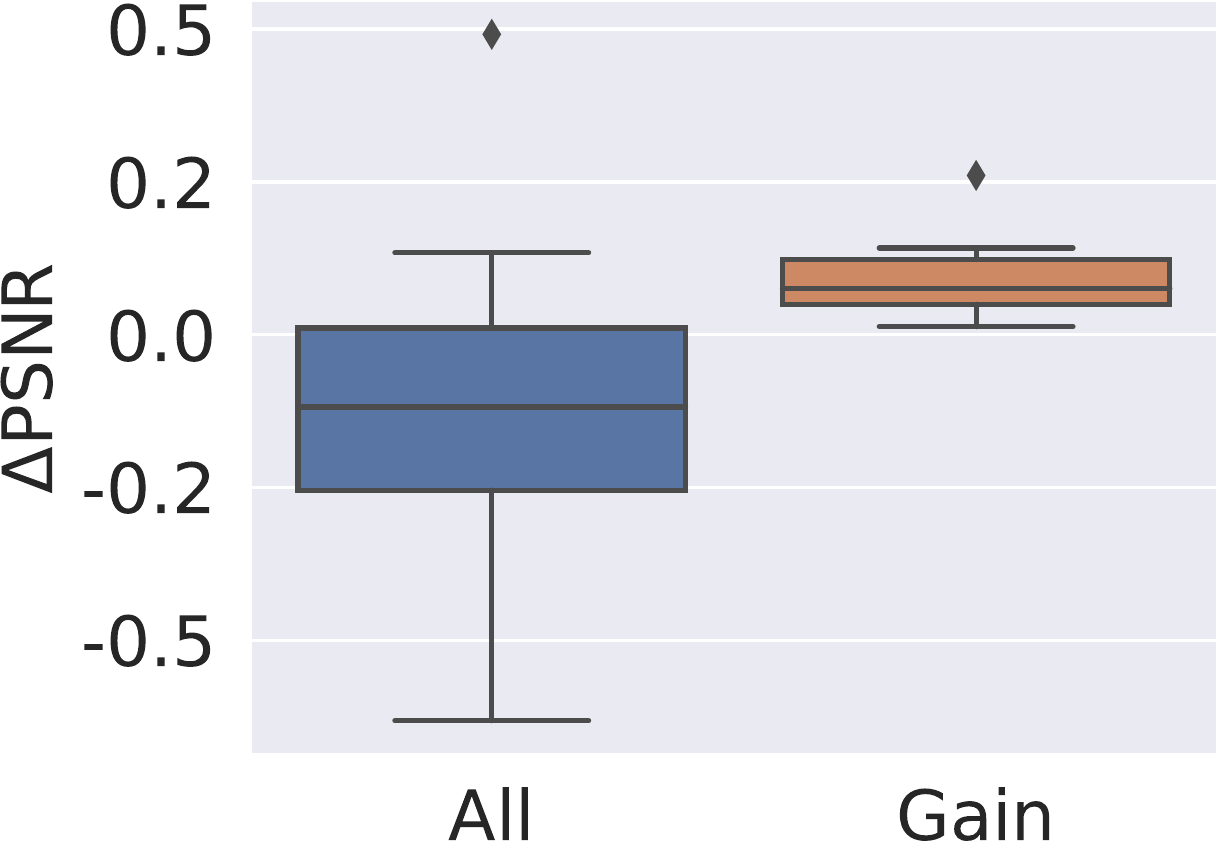}\\
    \includegraphics[height=1.35in]{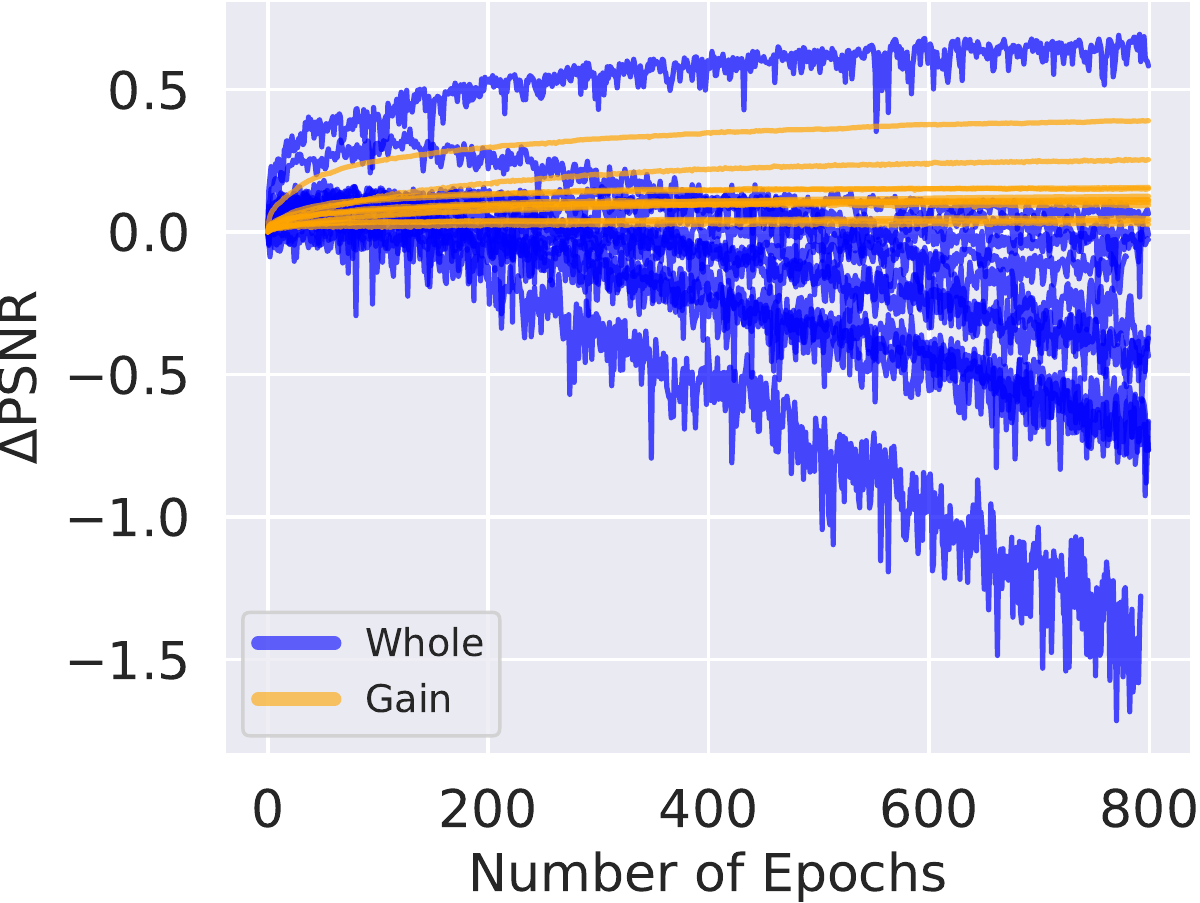}&
    \includegraphics[height=1.35in]{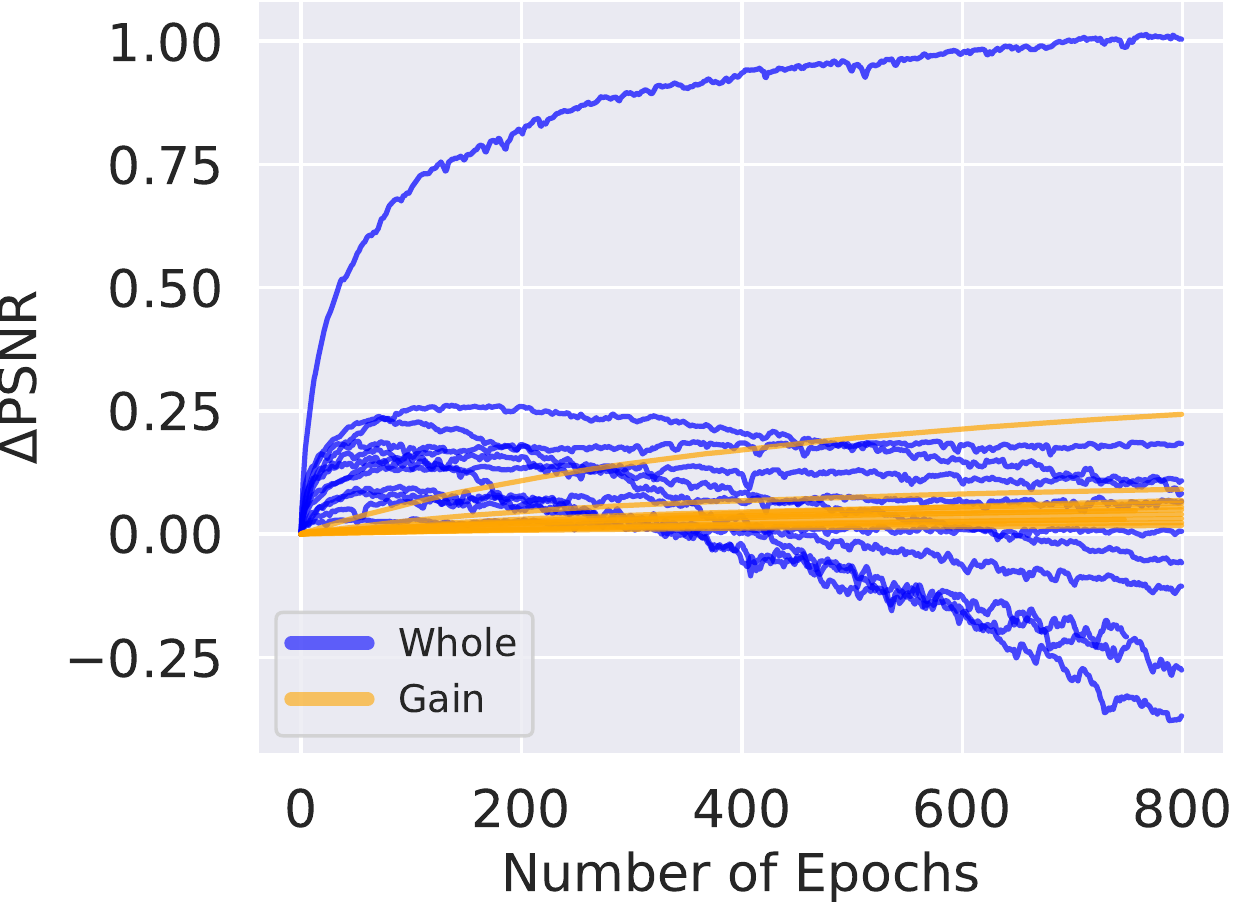}&
    \includegraphics[height=1.35in]{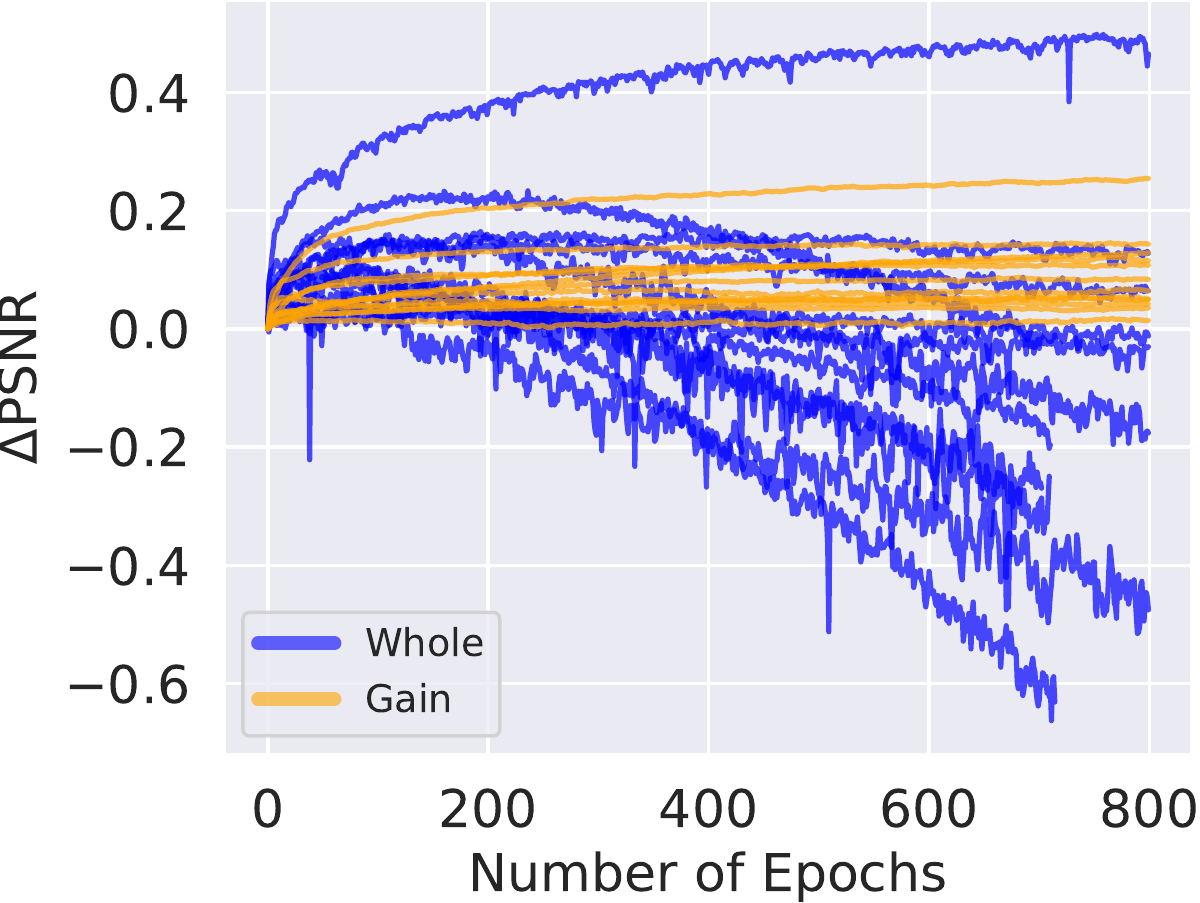}\\
    
    \midrule
    
    \multicolumn{3}{c}{Out-of-distribution noise.  Natural images ($\sigma \in [0, 55]$) $\rightarrow$ Set12 ($\sigma=70$)} \\[0.1 cm]
    \includegraphics[height=\f1ht]{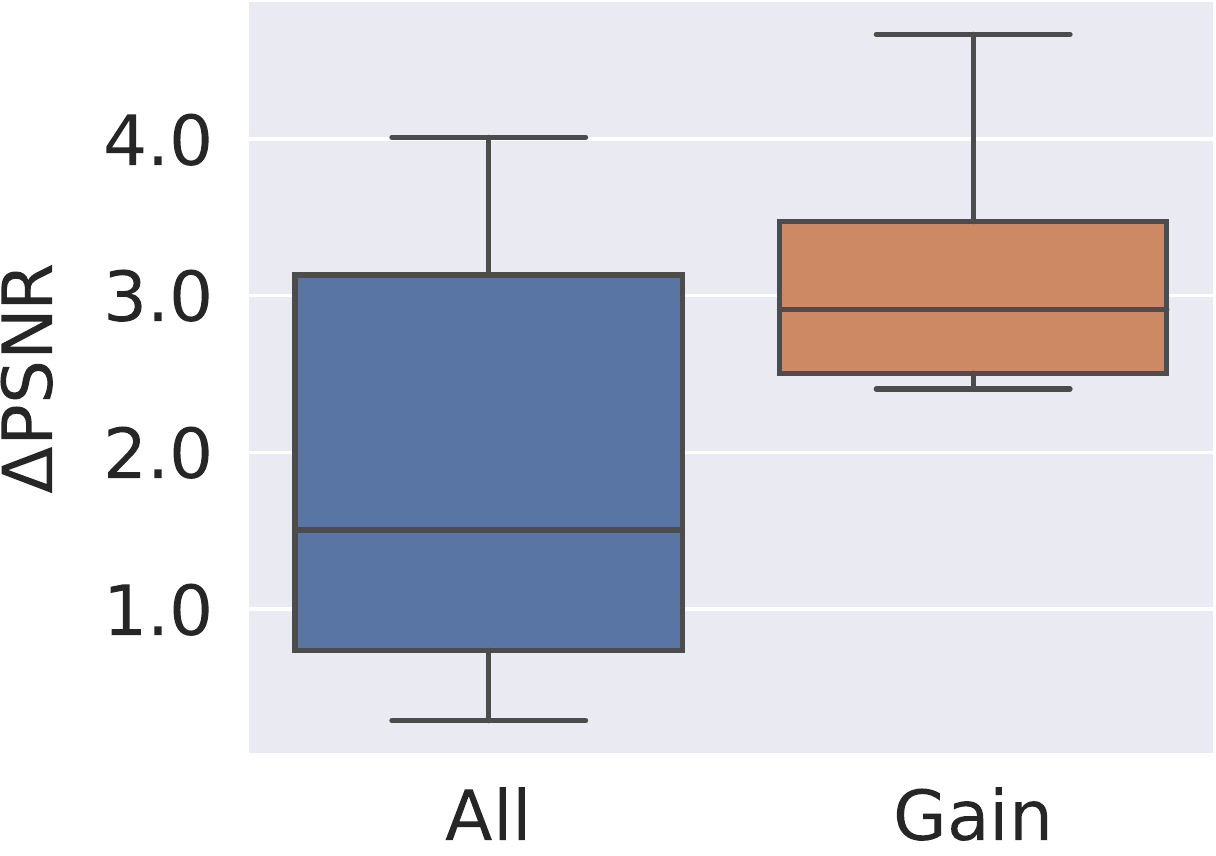}&
    \includegraphics[height=\f1ht]{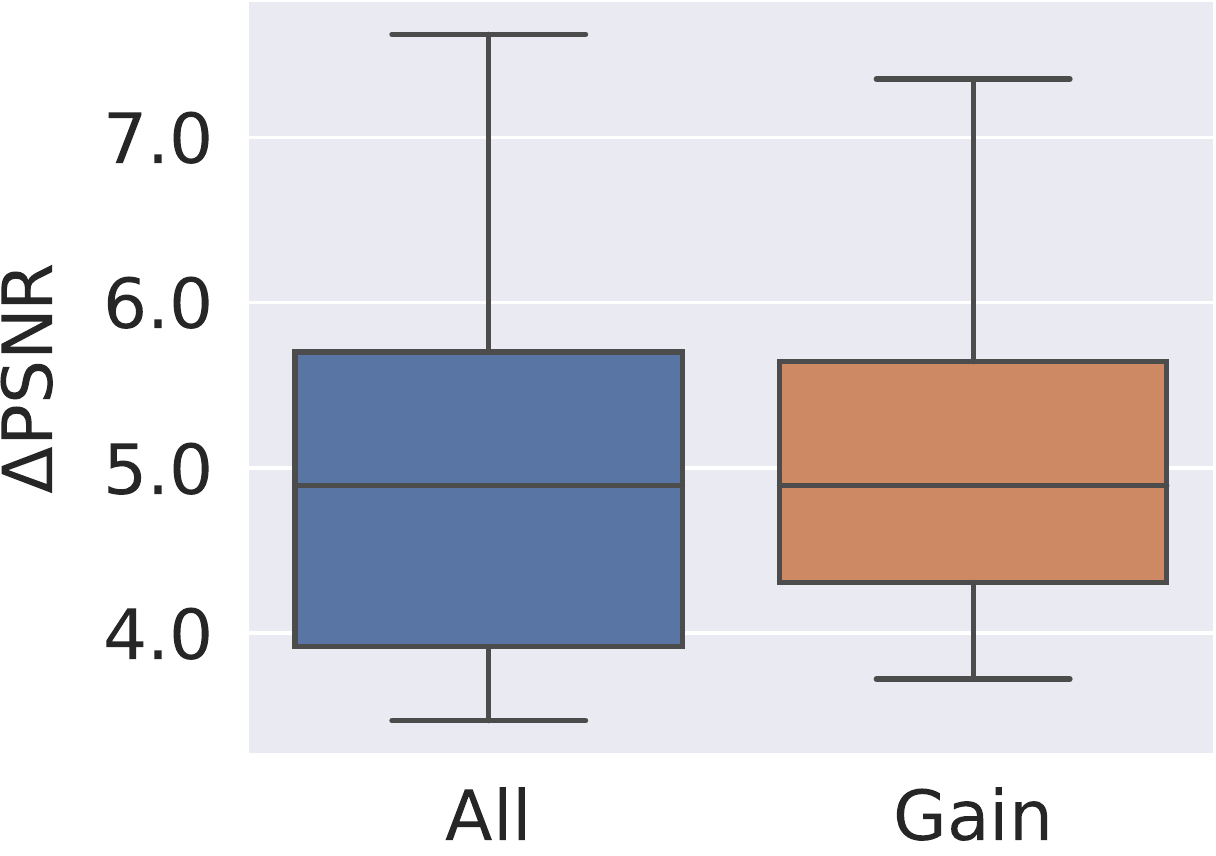}&
    \includegraphics[height=\f1ht]{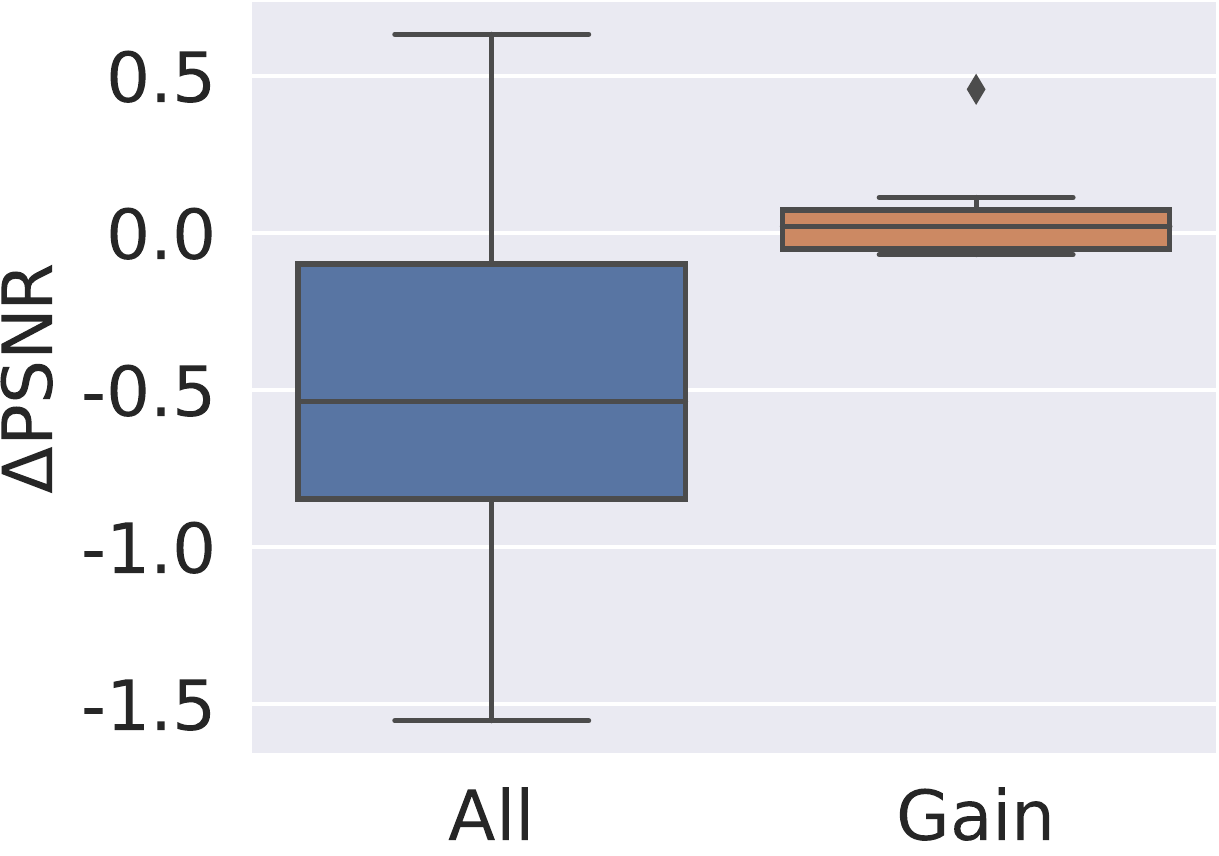}\\
    \includegraphics[height=1.35in]{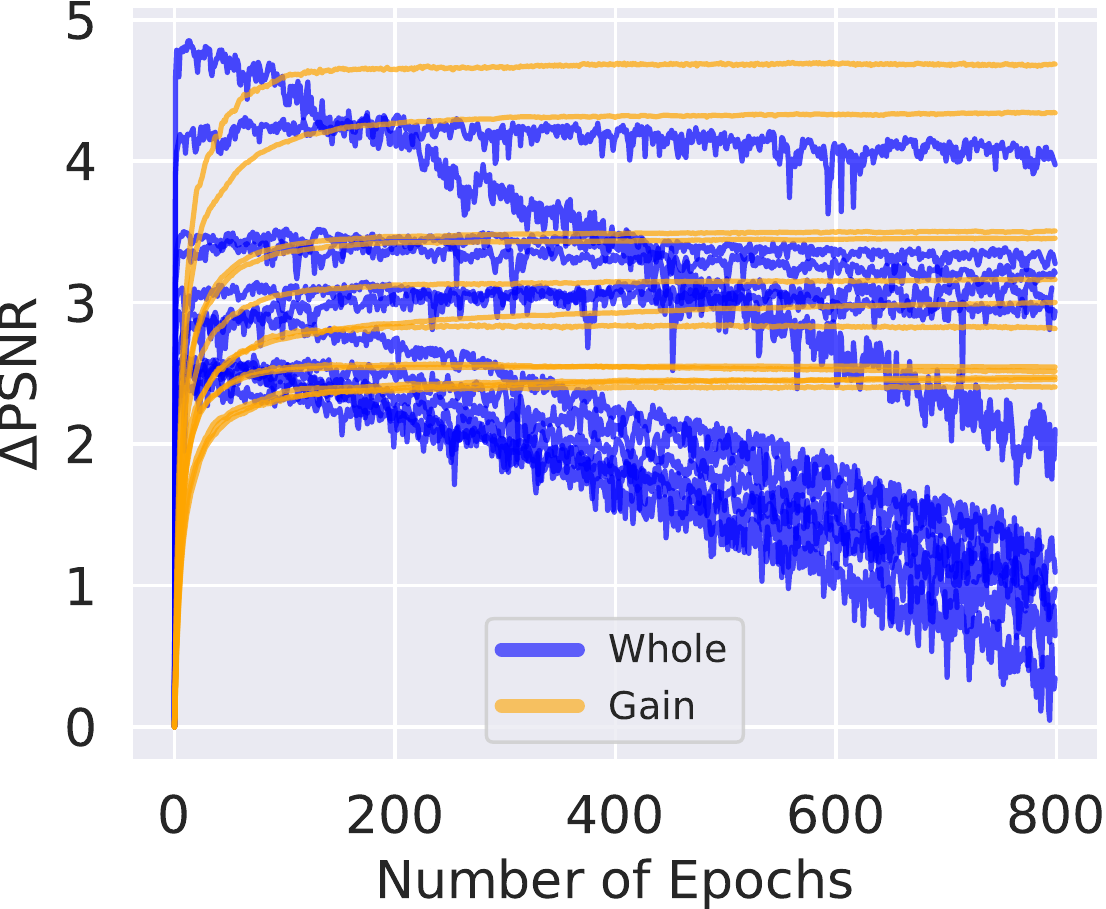}&
    \includegraphics[height=1.35in]{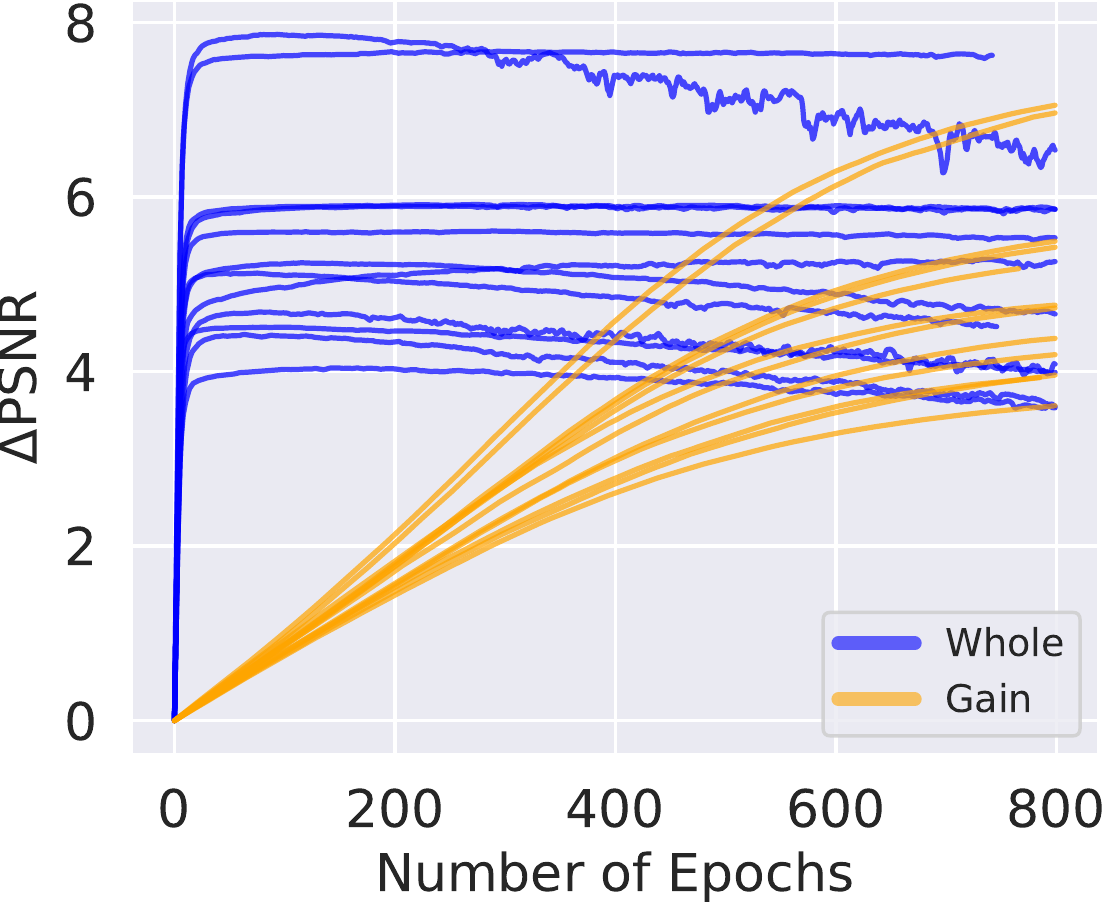}&
    \includegraphics[height=1.35in]{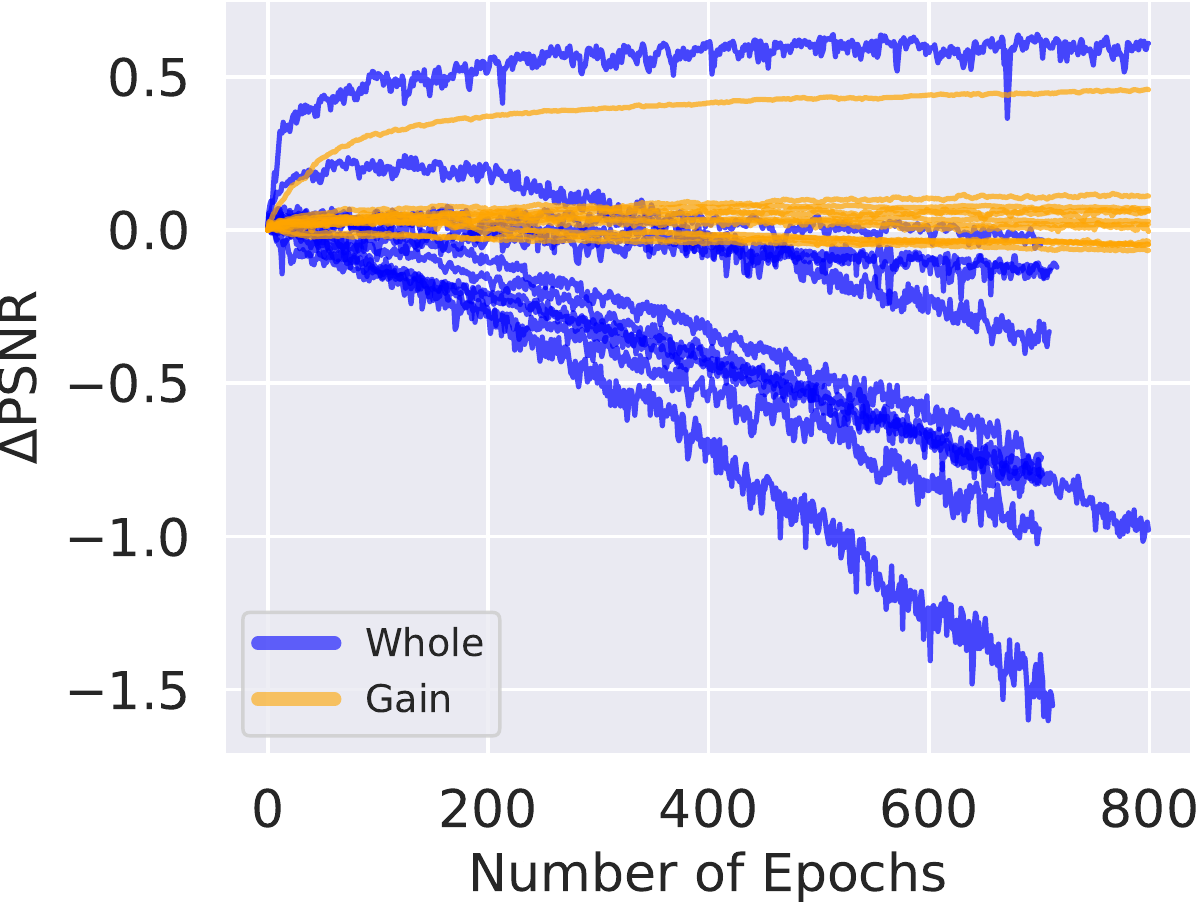}\\
    \midrule

    \multicolumn{3}{c}{Out-of-distribution signal. Natural images ($\sigma \in [0, 55]$) $\rightarrow$ Urban100 ($\sigma=30$)} \\[0.1 cm]
    \includegraphics[height=\f1ht]{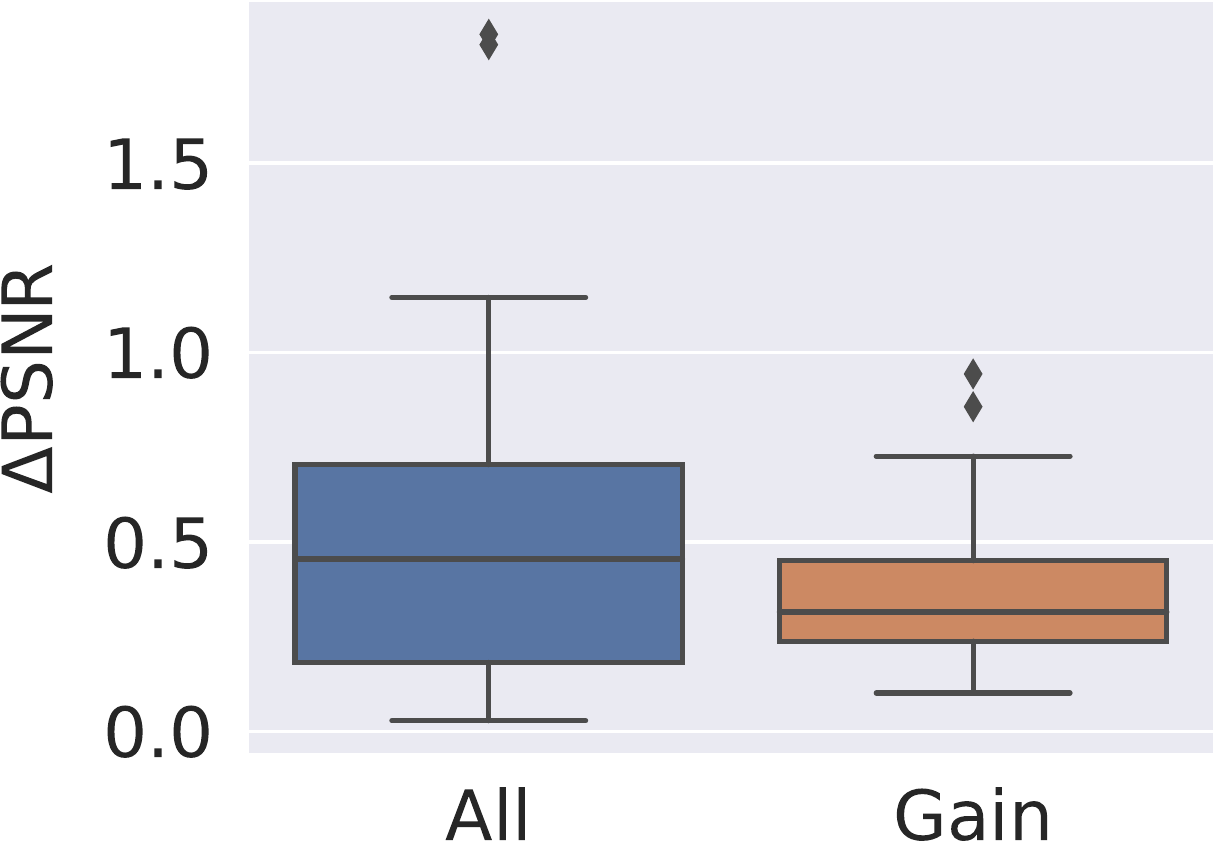}&
    \includegraphics[height=\f1ht]{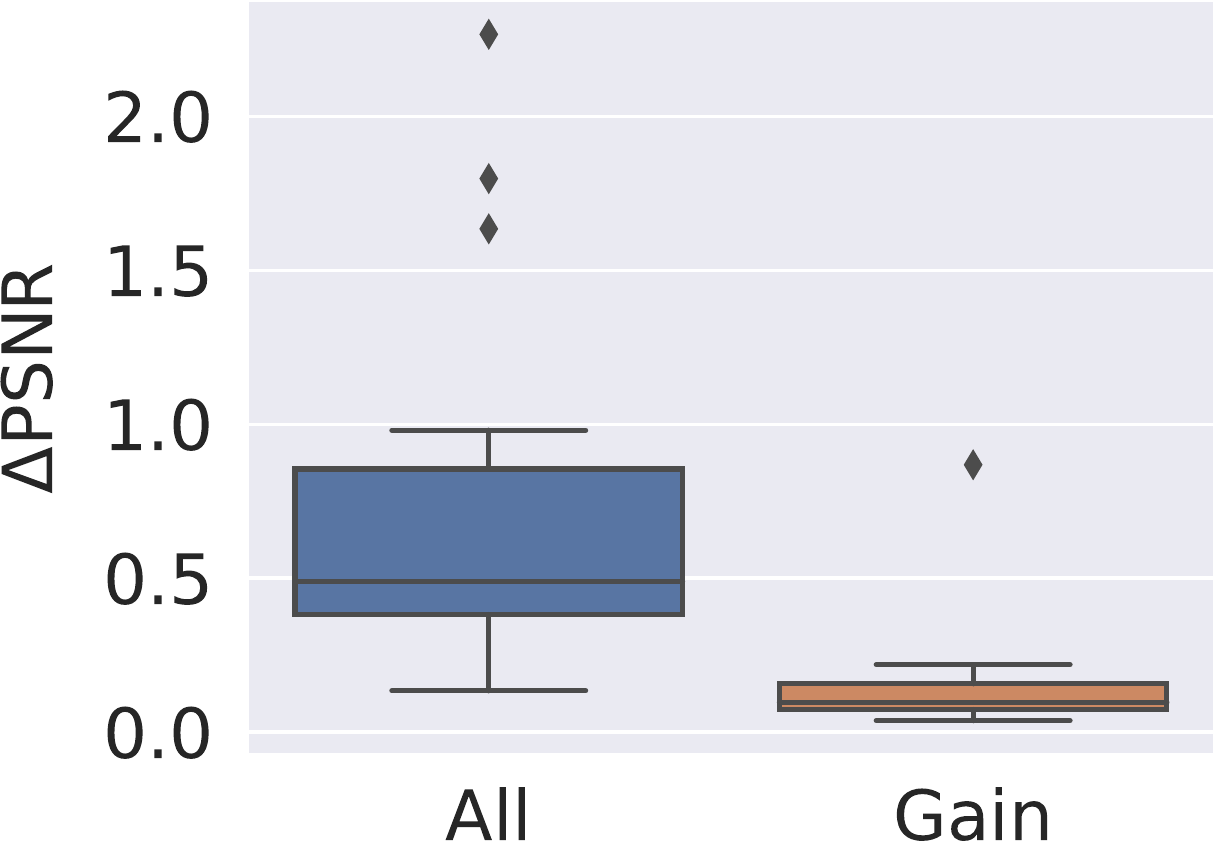}&
    \includegraphics[height=\f1ht]{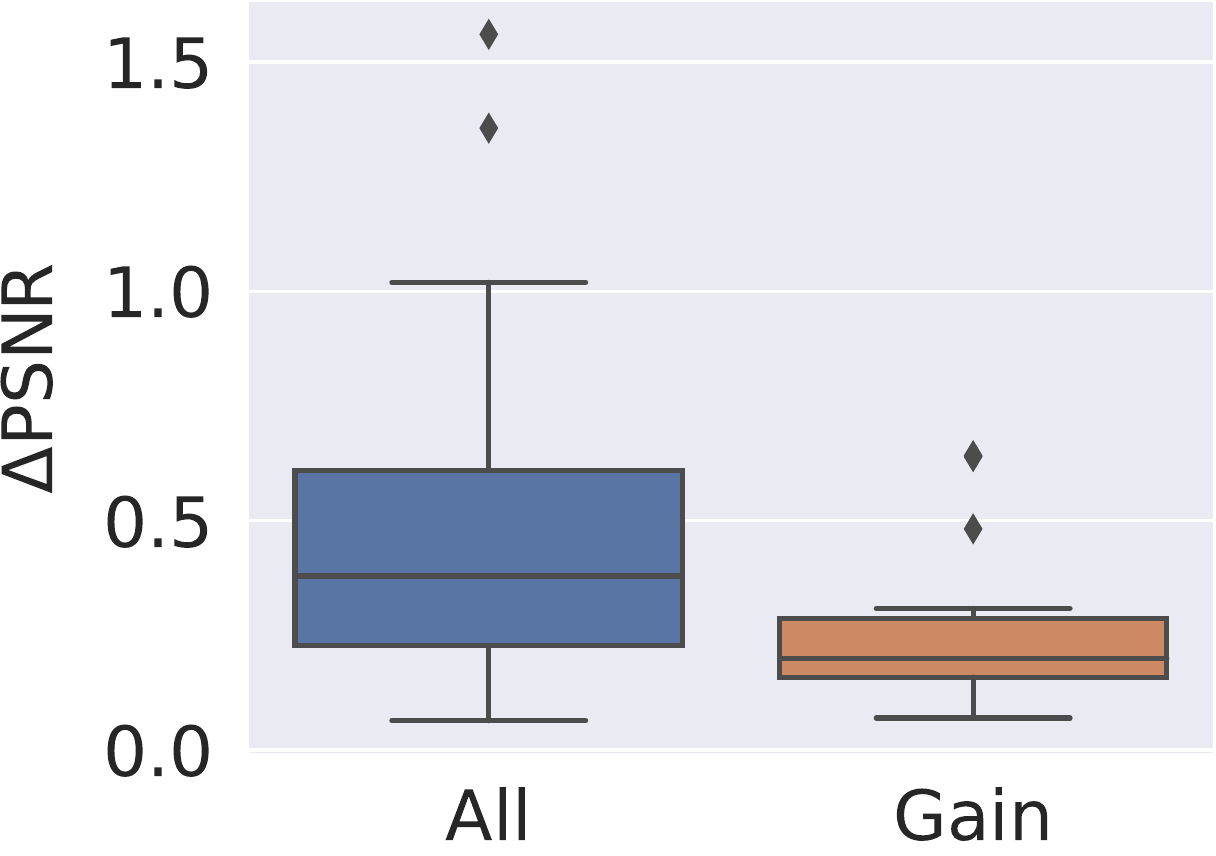}\\
    \includegraphics[height=1.35in]{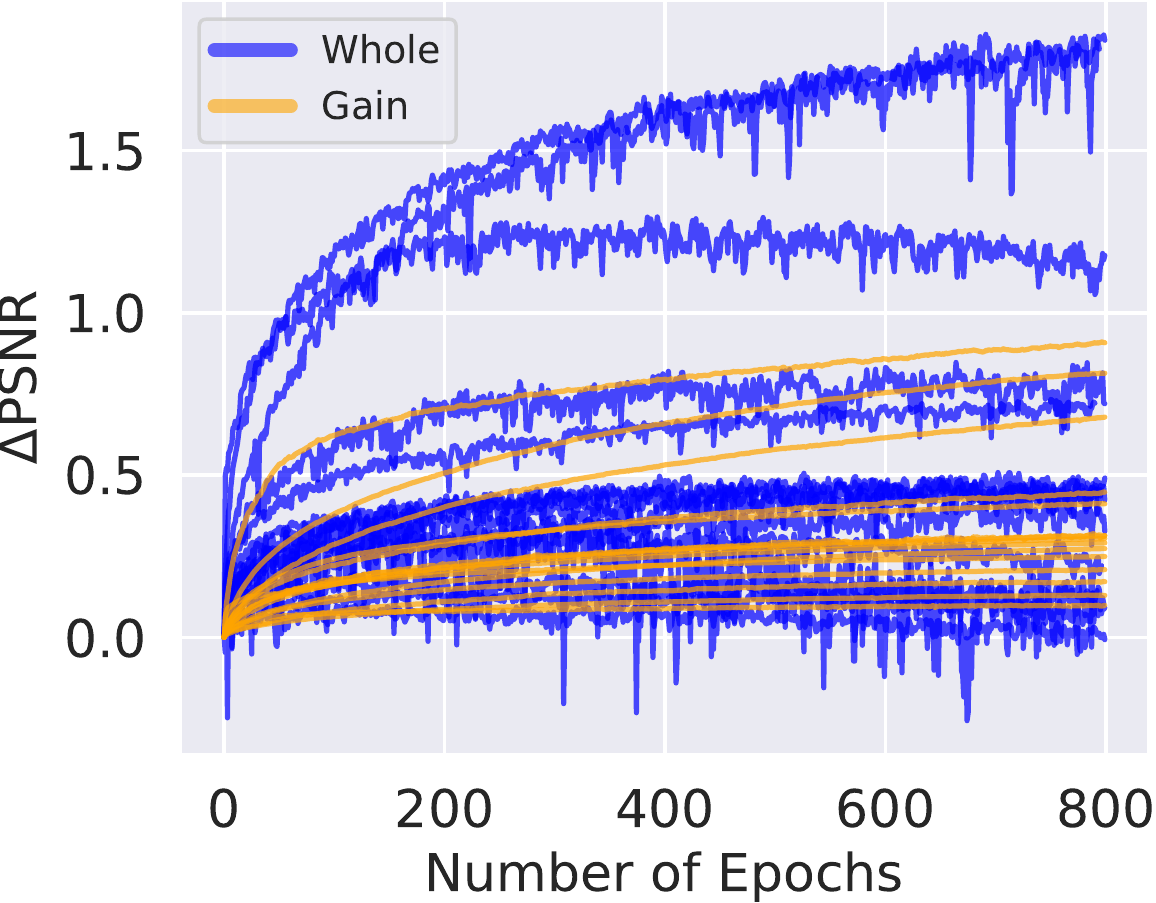}&
    \includegraphics[height=1.35in]{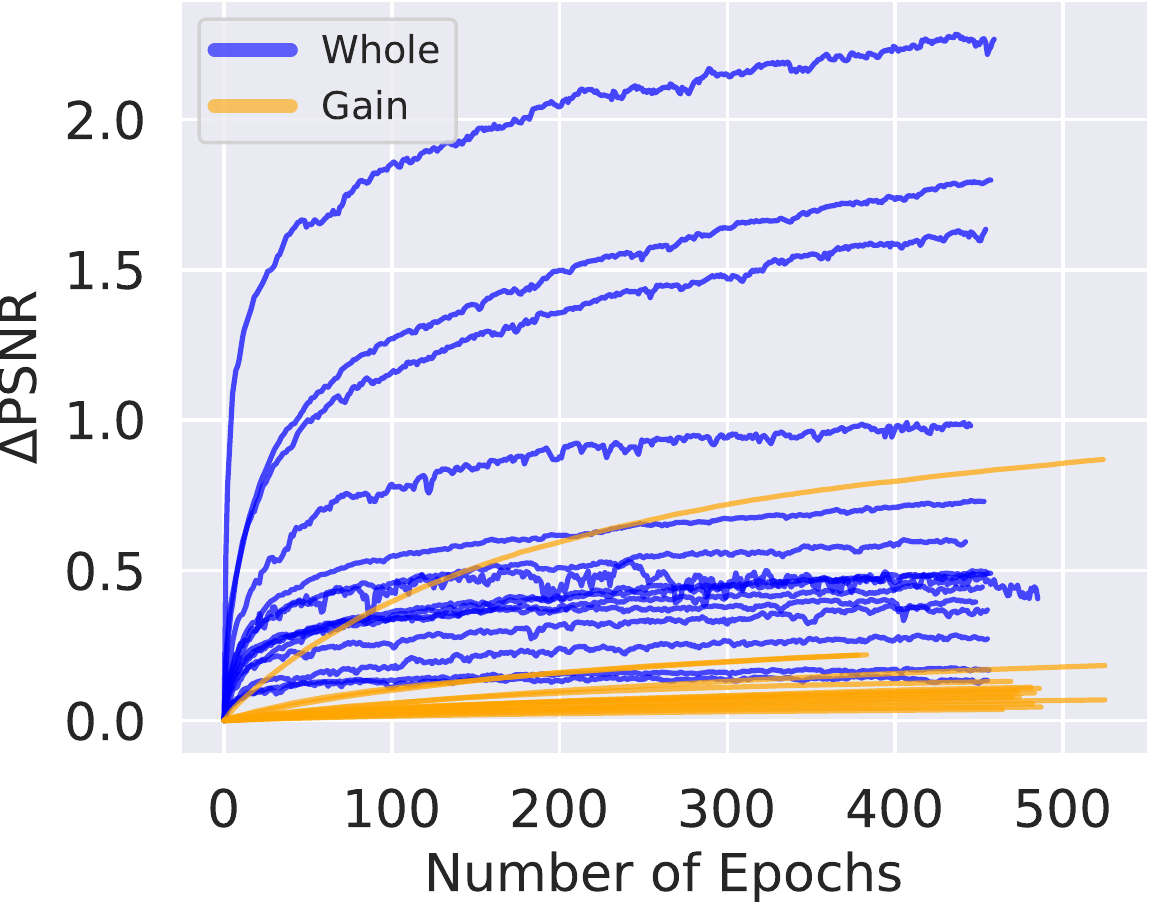}&
    \includegraphics[height=1.35in]{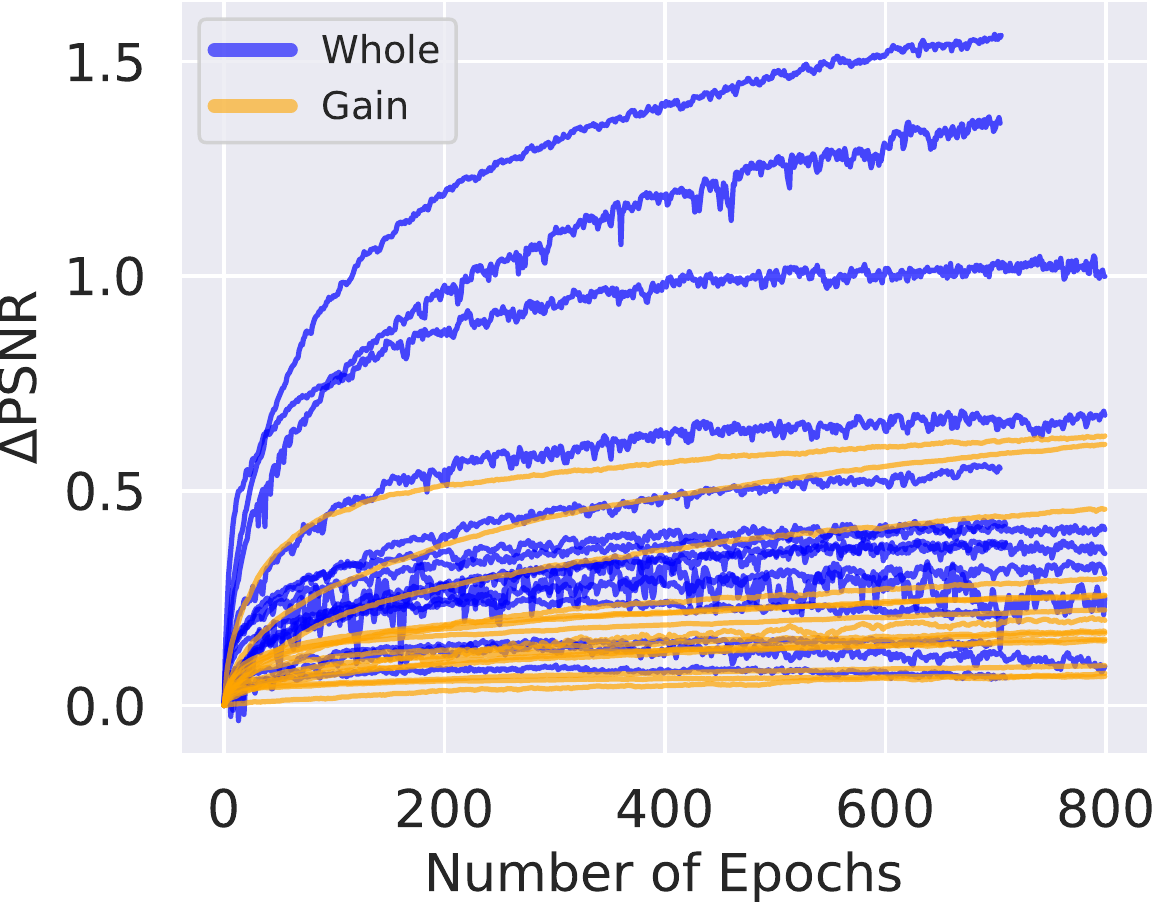}\\

     DnCNN  & UNet  & BFCNN \\[-0.25cm]
    
\end{tabular}
}
\caption{\textbf{\gt\ prevents overfitting}. We compare training all parameters of the network (blue) and only the gain parameters (orange) during the adaptation process.  All architectures are trained using the SURE cost function. }

\label{fig:all_vs_gain_suppl}
\end{figure}

\def\f1ht{2.2in}%
\begin{figure}
\centering 
\footnotesize{
\begin{tabular}{c@{\hskip 0.5in}c}
DnCNN  &  BFCNN \\[0.3cm]
\includegraphics[width=\f1ht]{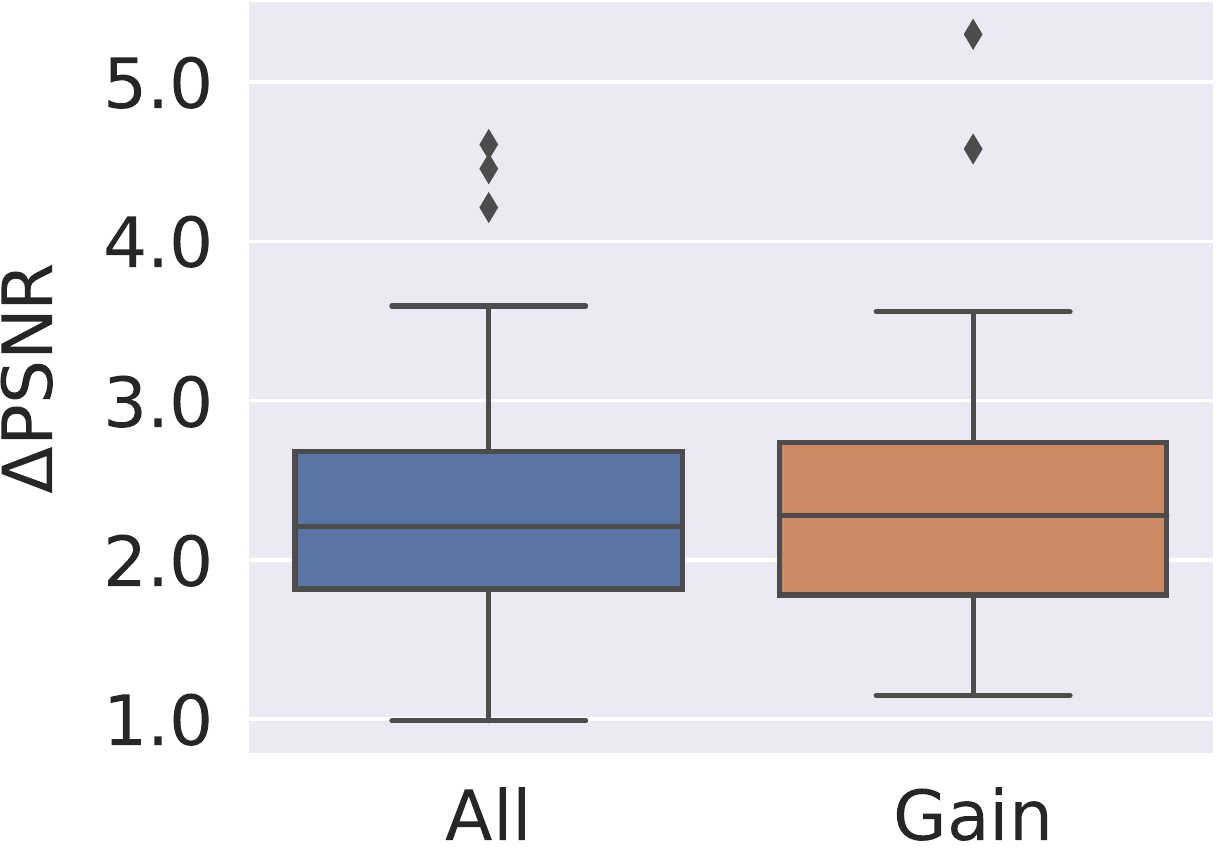}&
\includegraphics[width=\f1ht]{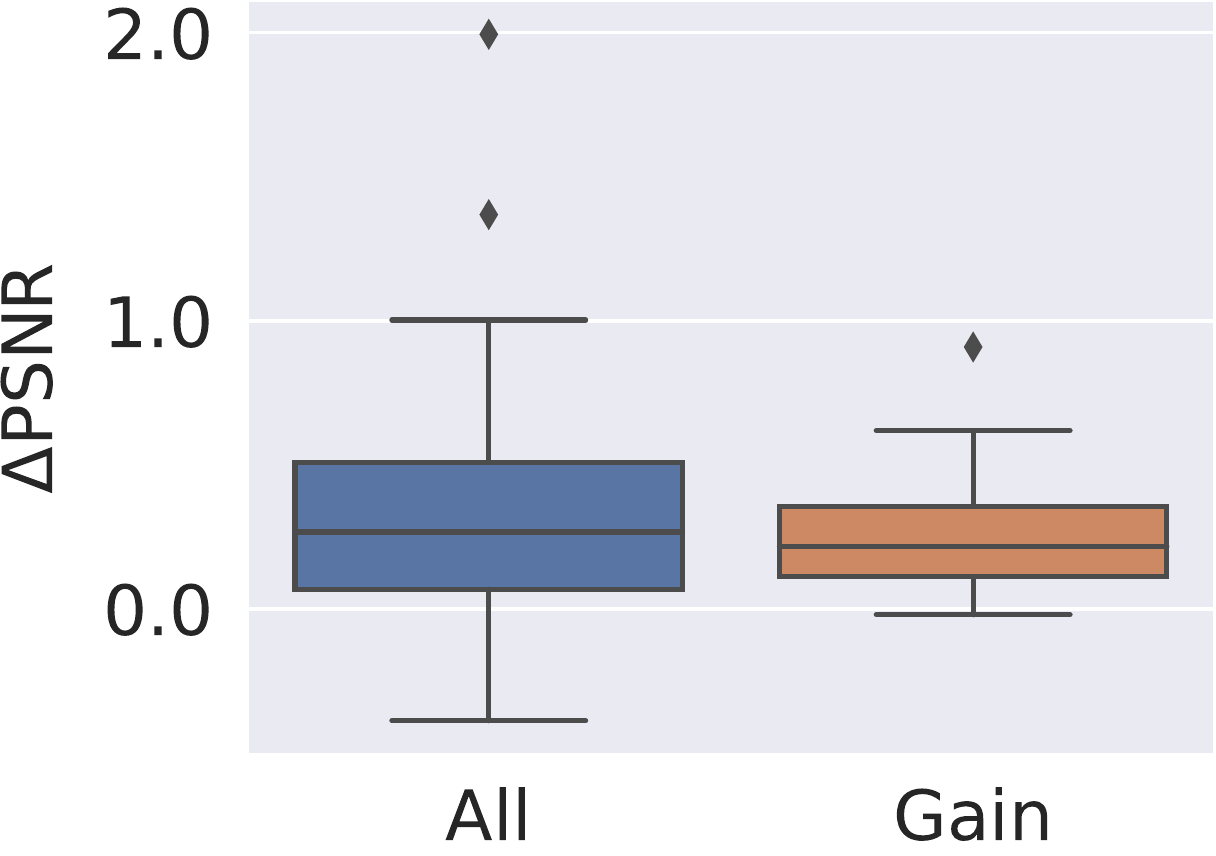}\\

 \includegraphics[width=\f1ht]{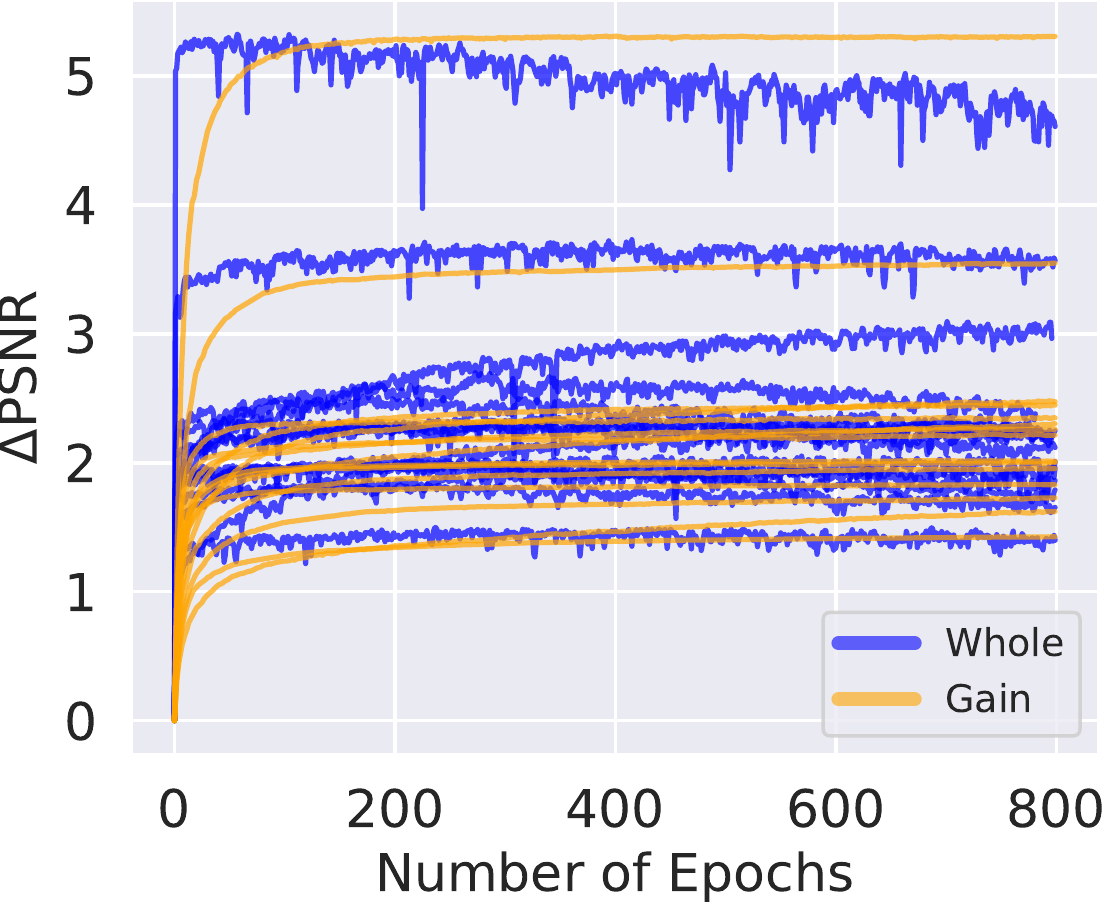}&
\includegraphics[width=\f1ht]{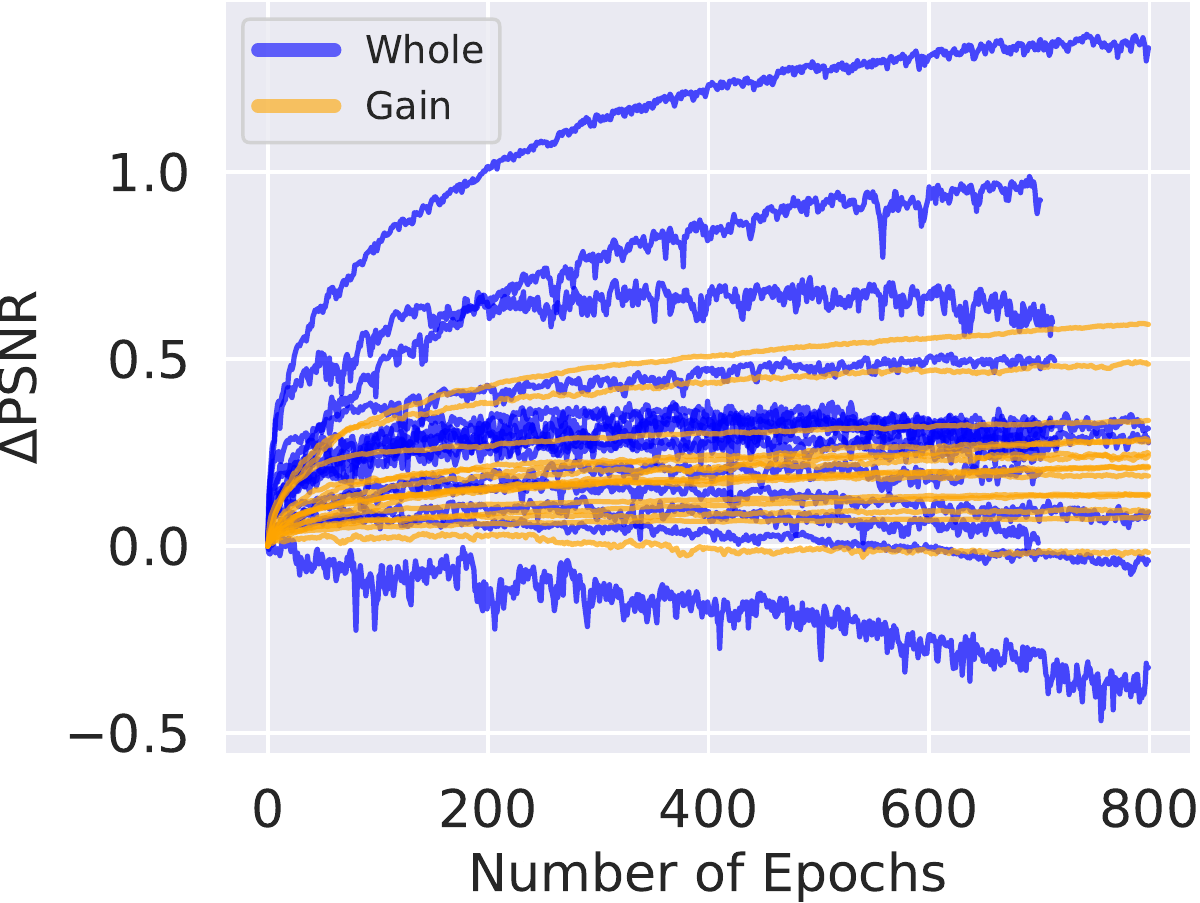}\\

\end{tabular}
}
\caption{\textbf{Out-of-distribution noise and signal}. We compare training all parameters of the network (blue), and only the gain parameters (orange) during the adaptation process. The CNN is pre-trained on generic natural images corrupted with Gaussian noise of standard deviation $\sigma \in [0, 55]$. We apply \gt\ to adapt it to images of urban scenes (high self-similarity, hence different signal characteristics from natural images) corrupted with $\sigma=70$ (which is outside the training range of noise). All architectures are trained using the SURE cost function. }

\label{fig:suppl_out_noise_signal}
\end{figure}

\begin{figure}
\def\f1ht{1.55in}%
\centering 
\footnotesize{
\begin{tabular}{c@{\hskip 0.1in}c@{\hskip 0.1in}c}
    \thead{In-distribution \\ Natural images ($\sigma \in [0, 55]$) $\rightarrow$ Set12 $\sigma=30$} & \thead{ Out-of-distribution noise \\ Natural images ($\sigma \in [0, 55]$) $\rightarrow$ Set12 $\sigma=70$} \\ 

    \includegraphics[height=\f1ht]{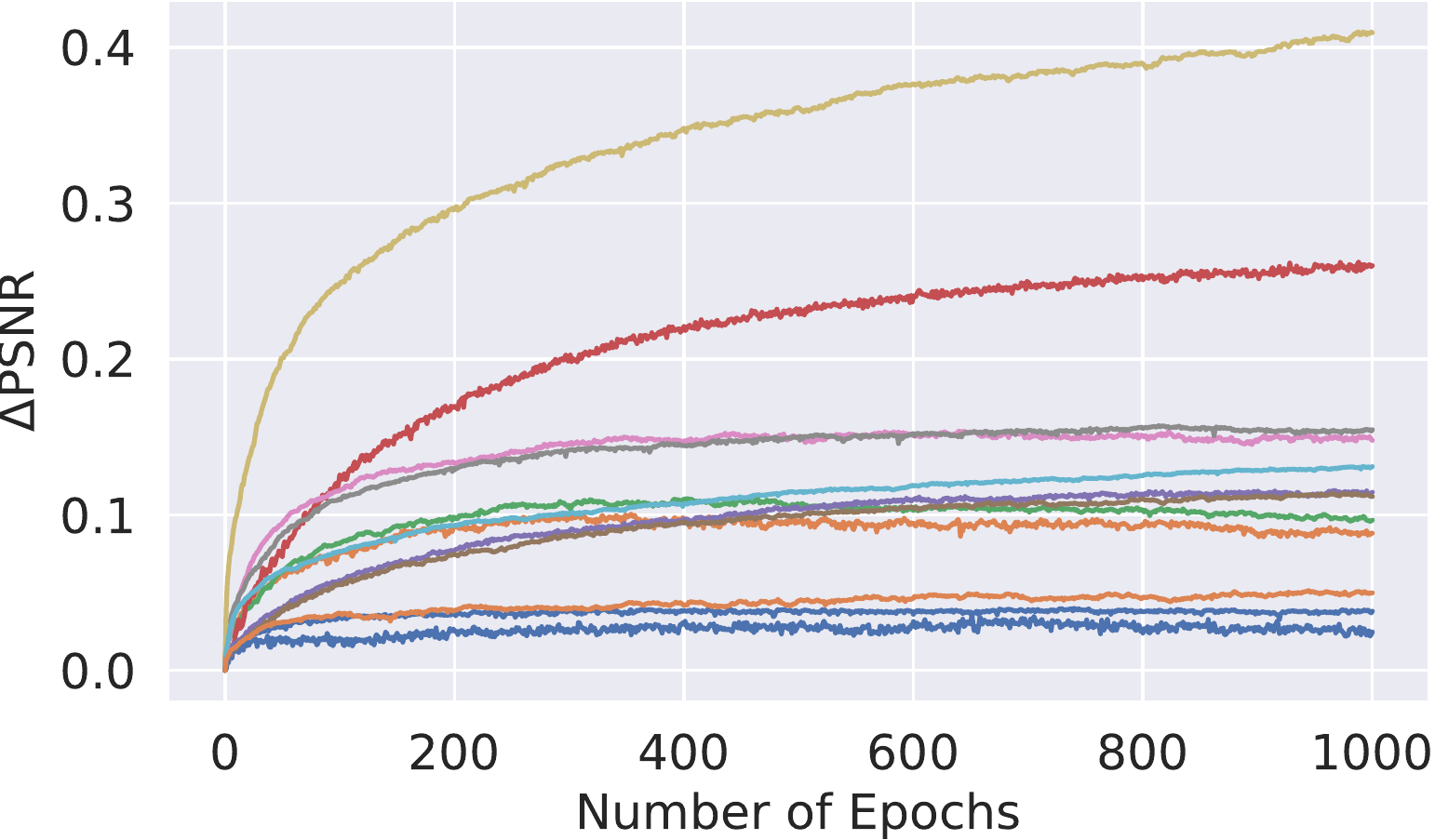}&
    \includegraphics[height=\f1ht]{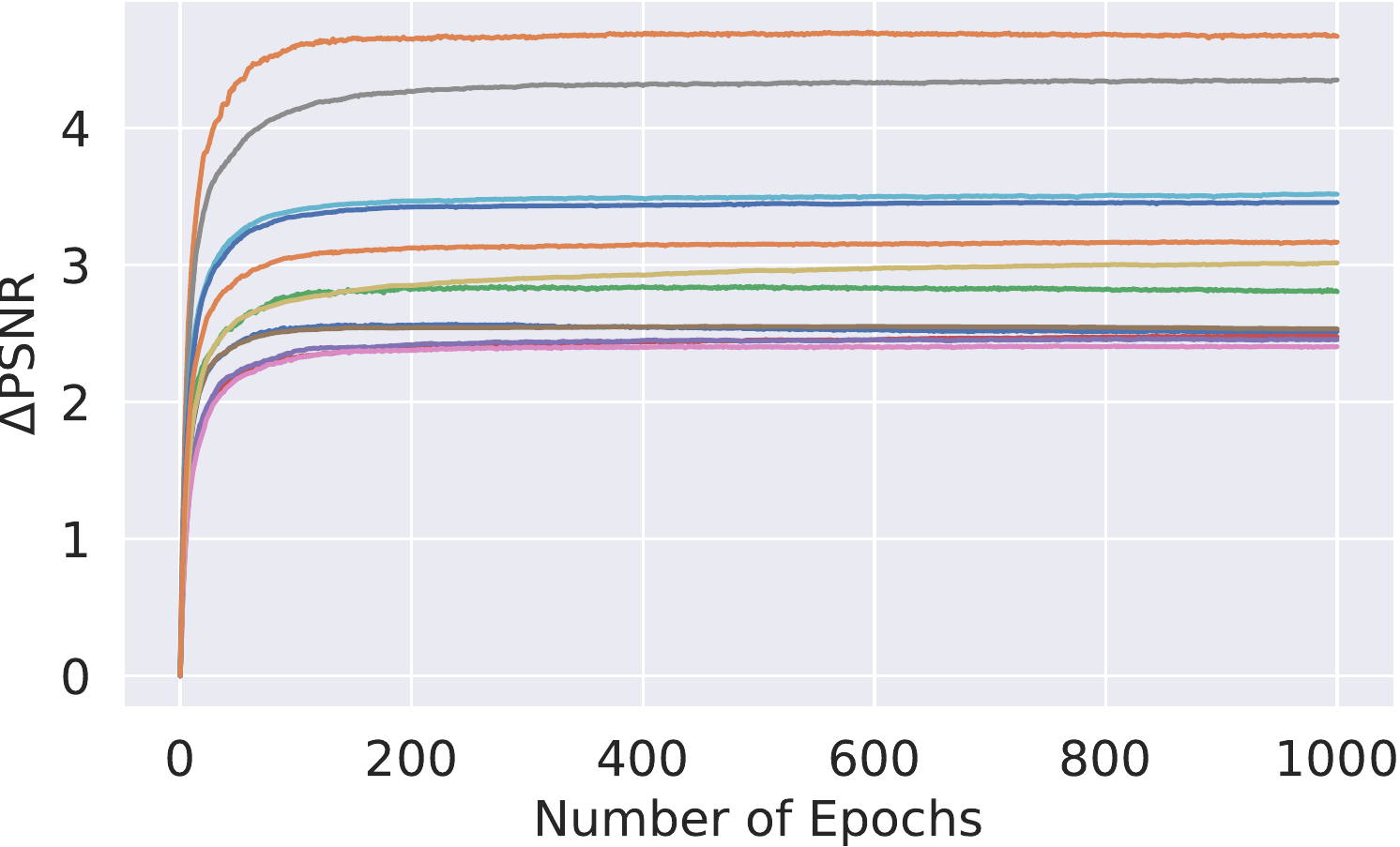} \\
    
    \thead{Out-of-distribution image \\  Natural images ($\sigma \in [0, 55]$) $\rightarrow$ Urban100 $\sigma=30$} & \thead{Out-of-distribution image and noise \\  Natural images ($\sigma \in [0, 55]$) $\rightarrow$ Urban100 $\sigma=70$ } \\
    \includegraphics[height=\f1ht]{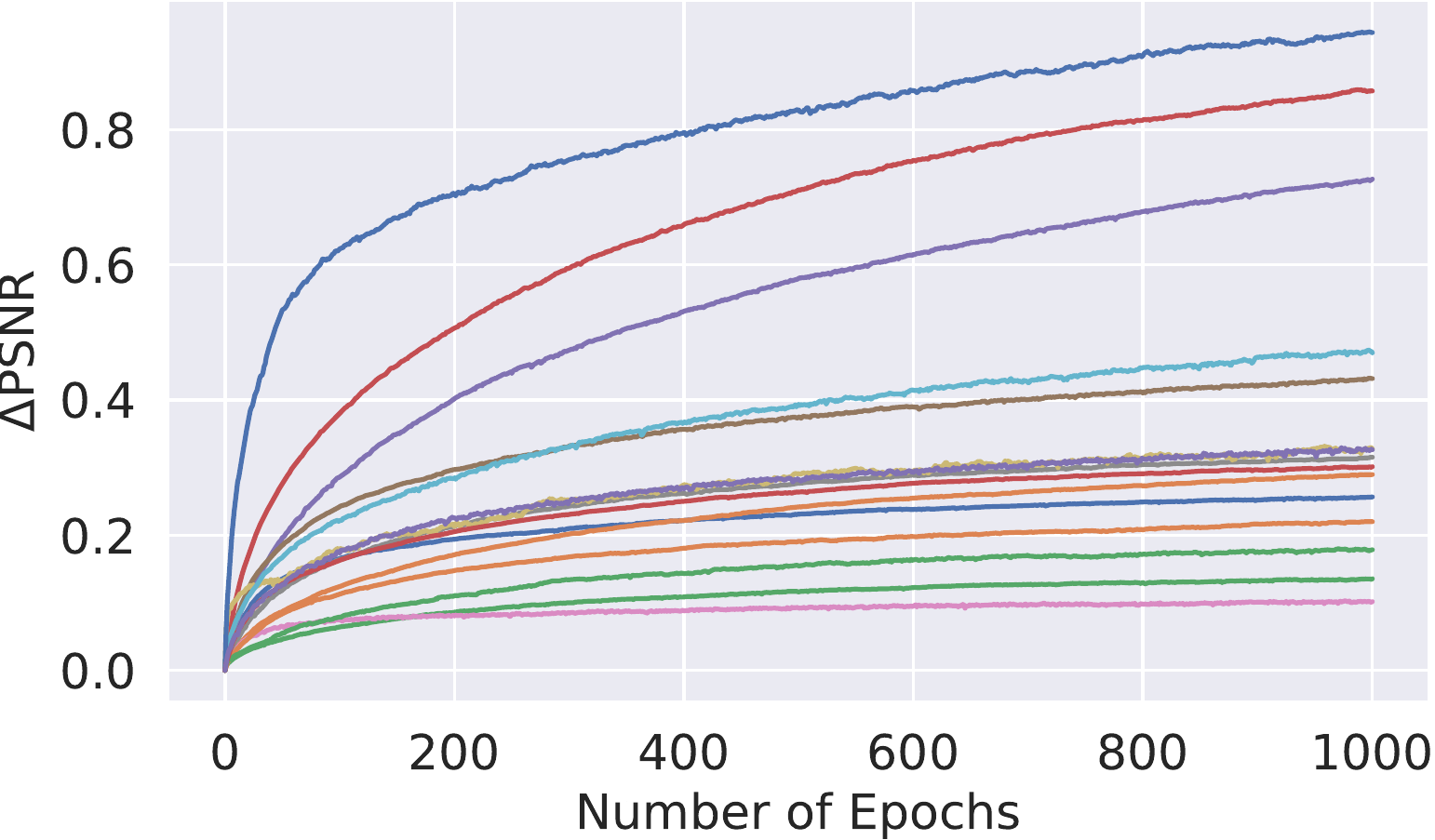}&
    \includegraphics[height=\f1ht]{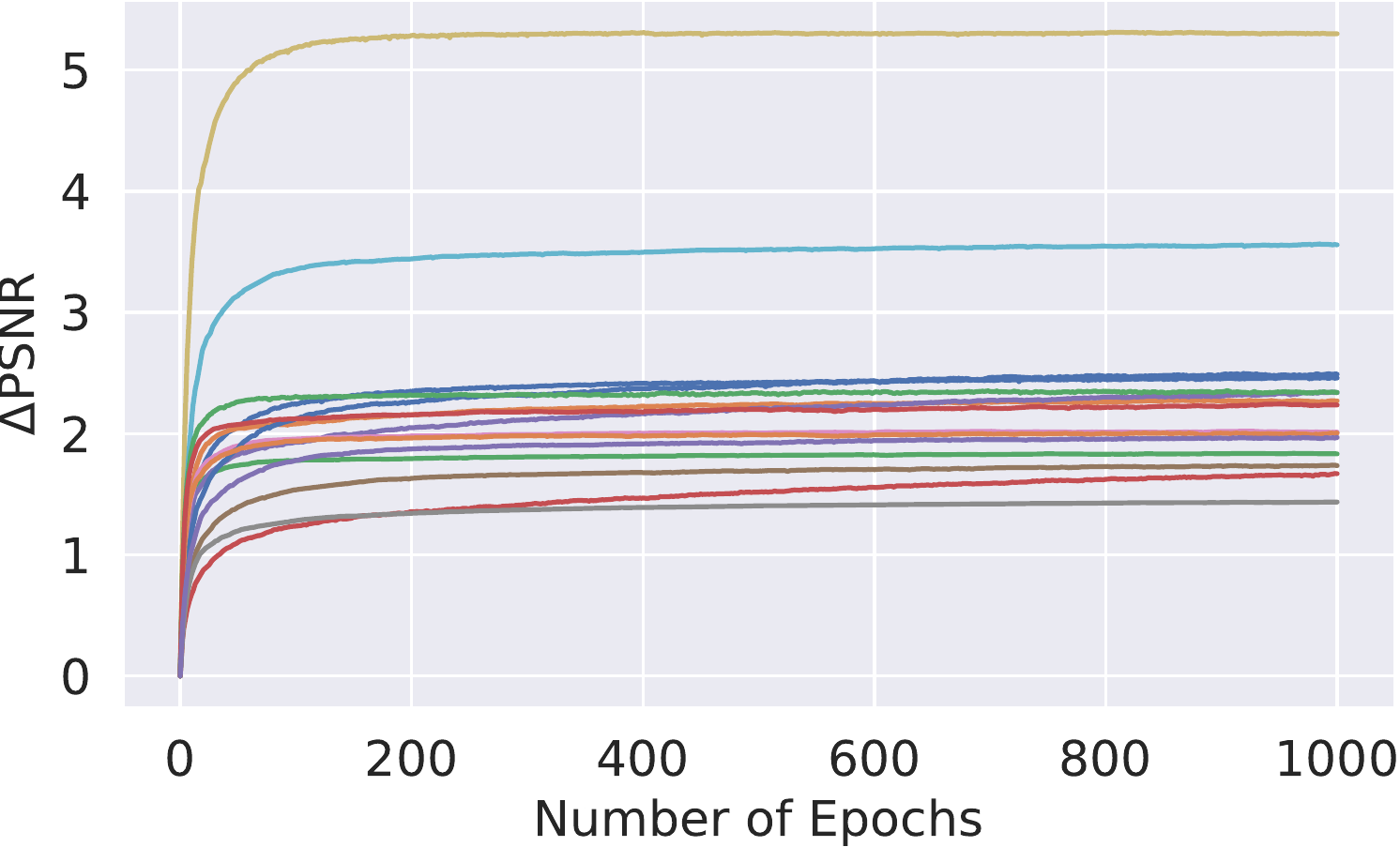}\\
\end{tabular}
}
\caption{\textbf{\gt\ does not require early stopping}. We plot the improvement in performance achieved by \gt\ with the number of iterations. Each iteration step is a pass through 10000 random $50 \times 50$ patch extracted from the image. The performance achieved by optimizing only the gain parameters remains constant or monotonically increases with iteration, while training all parameters often overfits (see Figure~\ref{fig:all_vs_gain_suppl})}
\label{fig:training_curve_dncnn}
\end{figure}

\begin{figure}
\def\f1ht{1.05in}%
\centering 
\footnotesize{
\begin{tabular}{ >{\centering\arraybackslash}m{.5in} >{\centering\arraybackslash}m{2in} >{\centering\arraybackslash}m{2in}}
\centering 
    Gradient Steps  &  All parameters  & Gain parameters \\[0.2cm]
    0 &
    \includegraphics[height=\f1ht]{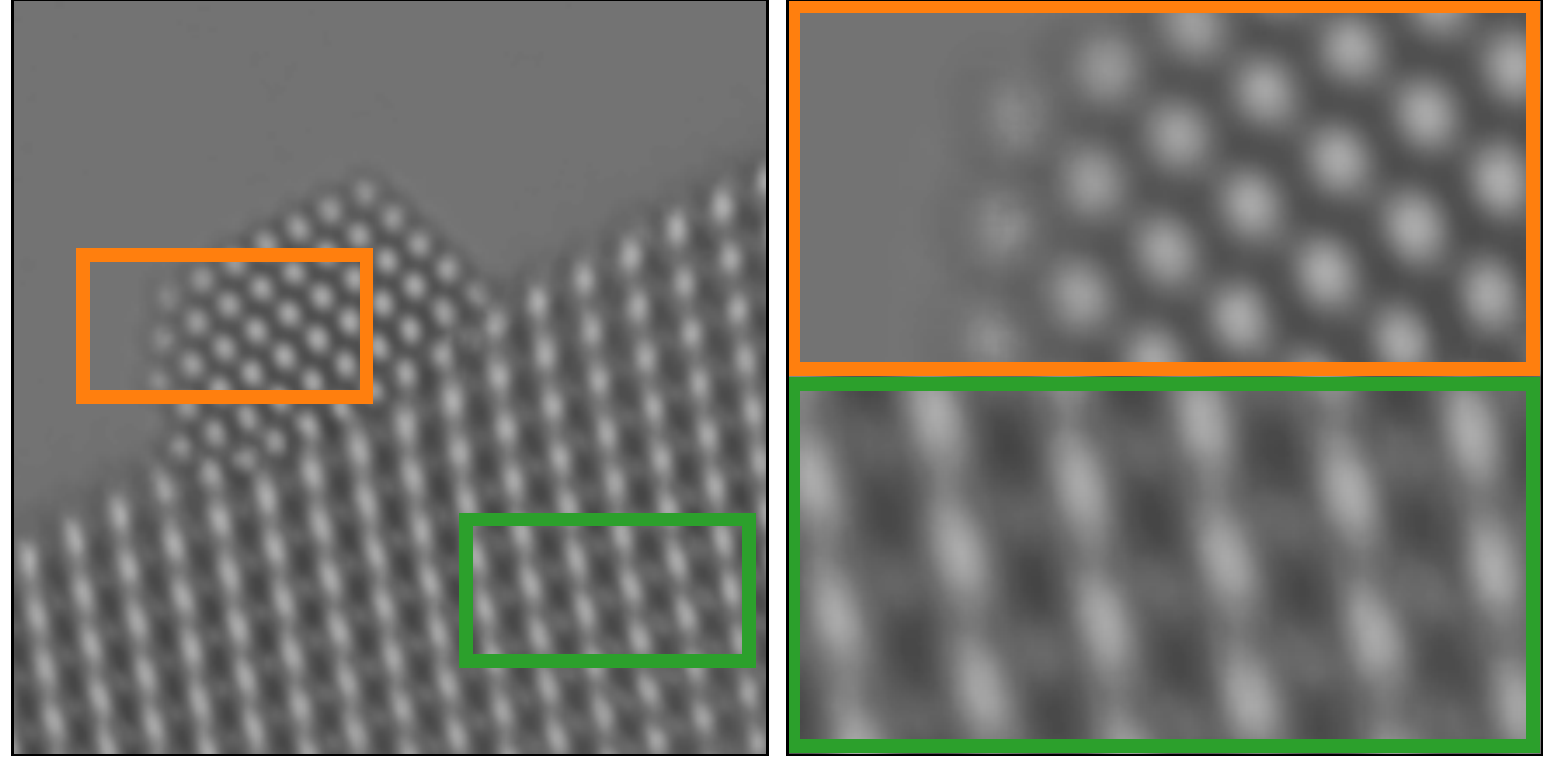}&
    \includegraphics[height=\f1ht]{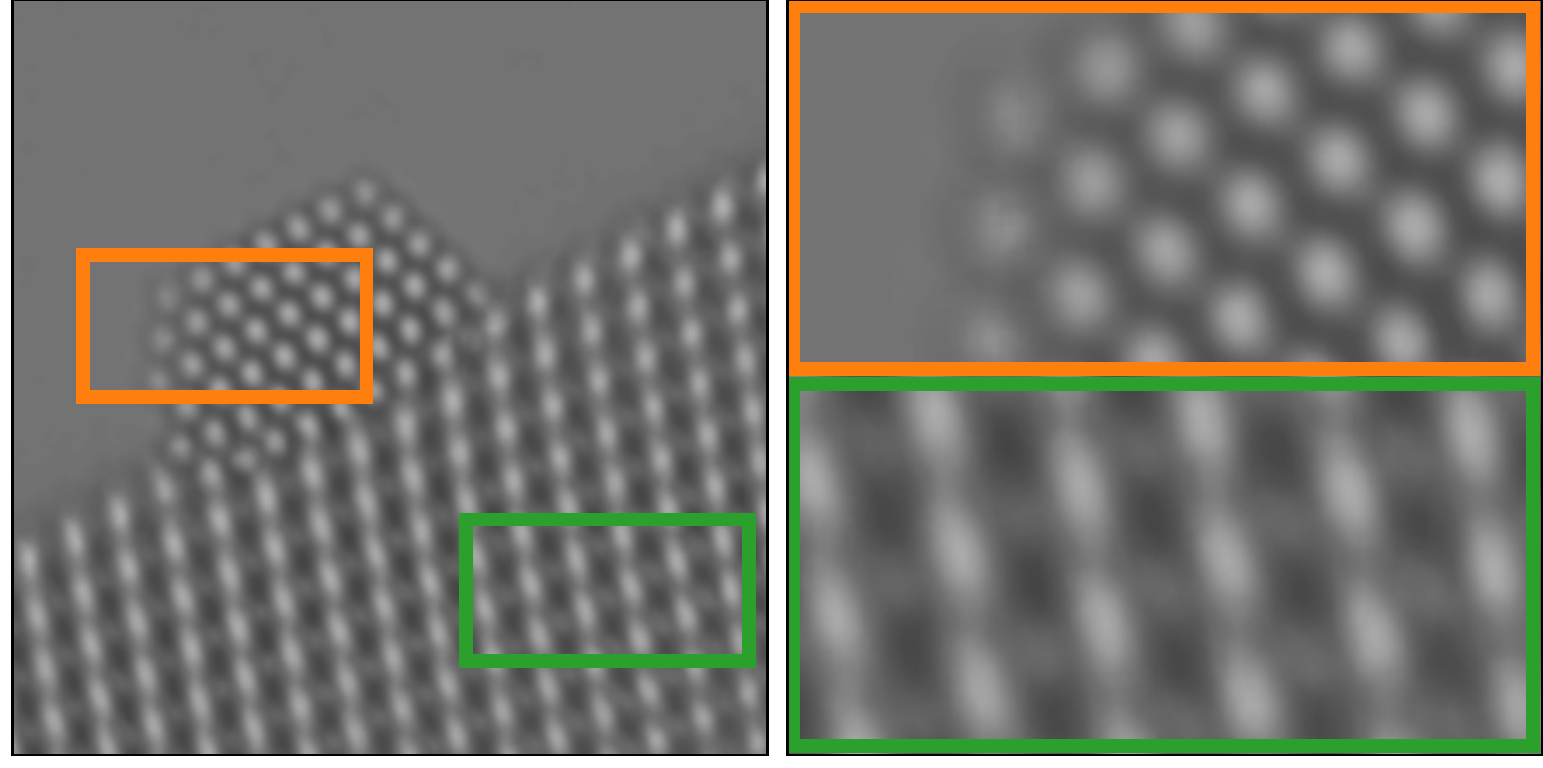}\\
    1.5K &
    \includegraphics[height=\f1ht]{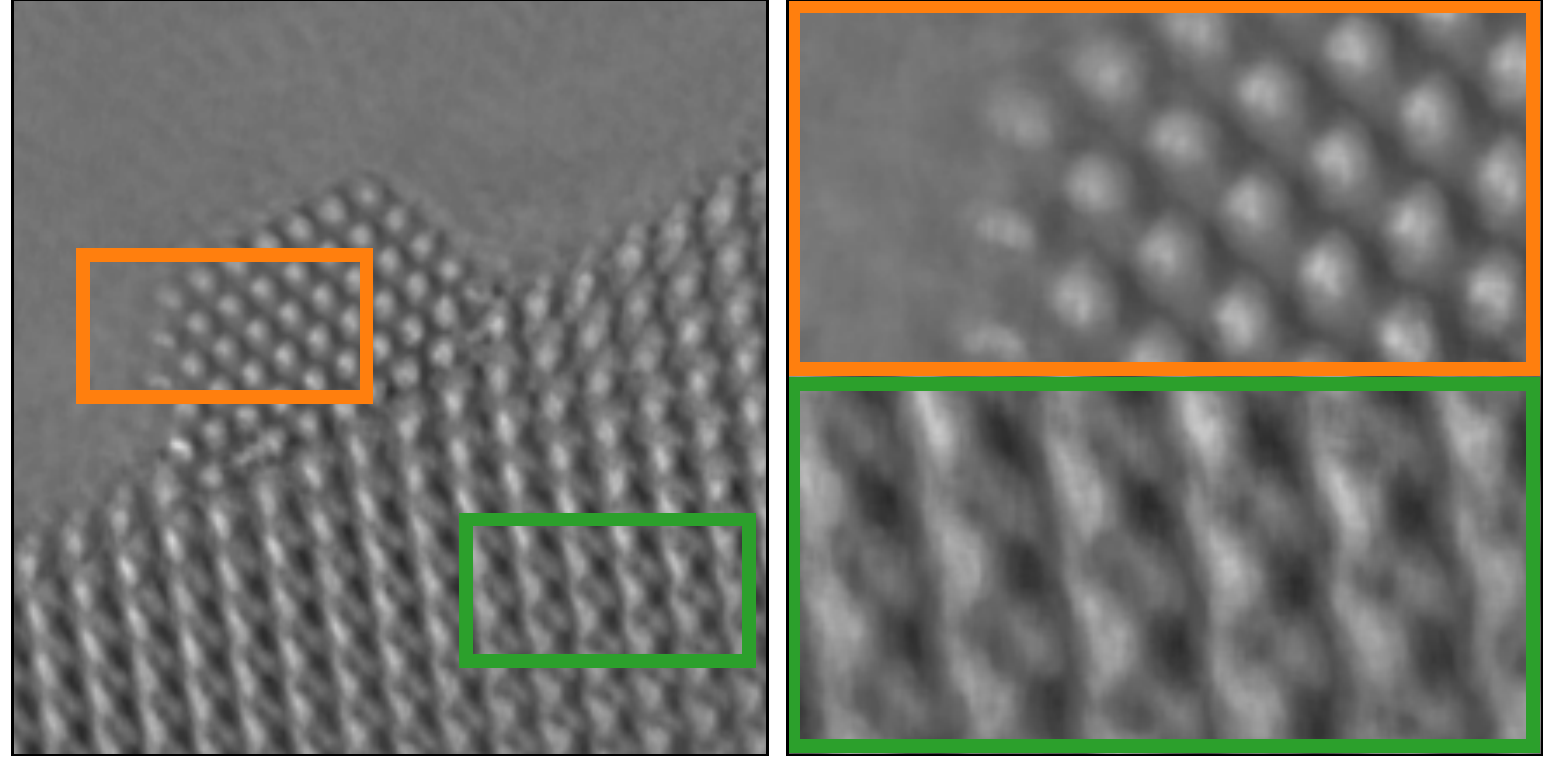}&
    \includegraphics[height=\f1ht]{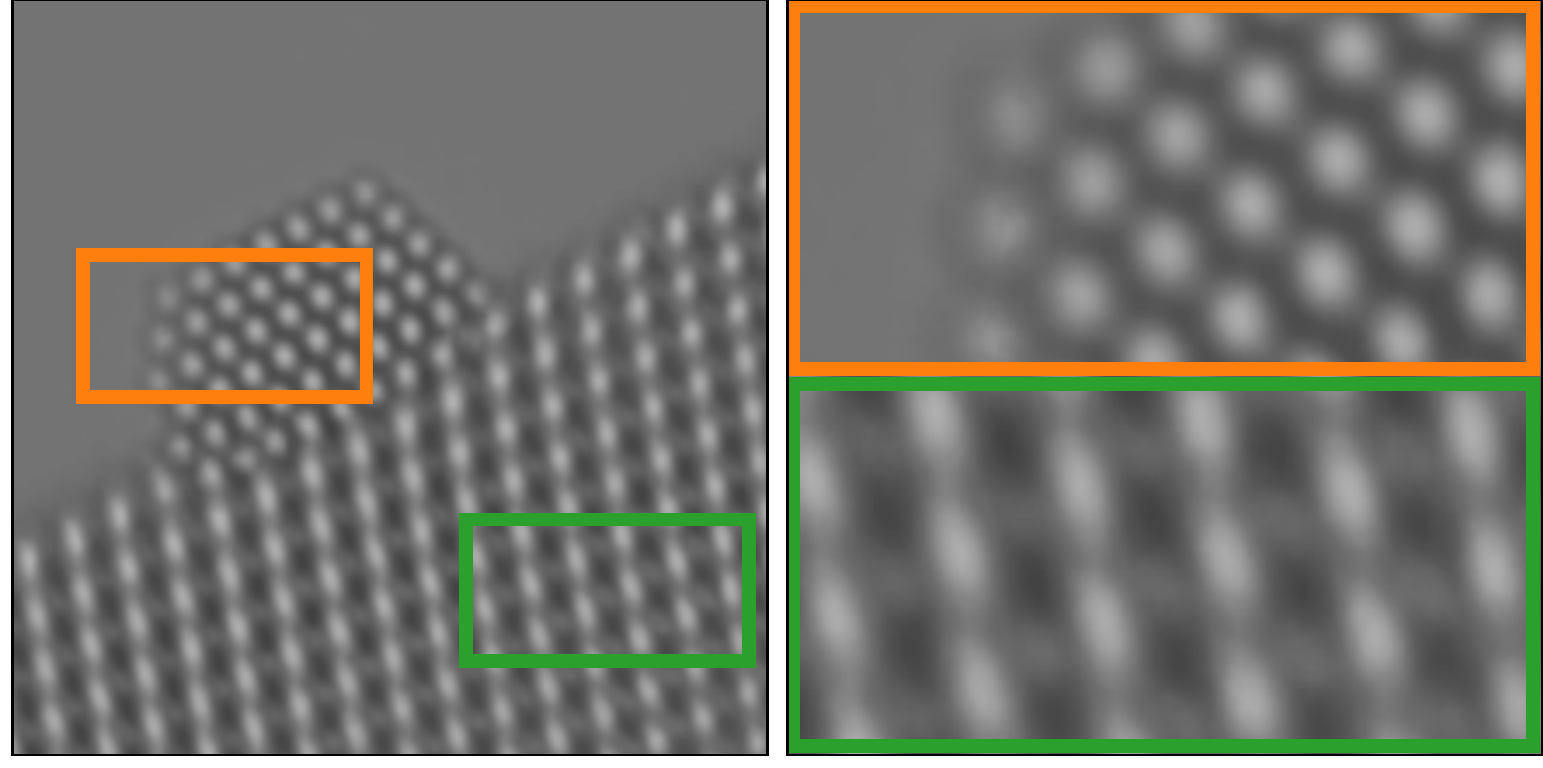}\\
    2.5K &
    \includegraphics[height=\f1ht]{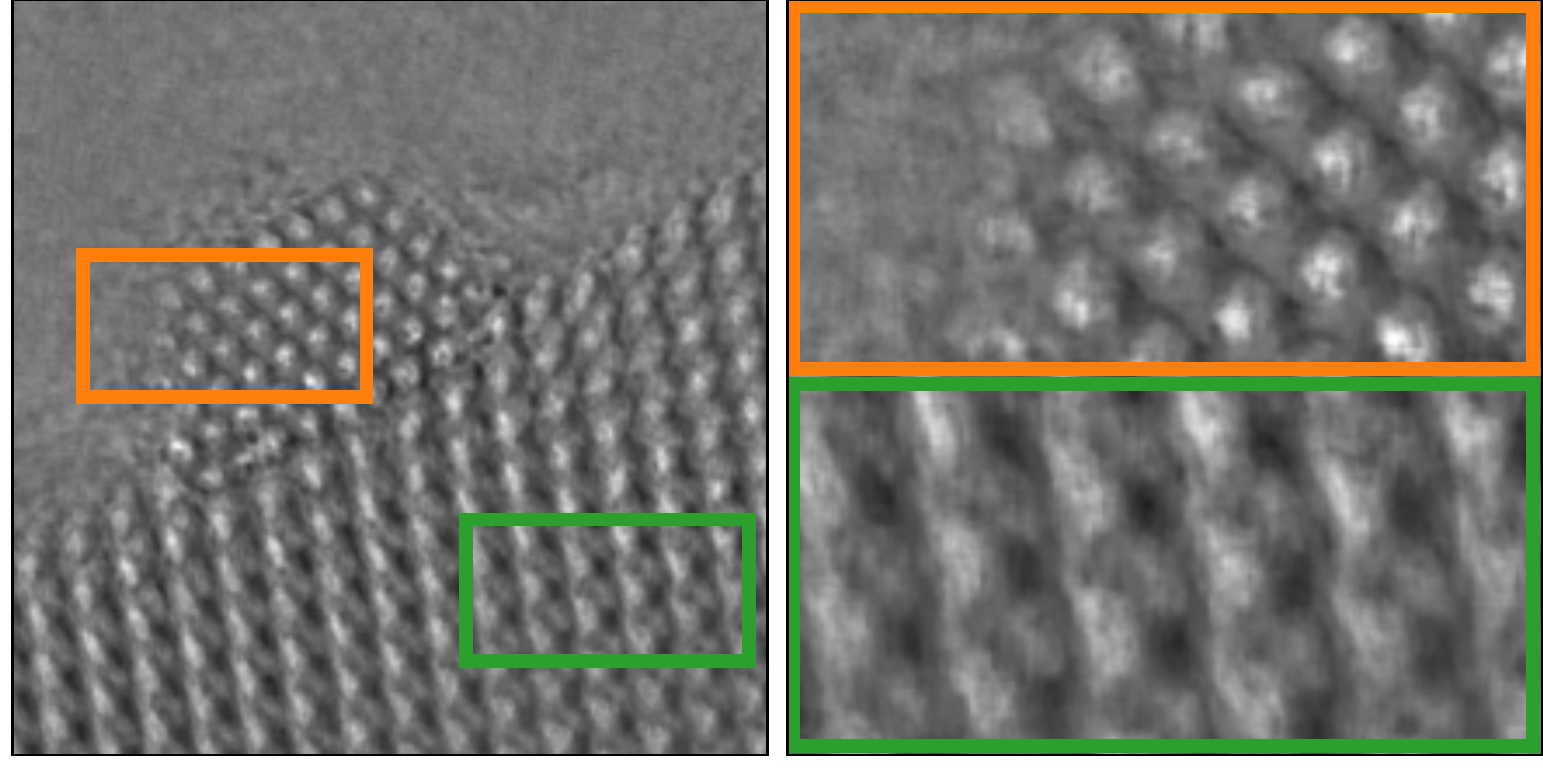}&
    \includegraphics[height=\f1ht]{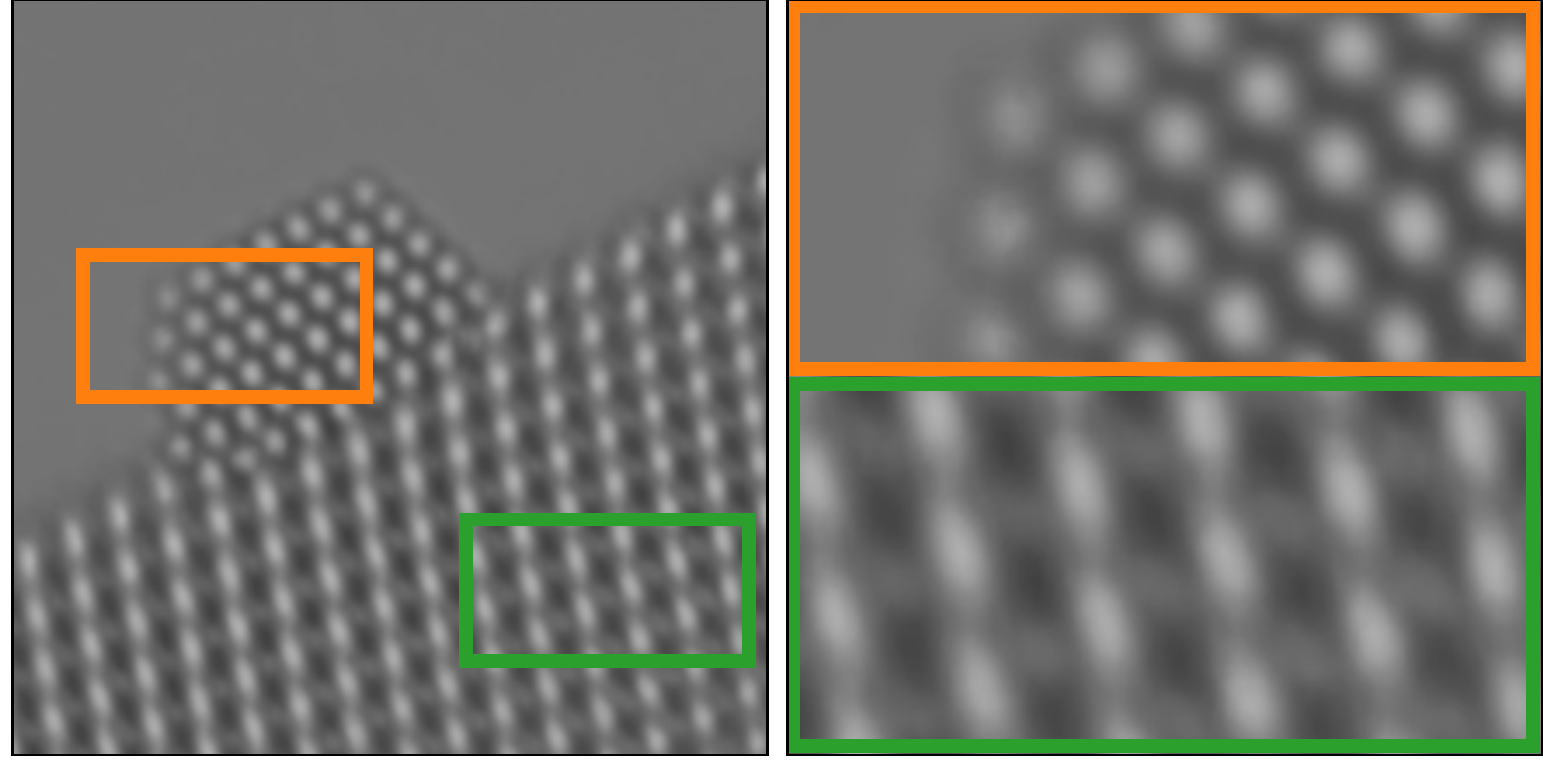}\\
    5K &
    \includegraphics[height=\f1ht]{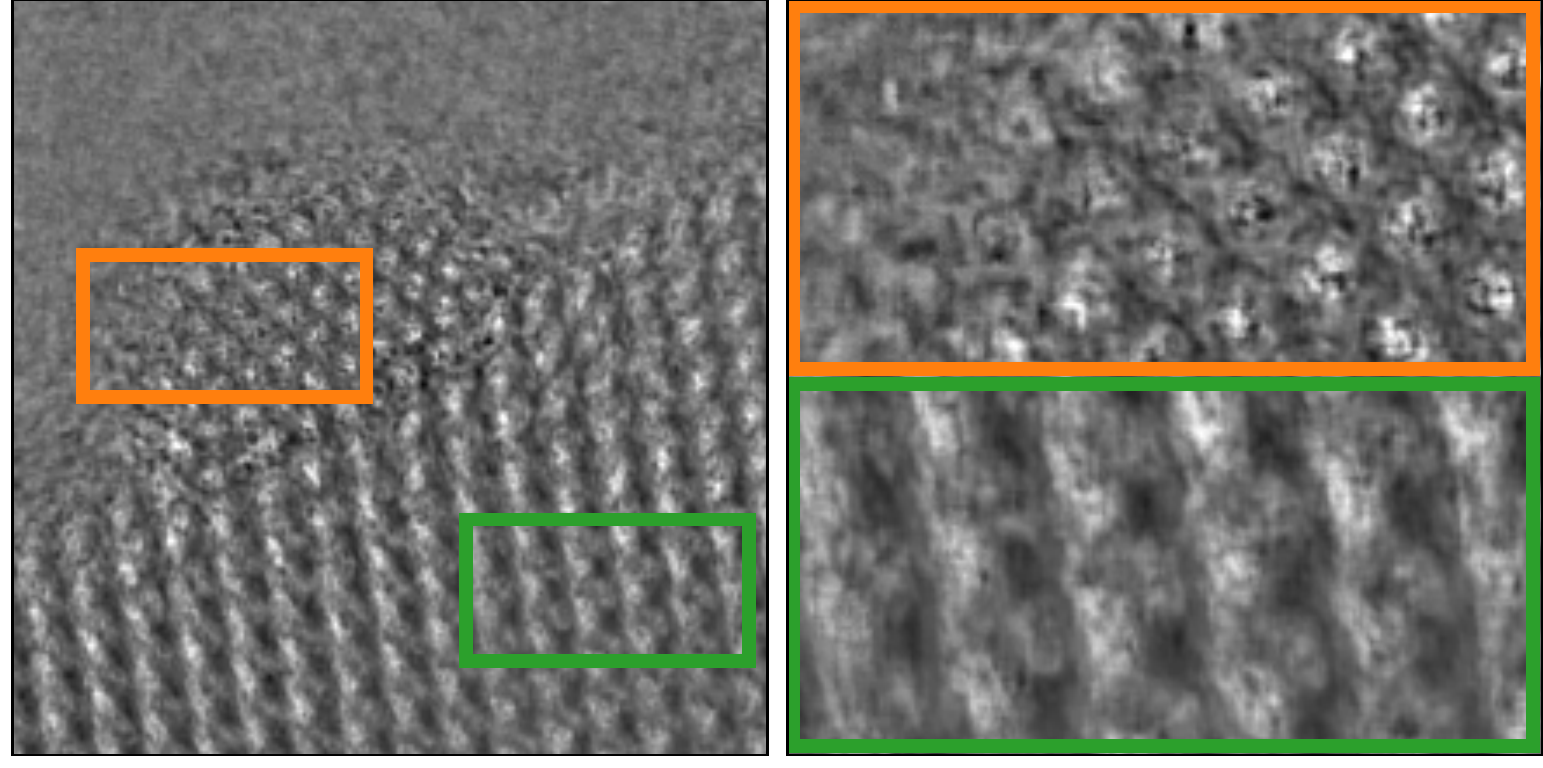}&
    \includegraphics[height=\f1ht]{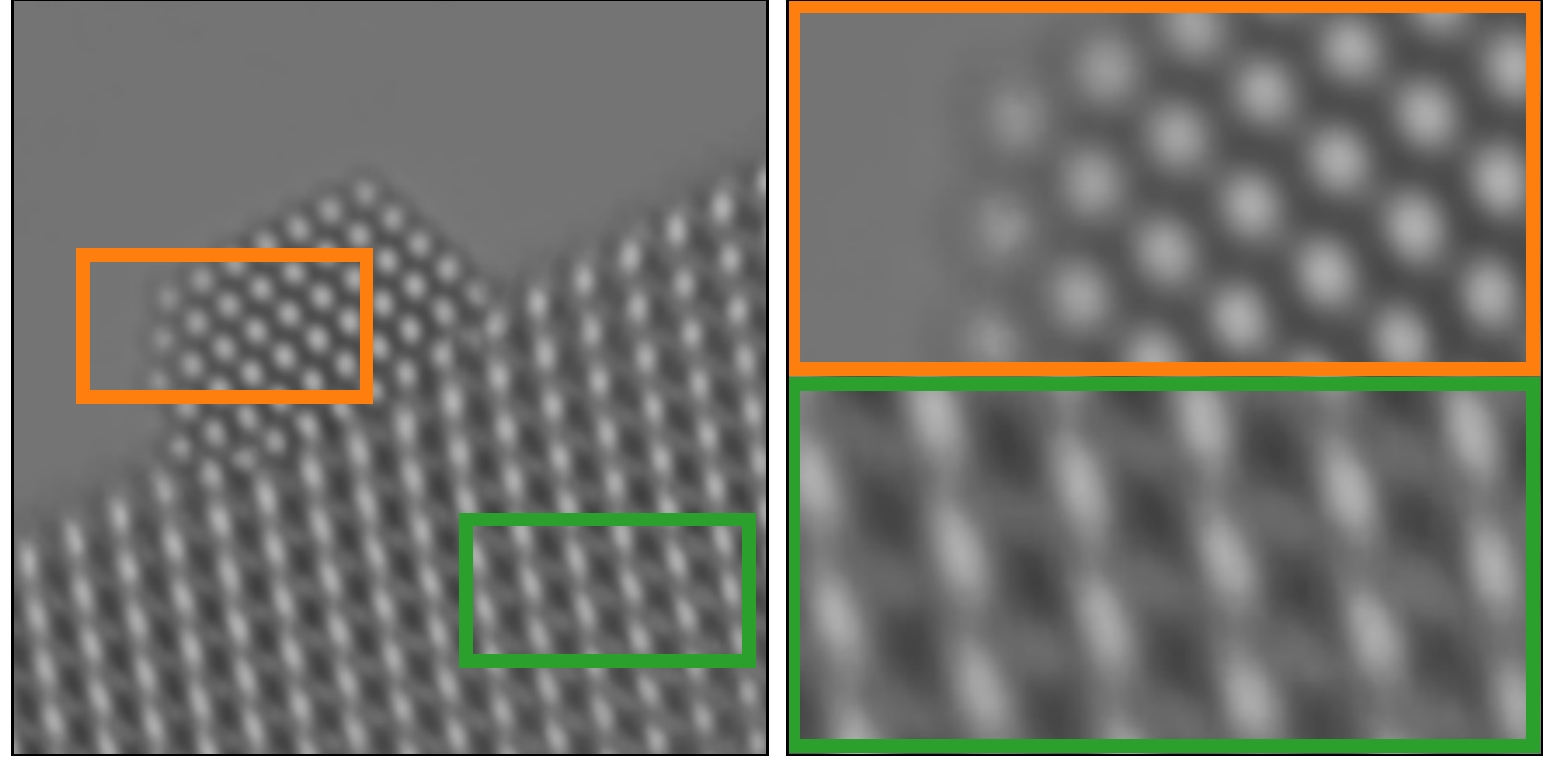}\\
    10K &
    \includegraphics[height=\f1ht]{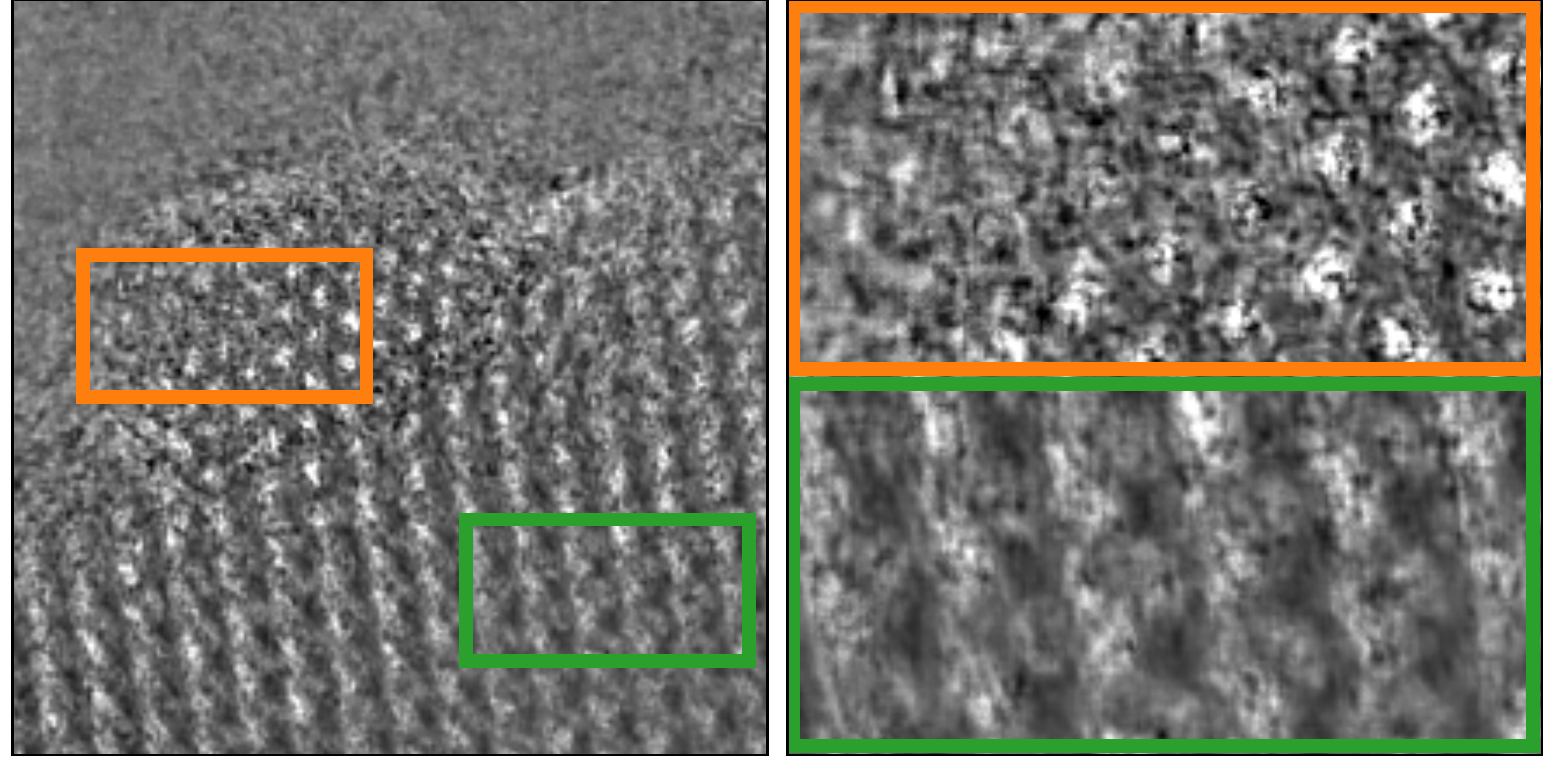}&
    \includegraphics[height=\f1ht]{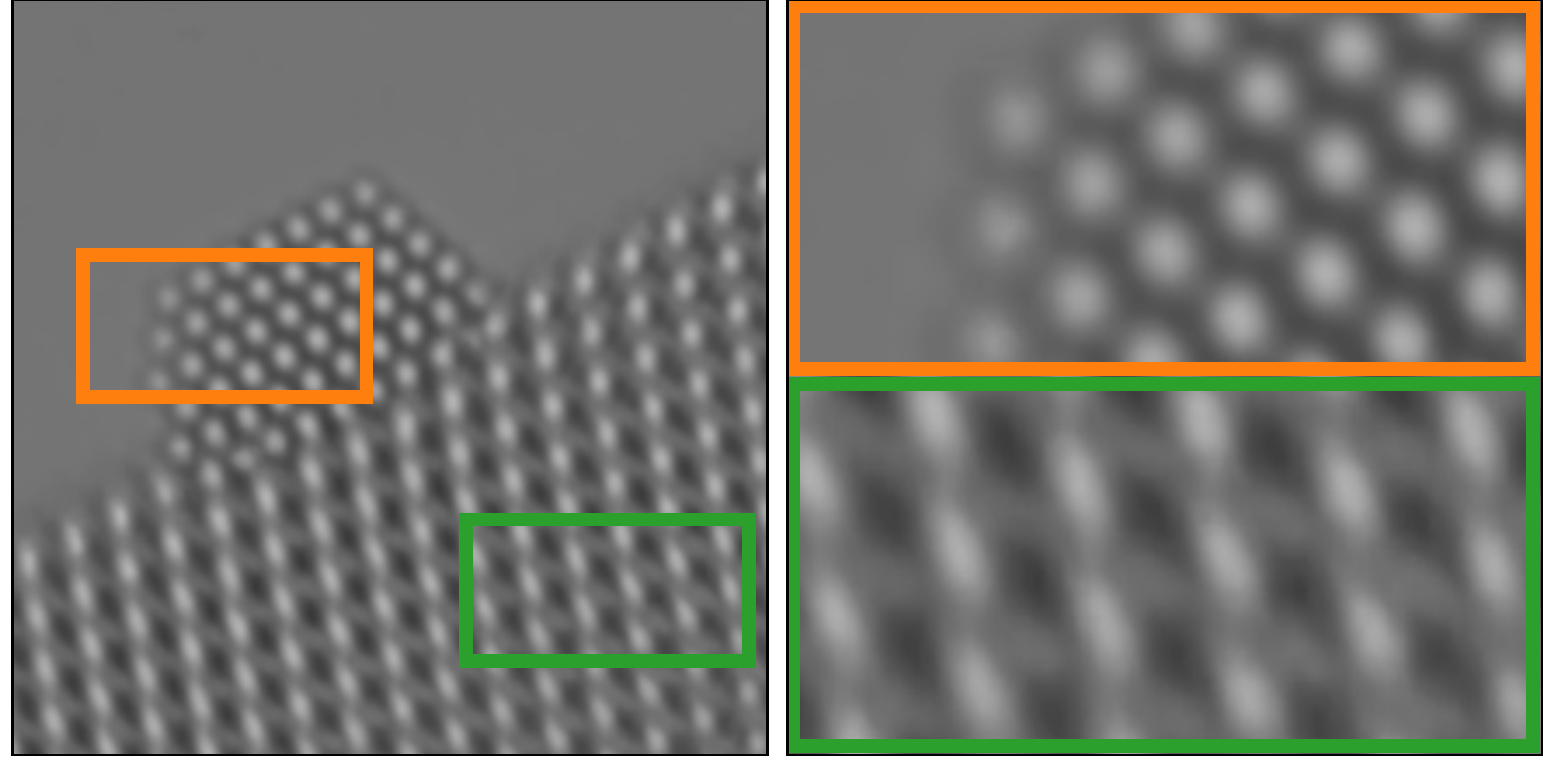}\\
    25K &
    \includegraphics[height=\f1ht]{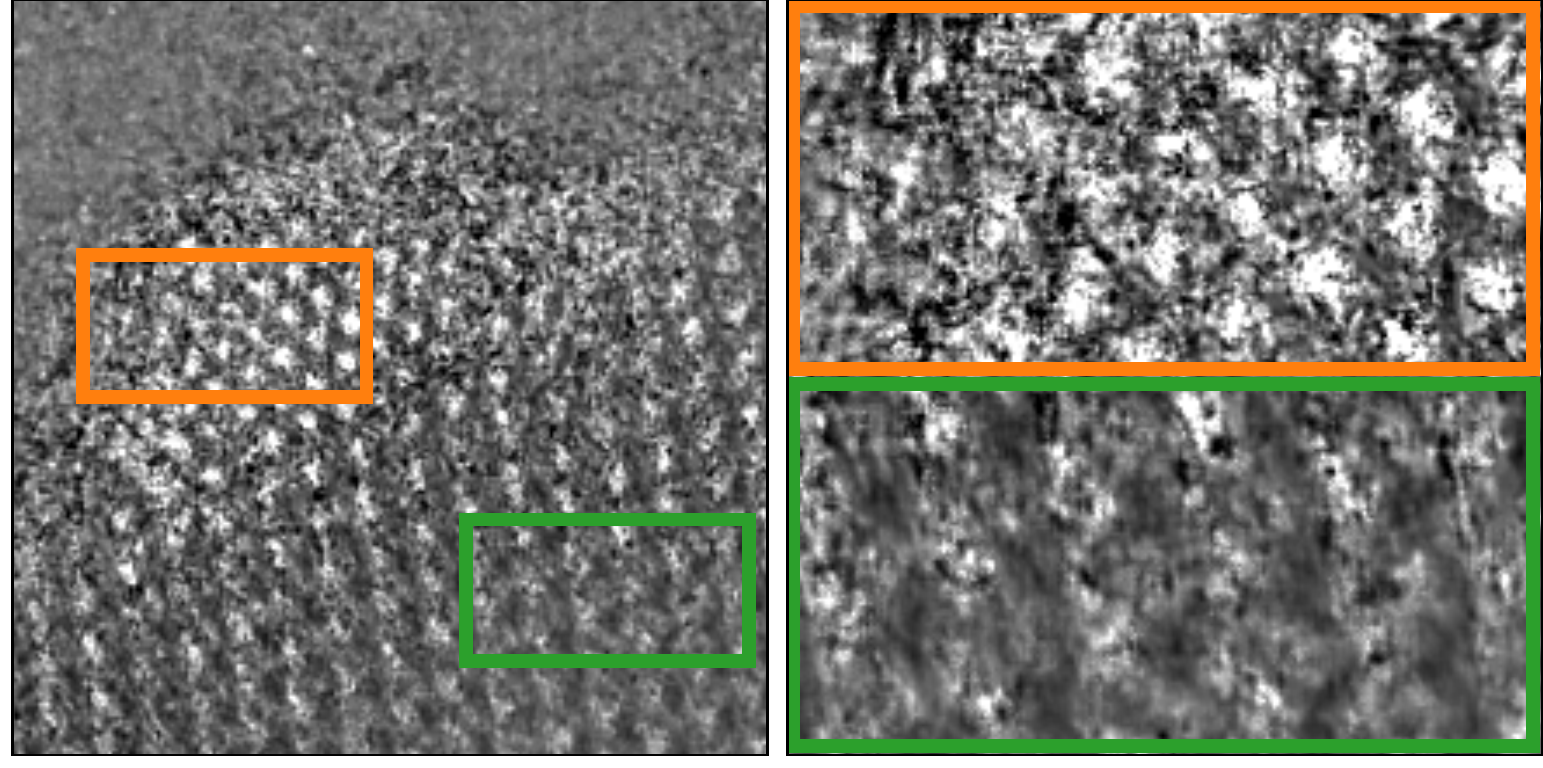}&
    \includegraphics[height=\f1ht]{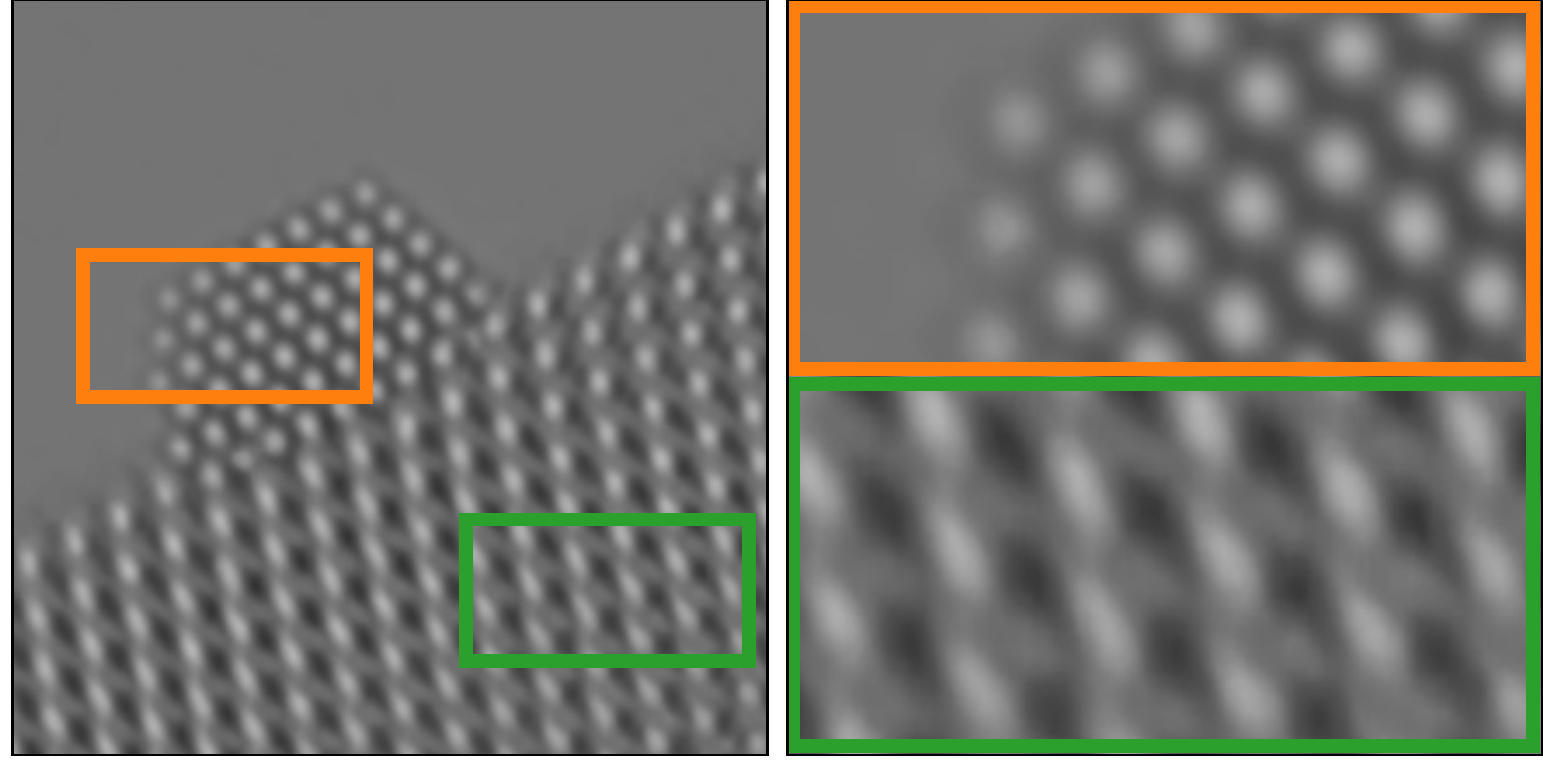}\\
    50K &
    \includegraphics[height=\f1ht]{figs/nano/whole-vs-gain/whole_50.pdf}&
    \includegraphics[height=\f1ht]{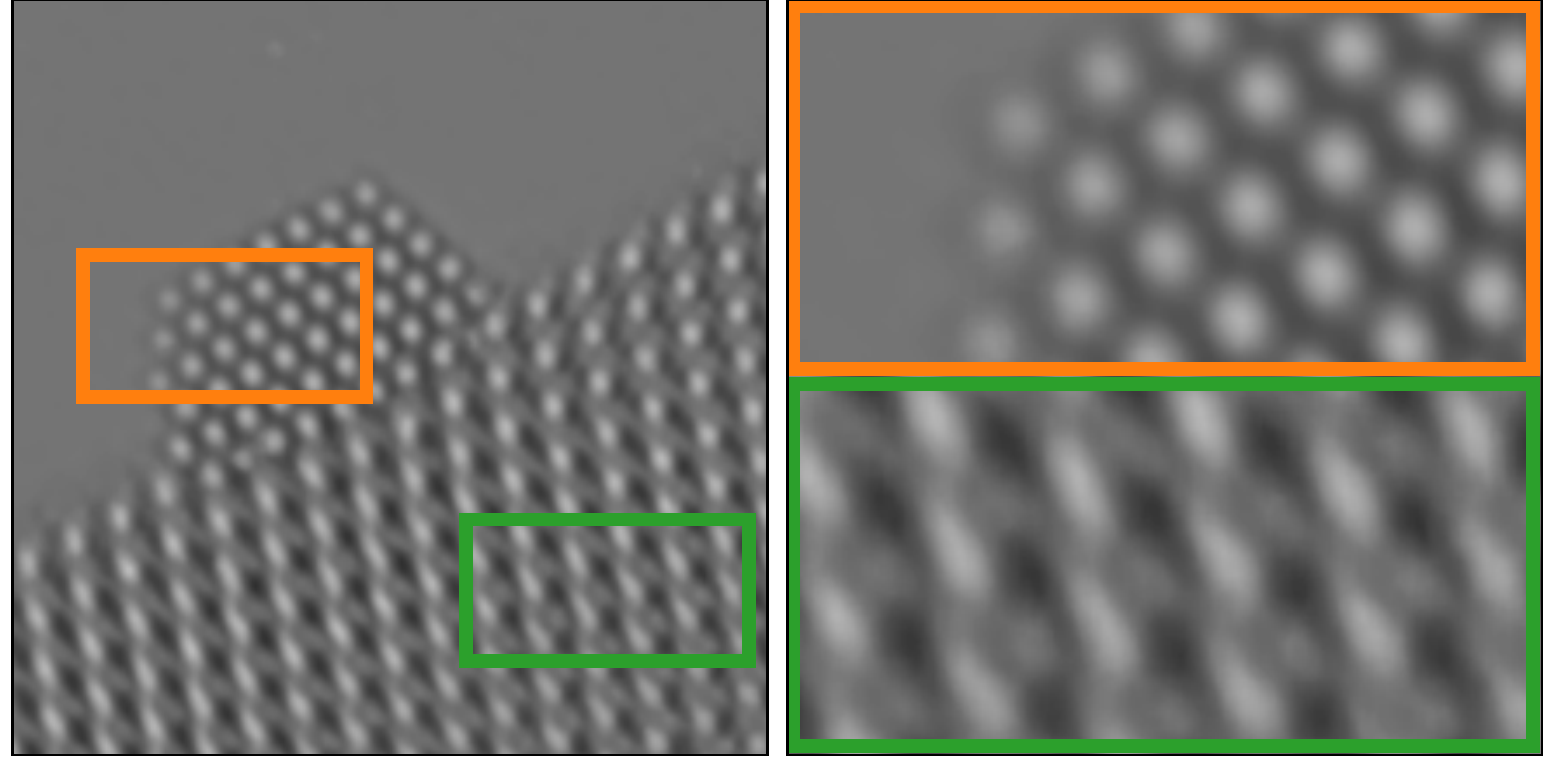}\\
\end{tabular}
}
\caption{\textbf{\gt\ prevents overfitting in TEM data}. We compare training all parameters and only the gain parameters while adapting a CNN pre-trained on simulated TEM data to real TEM data. Training all parameters clearly overfits to the noisy image. Each gradient step is updated over two random patches of size $400 \times 400$.}
\label{fig:all_vs_gain_nano_blindspot}
\end{figure}

\section{Performance of \gt}
\label{sec:suppl_experiments}

\subsection{In-distribution test image}
\label{sec:suppl_in_distr}

\textbf{Different architectures.} We evaluated DnCNN, UNet and BFCNN architectures for this task. All models were trained on denoising Gaussian white noise of standard deviation $\sigma \in [0, 55]$ from generic natural images. Results of DnCNN and UNet are presented in Figure~\ref{fig:sota} in the main paper. Results for the BFCNN architecture are provided in Table~\ref{tab:bfcnn_in_distr}.

\textbf{Different cost functions}. We provide the results of evaluating DnCNN architecture with different cost functions in Table~\ref{tab:lossfun_comparison}. 

\textbf{Distribution of improvements.} We visualize the distribution of improvements in denoising performance for different architectures after performing \gt\ using the SURE cost function in Figure~\ref{fig:hist_in_distr}. As discussed in Section~\ref{sec:limitations}, if the CNN is optimized well and the test image is in-distribution, \gt\ can degrade performance. This degradation is atypical (3 out of 408 total evaluations) and very small (maximum degradation of 0.02 dB in PSNR). 

\begin{table}[t]
\centering
\begin{tabular}{cccccc}
        \toprule
        \multirow{2}{*}{Model} & 
        \multirow{2}{*}{$\sigma$} & 
        \multicolumn{2}{c}{Set12}    &
        \multicolumn{2}{c}{Set68}  \\

        \cmidrule(lr){3-4}
        \cmidrule(lr){5-6}
        
        & &
        \multicolumn{1}{c}{Pre-trained} &
        \multicolumn{1}{c}{\gt} &
         \multicolumn{1}{c}{Pre-trained} &
        \multicolumn{1}{c}{\gt} \\

        \midrule
        BFCNN
        & $30$ & 29.52 & 29.61 & 28.36  & 28.45  \\
    
        \bottomrule
\end{tabular}\\[0.3cm]

\caption{\textbf{Results for BFCNN}. Results for BFCNN~\cite{biasfree} architecture trained on BSD400 dataset corrupted with Gaussian noise of standard deviation $\sigma \in [0, 55]$. Results for other architectures are provided in Section~\ref{sec:exp_in_distr}. }
\label{tab:bfcnn_in_distr}
\end{table}

\def\f1ht{2.4in}%
\begin{figure}
    \centering
    \footnotesize{
    \begin{tabular}{c@{\hskip 0.03in}c@{\hskip 0.01in}c}
    \toprule
    $\sigma$ & \multicolumn{1}{c}{DnCNN} &
        \multicolumn{1}{c}{UNet}\\
    \midrule
    
     \multirow{2}{*}[6em]{\textbf{30}}&  \multicolumn{1}{c}{\includegraphics[width=\f1ht]{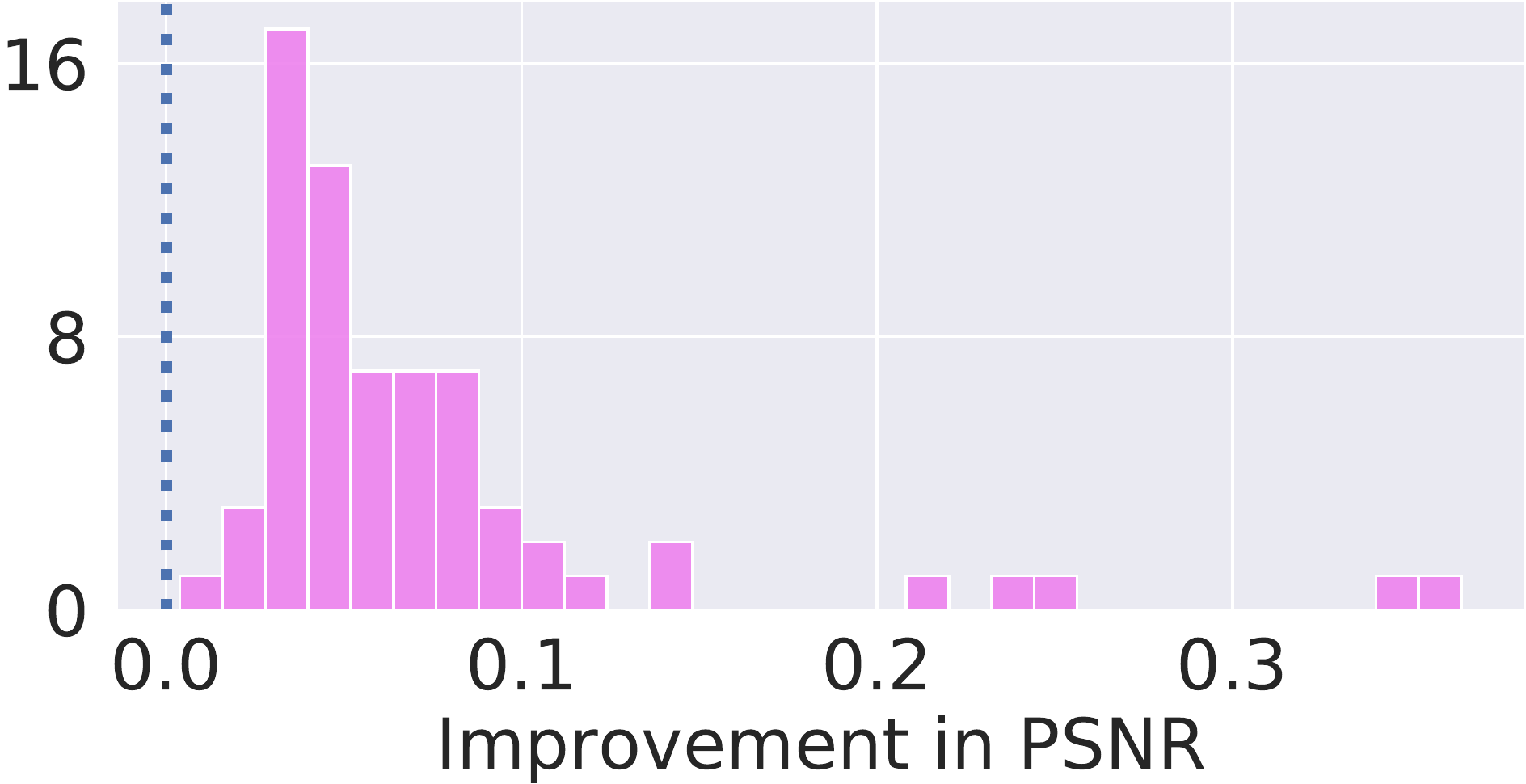}}&
    \multicolumn{1}{c}{\includegraphics[width=\f1ht]{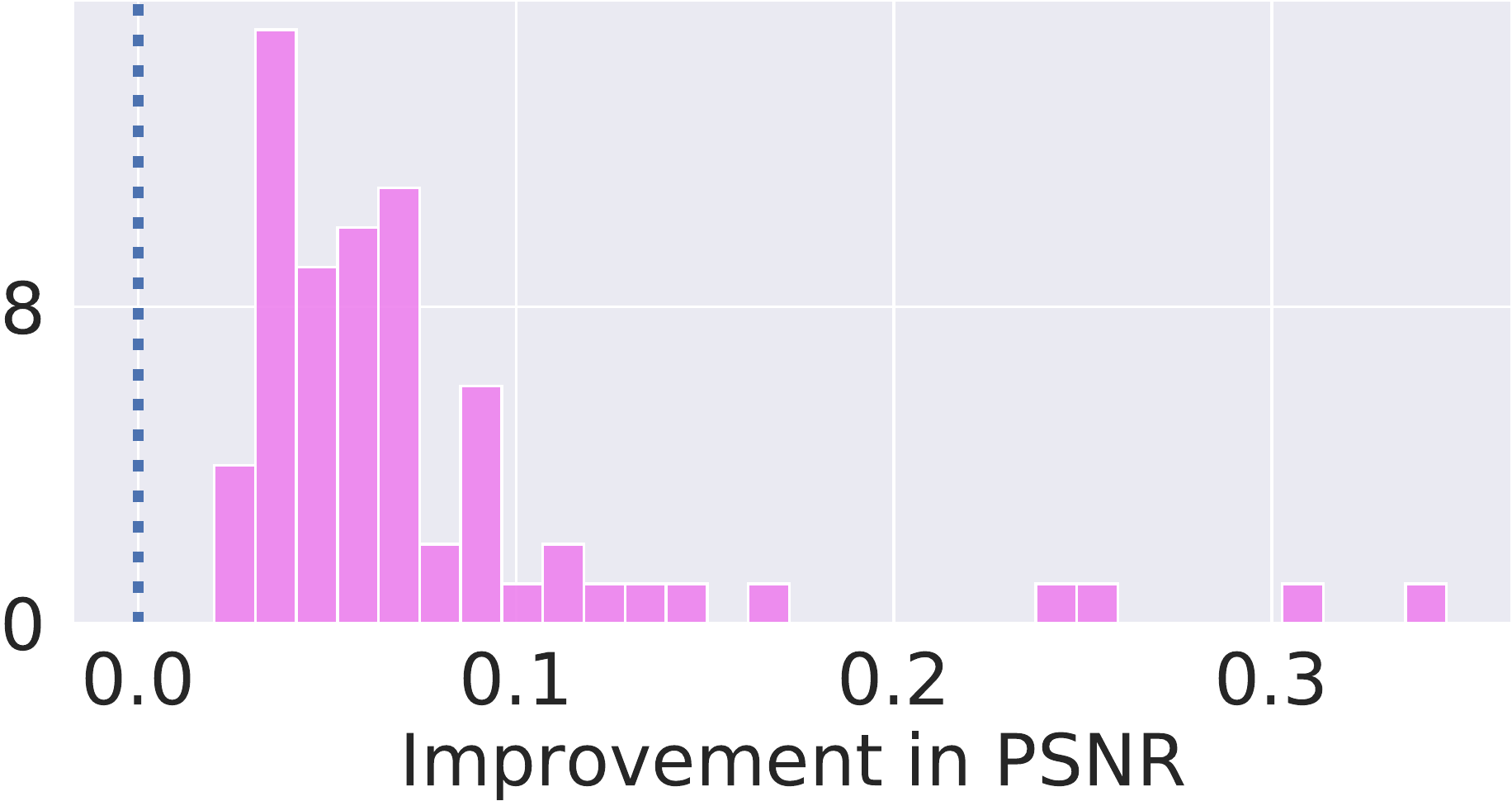}}\\
    
    & \begin{tabular}{|c|c|c|}
    \hline
         Max & Min & Num. of $\Delta$PSNR < 0 \\
         \hline
         0.364 & 0.004 & 0 \\
    \hline
    \end{tabular} &
    \begin{tabular}{|c|c|c|}
    \hline
         Max & Min & Num. of $\Delta$PSNR < 0 \\
         \hline
         0.346 & 0.020 & 0 \\
    \hline
    \end{tabular} \\[0.3cm]
    
    \midrule
    
          \multirow{2}{*}[6em]{\textbf{40}}&  \multicolumn{1}{c}{\includegraphics[width=\f1ht]{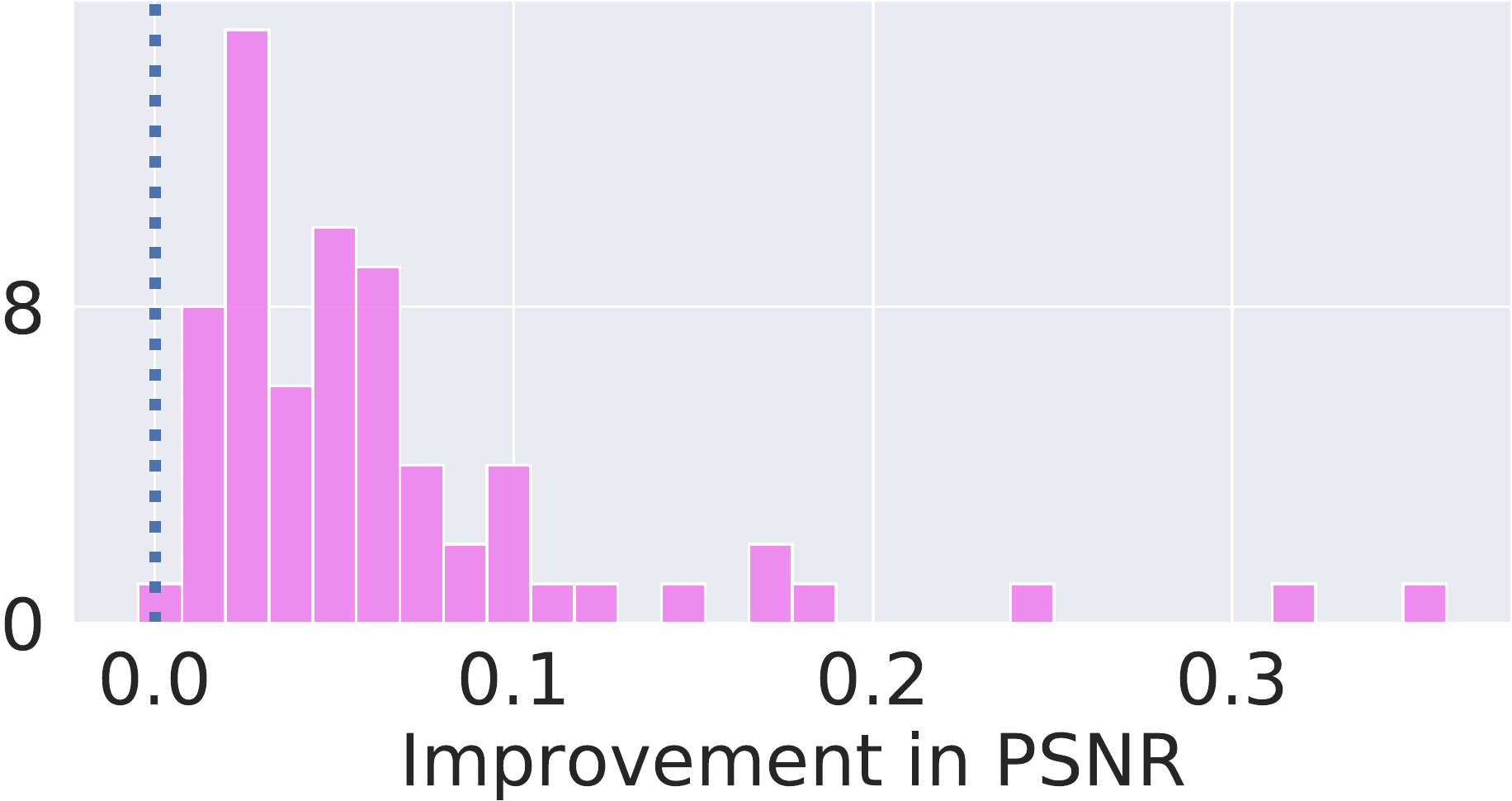}}&
    \multicolumn{1}{c}{\includegraphics[width=\f1ht]{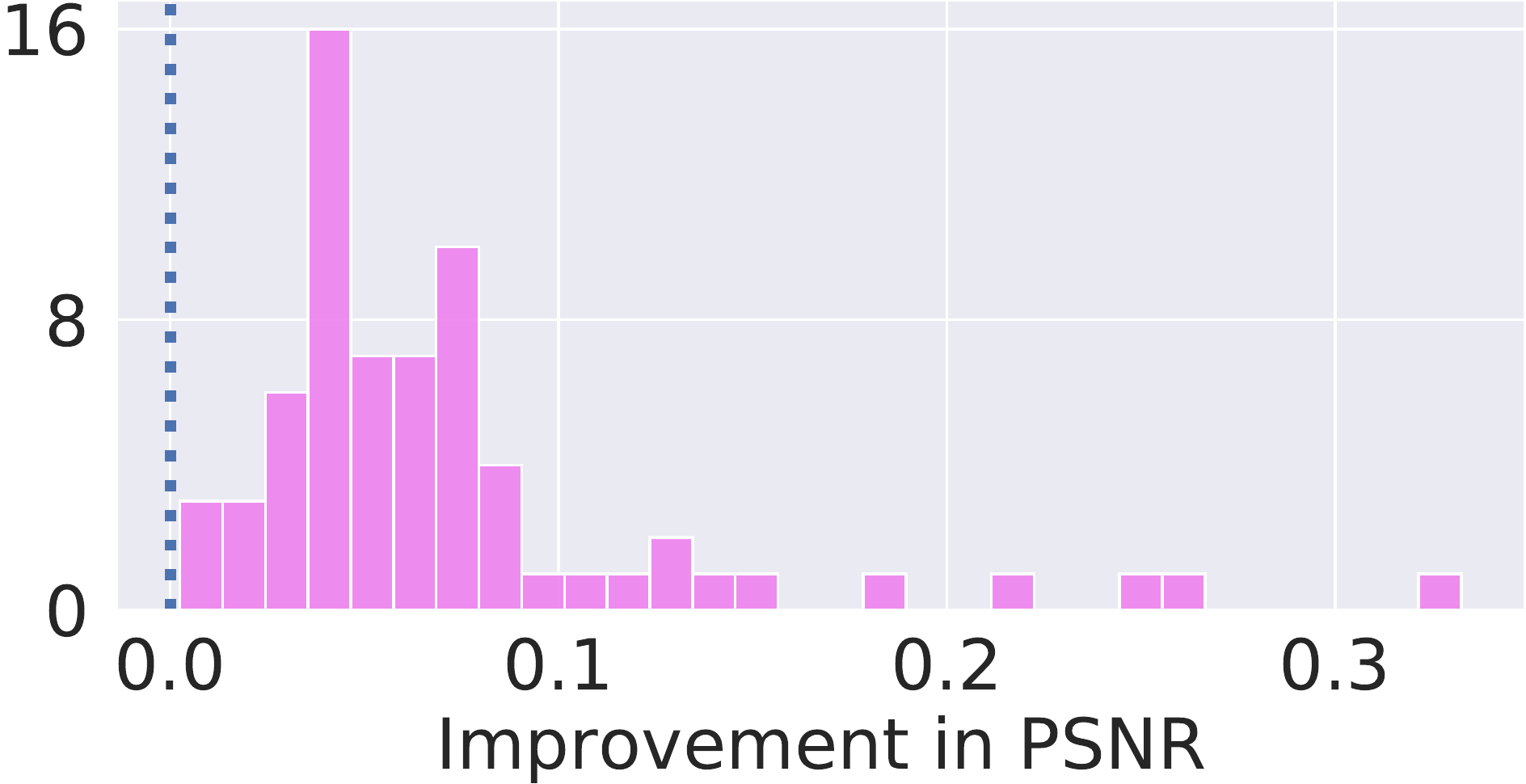}}\\
    
    & \begin{tabular}{|c|c|c|}
    \hline
         Max & Min & Num. of $\Delta$PSNR < 0 \\
         \hline
         0.360 & -0.004 & 1 \\
    \hline
    \end{tabular} &
    \begin{tabular}{|c|c|c|}
    \hline
         Max & Min & Num. of $\Delta$PSNR < 0 \\
         \hline
         0.332 & 0.002 & 0 \\
    \hline
    \end{tabular} \\[0.3cm]
    
    \midrule
    
         \multirow{2}{*}[6em]{\textbf{50}}&  \multicolumn{1}{c}{\includegraphics[width=\f1ht]{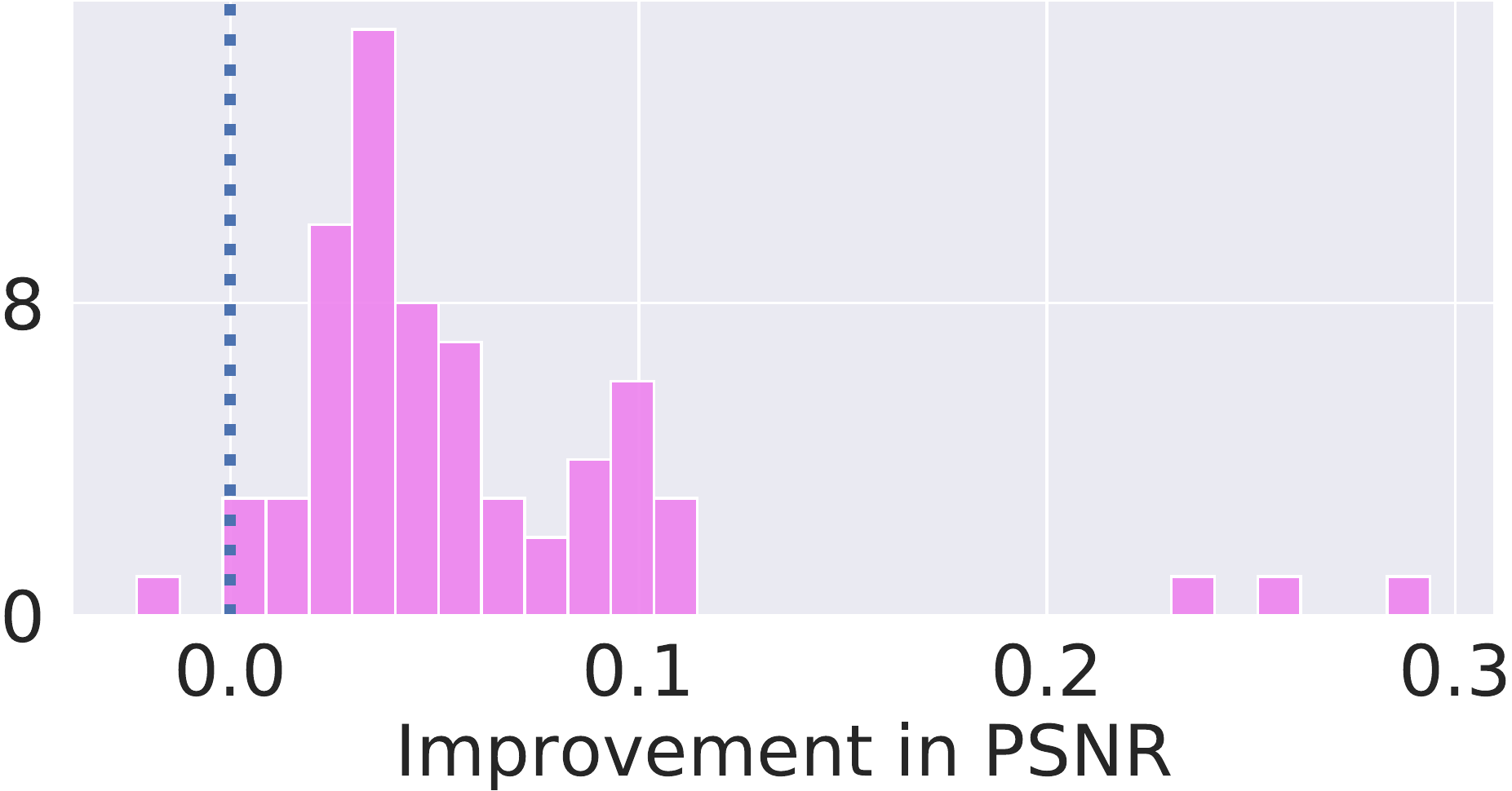}}&
    \multicolumn{1}{c}{\includegraphics[width=\f1ht]{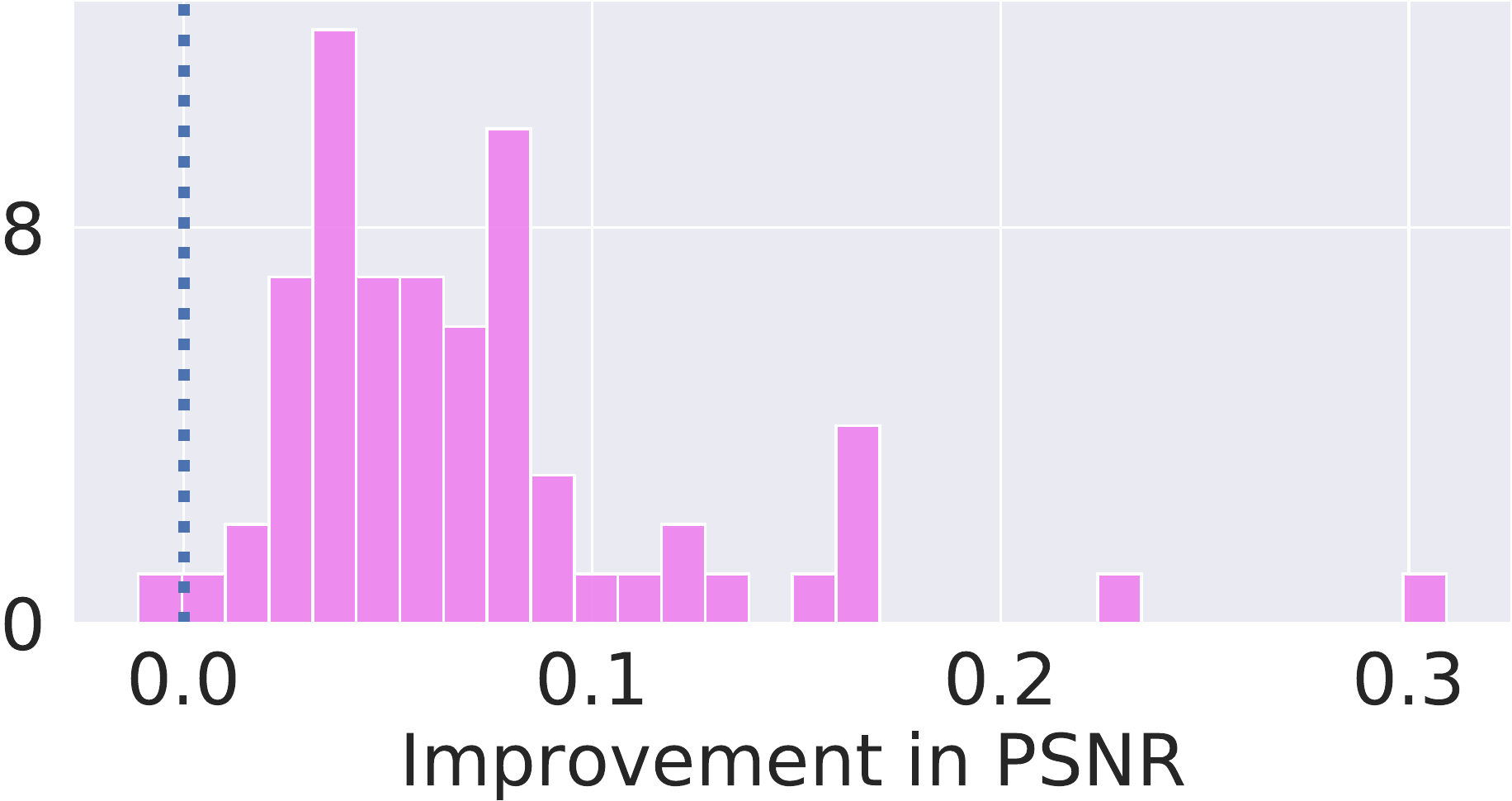}}\\
    
    & \begin{tabular}{|c|c|c|}
    \hline
         Max & Min & Num. of $\Delta$PSNR < 0 \\
         \hline
         0.294 & -0.022 & 1 \\
    \hline
    \end{tabular} &
    \begin{tabular}{|c|c|c|}
    \hline
         Max & Min & Num. of $\Delta$PSNR < 0 \\
         \hline
         0.309 & -0.011 & 1 \\
    \hline
    \end{tabular} \\[0.3cm]
    
    \bottomrule

    \end{tabular}
    }
    
\caption{\textbf{Distribution of PSNR improvement on in-distribution test set}. Distribution of improvement on BSD68 dataset at noise levels $\sigma = \{30, 40, 50\}$ (in-distribution). When the network is well optimized, and the test image is in-distribution, \gt\ can sometimes degrade the performance of the network. This degradation is atypical (in this figure, there are only 3 occurrences of degradation out of 408 experiments), and very small (in this figure, the maximum degradation is 0.022)}
\label{fig:hist_in_distr}

\end{figure}

\subsection{Out-of-distribution noise}
\label{sec:suppl_out_noise}

\begin{table}
    \footnotesize{
    \begin{tabular}{ccccccccc}
        \toprule
        
       \multirow{3}{*}{Test set} & 
       \multirow{3}{*}{$\sigma$} & 
       \multicolumn{2}{c}{\multirow{2}{*}{Trained on $\sigma \in [0, 55]$}} &
       \multicolumn{5}{c}{Baselines} \\
        \cmidrule(lr){5-9}
         &  & & &
        \multirow{2}{*}{\thead{\footnotesize{Bias Free} \\ \footnotesize{Model~\cite{biasfree}} } } & 
        \multirow{2}{*}{\thead{\footnotesize{Trained on} \\ \footnotesize{$\sigma \in [0, 100]$} } } &
        \multicolumn{2}{c}{LIDIA~\cite{lidia}}    &
        \multirow{2}{*}{S2S~\cite{self2self}} \\
        
        \cmidrule(lr){3-4}
        \cmidrule(lr){7-8}
        
        & &
        \multicolumn{1}{c}{Pre-trained} &
        \multicolumn{1}{c}{\gt} &
        & &
        \multicolumn{1}{c}{Pre-trained} &
        \multicolumn{1}{c}{Adapted} & \\

        \midrule
        \multirow{2}{*}{Set12} 
        & $70$ & 22.45 & 25.54 & 25.59  & 25.50 & 23.69 & 25.01 & 24.61  \\
        & $80$ & 18.48 & 24.57 & 24.94 & 24.88 & 22.12 & 24.17 & 23.64 \\
        \midrule
        \multirow{2}{*}{BSD68} 
        & $70$ & 22.15 & 24.89 & 24.87     & 24.88 & 23.28 & 24.57 & 24.29   \\
        & $80$ & 18.72  & 24.14 &  24.38     &  24.36  & 21.87 & 23.97 & 23.65 \\
        \bottomrule
    \end{tabular}
    }
    \caption{\textbf{\gt\ for out-of-distribution noise}. We evaluate a DnCNN trained on generic natural images for $\sigma \in [0, 55]$ on a test set of generic natural images corrupted with $\sigma = \{70, 80\}$, which is outside the training range of the network. \gt\ is able is generalize effectively to this out-of-distribution test set. \gt\ achieves comparable performance to a network trained with supervision on a large range of noise levels ($\sigma \in [0, 100]$) an bias-free models which is an architecture explicitly designed to generalize to noise levels outside the training range. \gt\ also outperforms LIDIA~\cite{lidia}, a specialized architecture and adaptation procedure, and Self2Self~\cite{self2self}, a method trained exclusively on the test image.}
    \label{tab:out_noise_suppl}
\end{table}

\textbf{Different Architectures}. We summarize the results using DnCNN in Table~\ref{fig:all_gen} in the main paper. Figure~\ref{fig:all_vs_gain_suppl} shows that the UNet architecture is also able to generalize to out-of-distribution noise.

\textbf{Different Loss Functions}. We provide the results of evaluating DnCNN architecture with different cost functions in Table~\ref{tab:lossfun_comparison}. 

\textbf{Comparison to baselines}. Table~\ref{tab:out_noise_suppl} summarizes the result of evaluating a DnCNN trained on generic natural images for $\sigma \in [0, 55]$ on a test set of generic natural images corrupted with $\sigma = \{70, 80\}$, which is outside the training range of the network. \gt\ is able to generalize effectively to this out-of-distribution test set. \gt\ achieves comparable performance to a network trained with supervision on a large range of noise levels ($\sigma \in [0, 100]$), and a bias-free model which is explicitly designed to generalize to noise levels outside the training range. \gt\ also outperforms LIDIA~\cite{lidia} (a specialized architecture and adaptation procedure). and Self2Self~\cite{self2self} (a method trained exclusively on the test image). 

\subsection{Out-of-distribution image}
\label{sec:suppl_out_image}

\begin{table}
    \centering
    \footnotesize{
    \begin{tabular}{cccccccc}
        \toprule
        
       & \multirow{3}{*}{\thead{\footnotesize{Training} \\ \footnotesize{Data}}} & 
       \multirow{3}{*}{\thead{\footnotesize{Test} \\ \footnotesize{Data}}} & 
       \multicolumn{2}{c}{\multirow{2}{*}{DnCNN~\cite{dncnn}}} &
       \multicolumn{3}{c}{Baselines} \\
       
        \cmidrule(lr){6-8}
         & &  & & &
        \multicolumn{2}{c}{LIDIA~\cite{lidia}}    &
        \multirow{2}{*}{S2S~\cite{self2self}} \\
        
        \cmidrule(lr){4-5}
        \cmidrule(lr){6-7}
        
        & & &
        \multicolumn{1}{c}{Pre-trained} &
        \multicolumn{1}{c}{\gt} &
        \multicolumn{1}{c}{Pre-trained} &
        \multicolumn{1}{c}{Adapted} & \\

        \midrule
        (a) & \thead{\footnotesize{Piecewise} \\ \footnotesize{constant}}  & \thead{\footnotesize{Natural} \\ \footnotesize{images}} & 27.31 & 28.60 &- & - & 29.21\\
        (b) & \thead{\footnotesize{Natural} \\ \footnotesize{images}} & \thead{\footnotesize{Urban} \\ \footnotesize{images}} & 28.35 & 28.79  & 28.54 & 28.71 & 29.08\\
        (c) & \thead{\footnotesize{Natural} \\ \footnotesize{images}} & \thead{\footnotesize{Scanned} \\ \footnotesize{documents}} & 30.02 & 30.73  & 30.05 & 30.23 & 30.86\\
        \bottomrule
    \end{tabular}
    }\\[0.2cm]
    
    \caption{\textbf{\gt\ for out-of-distribution images}. \gt\ generalizes robustly when the test image has different characteristics than the training data. We demonstrate this through three different experiments. (a) \gt\ provides an average of 1.3 dB in performance while adapting a CNN trained on simulated piecewise constant dataset to natural images. This controlled setting demonstrates the capability of \gt\ to adapt from a simple simulated training set to a significantly more complex real dataset. (b) \gt\ provides an average of 0.45 dB improvement in performance when a CNN trained on natural images is adapted to a dataset of images taken in urban settings. These images display a lot of repeating structure (see Section~\ref{sec:datasets}) and hence has different characters than generic natural images. Similarly, (c) \gt\ provides an average of 0.70 dB improvement in performance when a CNN pre-trained on natural images is adapted to images of scanned documents. While \gt\ outperforms LIDIA~\cite{lidia}, a specialized architecture designed for adapting, it does not match the performance of Self2Self (see Section~\ref{sec:limitations} for a discussion on this). As noted in Section~\ref{sec:exp_out_signal}, we did not train LIDIA for (a).} 
    \label{tab:out_signal_suppl}
\end{table}

\textbf{Different Architectures}. We summarize the results using DnCNN in Table~\ref{fig:all_gen} in the main paper. Figures~\ref{fig:all_vs_gain_suppl} show that the UNet and BFCNN architectures are also able to generalize to test data with different characteristics from the training data when adapted using \gt\ . 

\textbf{Different Loss Functions}. We provide the results of evaluating the DnCNN architecture with different cost functions in Table~\ref{tab:lossfun_comparison}. 

\textbf{Comparison to baselines}. Results of comparison to LIDIA~\cite{lidia}, a specialized architecture to perform adaptation, and Self2Self~\cite{self2self} a method trained exclusively on the test image is summarized in Table~\ref{tab:out_signal_suppl}. While \gt\ outperforms LIDIA, it does not match the performance of Self2Self (see Section~\ref{sec:limitations} for a discussion on this).

\subsection{Out-of-distribution noise and image}
\label{sec:suppl_out_noise_and_image}
We evaluated the ability of \gt\ to adapt to test images which have different characteristics from those in the training set, and are additionally corrupted with a noise distribution that is different from the noise in the training set. Figure~\ref{fig:suppl_out_noise_signal} shows that \gt\ is successful in this setting. The CNN was pre-trained on natural images corrupted with Gaussian white noise of standard deviation $\sigma \in [0, 55]$. We used \gt\ to adapt this CNN to a test set of images taken in urban setting (see Section~\ref{sec:datasets} for a discussion on how it is different from natural images), corrupted with Gaussian noise of standard deviation $\sigma=70$ (which is outside the training range of $[0, 55]$). 

\subsection{Application to Electron Microscopy}
\label{sec:suppl_nano}

\begin{figure}
    \centering
    \begin{tabular}{c@{\hskip 0.01in}c@{\hskip 0.01in}c@{\hskip 0.01in}c@{\hskip 0.01in}c@{\hskip 0.01in}c}
    \scriptsize{(a) Noisy image} &  
    \thead{ \scriptsize{(b) Unsupervised training} \\ \scriptsize{only on (a) \cite{self2self}}} &
    \thead{\scriptsize{(c) Supervised training} \\ \scriptsize{on simulated data~\cite{mohan2020deep}}} &
    \thead{\scriptsize{(d) \gt\ on CNN} \\ \scriptsize{trained on sim. data (c) }} &
    \thead{ \scriptsize{(e) Estimated reference} \\ \scriptsize{image} } \\
    \includegraphics[width=1.05in]{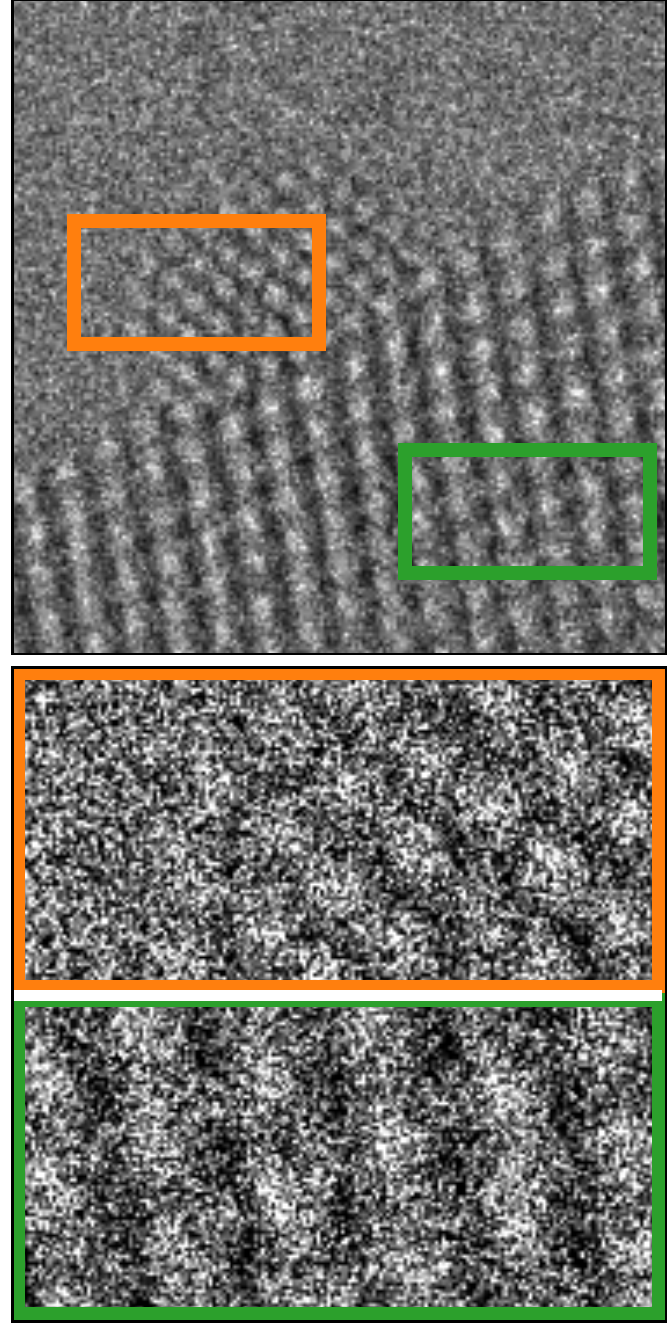}&
    \includegraphics[width=1.05in]{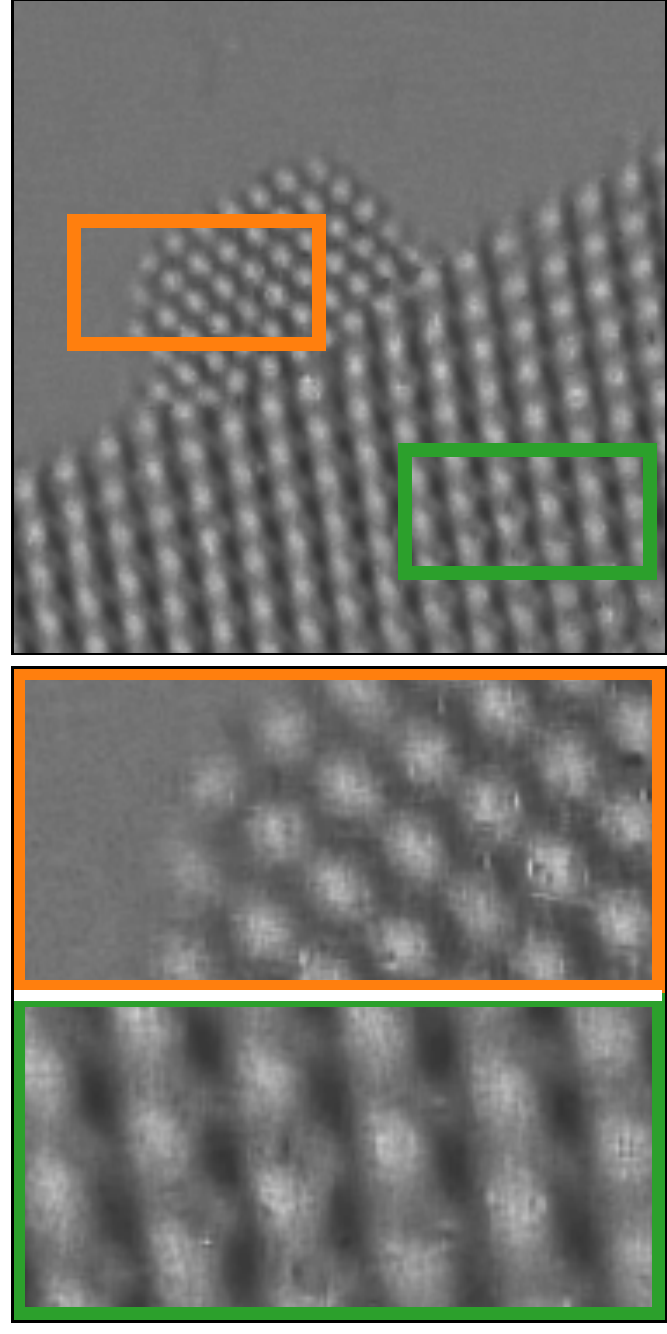}&
    \includegraphics[width=1.05in]{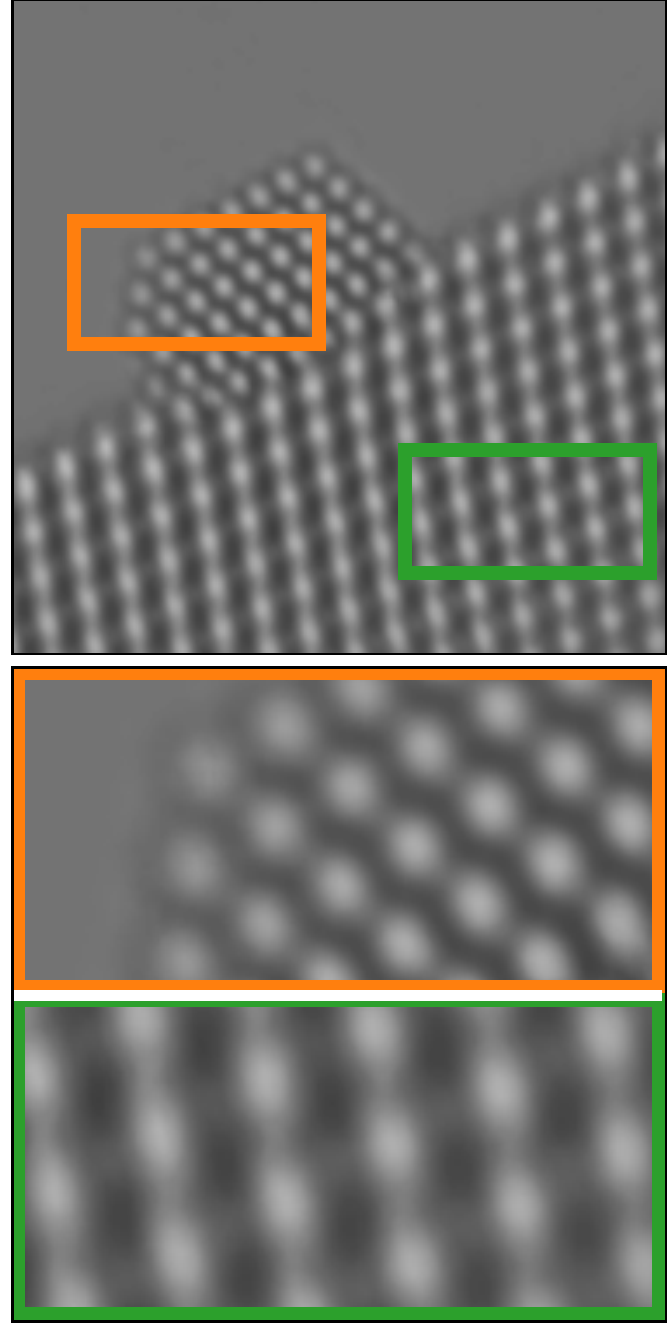}&
    \includegraphics[width=1.05in]{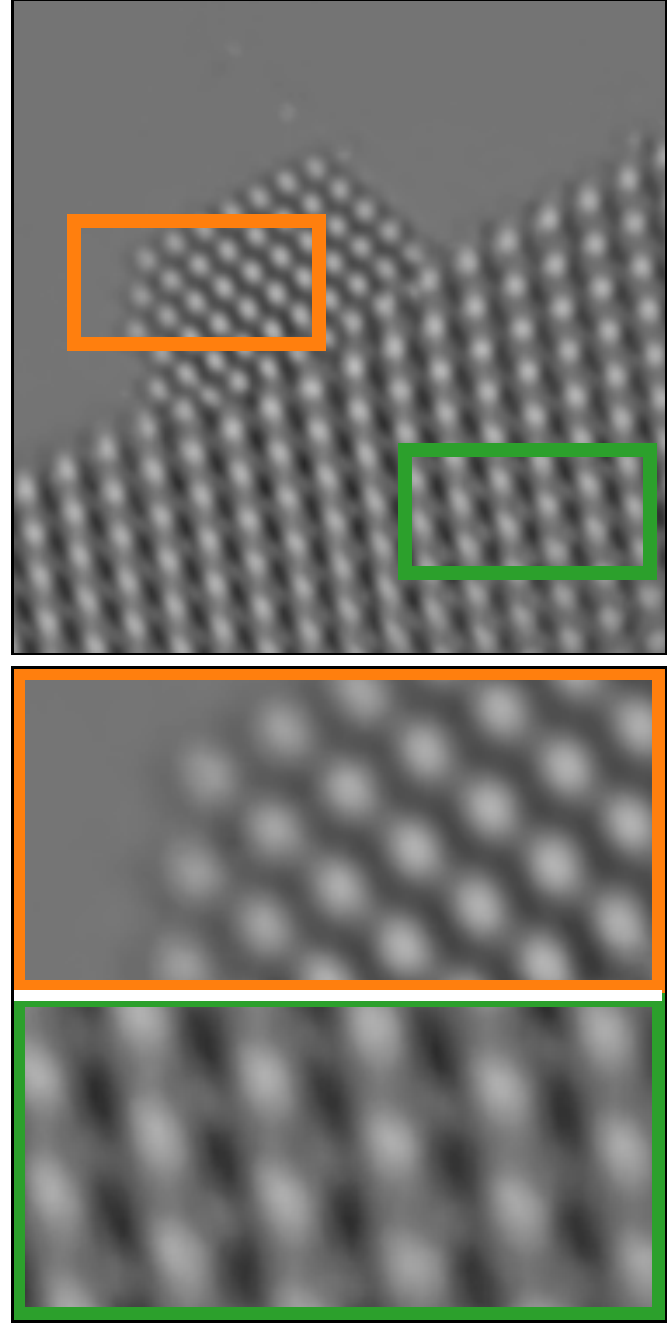}&
    \includegraphics[width=1.05in]{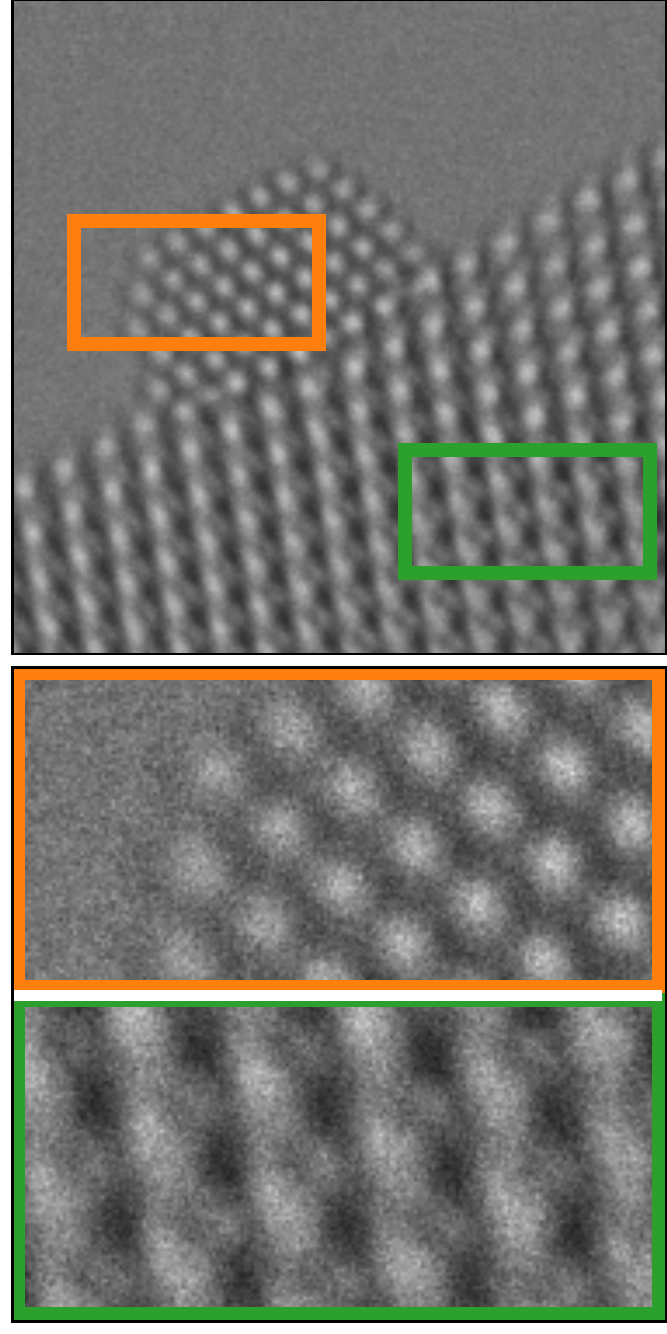}\\
    \end{tabular}
    \vspace{-0.2cm}
    \caption{\textbf{Denoising results for real-world data.} (a) An experimentally-acquired atomic-resolution transmission electron microscope image of a CeO2-supported Pt nanoparticle. The image has a very low signal to noise ratio (PSNR of $\approx 3 dB$). (b) Denoised image obtained using Self2Self~\cite{self2self}, which contains significant artefacts. (c) Denoised image obtained via a CNN trained on a simulated dataset, where the pattern of the supporting atoms is not recovered faithfully (third row). (d) Denoised image obtained by adapting the CNN in (c) to the noisy test image in (a) using \gt. Both the nanoparticle and the support are recovered without artefacts. (e) Reference image, estimated by averaging $40$ different noisy images of the same nanoparticle. See Figure~\ref{fig:nano} for an additional example.
    }
    \label{fig:nano_appendix}
\end{figure}

\begin{figure}
    \centering
    \begin{tabular}{c@{\hskip 0.01in}c@{\hskip 0.01in}c@{\hskip 0.01in}c@{\hskip 0.01in}c@{\hskip 0.01in}c}
    Noisy image &  Blind-spot~\cite{blindspotnet}  & \thead{Blind-spot \\ Early Stopping~\cite{udvd}} & UDVD~\cite{udvd} & \thead{UDVD \\ Early stopping~\cite{udvd}} \\
    \includegraphics[width=1.05in]{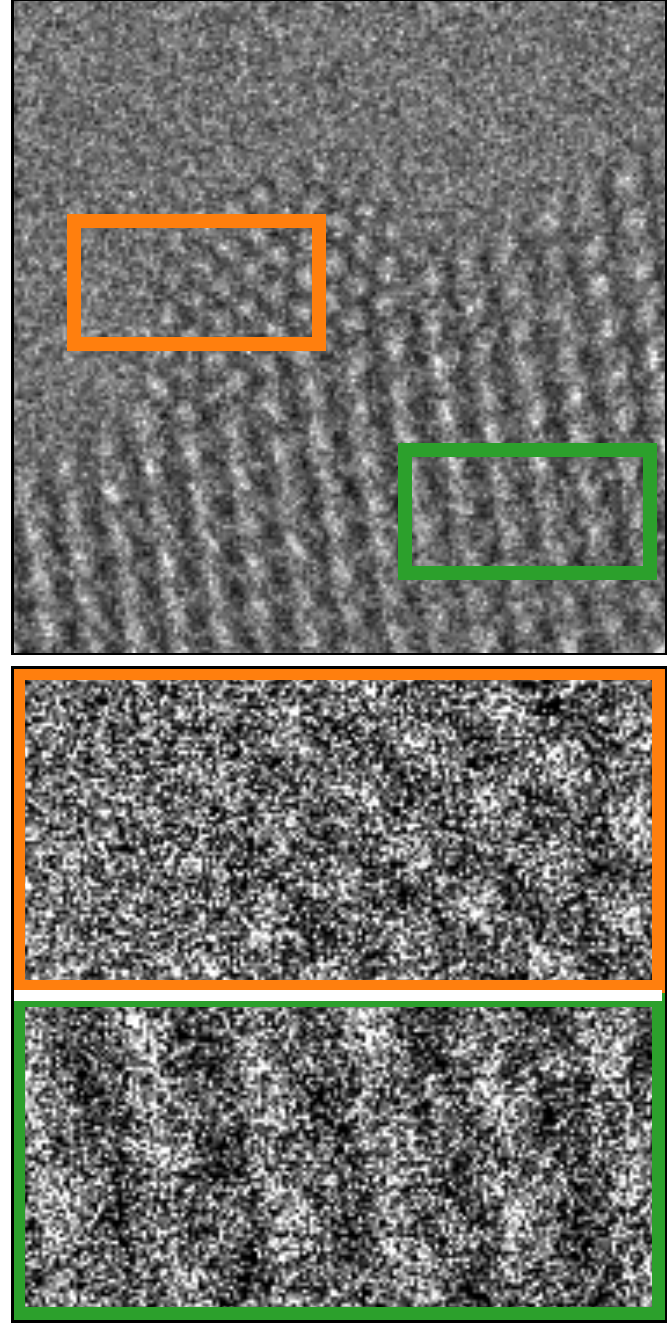}&
    \includegraphics[width=1.05in]{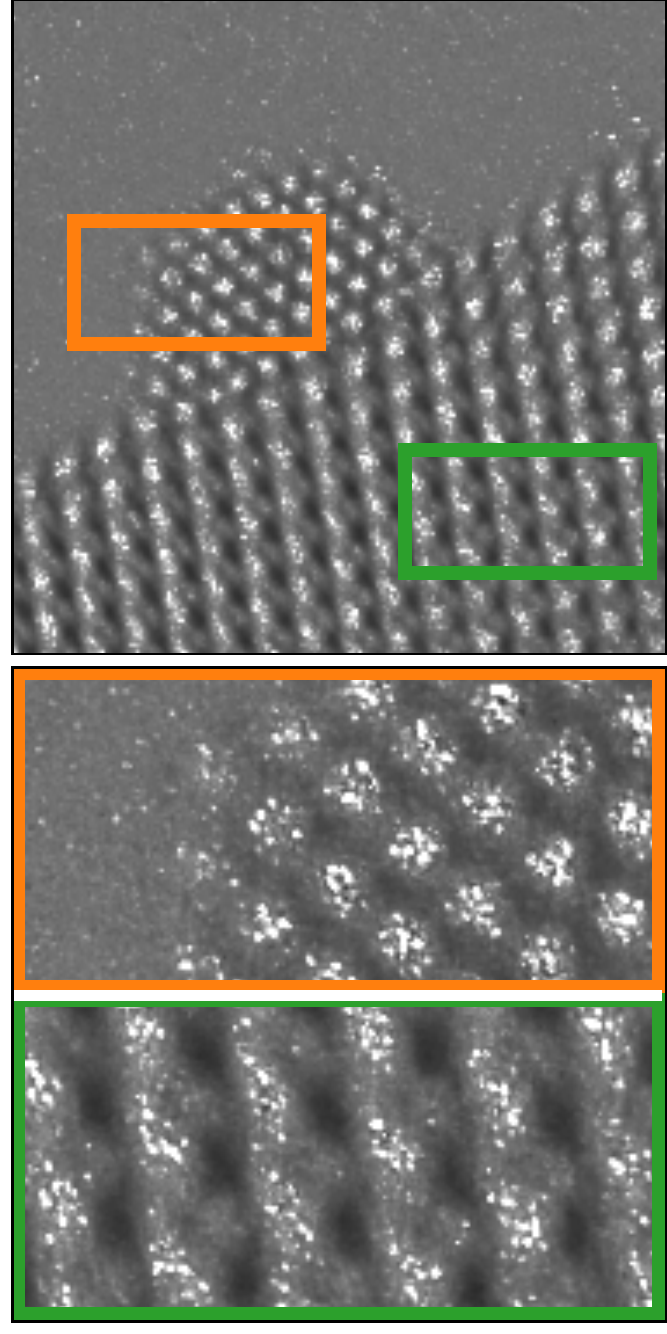}&
    \includegraphics[width=1.05in]{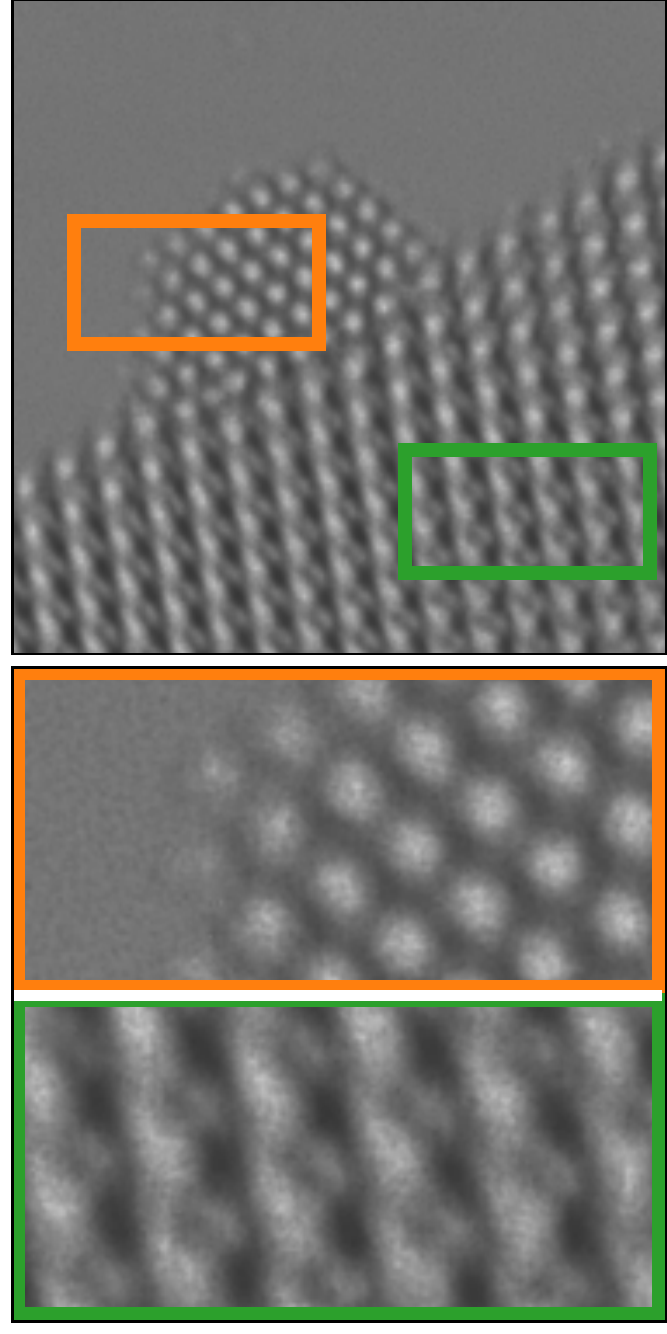}&
    \includegraphics[width=1.05in]{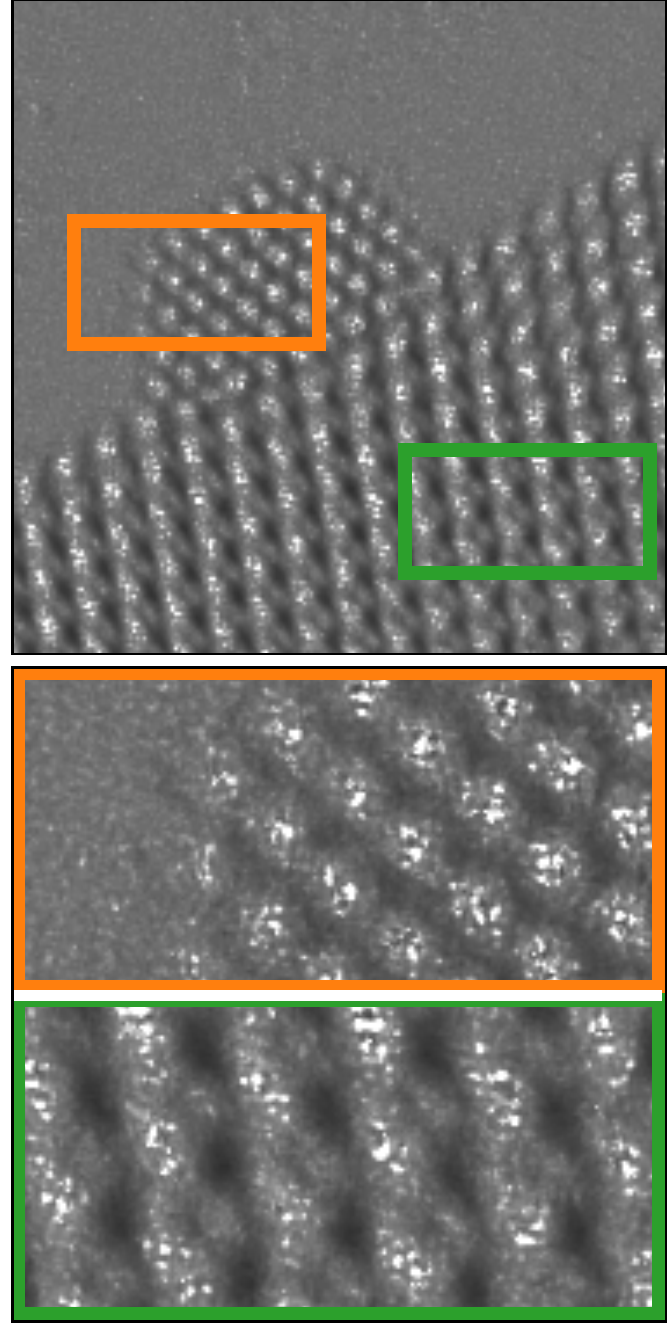}&
    \includegraphics[width=1.05in]{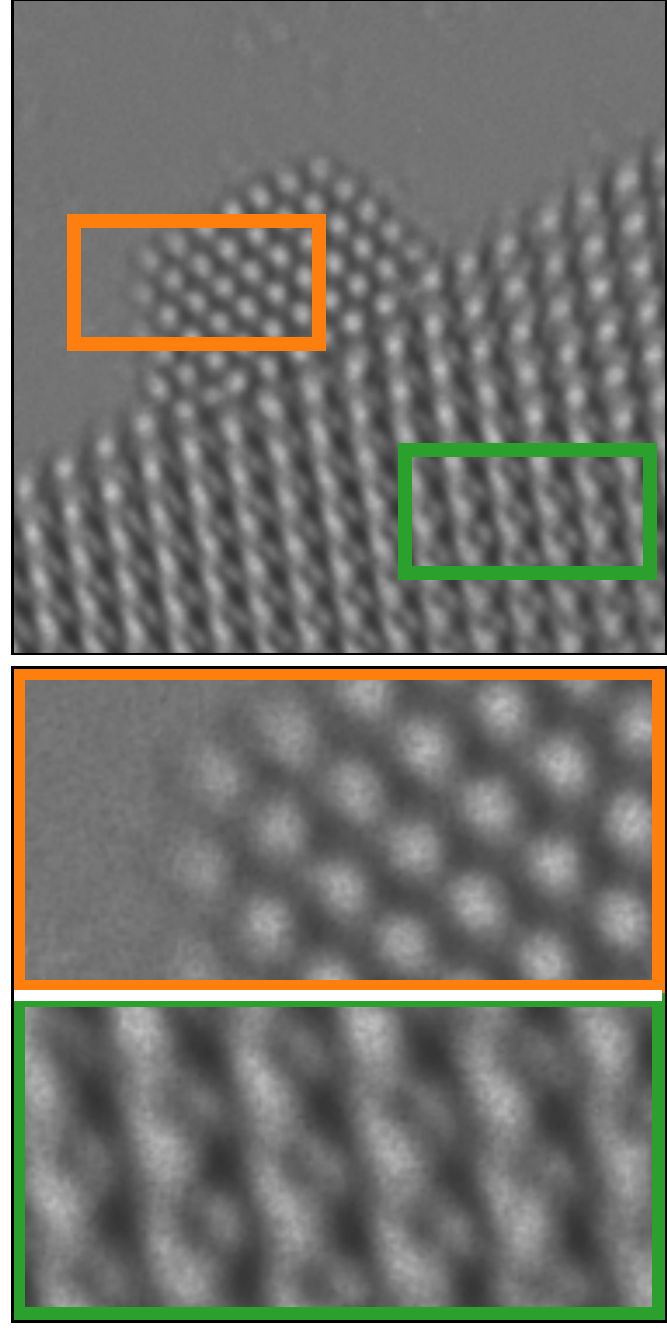}\\
     Self2Self~\cite{self2self}  & Pre-trained~\cite{mohan2020deep, vincent2021developing} & \thead{\gt \\ Noise resampling} & \thead{\gt \\ blind-spot} & \thead{Estimated \\ reference} \\
    \includegraphics[width=1.05in]{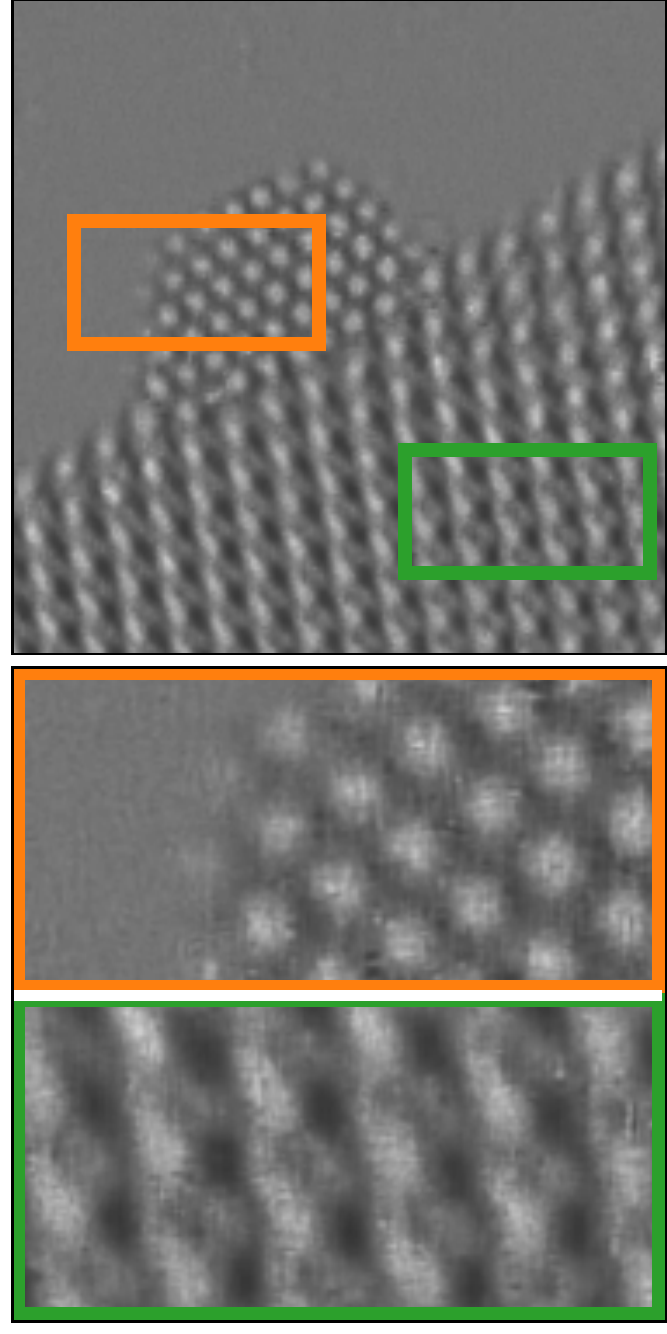}&
    \includegraphics[width=1.05in]{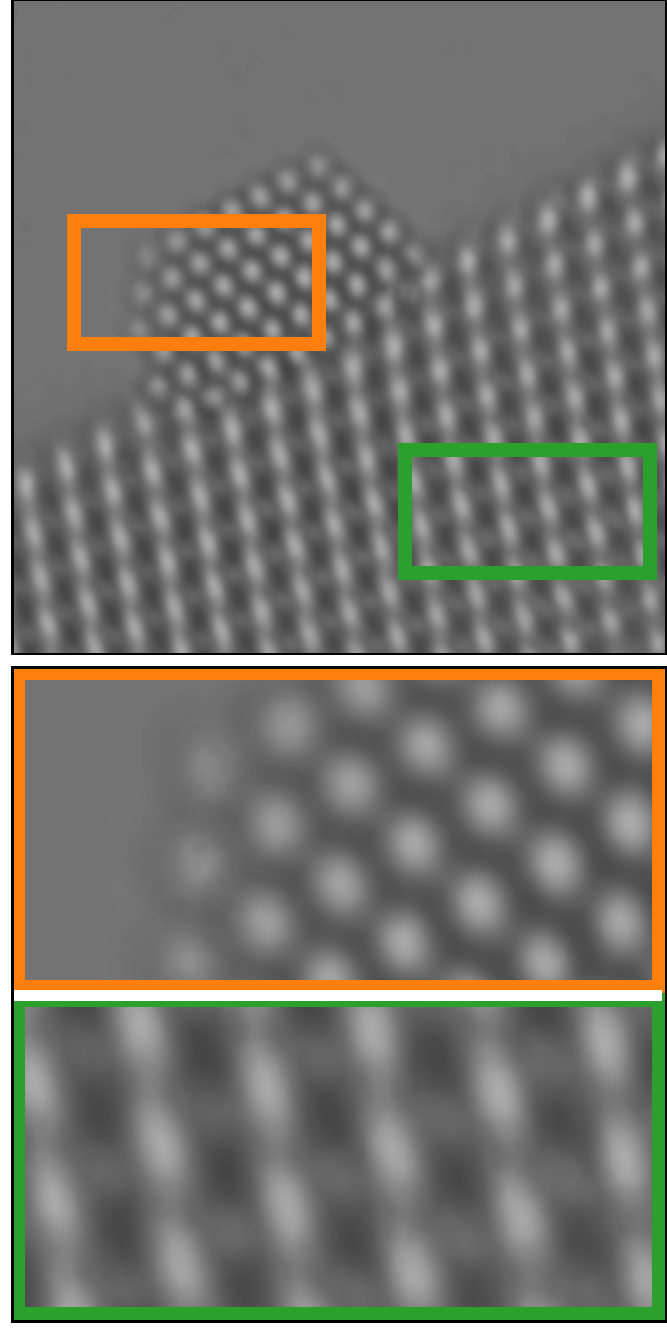}&
    \includegraphics[width=1.05in]{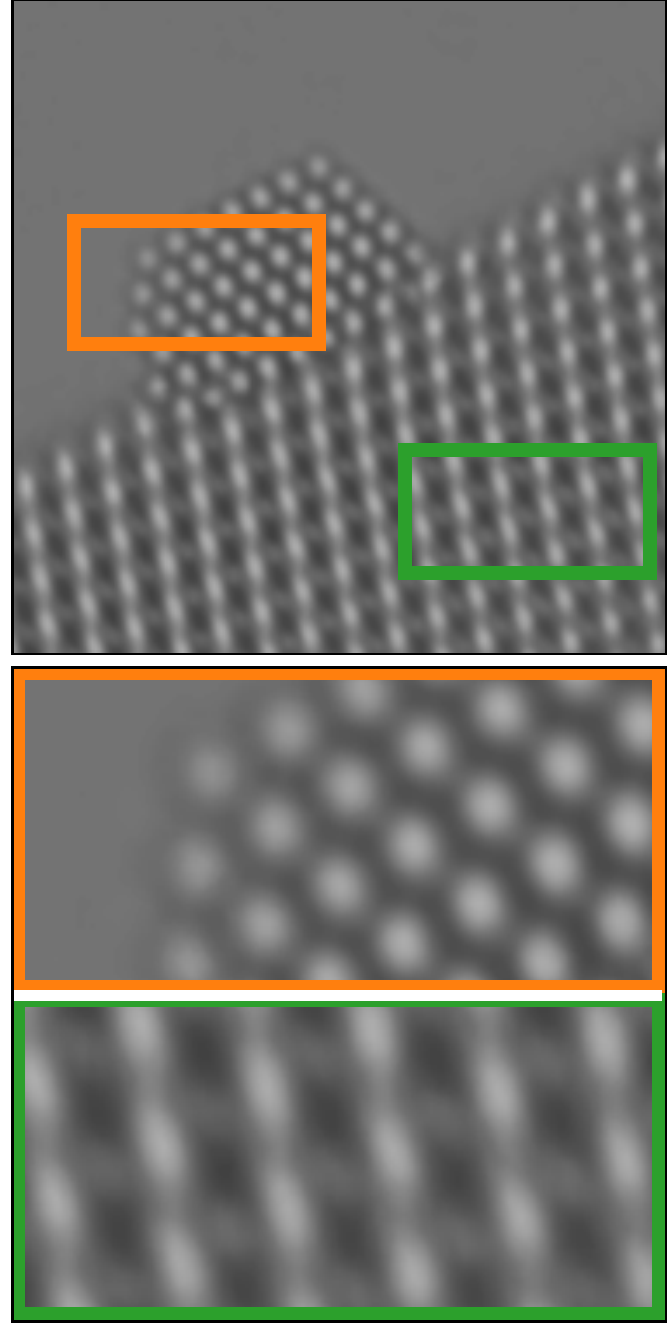}&
    \includegraphics[width=1.05in]{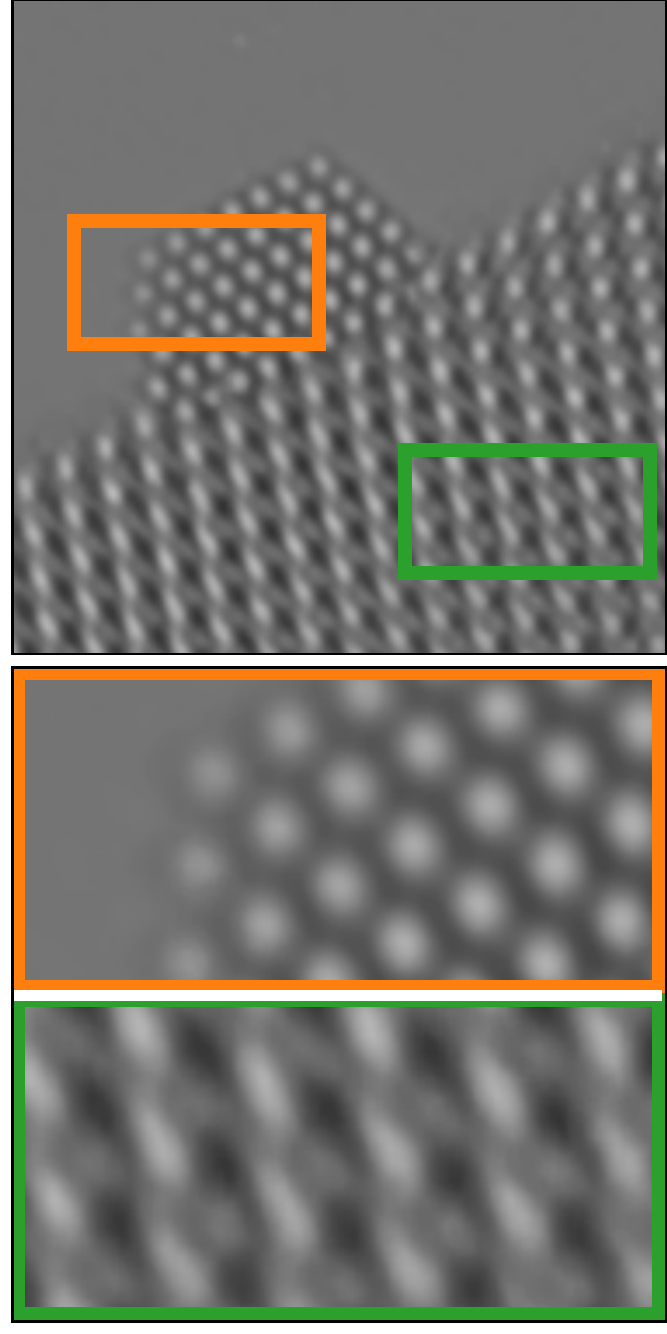}&
    \includegraphics[width=1.05in]{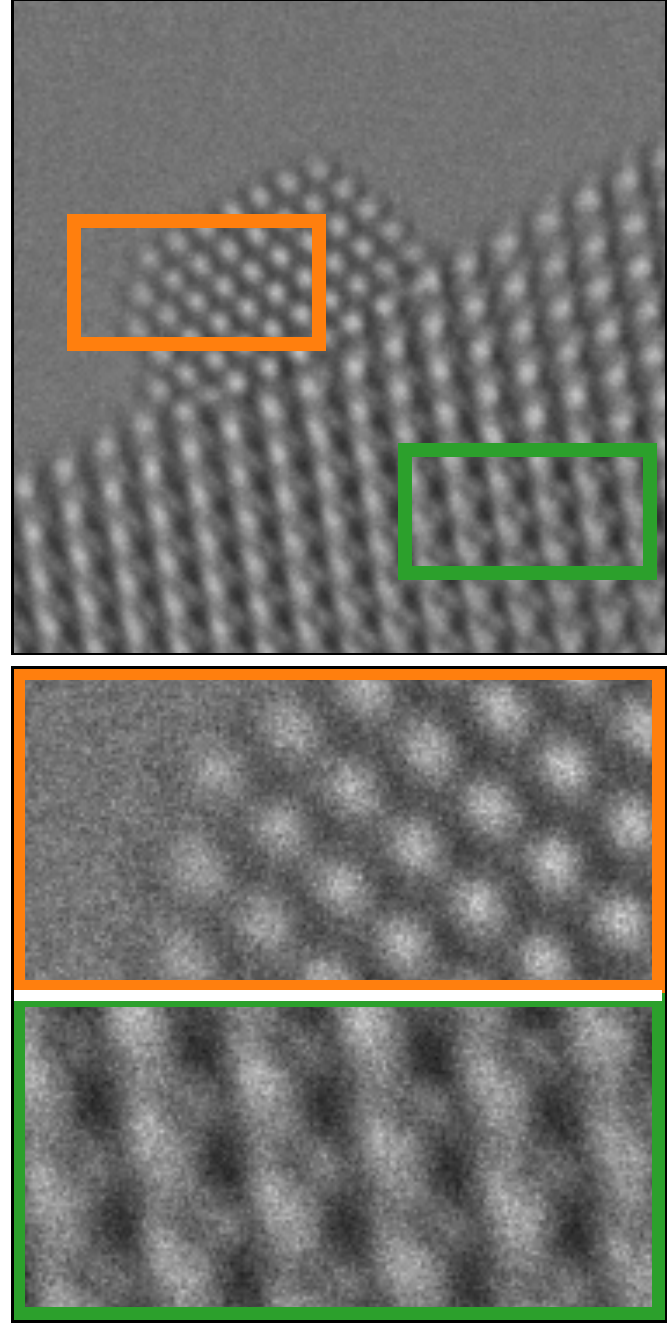}\\
    \end{tabular}
    \caption{\textbf{Comparison with baselines for electron microscopy}. \gt\ clearly outperforms Self2Self, which is trained exclusively on the real data. The denoised image from Self2Self shows missing atoms and substantial artefacts (see Figure~\ref{fig:nano_appendix} for another example). We also compare \gt\ dataset to blind-spot methods using the 40 test frames~\cite{blindspotnet,udvd}. \gt\ clearly outperforms these methods.}
    \label{fig:nano_unsup_comparison}
\end{figure}

\textbf{Comparison to pre-trained CNN}. As discussed in Section~\ref{sec:exp_nano}, a CNN~\cite{blindspotnet} pre-trained on the simulated data fails to reconstruct the pattern of atoms faithfully. We show an additional example (Figure~\ref{fig:nano_appendix}) to support this. \gt\ applied to the pre-trained CNN  using the blind-spot loss correctly recovers this pattern (green box in Figure~\ref{fig:nano_appendix} (d),~(e)) reconstructing the small oxygen atoms in the CeO$_2$ support.  \gt\ with noise resampling failed to reproduce the support pattern, probably because it is absent from the initial denoised estimate (see Figure~\ref{fig:nano_unsup_comparison}).

\textbf{Comparison to baselines.} Since no ground-truth images are available for this dataset (see Section~\ref{sec:exp_nano}), we average 40 different acquisitions of the same underlying image to obtain an estimated reference for visual reference. We also compare \gt\ to state-of-the-art dataset based unsupervised methods, which are trained on these 40 images. 
\begin{itemize}
    \item \textbf{Blind-spot net}~\cite{blindspotnet} is a CNN which is constrained to predict the intensity of a pixel as a function of the noisy pixels in its neighbourhood, without using the pixel itself. This method is competitive with the current supervised state-of-the-art CNN on photographic images. However, when applied to this dataset it produces denoised images with visible artefacts (see Figure~\ref{fig:nano_unsup_comparison}). Ref.~\cite{udvd} shows that this may be because of the limited amount of data ($40$ noisy images): They trained a blind-spot net on simulated training sets of different sizes, observing that the performance on held-out data is indeed poor when the training set is small, but improves to the level of supervised approaches for large training sets. 
    
    \item \textbf{Unsupervised Deep Video Denoising (UDVD)}~\cite{udvd} is an unsupervised method for denoising video data based on the blind-spot approach. It estimates a denoised frame using $5$ consecutive noisy frames around it. Our real data consists of $40$ frames acquired sequentially. UDVD produces better results than blind-spot net, but still contains visible artefacts, including missing atoms (see Figure~\ref{fig:nano_unsup_comparison}). Note that UDVD uses $5$ noisy images as input, and thus has more context to perform denoising than the other methods (including \gt\ ). 
    
    \item \textbf{Blind-spot net with early stopping}. 
    In Ref.~\cite{udvd} it is shown that early stopping based on noisy held-out data can boost the performance of blind-spot nets. 
    Here we used $35$ images for training the blind-spot net and the remaining $5$ images as a held-out validation set. We chose the model parameters that minimized the mean squared error between the noisy validation images and the corresponding denoised estimates.
    The results (shown in Figure~\ref{fig:nano_unsup_comparison}) are significantly better than those of the standard blind-spot network. However, there are still noticeable artefacts, which include missing atoms. This method is similar in spirit to \gt\ - but uses a different strategy to prevent overfitting. 
    
    \item \textbf{Unsupervised Deep Video Denoising (UDVD) with early stopping}. Similar to blind-spot net, performing early stopping on UDVD using 5 held-out frames greatly improves its performance~\cite{udvd} (Figure~\ref{fig:nano_unsup_comparison})). However, there are still noticeable artefacts in the denoised output. 
    
\end{itemize}

\subsection{Different loss functions}
\label{sec:suppl_exp_summary}

\gt\ can be used in conjunction with any unsupervised denoising cost function. We explore three different choices - SURE, noise resampling, and blind-spot cost functions (see Section~\ref{sec:unsup_loss}), and summarize our finding in Table~\ref{tab:lossfun_comparison}.  

SURE loss outperforms other choices in most experiments. Noise resampling has comparable performance to SURE when the test data is in-distribution, or when it is corrupted with out-of-distribution noise. However, noise resampling generally under-performs SURE when the test images have different features from the training images. A possible explanation for this is that noise resampling relies on the initial denoised image to fine-tune and, therefore, it may not be able to exploit features which are not present in the initial estimate. In contrast, the SURE cost function is computed on the noisy test image itself, thereby enabling it to adapt to features that the pre-trained network may be agnostic to. 

Finally, adapting using blind-spot cost function often under-performs both SURE and noise resampling. The difference in performance is reduced at higher noise levels (see also Section~\ref{sec:exp_nano} where we use blind-spot cost function for experiments with real TEM data with very high noise). %
The reason for this could be that at higher noise levels, the information contained in a single pixel becomes less relevant for computing the corresponding denoised estimate (in fact, the regularization penalty on ``self pixel` for SURE cost function (Section~\ref{sec:unsup_loss}) increases as the noise level increases). Therefore, the loss of performance incurred by the blind-spot cost function is diminished. At lower noise levels (particularly when the images are in-distribution), adapting using blind-spot cost function will force the pre-trained network to give up using the ``self pixel``, which results in a degraded performance. An alternative to adapting a generic pre-trained network using blind-spot architecture is to use a CNN that is architecturally constrained to include a blind-spot. In Table~\ref{tab:blindspotresults}, we show that adapting such a CNN using blind-spot loss improves the performance its performance. However, the overall performance of this architecture is in general lower than the networks which also use the ``self pixel``. We refer interested readers to Ref.~\cite{blindspotnet, pn2v, noise2same} for approaches to incorporate the noisy pixel into the denoised estimate.

\begin{table}[]
\centering

    \begin{tabular}{cccccc}
        \toprule
        \multicolumn{1}{c}{\phantom} &
        \multicolumn{1}{c}{\phantom} &
        \multicolumn{1}{c}{\phantom} &
        \multicolumn{3}{c}{\gt\ with} \\
        \cmidrule(lr){4-6}
         &  & Pre-training & SURE & \thead{Noise \\resampling} & \thead{Blind-spot \\(Noise2Self~\cite{noise2self})} \\
        \midrule
        \multirow{2}{*}{\thead{in distribution}}  & Set12 &  29.52 &  29.62 & \textbf{29.63} & 29.50 \\
        & BSD68 & 28.39 & \textbf{28.46} & 28.40 & 28.36 \\
        \midrule
        \multirow{2}{*}{\thead{out-of-distribution \\ noise}}  & Set12 &  18.48 &  \textbf{24.57} & 24.11 & 22.93  \\
        & BSD68 & 18.72 & \textbf{24.14} & 23.65 & 22.50  \\
        \midrule
        \multirow{3}{*}{\thead{out-of-distribution \\ image}}  & \thead{Piecewise constant $\rightarrow$ \\ Natural images} &  27.31 &  \textbf{28.60} & 28.29 & 27.39  \\
        & \thead{Natural images $\rightarrow$ \\ Urban100} &  28.35 &  \textbf{28.79} & \textbf{28.79} & 28.29 \\
        & \thead{Natural images $\rightarrow$ \\ Scanned documents} &  30.02 &  \textbf{30.73} & 30.57 & 29.23  \\
        \bottomrule
    \end{tabular}\\[0.3cm]

    \caption{\textbf{Different loss functions for \gt}. Comparison of the performance of \gt\ when used in conjunction with three different loss functions. SURE loss outperforms other choices in most experiments. Noise resampling has comparable performance to SURE when the test data is in-distribution, or when it is corrupted with out-of-distribution noise. However, noise resampling generally under-performs SURE when the test images have different features from the training images. This maybe because such features are absent from the initial denoised estimate (see Section~\ref{sec:unsup_loss} for a description of the different loss functions). Finally, optimizing using blind-spot cost functions often under-performs both SURE and noise resampling, but the difference in performance is reduced as the test noise increases (see also Section~\ref{sec:exp_nano} where we use blind-spot cost function for experiments with real TEM data with very high noise). This may be because, at lower noise levels, the information contained in a pixel is often crucially important to compute its denoised estimate, and blind-spot cost function ignores this information (see Section~\ref{sec:unsup_loss}). Here, we implemented blind-spot cost function through masking~\cite{noise2self}, see Table~\ref{tab:blindspotresults} for results where the implemented blind-spot cost function as an architectural constraint~\cite{blindspotnet}. We note that GainTuning with blind-spot cost function based on masking was very sensitive to the optimization hyper parameters used, and it is possible that the results will improve further with careful choice of optimization parameters.
    }
    \label{tab:lossfun_comparison}
\end{table}

\begin{table}[]
\centering
{ %
    \renewcommand{\arraystretch}{1.5}
    \begin{tabular}{c@{\hskip 0.65in}c@{\hskip 0.35in}c@{\hskip 0.35in}c@{\hskip 0.35in}c}
    \toprule
    & \multicolumn{2}{c}{in-distribution} &  \multicolumn{2}{c}{out-of-distribution image}\\
    \cmidrule(lr){2-3}
    \cmidrule(lr){4-5}
    & Set12 &  BSD68 &  \thead{Urban100 \\(urban scenes)} &  \thead{IUPR \\ (scanned documents)} \\
   \midrule
   Pre-trained & 27.92 & 26.47 &  26.59 & 28.25  \\
   \gt\ & 27.92 & 26.61 & 26.85 & 28.40 \\
    \bottomrule

    \end{tabular}\\[0.3cm]
}
    \caption{\textbf{\gt\ using architecturally constrained blind-spot cost function}. We perform \gt\ using blindspot network~\cite{blindspotnet} which is architecturally constrained to estimate a denoised pixel exclusively from its neighbouring pixels (excluding the pixel itself). The network was pre-trained on generic natural images corrupted with Gaussian noise of standard deviation $\sigma \in [0, 55]$. Performing \gt\ on this always increases its performance, unlike \gt\ on a generic architecture trained with supervision and adapted using blind-spot loss implemented via masking. However, note the overall performance of this architecture is in general lower than the networks which also use the ``self pixel``. We refer interested readers to Ref.~\cite{blindspotnet, pn2v, noise2same} for approaches to incorporate the information in noisy pixel back into the denoised output, thus potentially improving the performance. Our blind-spot architecture generalizes robustly to out-of-distribution noise (since it is bias-free~\cite{biasfree}), and therefore we do not include an out-of-distribution noise comparison in this table. 
    }
    \label{tab:blindspotresults}
\end{table}

\section{Analysis}
\label{sec:suppl_analysis}

\subsection{What kind of images benefit the most from adaptive denoising?}
\label{sec:suppl_analysis_top5}
\begin{figure}
\def\f1ht{0.9in}%
\centering 
\footnotesize{
\begin{tabular}{ >{\centering\arraybackslash}m{.1in}>{\centering\arraybackslash}m{.3in} >{\centering\arraybackslash}m{2.0in}  >{\centering\arraybackslash}m{2.0in}}
\toprule
\centering 
     &  & DnCNN & UNet \\[0.2cm]
    \midrule
    \multirow{2}{*}{\rotatebox[origin=c]{90}{In-distribution}} & Top 6 &
    \includegraphics[height=\f1ht]{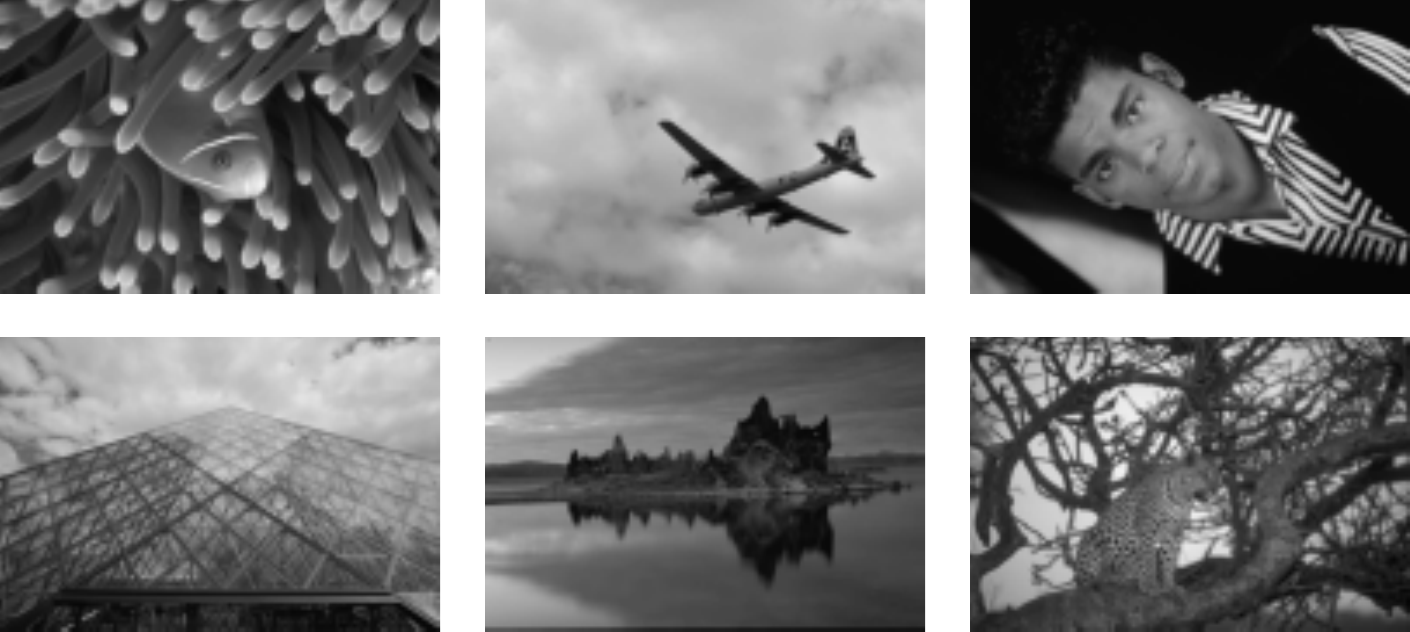}&
    \includegraphics[height=\f1ht]{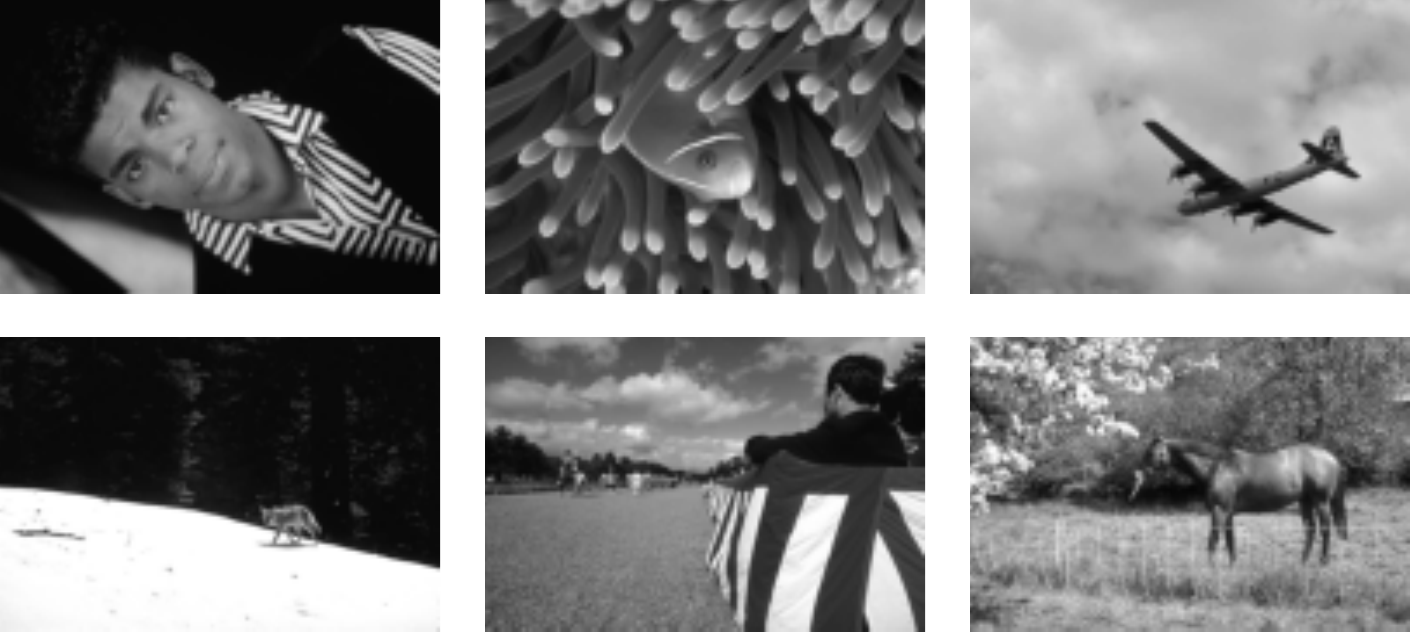}\\
    \cmidrule(lr){2-4}
    & Bottom 6  &
    \includegraphics[height=\f1ht]{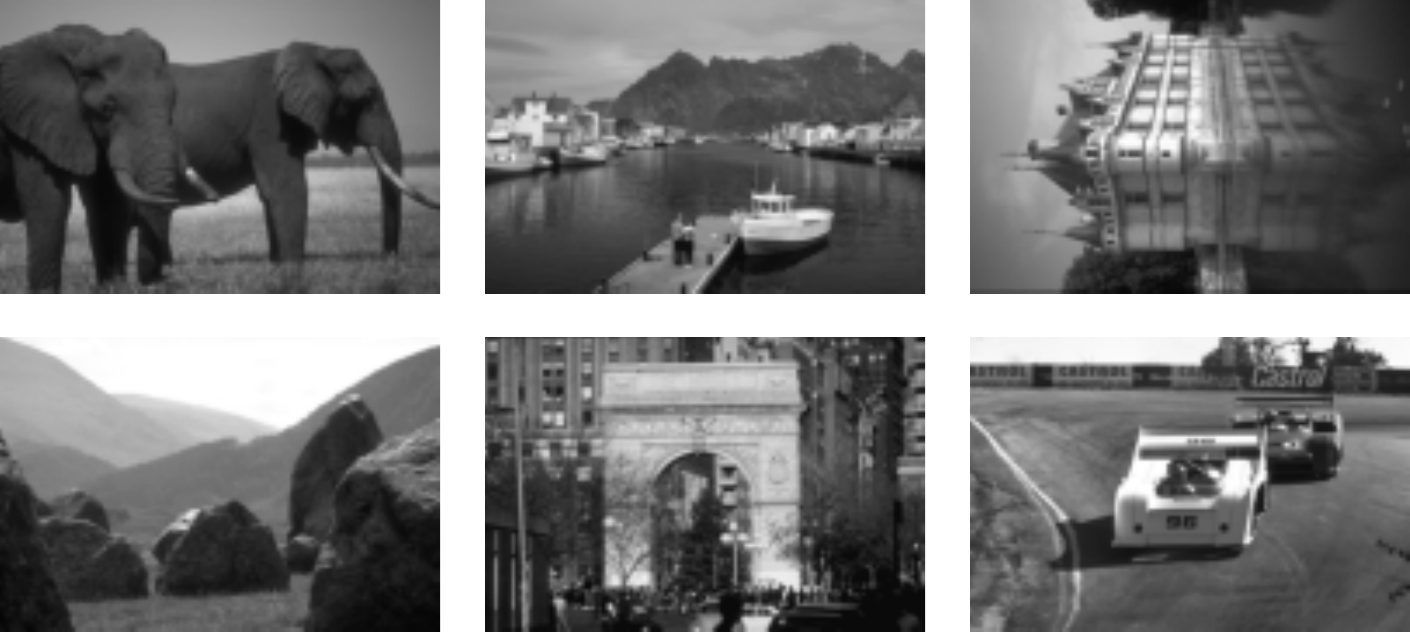}&
    \includegraphics[height=\f1ht]{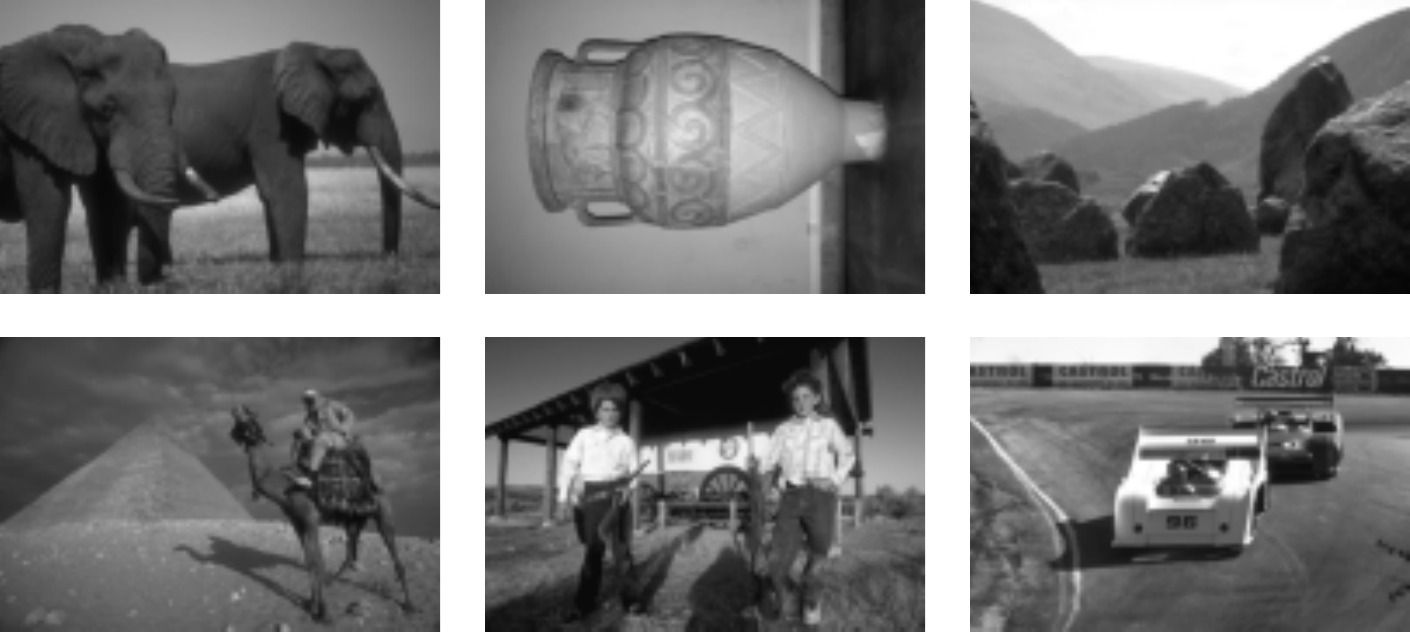}\\
    \midrule 
    \multirow{2}{*}{\rotatebox[origin=c]{90}{Out-of-distribution image}}&  Top 6 &
    \includegraphics[height=\f1ht]{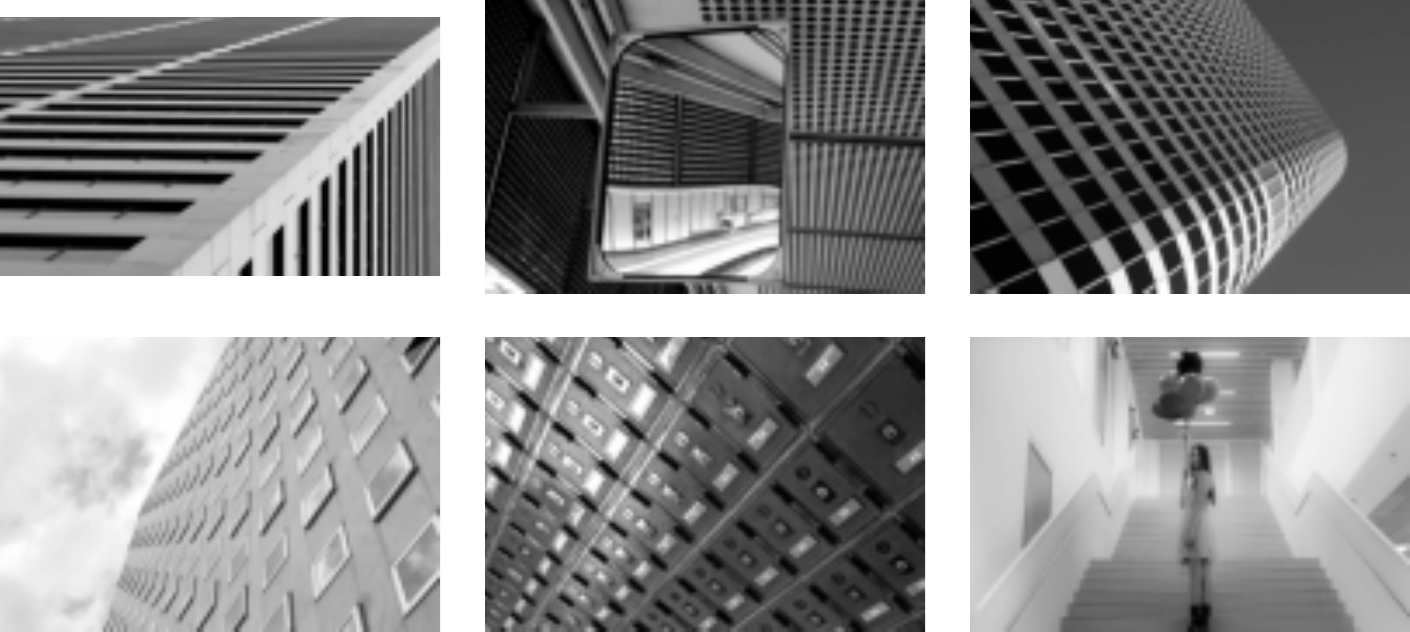}&
    \includegraphics[height=\f1ht]{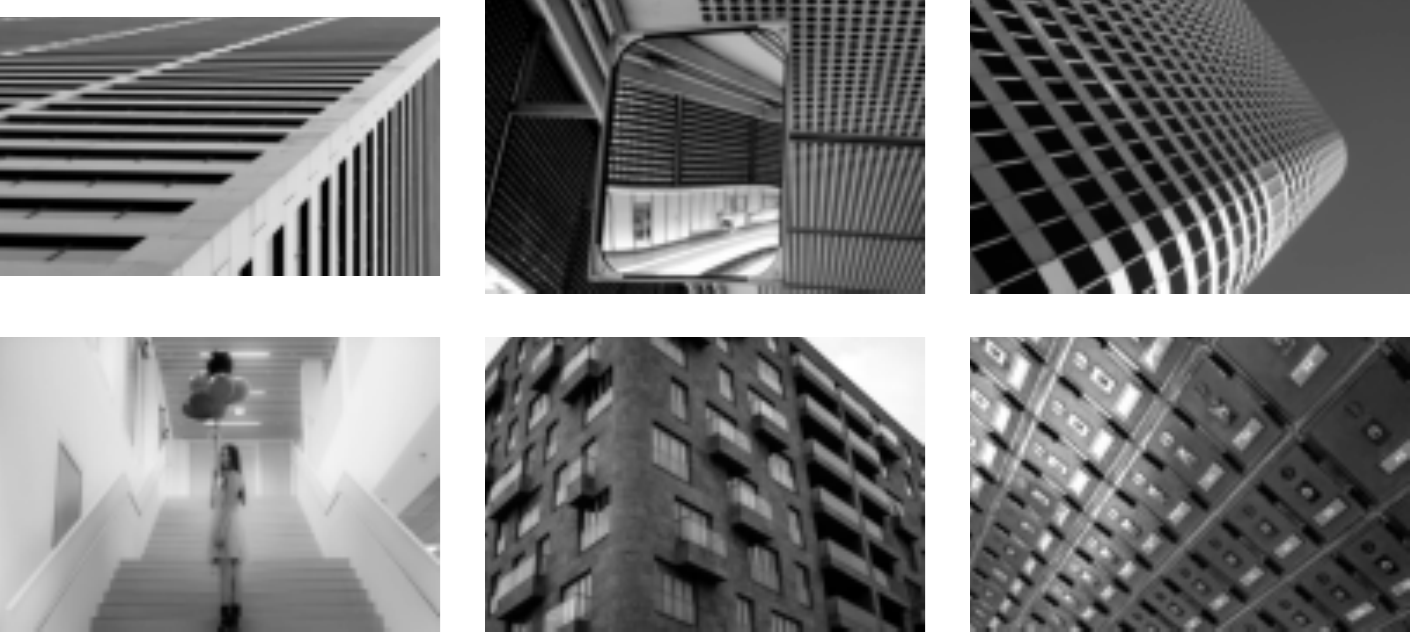}\\
    \cmidrule(lr){2-4}
    & Bottom 6 &
    \includegraphics[height=\f1ht]{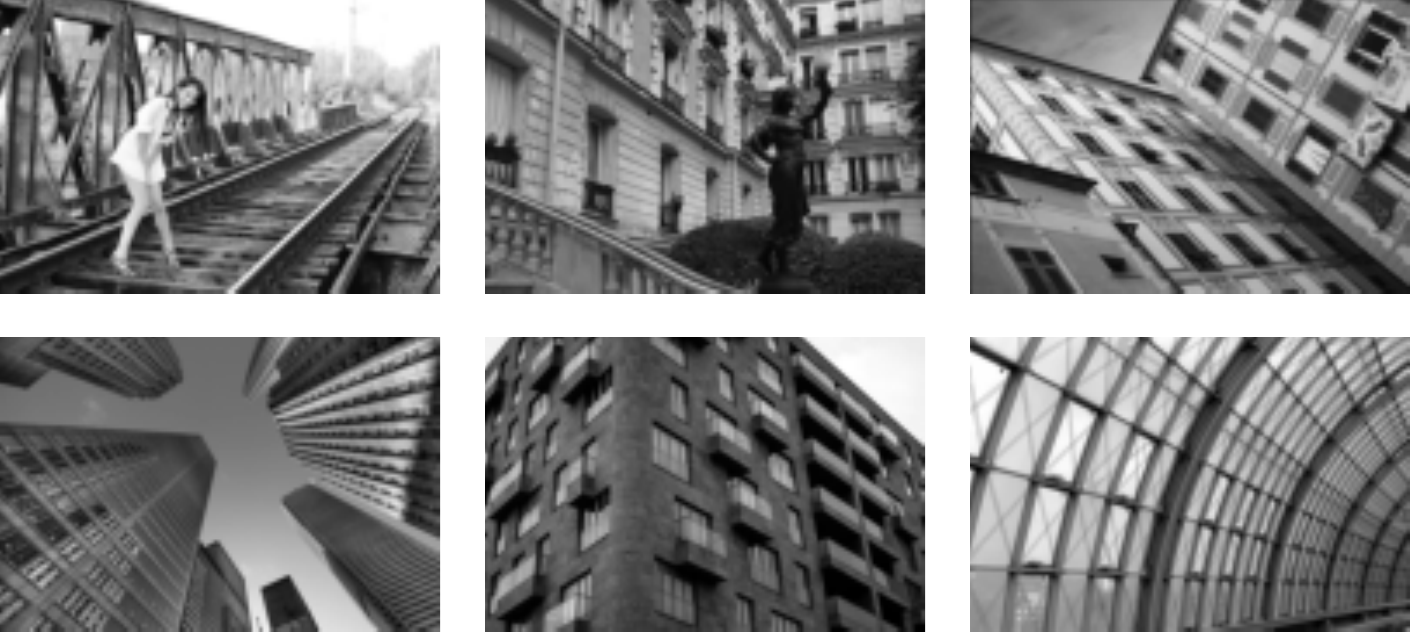}&
    \includegraphics[height=\f1ht]{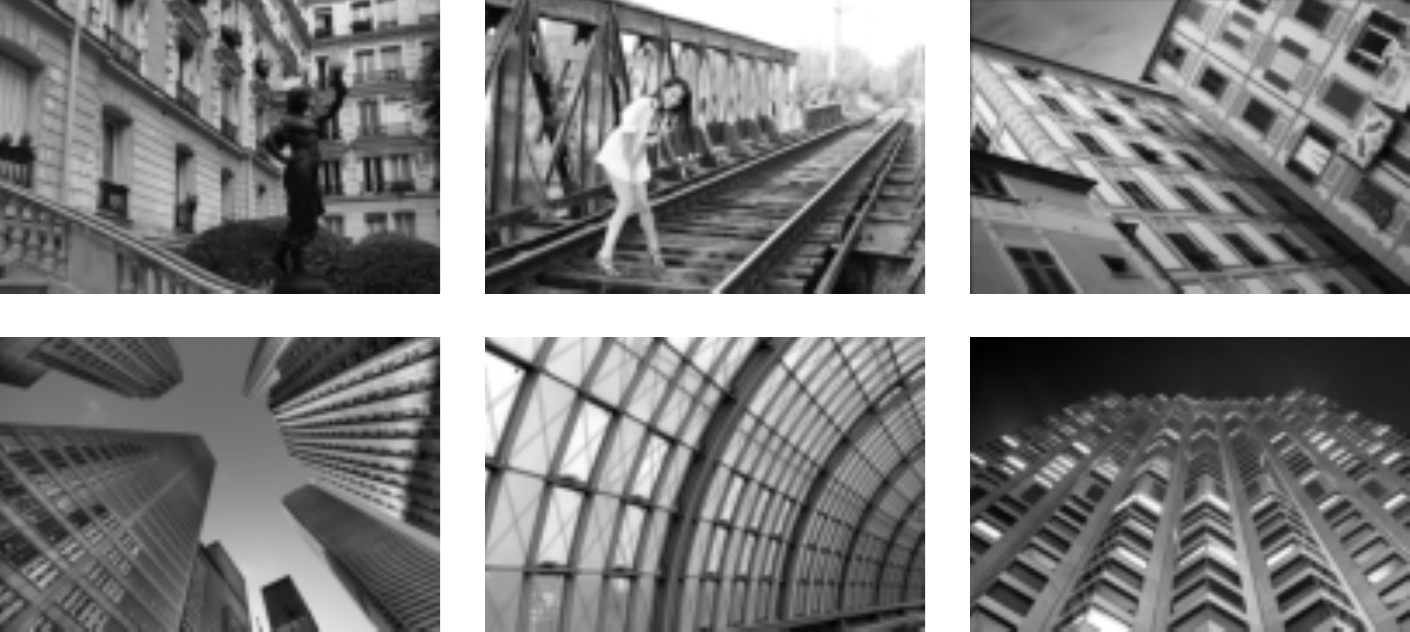}\\
    \bottomrule
\end{tabular}
}
\caption{\textbf{What kind of images benefit the most from adaptive denoising?} We visualize the images which achieve the top 6 and bottom 6 (left top to the right bottom of each grid) improvement in performance (in PSNR) after performing \gt\. Images with the largest improvement in performance often have highly repetitive patterns or large regions with constant intensity. Images with least improvement in performance tend to have more heterogeneous structure. Note that, in general, the distribution of improvements in performance is often skewed towards the images with minimal improvement in performance (See Figures~\ref{fig:sota}, \ref{fig:all_gen}, and \ref{fig:hist_in_distr}). }
\label{fig:top5bottom5}
\end{figure}

We sort images by the improvement in performance (PSNR) achieved with \gt. We observe that the ordering of images is similar for different models and cost functions (See Figure~\ref{fig:top5bottom5}), implying that the improvement in performance is mostly dependent on the image content. The images with largest improvement typically contain repeated patterns and are more structured. Repetition of patterns effectively provides multiple samples from which the unsupervised refinement can benefit.  

\subsection{Generalization via \gt }
\label{sec:suppl_analysis_gen}

\begin{figure}[t]
    \centering
    \begin{tabular}{c@{\hskip 0.1in}c@{\hskip 0.1in}c}
    {\thead{\scriptsize{(a) Dataset of} \\ \scriptsize{scanned documents}}} & \thead{\scriptsize{(b) Natural images with} \\ \scriptsize{out-of-distribution noise}} & \thead{\scriptsize{(c) Equivalent bias before} \\ \scriptsize{and after gaintuning}} \\
    \includegraphics[height=1.4in]{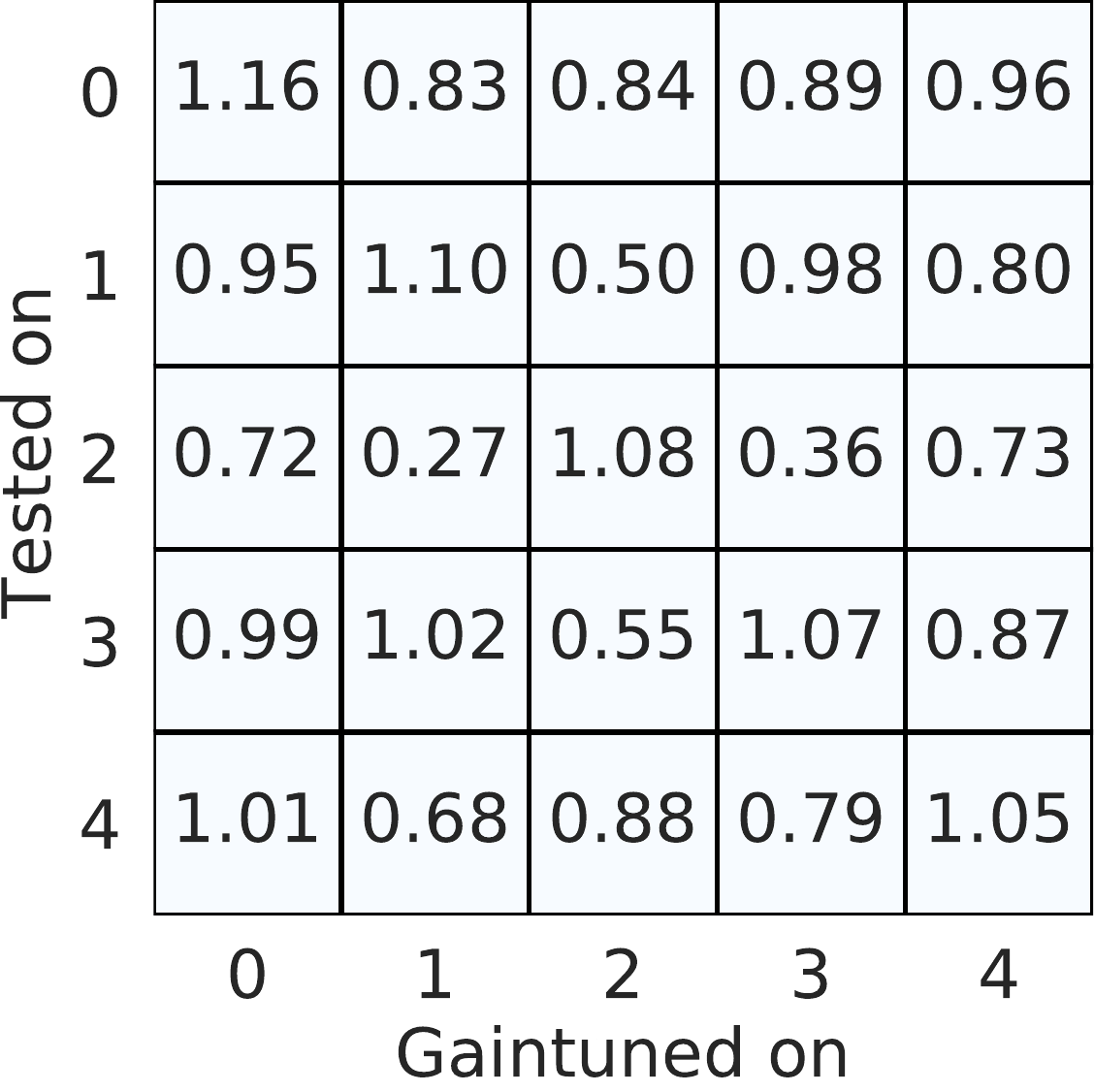}&
    \includegraphics[height=1.4in]{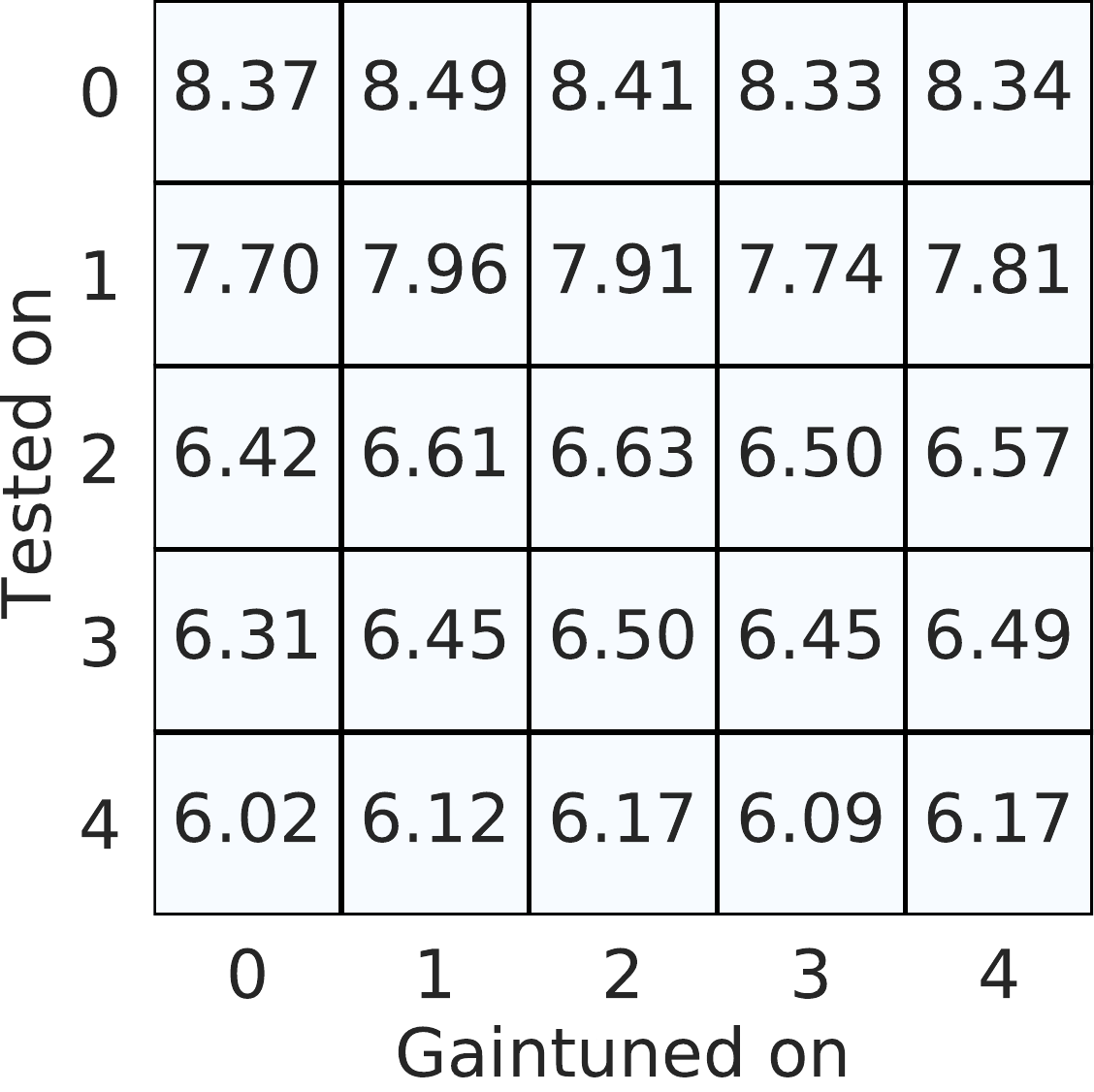}&
    \includegraphics[height=1.4in]{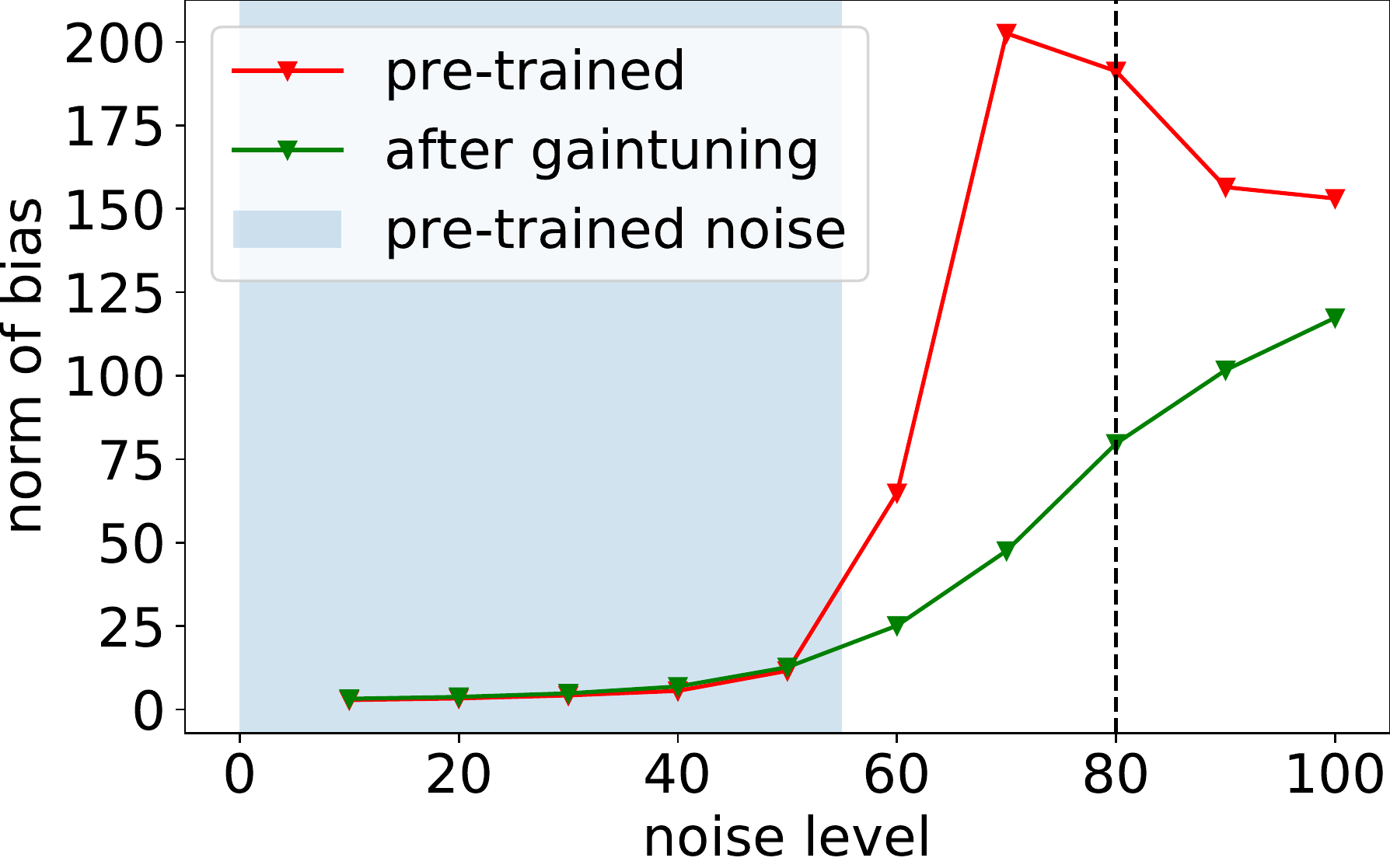}\\
    \end{tabular}
    \caption{\textbf{Analysis of \gt}. \gt\ can achieve generalization to images that are similar to the test image used for adaptation. We show this through two examples: (a) adapting a network to an image of a scanned document generalizes to other scanned documents, and (b) adapting a a network to an image with out-of-distribution noise generalizes to other images with similar noise statistics. The $(i,j)^{\text{th}}$ entry of the matrix in (a) and (b) represents the improvement in performance (measured in PNSR) when a CNN GainTuned on image $j$ is used to denoise image $i$. We use 5 images with the largest improvement in performance across the dataset for (a) and (b). Finally, (c) shows that generalization to noise levels outside the training range is enabled by reducing the \emph{equivalent bias} of the pre-trained CNN (see equation~\eqref{eq:local_affine}).  }
    \label{fig:analysis_bias}
\end{figure}

We investigate the generalization capability of \gt. We observe that a CNN adapted to a particular image via \gt\ generalizes effectively to other similar images. 
Figure~\ref{fig:analysis_bias} shows that \gt\ can achieve generalization to images that are similar to the test image used for adaptation on two examples: (1) adapting a network to an image of a scanned document generalizes to other scanned documents, and (2) adapting a a network to an image with out-of-distribution noise generalizes to other images with similar noise statistics. 

\subsection{How does \gt\ adapt to out-of-distribution noise?}
\label{sec:suppl_analysis_bias}

Let $y \in \mathbb{R}^{N}$ be a noisy image processed by a CNN. Using the first-order Taylor approximation, the function $f:\mathbb{R}^{N} \rightarrow \mathbb{R}^N$ computed by a denoising CNN may be expressed as an affine function
\begin{align}
\label{eq:local_affine}
    f(z) = f(y) + A_y(z - y) = A_y z + b_y,
\end{align}
where $A_y \in \mathbb{R}^{N \times N}$ is the Jacobian of $f(\cdot)$ evaluated at input $y$, and $b_y\in \mathbb{R}^{N}$ represents the \emph{net bias}. In~\cite{biasfree}, it was shown that the bias tends to be small for CNNs trained to denoise natural images corrupted by additive Gaussian noise, but is a primary cause of failures to generalize to noise levels not encountered during training. Figure~\ref{fig:analysis_bias} shows that \gt\ reduces the net bias of CNN, facilitating the generalization to new noise levels.

\subsection{How does \gt\ adapt to out-of-distribution images?}
\label{sec:suppl_analysis_filter}

In order to understand how \gt\ adapt to out-of-distribution images, we examine the adaptation of a CNN pre-trained on piecewise constant to natural images. Piecewise constant images have large areas with constant intensities, therefore, CNNs trained on these images tends to average over large areas. This is true even when the test image contains detailed structures. We verify this by forming the affine approximation of the network (eq. \ref{eq:local_affine}) and visualizing the equivalent linear filter~\cite{biasfree}, as explained below:

Let $y \in \mathbb{R}^{N}$ be a noisy image processed by a CNN. We process the test image using a Bias-Free CNN~\cite{biasfree} so that the net bias $b_y$ is zero in its first-order Taylor decomposition~\eqref{eq:local_affine}. When $b_y=0$,~\eqref{eq:local_affine} implies that the $i$th pixel of the output image is computed as an inner product between the $i$th row of $A_y$, denoted $a_y(i)$, and the input image:
\begin{align}
    f(y)(i) = \sum_{j=1}^{N}A_y(i,j)y(j) = a_y(i)^T y.
\end{align}
The vectors $a_y(i)$ can be interpreted as \emph{adaptive filters} that produce an estimate of the denoised pixel via a weighted average of noisy pixels. As shown in Figure~\ref{fig:filter} the denoised output of CNN pre-trained on piece wise constant images is over-smoothed and the filters average over larger areas. After \gt\ the model learns to preserve the fine features much better, which is reflected in the equivalent filters

\end{document}